\pdfoutput=1

\documentclass[11pt]{article}

\usepackage[final]{acl}

\usepackage{times}
\usepackage{latexsym}

\usepackage[T1]{fontenc}

\usepackage[utf8]{inputenc}

\usepackage{microtype}

\usepackage{inconsolata}

\usepackage{microtype}
\usepackage{hyperref}
\usepackage{url}
\usepackage{booktabs}
\definecolor{darkblue}{rgb}{0, 0, 0.5}
\hypersetup{colorlinks=true, citecolor=darkblue, linkcolor=darkblue, urlcolor=darkblue}
\usepackage{multirow}
\usepackage{multicol}
\usepackage{tablefootnote}

\usepackage{amsmath}
\usepackage{caption}
\usepackage{subcaption}
\usepackage{wrapfig}
\usepackage{xcolor} 
\usepackage{graphicx}
\usepackage{float}

\usepackage{wrapfig}

\usepackage{algorithmic}
\usepackage[linesnumbered,ruled]{algorithm2e}

\title{In-Context Learning (and Unlearning) of Length Biases}

\author{Stephanie Schoch\quad Yangfeng Ji\\
  Department of Computer Science \\
  University of Virginia \\
  Charlottesville, VA 22903 \\
  \texttt{\{sns2gr,yangfeng\}@virginia.edu} 
  }

\begin{document}
\maketitle
\begin{abstract}
Large language models have demonstrated strong capabilities to learn in-context, where exemplar input-output pairings are appended to the prompt for demonstration. However, existing work has demonstrated the ability of models to learn lexical and label biases in-context, which negatively impacts both performance and robustness of models. The impact of other statistical data biases remains under-explored, which this work aims to address. We specifically investigate the impact of length biases on in-context learning. 
We demonstrate that models do learn length biases in the context window for their predictions, and further empirically analyze the factors that modulate the level of bias exhibited by the model. 
In addition, we show that learning length information in-context can be used to counter the length bias that has been encoded in models (e.g., via fine-tuning).
This reveals the power of in-context learning in debiasing model prediction behaviors without the need for costly parameter updates.
\end{abstract}

\section{Introduction}
In-context learning (ICL) has emerged as a new ability in large language models (LLMs), representative of a novel learning paradigm \citep{wei2022emergent}. With in-context learning, an LLM learns to perform an unseen task by seeing a number of demonstrations in the context window \citep{NEURIPS2020_1457c0d6}. Whereas previous methods such as fine-tuning update the model parameters to teach the model a desired task, ICL provides the model with input-output pairs as task exemplars directly at inference, with no parameter updates. While the goal of increased task accuracy is the same, the underlying mechanisms contributing to in-context learning are still being understood. 

\begin{figure}
    \centering
        \includegraphics[width=0.49 \textwidth]{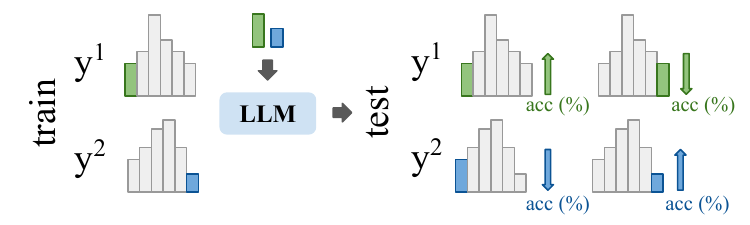}
        \caption{An illustration of our experiment setup and hypothesis. When sampling from the tails of the distribution (left of image), we introduce a data length bias. If the model can learn this shortcut feature in-context, we expect class performance on the data of similar length to be higher than data of the opposite length than what was seen in the context window (right of image).}\label{fig:sampling illlustration}
\end{figure}

This has motivated a body of work aiming to understand how in-context learning works. Some works have demonstrated simlarities between fine-tuning and in-context learning. For example, instability due to the choice of examples occurs both in few-shot finetuning \citep{schick-schutze-2021-just, gao-etal-2021-making} and in-context learning \citep{rubin-etal-2022-learning, liu-etal-2022-makes, wu-etal-2023-self}. However, other work has shown counterintuitive results when comparing the apparent learning mechanisms of in-context learning and finetuning \citep{min-etal-2022-rethinking}. 

A key area that is underexplored is whether in-context learning exhibits similar biases to finetuning with regard to statistical data biases. Statistical data biases can be defined as correlations between features and class labels. Under traditional learning paradigms such as fine-tuning, language models can learn exploitable statistical biases in the data. Such biases, or shallow features, can be exploited by a model as discriminatory features when they exhibit biased distributions across classes or are correlated with a specific class. 
This can lead to overestimates of a model's performance on the underlying task \citep{poliak-etal-2018-hypothesis, gururangan-etal-2018-annotation}. 

Prior work has identified length as an exploitable statistical bias in natural inference datasets. For example, in the MultiNLI and SNLI datasets, length has been shown to be a discriminatory feature \citep{gururangan-etal-2018-annotation}, and on the ROC story cloze task choosing the longer ending performs above random baseline \citep{cai-etal-2017-pay}.
However, length biases have been largely ignored in prior work on ICL, and some existing studies on which factors impact ICL have treated length as a static variable, selecting examples with similar lengths to test inputs \citep{min-etal-2022-rethinking}.
It is therefore unclear whether models can exploit length biases in the data under an in-context learning setting (similar to finetuning) or whether length is another factor with counterintuitive tendencies. 
\begin{figure}
    \centering
        \includegraphics[width=0.49\textwidth]{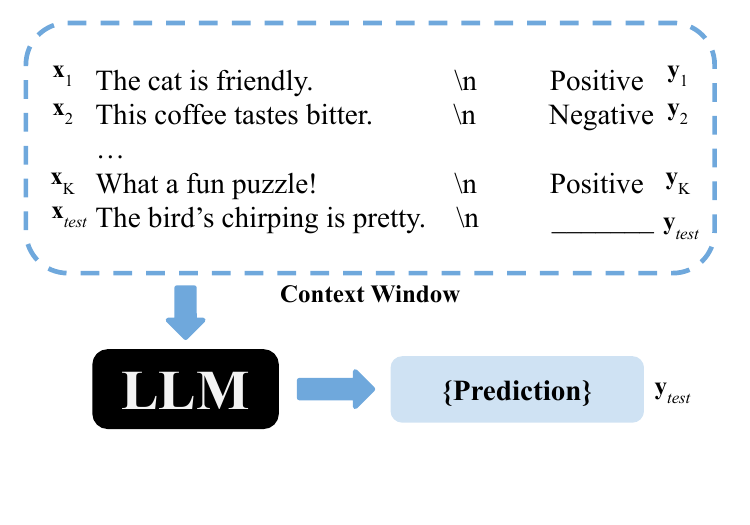}
        \vspace{-12mm}
        \caption{An overview of in-context learning using $K$ input-output demonstrations concatenated to the test input $\{x_{test}, y_{test}\}$.}\label{fig:icl_desc}
\end{figure}

In this work, we perform a series of empirical studies to investigate the ability of LLMs to learn statistical data biases in the context window during ICL (\autoref{fig:sampling illlustration}). 
This has been studied in the finetuning literature, yet is underexplored in the ICL literature. We demonstrate empirically the ability of LLMs to learn length biases in-context. In the sections to follow, we analyze which factors influence these results, and we show the efficacy of ICL in debiasing finetuned models. Our results show that ICL can introduce biases to LLMs that negatively influence task performance. Specifically, our contributions are as follows:
\begin{enumerate}
    \item We empirically demonstrate the ability of a range of LLM families to learn length biases in-context.
    \item We investigate the influence of number of examples, number of model parameters, and class-length difference on how models learn biases.
    \item We show that ICL can debias a model that contains existing length biases.
\end{enumerate}

\section{Background}
\paragraph{In-context learning}
In-context learning is an emergent ability of LLMs that enables pre-trained models to learn an unseen task using a set of exemplars concatenated in the context window (see \autoref{fig:icl_desc}). Formally, given a test example $x$, in-context learning concatenates $K$ demonstration examples to the task instruction $I$, where $S = \{x_{i}, y_{i}\}_{i=1}^K$ denotes the example set.
The performance of in-context learning, however, is highly dependent on both the selected examples \citep{rubin-etal-2022-learning, liu-etal-2022-makes, wu-etal-2023-self, pmlr-v202-ye23c} and their orderings \citep{lu-etal-2022-fantastically, chen-etal-2023-relation}, resulting in performance variation from nearly random to comparable with finetuned models.

\paragraph{In-Context Learning \& Bias}
While in-context learning has shown significant potential as a way to extract relevant information from an LLM and align the model with user expectations, it has also exhibited brittleness to an assortment of factors. These include selected examples \citep{rubin-etal-2022-learning, liu-etal-2022-makes, wu-etal-2023-self, pmlr-v202-ye23c} and their orderings \citep{lu-etal-2022-fantastically, chen-etal-2023-relation}, which have recently been categorized under the umbrella of demonstration biases \citep{li-etal-2024-debiasing}. 

Beyond demonstration bias, instability of ICL has been attributed to biases in the model toward predicting certain answers due to majority label bias, recency bias, and common token bias \citep{pmlr-v139-zhao21c}. Correspondingly, several works have looked at identifying and mitigating label bias \citep{pmlr-v139-zhao21c, fei-etal-2023-mitigating} \cite{fei-etal-2023-mitigating} with respect to lexical information, and \citet{ali2024mitigating} have looked at mitigating ``copy bias'', where LLMs copy lexical information from demonstrations rather than learning underlying task information. 

However, statistical data biases such as length information have been largely ignored in the in-context learning literature, yet received wide attention in the natural language inference literature with respect to traditional finetuned models \citep{mccoy-etal-2019-right, poliak-etal-2018-hypothesis, cai-etal-2017-pay, gururangan-etal-2018-annotation}. Our work bridges this gap by looking at in-context learning with relation to a specific statistical bias: length bias.

\section{Experiment Setup}

\begin{table*}[ht]
\centering
\begin{tabular}{lllll}
  \toprule
  \textbf{Category} & \textbf{Dataset} & \textbf{Task} & \textbf{\#Train} & \textbf{\#Val} \\
  \midrule
  \multirow{4}{*}{Inference} & QNLI \citep{wang-etal-2018-glue} & Natural Language Inference & 105k & 5.46k \\
  & RTE \tablefootnote{\citep{dagan2006pascal, bar2006second, giampiccolo2007third, bentivogli2009fifth}} & Natural Language Inference & 2.49k & 277 \\
  & WNLI \citep{levesque2011winograd} & Natural Language Inference & 635 & 71 \\
  & HANS \citep{DBLP:journals/corr/abs-1902-01007} & Natural Language Inference & 30k & 30k \\
  \midrule
  Single Sentence & SST-2 \citep{socher2013recursive} & Sentiment Analysis & 67.3k & 872 \\
  \midrule
  \multirow{1}{*}{Paraphrase} & MRPC \citep{dolan2004unsupervised} & Paraphrase Detection & 3.67k & 408 \\
  Detection & PAWS-X$_{\textsc{en}}$ \citep{yang-etal-2019-paws} & Paraphrase Detection & 49.4k & 2k \\
  \bottomrule
\end{tabular}
\caption{\label{table:datasets} Datasets used in our experiments. We use the distributions available from Huggingface \citep{lhoest-etal-2021-datasets}, and use the respective validation sets to measure performance. Dataset descriptions can be found in \autoref{table:dataset_descriptions}.}
\end{table*}

\begin{table}[ht]
\centering
\begin{tabular}{ll}
  \toprule
  \textbf{Model} & \textbf{Parameters} \\
  \midrule
  \multirow{1}{*}{LLaMa 3} \citep{dubey2024llama3herdmodels} & 8B \\
  \multirow{1}{*}{LLaMa 2 \citep{touvron2023llama}} & 7B \\
  Mistral \citep{jiang2023mistral7b} & 7B \\
  OPT \citep{zhang2022opt} & 6.7B \\
\multirow{1}{*}{GPT-Neo \citep{gpt-neo}} & 2.7B \\
  \bottomrule
\end{tabular}
\caption{Models used in \autoref{sec:q1}. \label{table:models}}
\end{table}

In this section, we describe the experiment setup used in our analyses. 

\paragraph{Datasets} We use $7$ binary classification datasets, representing natural language inference, sentiment analysis, and paraphrase detection tasks. As we sample from the tails of the length distributions, binary classification is ideal for our setting. For each dataset, we utilize the splits available from Huggingface. Dataset statistics are provided in \autoref{table:datasets}, with detailed descriptions in \autoref{app:datasets}. To count the length of each input, we use the \textsc{nltk} word-tokenize package \citep{bird-loper-2004-nltk} rather than the LLM-specific tokenizers, to maintain consistency across experiments.
Prompts are adapted from \cite{eval-harness} and provided in \autoref{app:prompts}.

\paragraph{Models} Experiments in \autoref{sec:q1} are run using five models from different LLM families, listed in \autoref{table:models}. The selected models vary in size from 2.7B parameters to 8B parameters. 
Notably, the upper bound of the parameter range is due to our resource constraint, as each experiment is run using a single NVIDIA A100 GPU. For experiments in \autoref{sec:factors} and \autoref{sec:int}, we use a subset of these models, Llama3 and GPT-Neo. 
For experiments in \autoref{sec:v1}, we use the OPT model family. 

\paragraph{Other Details} Following \citet{min-etal-2022-rethinking}, unless otherwise noted, all experiments use $k=16$ demonstrations. For finetuning experiments, we use $k=200$ finetuning examples. To minimize the impact of ordering effects, each result represents the mean of $4$ trials, with standard deviation shown using error bars. Results are all run on the full validation split of each dataset.

In \autoref{sec:q1}, we investigate whether LLMs can learn length biases in-context. To further analyze these results, in \autoref{sec:factors} we look at the impact of model parameter size, number of examples, and length distribution. Finally, \autoref{sec:int} demonstrates the utility of ICL to debias finetuned models that exhibit length biases.

\section{Length Biases in Finetuning and ICL}\label{sec:q1}
In this section, we investigate the question \textbf{\textit{do models learn length biases in-context?}} We demonstrate empirically the ability of LLMs to learn length biases in-context.

\subsection{Method}

\begin{figure*}[t!]
    \centering
    \begin{minipage}[t]{\linewidth}
        \begin{subfigure}{0.19\linewidth}
            \centering
            \includegraphics[width=\textwidth]{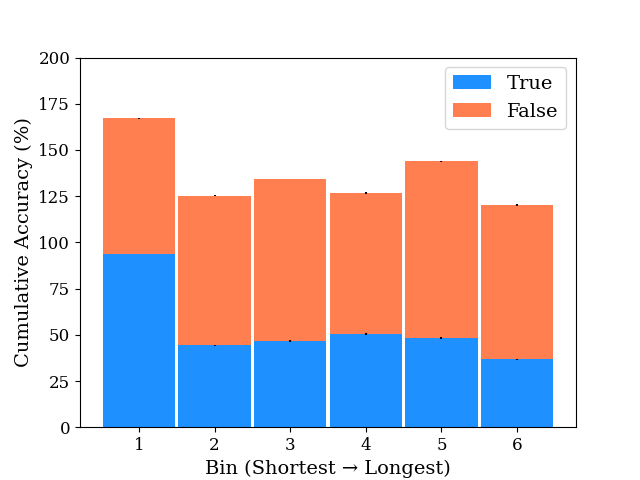}
            \caption{Llama3 8B}\label{fig:image1}
        \end{subfigure}%
        \hfill
        \begin{subfigure}{0.19\linewidth}
            \centering
            \includegraphics[width=\textwidth]{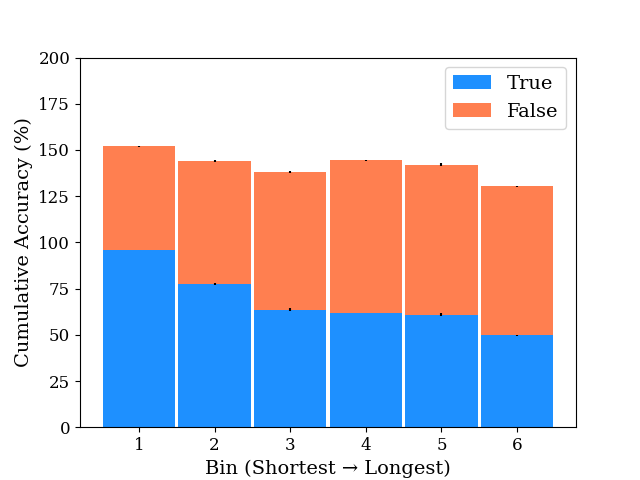}
            \caption{Llama2 7B}\label{fig:image2}
        \end{subfigure}
        \hfill
        \begin{subfigure}{0.19\linewidth}
            \centering
            \includegraphics[width=\textwidth]{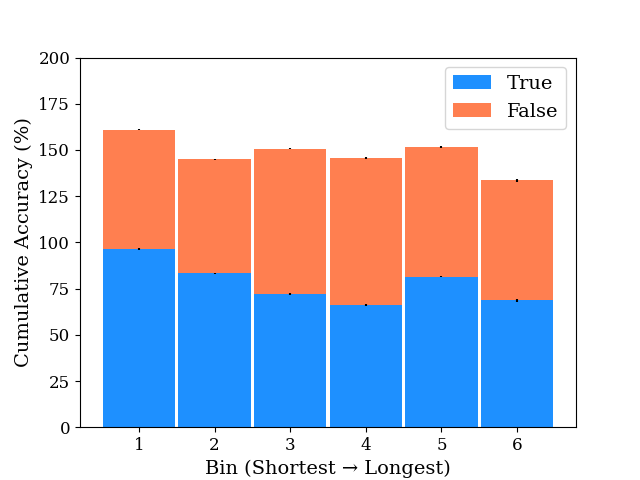}
            \caption{Mistral 7B}\label{fig:image3}
        \end{subfigure}
        \begin{subfigure}{0.19\linewidth}
            \centering
            \includegraphics[width=\textwidth]{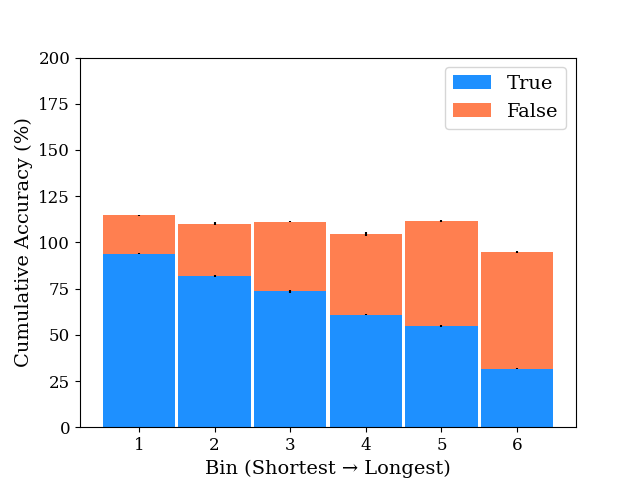}
            \caption{OPT 6.7B}\label{fig:image4}
        \end{subfigure}%
        \begin{subfigure}{0.19\linewidth}
        \centering
        \includegraphics[width=\textwidth]{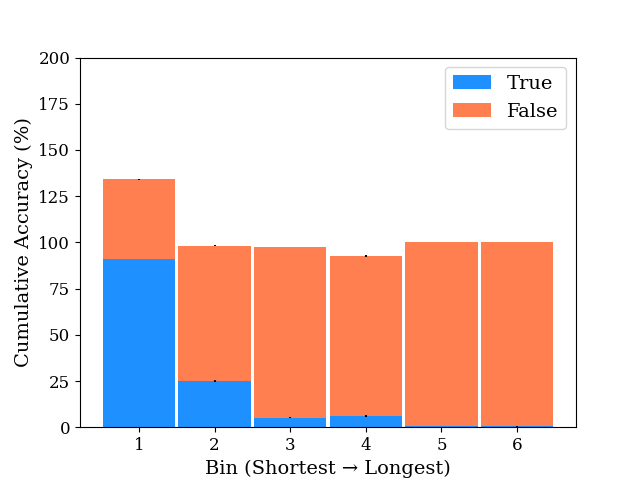}
        \caption{GPT Neo 2.7B}\label{fig:image4}
        \end{subfigure}%
    \end{minipage}%
    \hfill
    \begin{minipage}[c]{\linewidth}
        \caption{\label{fig:q1-icl-hans}In-context learning validation performance across different models on the Hans dataset. For each graph, $y_1$ (Blue) was sampled from the short instances, and $y_2$ (Orange) was sampled from the long instances.}
    \end{minipage}
\end{figure*}
\begin{figure*}[t!]
    \centering
    \begin{minipage}[t]{\linewidth}
        \begin{subfigure}{0.19\linewidth}
            \centering
            \includegraphics[width=\textwidth]{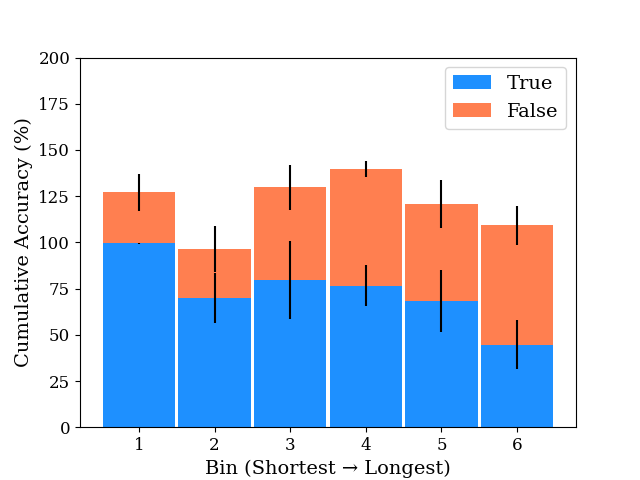}
            \caption{Llama3 8B}\label{fig:image1}
        \end{subfigure}%
        \hfill
        \begin{subfigure}{0.19\linewidth}
            \centering
            \includegraphics[width=\textwidth]{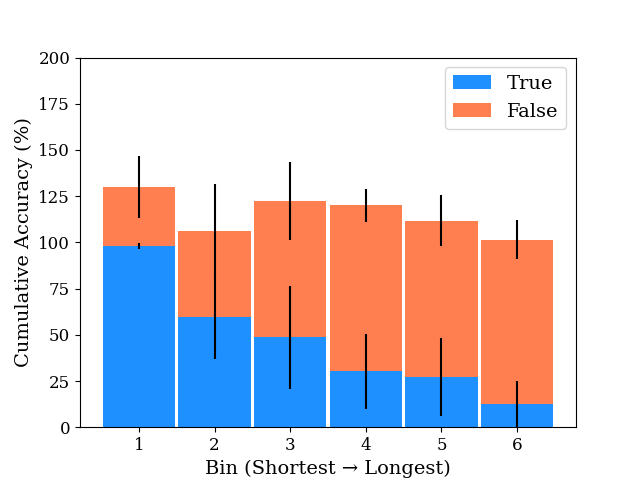}
            \caption{Llama2 7B}\label{fig:image2}
        \end{subfigure}
        \hfill
        \begin{subfigure}{0.19\linewidth}
            \centering
            \includegraphics[width=\textwidth]{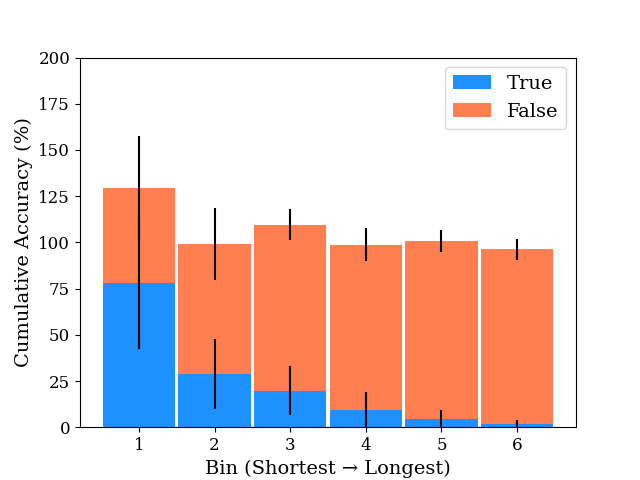}
            \caption{Mistral 7B}\label{fig:image3}
        \end{subfigure}
        \begin{subfigure}{0.19\linewidth}
            \centering
            \includegraphics[width=\textwidth]{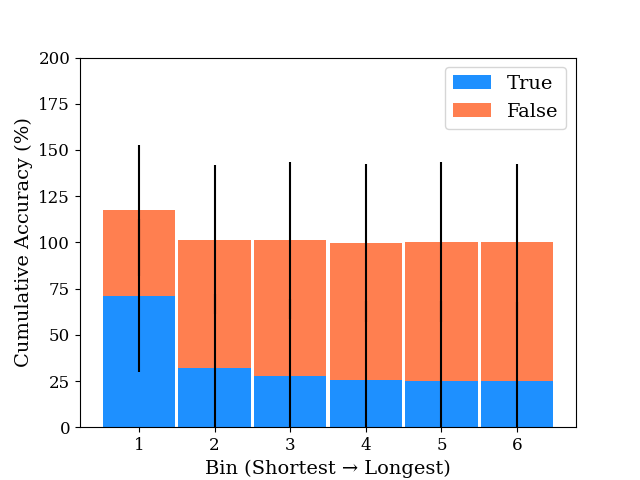}
            \caption{OPT 6.7B}\label{fig:image4}
        \end{subfigure}%
        \begin{subfigure}{0.19\linewidth}
        \centering
        \includegraphics[width=\textwidth]{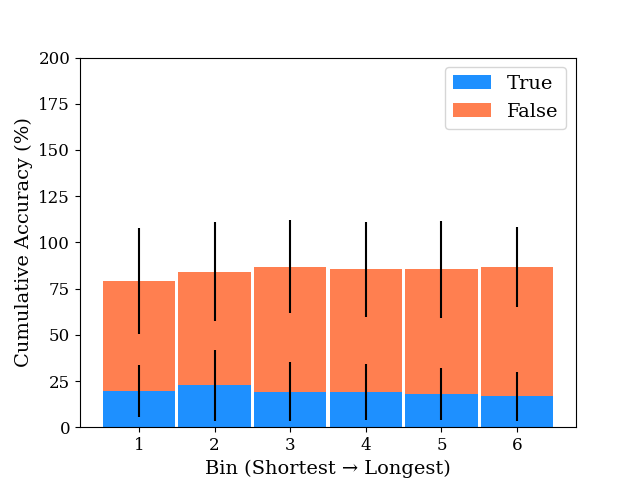}
        \caption{GPT Neo 2.7B}\label{fig:image4}
        \end{subfigure}%
    \end{minipage}%
    \hfill
    \begin{minipage}[c]{\linewidth}
        \caption{\label{fig:q1-ft-hans}Finetuning validation performance across different models on the Hans dataset. For each graph, $y_1$ (Blue) was sampled from the short instances, and $y_2$ (Orange) was sampled from the long instances.}
    \end{minipage}
\end{figure*}
\begin{figure*}[t!]
    \centering
    \begin{minipage}[t]{\linewidth}
        \begin{subfigure}{0.19\linewidth}
            \centering
            \includegraphics[width=\textwidth]{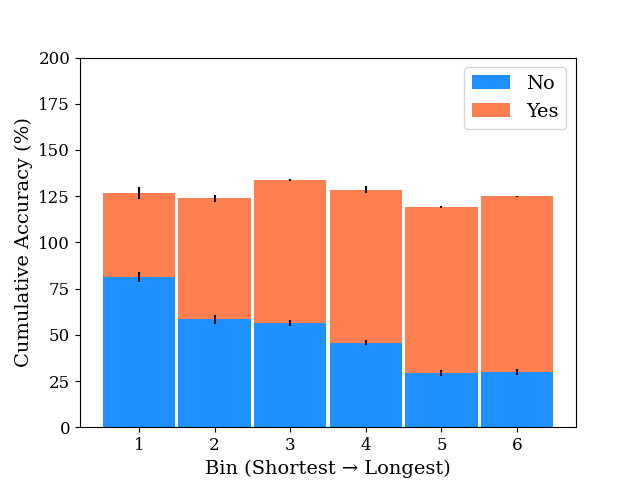}
            \caption{Llama3 8B}\label{fig:image1}
        \end{subfigure}%
        \hfill
        \begin{subfigure}{0.19\linewidth}
            \centering
            \includegraphics[width=\textwidth]{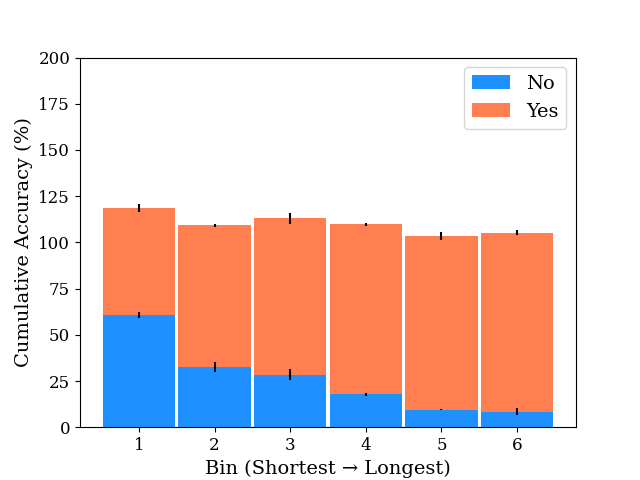}
            \caption{Llama2 7B}\label{fig:image2}
        \end{subfigure}
        \hfill
        \begin{subfigure}{0.19\linewidth}
            \centering
            \includegraphics[width=\textwidth]{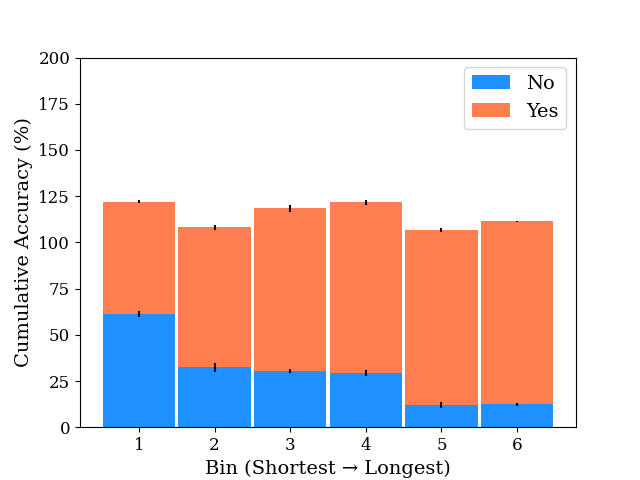}
            \caption{Mistral 7B}\label{fig:image3}
        \end{subfigure}
        \begin{subfigure}{0.19\linewidth}
            \centering
            \includegraphics[width=\textwidth]{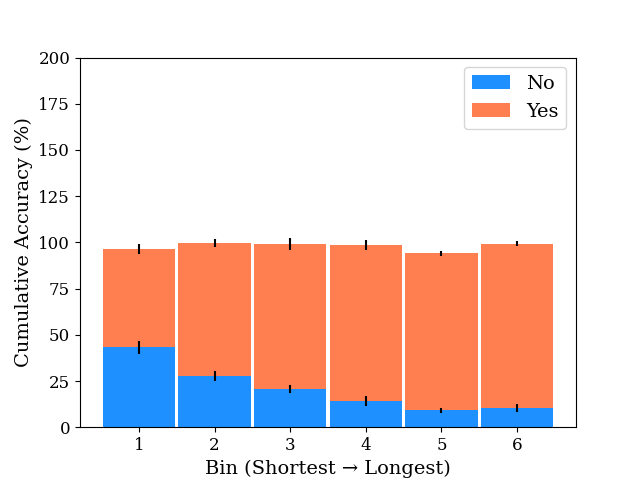}
            \caption{OPT 6.7B}\label{fig:image4}
        \end{subfigure}%
        \begin{subfigure}{0.19\linewidth}
        \centering
        \includegraphics[width=\textwidth]{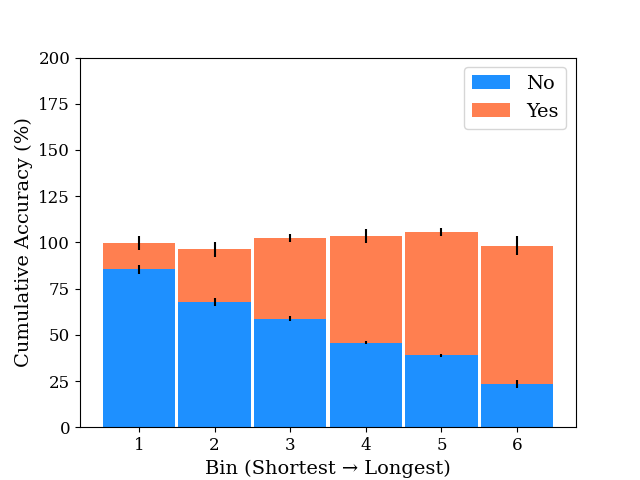}
        \caption{GPT Neo 2.7B}\label{fig:image4}
        \end{subfigure}%
    \end{minipage}%
    \hfill
    \begin{minipage}[c]{\linewidth}
        \caption{\label{fig:q1-icl-en}In-context learning validation performance across different models on the PAWS-X$_{\textsc{EN}}$ dataset. For each graph, $y_1$ (Blue) was sampled from the short instances, and $y_2$ (Orange) was sampled from the long instances.}
    \end{minipage}
\end{figure*}
\begin{figure*}[t!]
    \centering
    \begin{minipage}[t]{\linewidth}
        \begin{subfigure}{0.19\linewidth}
            \centering
            \includegraphics[width=\textwidth]{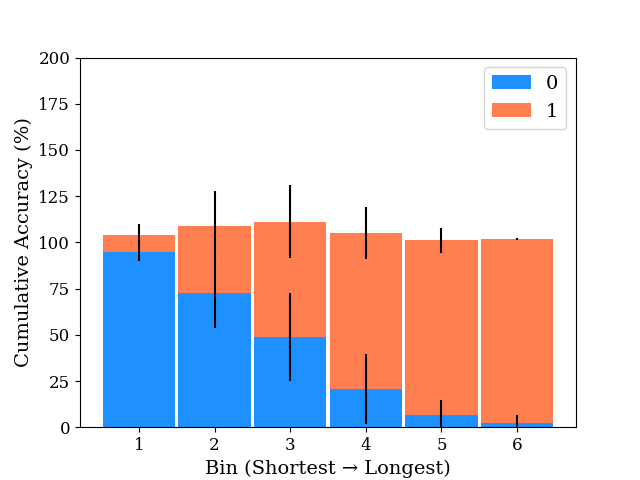}
            \caption{Llama3 8B}\label{fig:image1}
        \end{subfigure}%
        \hfill
        \begin{subfigure}{0.19\linewidth}
            \centering
            \includegraphics[width=\textwidth]{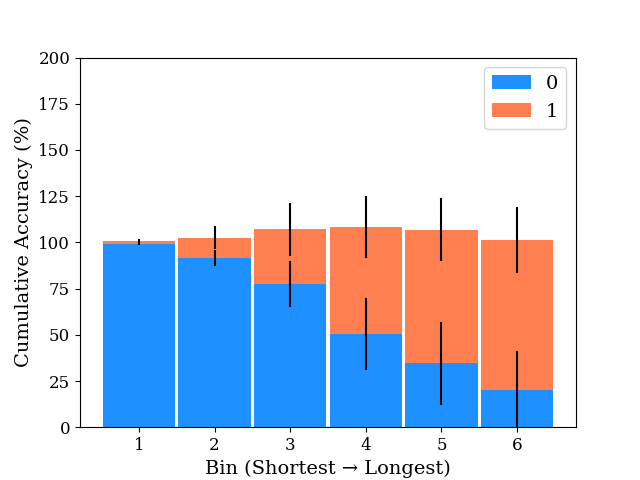}
            \caption{Llama2 7B}\label{fig:image2}
        \end{subfigure}
        \hfill
        \begin{subfigure}{0.19\linewidth}
            \centering
            \includegraphics[width=\textwidth]{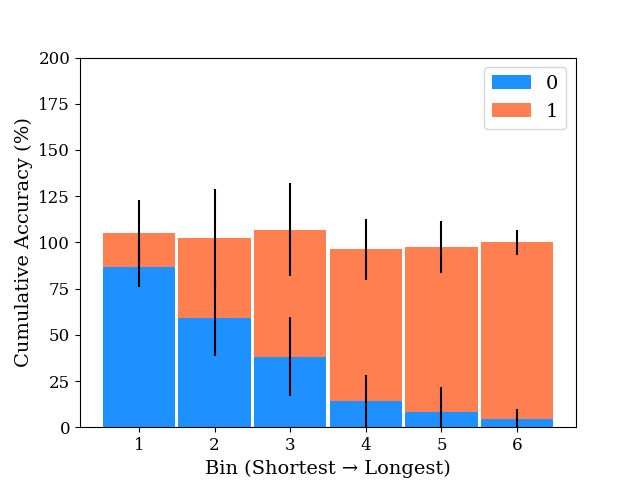}
            \caption{Mistral 7B}\label{fig:image3}
        \end{subfigure}
        \begin{subfigure}{0.19\linewidth}
            \centering
            \includegraphics[width=\textwidth]{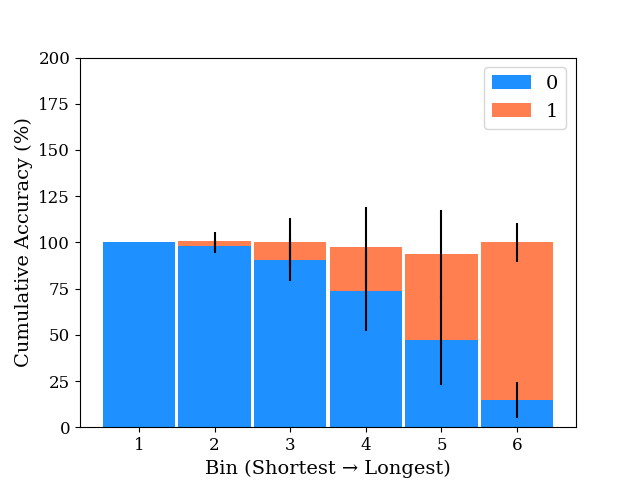}
            \caption{OPT 6.7B}\label{fig:image4}
        \end{subfigure}%
        \begin{subfigure}{0.19\linewidth}
        \centering
        \includegraphics[width=\textwidth]{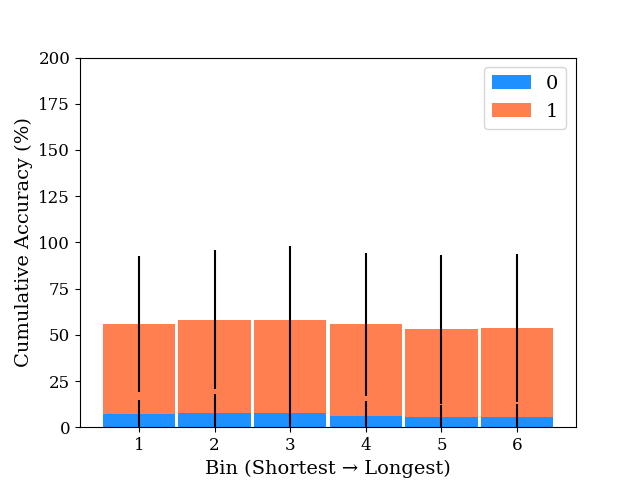}
        \caption{GPT Neo 2.7B}\label{fig:image4}
        \end{subfigure}%
    \end{minipage}%
    \hfill
    \begin{minipage}[c]{\linewidth}
        \caption{\label{fig:q1-ft-en}Finetuning validation performance across different models on the PAWS-X$_{\textsc{EN}}$ dataset. For each graph, $y_1$ (Blue) was sampled from the short instances, and $y_2$ (Orange) was sampled from the long instances.}
    \end{minipage}
\end{figure*}

Consider a dataset $D=\{(x_i ,y_i)\}_{i=1}^n$  that contains $n$ training instances. In this work, we consider binary classification datasets, where $Y=\{y^1, y^2\}$. We aim to introduce a distributional bias in the input lengths with respect to class.  To introduce a length bias in $k$ demonstrations drawn from $D$, we sample from the tails of each class's input length distribution. Specifically, we sample the top-$\frac{k}{2}$ examples belonging to $y^1$ and the bottom-$\frac{k}{2}$ from $y^2$ (and vice versa). This effectively produces a ``worst-case scenario'' in maximizing the distance between the classes under the given length distributions. 

To provide a baseline for comparison, we compare against finetuning. Specifically, we finetune each model (using LoRA adapters \citep{hu2022lora}) on $k=200$ training instances selected using the same procedure as above. As an additional baseline, for all experiments, we compare against randomly sampling the demonstrations and finetuning examples.

For results, we utilize a binning procedure. 
Specifically, we bin the validation set based on length, with $b=6$ bins. In this setting, bin $1$ represents the shortest $16.67\%$ of validation instances and bin $6$ represent the longest $16.67\%$ validation instances across both classes. If a model has learned a length bias, for the validation class with the training set drawn from the shortest instances, we expect performance on bin $1$ to be higher than bin $6$, and vice versa for the validation class where the training set is drawn from the longest instances. 

As a performance measurement, we report the sum of the individual accuracy from each class. 
As there may be a slight imbalance across classes in each bin, reporting individual class accuracy rather than the percentage of the entire bin ensures we account for class imbalances across bins.

\subsection{Results}
We report results on HANS and PAWS-X$_{\textsc{EN}}$ under finetuning (\autoref{fig:q1-ft-hans} and \autoref{fig:q1-ft-en}) and ICL (\autoref{fig:q1-icl-hans} and \autoref{fig:q1-icl-en}), where $y_1$ demonstrations were sampled from short instances and $y_2$ demonstrations were sampled from long instances. $y_1$ and $y_2$ correspond to the Blue and Orange bars, respectively.
Our results show decreased performance on validation examples that do not have a similar length as the demonstrations belonging to each respective class. This indicates that models can pick up length biases in-context. Additional results can be found in the Appendix.

\section{Analysis of Influencing Factors}\label{sec:factors}
In this section, we investigate a further question of \textbf{\textit{what factors influence how LLMs learn length biases in-context?}} We find that increased numbers of examples can exacerbate learned biases, and models across a range of sizes can learn length biases. Further, we find that length bias can be learned from as little as a few tokens of difference in average length between classes. 

\subsection{Number of Model Parameters}\label{sec:v1}
Existing work has suggested that the number of model parameters influences the ability of models to learn in-context, with larger models performing better \citep{milios-etal-2023-context, lu-etal-2022-fantastically}. In this section, we investigate whether the number of parameters also influences the ability of models to learn length biases in-context. For example, if larger models are better at learning in-context, are they more susceptible or more resilient to learning statistical biases in the data? 

We use the OPT model family \citep{zhang2022opt} across $p=\{350\mathrm{M}, 1.3\mathrm{B}, 2.7\mathrm{B}, 6.7\mathrm{B}\}$ parameters with $k=16$ in-context examples. Note that the parameter count is upper-bounded based on computational resources. We use the procedure described in \autoref{sec:q1} to introduce a length bias in the in-context demonstrations.

\paragraph{Results}
\begin{figure}[t!]
    \centering
    \begin{minipage}[t]{\linewidth}
        \begin{subfigure}{0.49\linewidth}
            \centering
            \includegraphics[width=\textwidth]{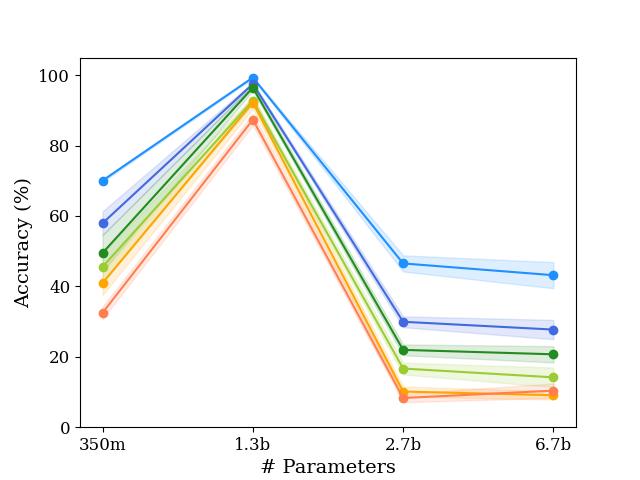}
            \caption{PAWS-X$_{\textsc{EN}}$: \\ $y_1$ (biased sampling)}\label{fig:image1}
        \end{subfigure}%
        \hfill
        \begin{subfigure}{0.49\linewidth}
            \centering
            \includegraphics[width=\textwidth]{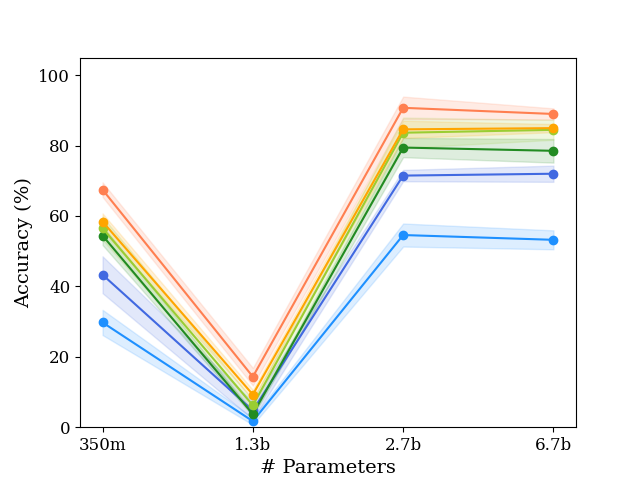}
            \caption{PAWS-X$_{\textsc{EN}}$: \\ $y_2$ (biased sampling)}\label{fig:image2}
        \end{subfigure}
        \hfill
        \begin{subfigure}{0.49\linewidth}
            \centering
            \includegraphics[width=\textwidth]{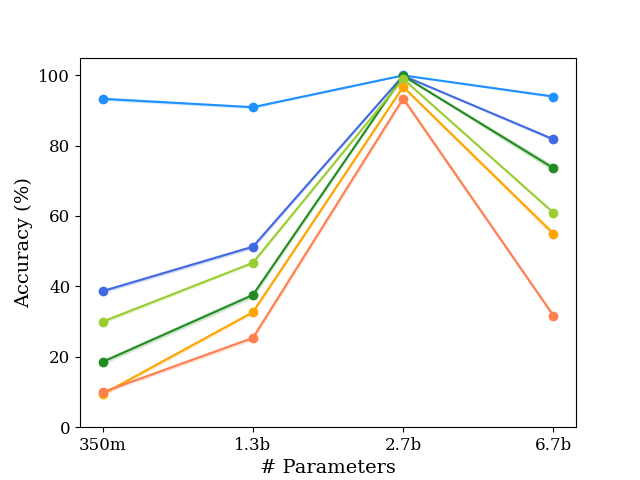}
            \caption{Hans: \\ $y_1$ (biased sampling)}\label{fig:image3}
        \end{subfigure}
        \hfill
        \begin{subfigure}{0.49\linewidth}
            \centering
            \includegraphics[width=\textwidth]{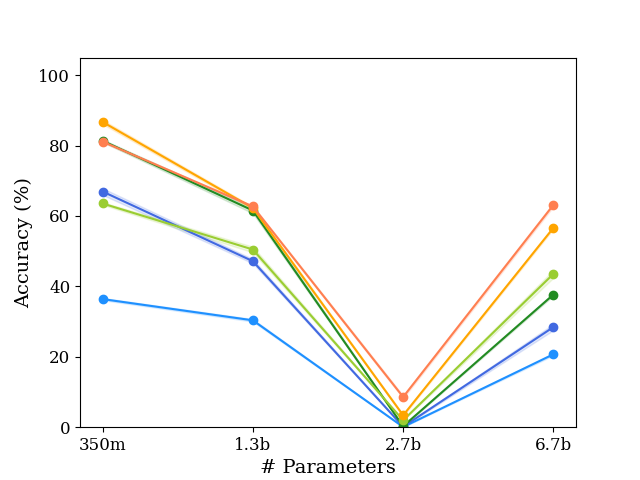}
            \caption{Hans: \\ $y_2$ (biased sampling)}\label{fig:image4}
        \end{subfigure}%
    \end{minipage}%
    \hfill
    \begin{minipage}[c]{\linewidth}
        \centering
        \vspace{2mm}
            \begin{subfigure}{0.7 \linewidth}
            \centering
            \includegraphics[width=\textwidth]{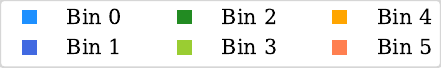}
        \end{subfigure}%
    \end{minipage}
    \hfill
    \begin{minipage}[c]{\linewidth}
        \caption{\label{fig:num_params} Validation performance across different numbers of model parameters using the OPT model family, on the PAWS-X$_{\textsc{EN}}$ and Hans datasets. In this figure, in-context examples from $y_2$ were sampled from long instances, and in-context examples from $y_1$ were sampled from short instances. Each subfigure shows results on the validation instances in the respective class, with Bin 0 containing the shortest demonstrations and Bin 5 containing the longest demonstrations. Additional results can be found in the Appendix.}
    \end{minipage}
\end{figure}

We report results on HANS and PAWS-X$_\textsc{EN}$ in \autoref{fig:num_params}. Notably, both datasets are designed to be challenging (see \autoref{app:datasets} for descriptions). Remaining datasets and conditions are reported in the Appendix. While we do observe length bias across varying model parameter sizes, there is not a consistent pattern of increased or decreased bias with increased model parameter sizes. Accordingly, we observe a dataset-model dependence with regard to the degree of length bias a model may learn.

\subsection{Number of Examples}\label{sec:v2}
The performance of ICL when using various numbers of examples has been studied in prior work \citep{wu-etal-2023-self, min-etal-2022-rethinking, lu-etal-2022-fantastically}. As such, we investigate the sensitivity of LLM's ability to learn length bias across different numbers of in-context examples. 

We use $k=\{2,4,8,16,24,32\}$ in-context examples on the datasets in \autoref{table:datasets} using Llama3 8B and GPT Neo 2.7B. Following the procedure from \autoref{sec:q1}, we select the longest $\frac{k}{2}$ examples from $y^1$ and shortest $\frac{k}{2}$ examples from the $y^2$ (and vice versa), thereby introducing a bias in the length distribution of inputs across classes. 

\begin{figure}[t!]
    \centering
    \begin{minipage}[t]{\linewidth}
        \begin{subfigure}{0.49\linewidth}
            \centering
            \includegraphics[width=\textwidth]{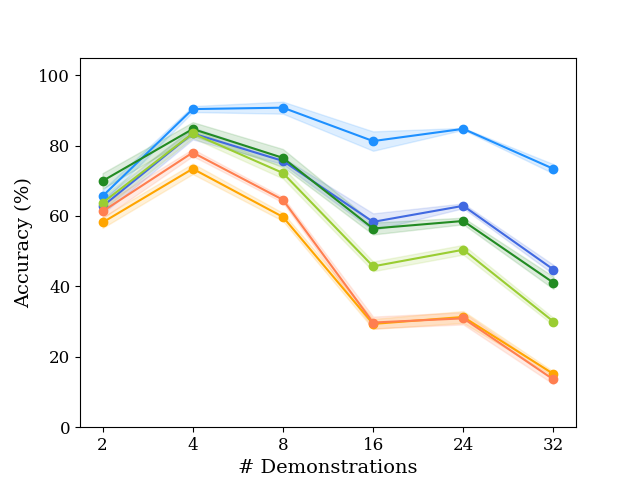}
            \caption{$y_1$ (biased sampling)}\label{fig:image1}
        \end{subfigure}%
        \hfill
        \begin{subfigure}{0.49\linewidth}
            \centering
            \includegraphics[width=\textwidth]{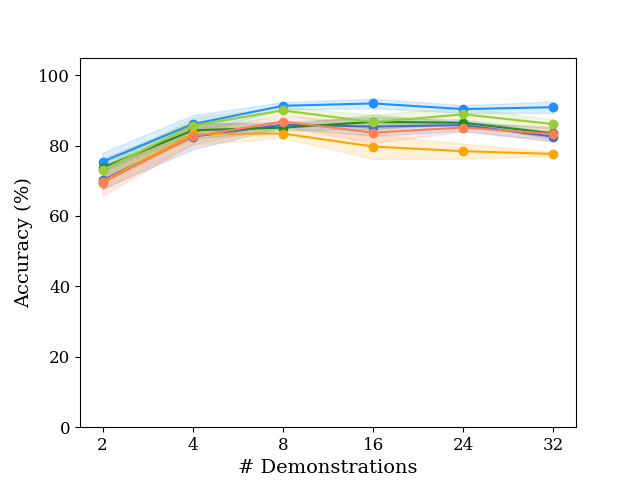}
            \caption{$y_1$ (random sampling)}\label{fig:image2}
        \end{subfigure}
        \hfill
        \begin{subfigure}{0.49\linewidth}
            \centering
            \includegraphics[width=\textwidth]{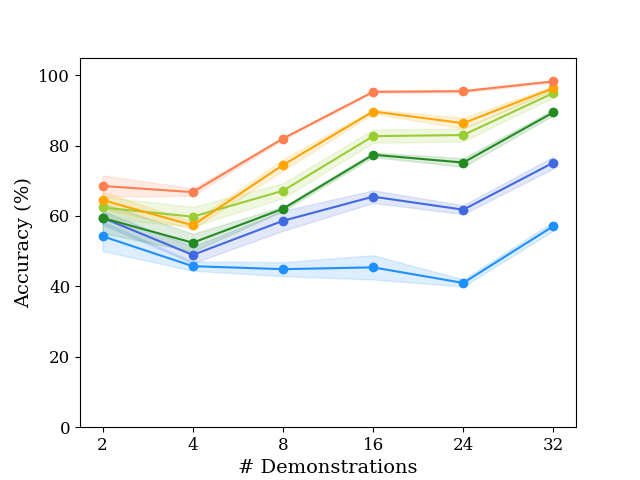}
            \caption{$y_2$ (biased sampling)}\label{fig:image3}
        \end{subfigure}
        \hfill
        \begin{subfigure}{0.49\linewidth}
            \centering
            \includegraphics[width=\textwidth]{latex/figures/lineplots/examples/en_llama3_8b_random_cls1.png}
            \caption{$y_2$ (random sampling)}\label{fig:image4}
        \end{subfigure}%
    \end{minipage}%
    \hfill
    \begin{minipage}[c]{\linewidth}
        \centering
        \vspace{2mm}
            \begin{subfigure}{0.7 \linewidth}
            \centering
            \includegraphics[width=\textwidth]{latex/figures/lineplots/legend.pdf}
        \end{subfigure}%
    \end{minipage}
    \hfill
    \begin{minipage}[c]{\linewidth}
        \caption{\label{fig:num_ex} Validation performance of Llama3 (8B) across different numbers of demonstrations on the PAWS-X$_{\textsc{EN}}$ dataset. In this figure, in-context examples from $y_2$ were sampled from long instances, and in-context examples from $y_1$ were sampled from short instances. Each subfigure shows results on the validation instances in the respective class. The conditions that introduce length bias in the context window (a and c subfigures) demonstrate a larger spread between performance on short and long validation instances, indicating greater potential to learn bias with longer contexts. Additional results can be found in the Appendix.}
    \end{minipage}
\end{figure}

\paragraph{Results} We report on the PAWS-X$_{EN}$ dataset using Llama3 (8B) in \autoref{fig:num_ex} and provide the average length for each class in \autoref{app:details}. Our results show that models can generally begin learning biases around 8 in-context examples, with the effect typically strengthening with increased numbers of examples. 

Longer context models are gaining traction, with a recent line of work focusing on scaling in-context learning to larger numbers of demonstrations. Longer contexts can increase performance and decrease sensitivity to ordering effects \citep{cai2023scaling, hao2022structured}, and contexts (beginning around $k=8$) can decrease model calibration errors, where calibration is a measure of the faithfulness of a model's predictive uncertainty \citep{zhang-etal-2024-study}. Our results demonstrate that longer contexts exhibit a greater potential for statistical data biases being learned in-context, and underscore the need for balanced selection methods with regard to potential data biases.

\subsection{Difference in Average Demonstration Length Between Classes}\label{sec:v3}

Given the results from the previous section, we investigate whether the difference of the average demonstration length between classes influences the ability of LLMs to identify a length bias. We keep the number of examples consistent at $k=16$ and sample from $p=\{0.25\%, 0.5\%, 0.75\%\}$ of the longest and shortest inputs for each class, respectively. For example, if $y_1$ is the long class, we sample $\frac{k}{2}$ instances from the longest $p=\{0.25\%, 0.5\%, 0.75\%\}$ of the instances belonging to $y_1$, and sample $\frac{k}{2}$ instances from the shortest $p=\{0.25\%, 0.5\%, 0.75\%\}$ of the instances belonging to $y_2$.

\paragraph{Results}
\begin{figure}[t!]
    \centering
    \begin{minipage}[t]{\linewidth}
        \begin{subfigure}{0.49\linewidth}
            \centering
            \includegraphics[width=\textwidth]{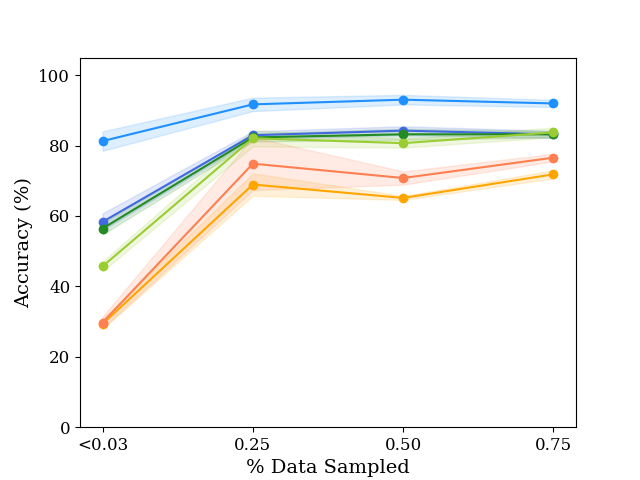}
            \caption{PAWS-X$_{\textsc{EN}}$: \\ $y_1$ (biased sampling)}\label{fig:image1}
        \end{subfigure}%
        \hfill
        \begin{subfigure}{0.49\linewidth}
            \centering
            \includegraphics[width=\textwidth]{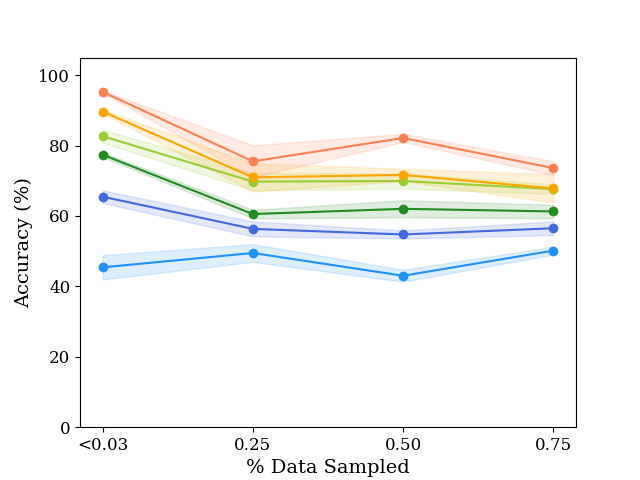}
            \caption{PAWS-X$_{\textsc{EN}}$: \\ $y_2$ (biased sampling)}\label{fig:image2}
        \end{subfigure}
        \hfill
        \begin{subfigure}{0.49\linewidth}
            \centering
            \includegraphics[width=\textwidth]{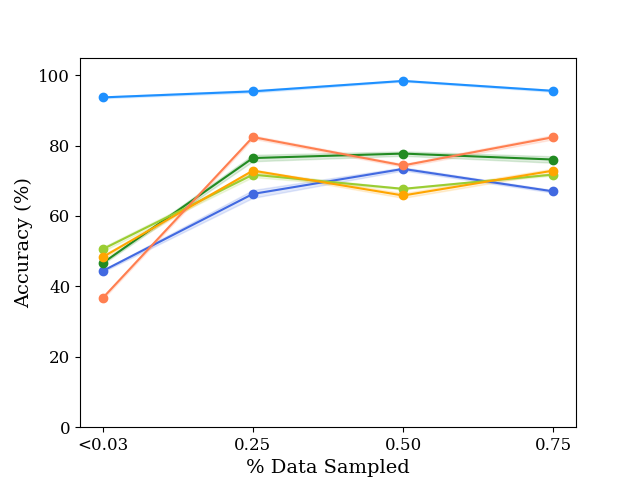}
            \caption{Hans: \\ $y_1$ (biased sampling)}\label{fig:image3}
        \end{subfigure}
        \hfill
        \begin{subfigure}{0.49\linewidth}
            \centering
            \includegraphics[width=\textwidth]{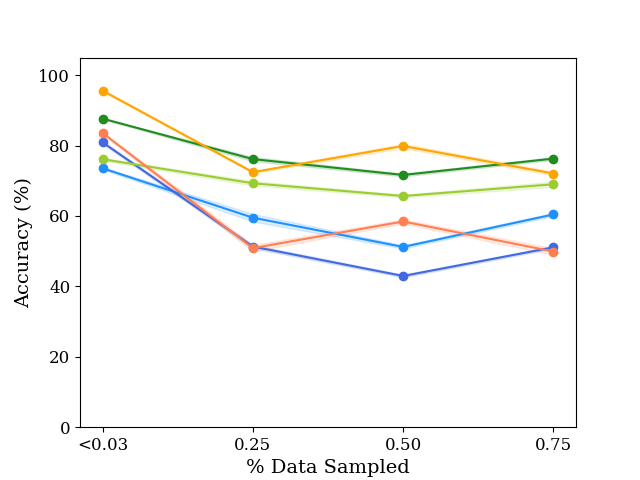}
            \caption{Hans: \\ $y_2$ (biased sampling)}\label{fig:image4}
        \end{subfigure}%
    \end{minipage}%
    \hfill
    \begin{minipage}[c]{\linewidth}
        \centering
        \vspace{2mm}
            \begin{subfigure}{0.7 \linewidth}
            \centering
            \includegraphics[width=\textwidth]{latex/figures/lineplots/legend.pdf}
        \end{subfigure}%
    \end{minipage}
    \hfill
    \begin{minipage}[c]{\linewidth}
        \caption{\label{fig:len} Validation performance across different data sampling percentages using Llama 3 (8B), on the PAWS-X$_{\textsc{EN}}$ and Hans datasets. In this figure, in-context examples from $y_2$ were sampled from long instances, and in-context examples from $y_1$ were sampled from short instances. Each subfigure shows results on the validation instances in the respective class, with Bin 0 containing the shortest demonstrations and Bin 5 containing the longest demonstrations. Additional results can be found in the Appendix.}
    \end{minipage}
\end{figure}

We report results using Llama 3 (8B) on the PAWS-X$_{\textsc{EN}}$ dataset in \autoref{fig:len},  where $0.03$ corresponds to an approximate sampling percentage from the previous experiment setup. We observe a length bias across different sampling percentages, despite the decrease in difference between average class lengths (see \autoref{table:avg_len_diff}). Intuitively, as the difference increases, so does the spread between performance across bins of different lengths. This indicates that while models can learn length biases from a few tokens difference (approximately $3$ tokens on HANS under $0.75$ sampling), the biases are amplified in the model as they are amplified in the demonstrations.

\begin{figure*}[t!]
    \centering
    \begin{minipage}[t]{\linewidth}
        \begin{subfigure}{0.31\linewidth}
            \centering
            \includegraphics[width=\textwidth]{latex/figures/ft/hans_200_llama3_8b_class2_6bins.png}
            \caption{Finetuning: $y_1$ (Blue) short demonstrations, $y_2$ (Orange) long demonstrations.}\label{fig:ft_og_hans}
        \end{subfigure}%
        \hfill
        \begin{subfigure}{0.31\linewidth}
            \centering
            \includegraphics[width=\textwidth]{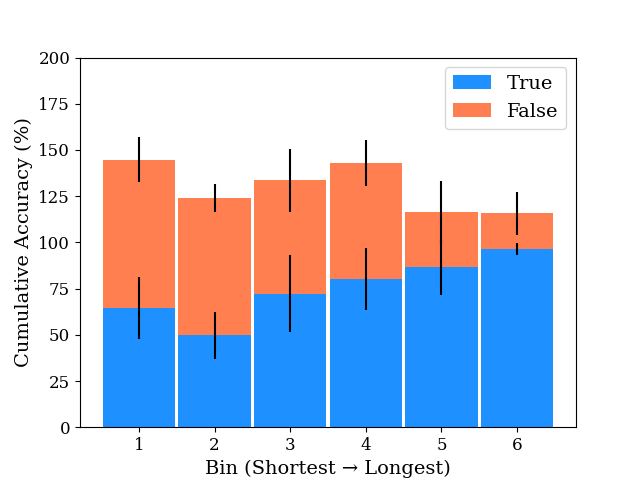}
            \caption{Intervention: $y_1$ (Blue) long demonstrations, $y_2$ (Orange) short demonstrations.}\label{fig:int_cls_hans}
        \end{subfigure}
        \hfill
        \begin{subfigure}{0.31\linewidth}
            \centering
            \includegraphics[width=\textwidth]{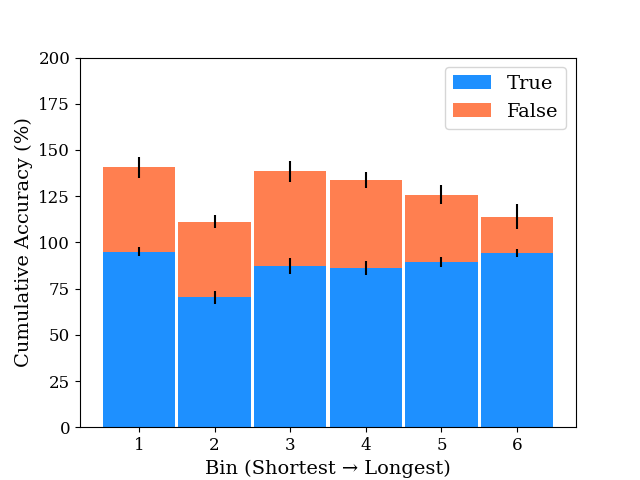}
            \caption{Intervention: $y_1$ (Blue) and $y_2$ (Orange) demonstrations randomly sampled.}\label{fig:int_ran_hans}
        \end{subfigure}
    \end{minipage}%
    \hfill
    \begin{minipage}[c]{\linewidth}
        \caption{\label{fig:int-hans} HANS validation set performance on a finetuned Llama 3 (8B) model exhibiting a length bias (see \autoref{fig:ft_og_hans} for finetuning performance prior to intervention). \autoref{fig:int_cls_hans} and \autoref{fig:int_ran_hans} (respectively) show results on two debiasing conditions: ICL demonstrations ($k=16$) sampled from the opposite lengths from what the model saw during finetuning (i.e. $y_1$ long demonstrations, $y_2$ short demonstrations), and random sampling.}
    \end{minipage}
\end{figure*}
\begin{figure*}[t!]
    \centering
    \begin{minipage}[t]{\linewidth}
        \begin{subfigure}{0.31\linewidth}
            \centering
            \includegraphics[width=\textwidth]{latex/figures/ft/paws_200_llama3_8b_class2_6bins.png}
            \caption{Finetuning: $y_1$ (Blue) short demonstrations, $y_2$ (Orange) long demonstrations.}\label{fig:ft_og_paws}
        \end{subfigure}%
        \hfill
        \begin{subfigure}{0.31\linewidth}
            \centering
            \includegraphics[width=\textwidth]{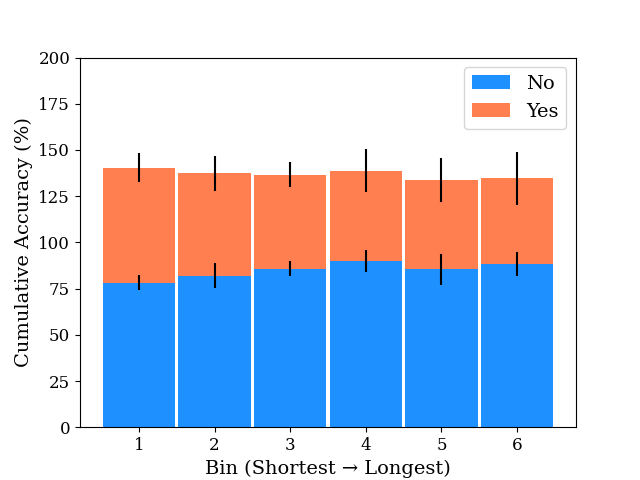}
            \caption{Intervention: $y_1$ (Blue) long demonstrations, $y_2$ (Orange) short demonstrations.}\label{fig:int_cls_paws}
        \end{subfigure}
        \hfill
        \begin{subfigure}{0.31\linewidth}
            \centering
            \includegraphics[width=\textwidth]{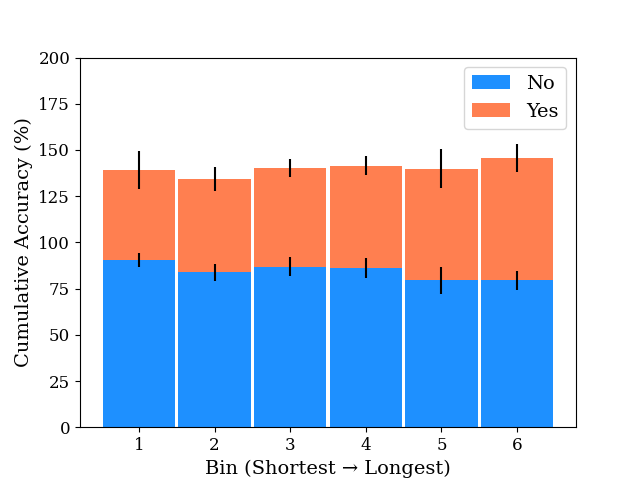}
            \caption{Intervention: $y_1$ (Blue) and $y_2$ (Orange) demonstrations randomly sampled.}\label{fig:int_ran_paws}
        \end{subfigure}
    \end{minipage}%
    \hfill
    \begin{minipage}[c]{\linewidth}
        \caption{\label{fig:int-paws} PAWS-X$_{\textsc{EN}}$ validation set performance on a finetuned Llama 3 (8B) model exhibiting a length bias (see \autoref{fig:ft_og_paws} for finetuning performance prior to intervention). \autoref{fig:int_cls_paws} and \autoref{fig:int_ran_paws} (respectively) show results on two debiasing conditions: ICL demonstrations ($k=16$) sampled from the opposite lengths from what the model saw during finetuning (i.e. $y_1$ long demonstrations, $y_2$ short demonstrations), and random sampling.}
    \end{minipage}
\end{figure*}
\begin{figure*}[t!]
    \centering
    \begin{minipage}[t]{\linewidth}
        \begin{subfigure}{0.31\linewidth}
            \centering
            \includegraphics[width=\textwidth]{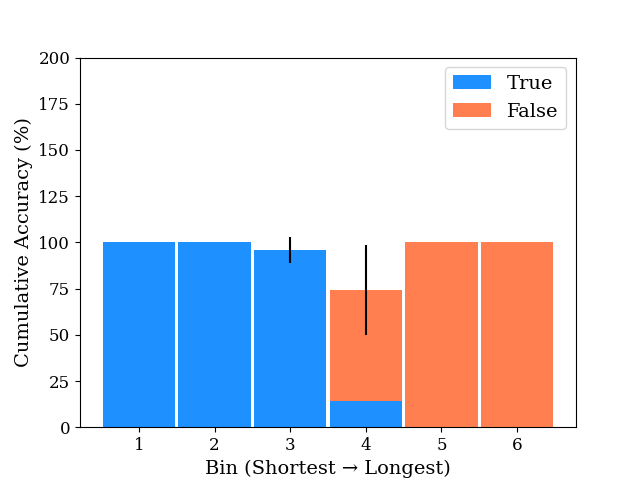}
            \caption{Finetuning: $y_1$ (Blue) short demonstrations, $y_2$ (Orange) long demonstrations.}\label{fig:ft_og_wnli}
        \end{subfigure}%
        \hfill
        \begin{subfigure}{0.31\linewidth}
            \centering
            \includegraphics[width=\textwidth]{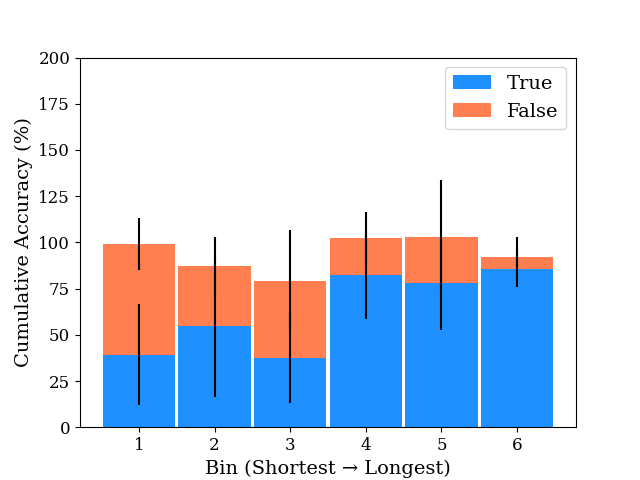}
            \caption{Intervention: $y_1$ (Blue) long demonstrations, $y_2$ (Orange) short demonstrations.}\label{fig:int_cls_wnli}
        \end{subfigure}
        \hfill
        \begin{subfigure}{0.31\linewidth}
            \centering
            \includegraphics[width=\textwidth]{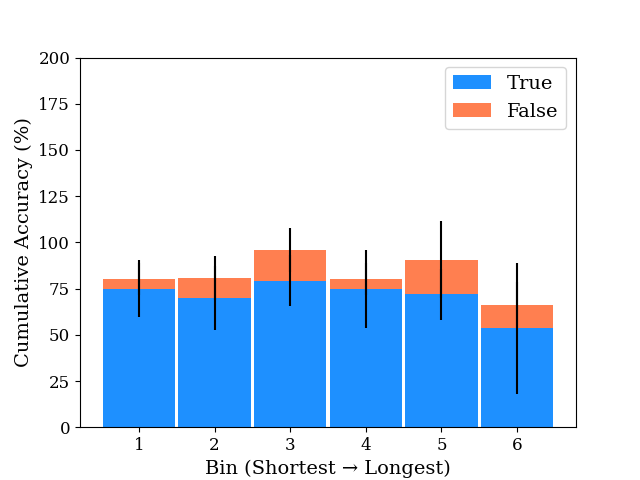}
            \caption{Intervention: $y_1$ (Blue) and $y_2$ (Orange) demonstrations randomly sampled.}\label{fig:int_ran_wnli}
        \end{subfigure}
    \end{minipage}%
    \hfill
    \begin{minipage}[c]{\linewidth}
        \caption{\label{fig:int-wnli} WNLI validation set performance on a finetuned Llama 3 (8B) model exhibiting a length bias (see \autoref{fig:ft_og_wnli} for finetuning performance prior to intervention). \autoref{fig:int_cls_wnli} and \autoref{fig:int_ran_wnli} (respectively) show results on two debiasing conditions: ICL demonstrations ($k=16$) sampled from the opposite lengths from what the model saw during finetuning (i.e. $y_1$ long demonstrations, $y_2$ short demonstrations), and random sampling.}
    \end{minipage}
\end{figure*}

\section{ICL for Debiasing Finetuned Models}\label{sec:int}
In-context learning eliminates the need for expensive model parameter updates incurred when finetuning. However, it is often the case that a model may have encoded biases picked up from the pretraining and/or finetuning. As our previous experiments show that in-context learning can learn length information, a natural extension is to question whether ICL can be used to ``unlearn'' or mitigate previously learned length biases.
In this section, we answer the question \textbf{\textit{can ICL be used as an intervention to mitigate biases in finetuned models?}}

We use previously finetuned models from \autoref{sec:q1} and modify the length distribution to try to counteract the biases. Specifically we experiment with two conditions: 1) using in-context demonstrations drawn from the opposite tail of the length distribution from what was seen during finetuning, and 2) using randomly sampled in-context demonstrations. We again use Llama3 8B and GPT-Neo 2.7B for these experiments using the datasets in \autoref{table:datasets}.

\paragraph{Results}

Results using Llama3 (8B) on HANS, PAWS-X$_{\textsc{EN}}$, and WNLI are reported in \autoref{fig:int-hans}, \autoref{fig:int-paws} and \autoref{fig:int-wnli}, respectively. We find that random sampling was able to counteract the bias, essentially ``unlearning'' the finetuned bias. This suggests that balanced data sampling is critical to minimize the likelihood of learning biases in-context. Further, if a dataset is balanced, random sampling may be sufficient. However, if a dataset contains shortcut features, more sophisticated sampling methods to mitigate the bias may be necessary.

Moreover, our results suggest balanced sampling over showing the models demonstrations of opposite lengths with respect to the finetuned bias. Specifically, the models learned the bias in the length information in the context window, regardless of how it contradicted what was seen during finetuning. One possible explanation is that the task may be implicitly encoded during pretraining and ICL extracts the knowledge \citep{xie2022an, min-etal-2022-rethinking}, however, further study is warranted on whether knowledge-extraction from ICL overrides knowledge-gain during finetuning.

\section{Discussion}
In this work, we investigate the impact of demonstration length bias on model performance when learning in-context. Under this setting, length is a statistical data bias, where the shallow feature (length) is correlated with class labels. It is important to make the distinction, however, between length as a linguistic feature containing information relevant to the underlying task, and length as an artifact of the data collection protocol. For example, length is an informative syntactic feature in classifying truthfulness vs. deceptiveness \citep{yancheva-rudzicz-2013-automatic} and authorship detection \citep{yule1939sentence}, however, length biases can also arise from artifacts reflective of heuristics used by human data annotators \citep{gururangan-etal-2018-annotation}. Our work pertains to the latter settings where length is an artifact rather than a task-informative natural language feature.

\paragraph{Which variables have the greatest impact on models learning length biases?} In \autoref{app:details}, we observe that when varying the number of in-context examples, the distance between classes is greater with fewer in-context examples. However, the amount of bias increases with  increased numbers of examples. Further, while we observe that bias increases with increased length difference, we still observe learned bias when class length difference is reduced to as few as $3$ tokens on the HANS dataset. This suggests that a key factor in learning bias is the number of examples the model sees. Additionally, our results suggest that any model can learn bias, and model parameter size is not necessarily correlated with increased ability to learn biases in-context.

\section{Conclusion} In this work, we empirically investigated the ability of LLMs to learn length biases under an in-context learning paradigm. Our results show that LLMs can learn statistical biases in the data. We further show the impact of model parameter sizes, number of examples, and class length difference on length biases. Finally, we demonstrate the potential for ICL to be used as a tool to debias fine-tuned models with previously learned length biases.

\section{Limitations}
While we test models up to $8\mathrm{B}$ parameters, we acknowledge a limitation of this work is the parameter threshold due to available computational resources. We believe our results scale to larger models.

\section*{Acknowledgments}
We thank the reviewers for their insightful feedback, which has helped us improve this paper significantly. We also thank Jason Stock for his helpful feedback and suggestions. This research was supported in part by NSF SaTC \#2124538 and NSF III \#2007492.

\bibliography{custom}
\clearpage
\appendix

\section{Appendix}
\subsection{Datasets}\label{app:datasets}
\begin{table*}[h!]
\centering
\begin{tabular}{p{0.29\textwidth}p{0.65\textwidth}}
  \toprule
  \multicolumn{2}{c}{\textbf{Binary NLI Datasets}} \\
  \midrule
  \textbf{Dataset} & \textbf{Description} \\
  \midrule
  QNLI \citep{wang-etal-2018-glue} & The Stanford Question Answering Dataset. A corpus of question-sentence pairs, with context sentences extracted from Wikipedia and questions written by a human annotator. The model is tasked with determining whether the context contains the answer to (entails) the question. \\
  \midrule
  RTE \tablefootnote{\citep{dagan2006pascal, bar2006second, giampiccolo2007third, bentivogli2009fifth}} & The Recognizing Textual Entailment (RTE) datasets. A corpus constructed from annual textual entailment challenges based on news and Wikipedia text.\\
  \midrule
  WNLI \citep{levesque2011winograd} & The Winograd Schema Challenge. A corpus of reading comprehension sentence pairs, where ambiguous pronouns are replaced with each possible referent. The task is to predict if the substituted sentence is entailed by the original sentence.\\
  \midrule
  HANS \citep{DBLP:journals/corr/abs-1902-01007} & Heuristic Analysis for NLI Systems. A corpus of challenging premise and hypothesis pairs designed to target evaluation of lexical overlap, sub-sequence, and constituent heuristics. \\
  \midrule
  \multicolumn{2}{c}{\textbf{Single-Sentence Datasets}} \\
  \midrule
  \textbf{Dataset} & \textbf{Description} \\
  \midrule
  SST-2 \citep{socher2013recursive} & The Stanford Sentiment Treebank. A corpus of sentences extracted from movie reviews, with human judgments of positive or negative sentiment. \\
  \midrule
  \multicolumn{2}{c}{\textbf{Similarity \& Paraphrase Detection Datasets}} \\
  \midrule
  \textbf{Dataset} & \textbf{Description} \\
  \midrule
  MRPC \citep{dolan2004unsupervised} & The Microsoft Research Paraphrase Corpus. A corpus of sentence pairs extracted from online news sources. Human raters judged semantic equivalence.\\
  \midrule
  PAWS-X$_{\textsc{en}}$ \citep{yang-etal-2019-paws} & Paraphrase Adversaries from Word Scrambling - Cross-lingual. A corpus of challenging paraphrase and non-paraphrase pairs created using data from Wikipedia. Sentence pairs were generated using controlled word swapping and back translation to ensure high lexical overlap, and human raters judged semantic equivalence. Our experiments utilize the English data split.\\
  \bottomrule
\end{tabular}
\caption{\label{table:dataset_descriptions} Dataset descriptions.}
\end{table*}

\subsection{Prompts}\label{app:prompts}
\begin{table*}[h!]
\centering
\begin{tabular}{p{0.29\textwidth}p{0.65\textwidth}}
  \toprule
  \multicolumn{2}{c}{\textbf{Binary NLI Datasets}} \\
  \midrule
  \textbf{Dataset} & \textbf{Prompt} \\
  \midrule
  \multirow{3}{*}{QNLI \citep{wang-etal-2018-glue}} & \{\textsc{sentence}\}\\
  & \textsc{Question: } \{\textsc{question}\} \textsc{True or False?} \\
  & \textsc{Answer: } \\ 
  \midrule
  \multirow{3}{*}{RTE \tablefootnote{\citep{dagan2006pascal, bar2006second, giampiccolo2007third, bentivogli2009fifth}}} & \{\textsc{sentence1}\} \\
  & \textsc{Question: } \{\textsc{sentence2}\} \textsc{True or False?} \\
  & \textsc{Answer: } \\
  \midrule
  \multirow{3}{*}{WNLI \citep{levesque2011winograd}} & \{\textsc{sentence1}\} \\
  & \textsc{Question: } \{\textsc{sentence2}\} \textsc{True or False?} \\
  & \textsc{Answer: } \\
  \midrule
  \multirow{3}{*}{HANS \citep{DBLP:journals/corr/abs-1902-01007}} & \{\textsc{premise}\} \\
  & \textsc{Question: } \{\textsc{hypothesis}\} \textsc{True or False?} \\
  & \textsc{Answer: } \\
  \midrule
  \multicolumn{2}{c}{\textbf{Single-Sentence Datasets}} \\
  \midrule
  \textbf{Dataset} & \textbf{Description} \\
  \midrule
  \multirow{3}{*}{SST-2 \citep{socher2013recursive}} & \textsc{\{sentence\}} \\
  & \textsc{Question: Is this sentence positive or negative?} \\
  & \textsc{Answer: }\\
  \midrule
  \multicolumn{2}{c}{\textbf{Similarity \& Paraphrase Detection Datasets}} \\
  \midrule
  \textbf{Dataset} & \textbf{Description} \\
  \midrule
  \multirow{4}{*}{MRPC \citep{dolan2004unsupervised}} &
  \textsc{Sentence 1: \{sentence1\}} \\ 
  & \textsc{Sentence 2: \{sentence2\}} \\
  & \textsc{Question: Do both sentences mean the same thing?} \\
  & \textsc{Answer: } \\
  \midrule
  \multirow{4}{*}{PAWS-X$_{\textsc{en}}$ \citep{yang-etal-2019-paws}} &
  \textsc{Sentence 1: \{sentence1\}} \\ 
  & \textsc{Sentence 2: \{sentence2\}} \\
  & \textsc{Question: Do both sentences mean the same thing?} \\
  & \textsc{Answer: } \\
  \bottomrule
\end{tabular}
\caption{\label{table:prompts} Prompts used in our experiments.}
\end{table*}

\subsection{Other Details}\label{app:details}
\begin{table*}[h!]
\centering
\small
\begin{tabular}{lllllllll}
  \toprule
  \multirow{2}{*}{\textbf{Dataset}} & \multirow{2}{*}{\textbf{Condition}} & \multirow{2}{*}{\textbf{Class}} & \multicolumn{6}{c}{\textbf{\# Examples}}\\
  \cmidrule{4-9}
  & & & 2 & 4 & 8 & 16 & 24 & 32 \\
  \midrule
  \multirow{6}{*}{QNLI} & \multirow{2}{*}{Random} & True & 51.33 & 42.5.00 & 47.67 & 44.81 & 48.48 & 45.49 \\
  \cmidrule{3-9}
  & & False & 48.60 & 36.5 & 49.94 & 47.70 & 46.86 & 48.44 \\
  \cmidrule{2-9}
  & \multirow{2}{*}{False-L} & True & 14.00 & 14.00 & 14.75 & 15.38 & 15.75 & 16.06 \\
  \cmidrule{3-9}
  & & False & 227.00 & 207.50 & 191.50 & 175.00 & 165.75 & 159.50 \\
  \cmidrule{2-9}
  & \multirow{2}{*}{True-L} & True & 446.00 & 445.40 & 445.00 & 358.63 & 305.92 & 276.00 \\
  \cmidrule{3-9}
  & & False & 15.00 & 15.50 & 15.75 & 15.88 & 16.00 & 16.25 \\  
\midrule
  \multirow{6}{*}{RTE} & \multirow{2}{*}{Random} & True & 60.00 & 59.33 & 61.25 & 66.57 & 63.02 & 65.77 \\
  \cmidrule{3-9}
  & & False & 44.00 & 57.71 & 53.00 & 63.06 & 56.10 & 63.36 \\
  \cmidrule{2-9}
  & \multirow{2}{*}{False-L} & True & 17.00 & 17.50 & 18.5 & 19.63 & 20.17 & 20.75 \\
  \cmidrule{3-9}
  & & False & 277.00 & 253.00 & 233.75 & 216.75 & 206.83 & 200.75 \\
  \cmidrule{2-9}
  & \multirow{2}{*}{True-L} & True & 195.00 & 194.00 & 192.00 & 187.25 & 183.67 & 181.19 \\
  \cmidrule{3-9}
  & & False & 16.00 & 17.50 & 18.75 & 19.75 & 20.75 & 21.50 \\  
  \midrule
  \multirow{6}{*}{WNLI} & \multirow{2}{*}{Random} & True & 37.75 & 45.50 & 41.07 & 41.33 & 38.20 & 39.62 \\
  \cmidrule{3-9}
  & & False & 41.75 & 38.17 & 39.50 & 35.18 & 36.65 & 36.72 \\
  \cmidrule{2-9}
  & \multirow{2}{*}{False-L} & True & 19.00 & 19.50 & 19.75 & 20.38 & 20.83 & 21.25 \\
  \cmidrule{3-9}
  & & False & 92.00 & 91.50 & 87.75 & 82.38 & 79.67 & 77.13 \\
  \cmidrule{2-9}
  & \multirow{2}{*}{True-L} & True & 93.00 & 91.50 & 89.50 & 84.88 & 82.42 & 80.63 \\
  \cmidrule{3-9}
  & & False & 19.00 & 19.00 & 19.50 & 20.13 & 20.58 & 21.00 \\  
  \midrule
  \multirow{6}{*}{HANS} & \multirow{2}{*}{Random} & True & 18.80 & 19.25 & 21.50 & 20.21 & 20.17 & 20.91 \\
  \cmidrule{3-9}
  & & False & 20.67 & 21.38 & 21.22 & 21.31 & 21.41 & 21.05 \\
  \cmidrule{2-9}
  & \multirow{2}{*}{False-L} & True & 13.00 & 13.00 & 13.00 & 13.00 & 13.00 & 13.00 \\
  \cmidrule{3-9}
  & & False & 27.00 & 27.00 & 27.00 & 27.00 & 27.00 & 27.00 \\
  \cmidrule{2-9}
  & \multirow{2}{*}{True-L} & True & 27.00 & 27.00 & 27.00 & 27.00 & 27.00 & 27.00 \\
  \cmidrule{3-9}
  & & False & 15.00 & 15.00 & 15.00 & 15.00 & 15.00 & 15.00 \\  
  \midrule
  \multirow{6}{*}{SST-2} & \multirow{2}{*}{Random} & True & 21.00 & 21.00 & 20.85 & 19.21 & 19.62 & 19.30 \\
  \cmidrule{3-9}
  & & False & 22.33 & 18.00 & 19.16 & 18.00 & 17.78 & 19.16 \\
  \cmidrule{2-9}
  & \multirow{2}{*}{False-L} & True & 10.00 & 10.00 & 10.00 & 10.00 & 10.00 & 10.00 \\
  \cmidrule{3-9}
  & & False & 61.00 & 61.00 & 60.50 & 60.00 & 59.58 & 59.19 \\
  \cmidrule{2-9}
  & \multirow{2}{*}{True-L} & True & 62.00 & 61.50 & 61.00 & 60.00 & 59.08 & 58.44 \\
  \cmidrule{3-9}
  & & False & 10.00 & 10.00 & 10.00 & 10.00 & 10.00 & 10.00 \\  
  \midrule
  \multirow{6}{*}{MRPC} & \multirow{2}{*}{Random} & True & 59.00 & 57.00 & 59.33 & 58.67 & 59.13 & 60.22 \\
  \cmidrule{3-9}
  & & False & 57.29 & 59.91 & 58.61 & 63.06 & 62.06 & 60.46 \\
  \cmidrule{2-9}
  & \multirow{2}{*}{False-L} & True & 34.00 & 34.00 & 34.50 & 35.13 & 35.67 & 36.19 \\
  \cmidrule{3-9}
  & & False & 97.00 & 91.50 & 88.25 & 86.00 & 85.00 & 84.25 \\
  \cmidrule{2-9}
  & \multirow{2}{*}{True-L} & True & 82.00 & 81.50 & 81.25 & 80.25 & 79.50 & 79.00 \\
  \cmidrule{3-9}
  & & False & 32.00 & 32.00 & 33.00 & 33.50 & 34.00 & 34.69 \\  
  \midrule
  \multirow{6}{*}{PAWS-X$_{\textsc{EN}}$} & \multirow{2}{*}{Random} & True & 63.33 & 57.50 & 58.50 & 57.84 & 58.11 & 58.00 \\
  \cmidrule{3-9}
  & & False & 61.60 & 58.60 & 58.06 & 54.31 & 58.79 & 59.42 \\
  \cmidrule{2-9}
  & \multirow{2}{*}{False-L} & True & 26.00 & 27.00 & 27.50 & 28.25 & 28.50 & 28.63 \\
  \cmidrule{3-9}
  & & False & 85.00 & 85.00 & 84.50 & 83.50 & 83.00 &  82.50 \\
  \cmidrule{2-9}
  & \multirow{2}{*}{True-L} & True & 86.00 & 86.00 & 86.00 & 85.63 & 85.25 & 84.94 \\
  \cmidrule{3-9}
  & & False & 26.00 & 26.00 & 26.25 & 27.63 & 28.33 & 28.75 \\  
\bottomrule
\end{tabular}
\caption{Average input length (including prompt length) across different numbers of examples. 
\label{table:avg_len_num_ex}}
\end{table*}

\begin{table*}[t]
\centering
\small
\begin{tabular}{lll}
  \toprule
  \textbf{Dataset} & \textbf{Class} & \textbf{Average Length}\\
  \midrule
  \multirow{2}{*}{QNLI} & True & 50.03 \\
  \cmidrule{2-3}
  & False & 47.50 \\
  \midrule
  \multirow{2}{*}{RTE} & True & 63.25 \\
  \cmidrule{2-3}
  & False & 64.99 \\
  \midrule
  \multirow{2}{*}{WNLI} & True & 40.23 \\
  \cmidrule{2-3}
  & False & 37.00 \\
  \midrule
  \multirow{2}{*}{HANS} & True & 20.53 \\
  \cmidrule{2-3}
  & False & 20.99 \\
  \midrule
  \multirow{2}{*}{SST2} & True & 28.17 \\
  \cmidrule{2-3}
  & False & 28.93 \\
  \midrule
  \multirow{2}{*}{MRPC} & True & 57.72 \\
  \cmidrule{2-3}
  & False & 61.16 \\
  \midrule
  \multirow{2}{*}{PAWS-X$_{\textsc{EN}}$} & True & 58.63 \\
  \cmidrule{2-3}
  & False & 58.63 \\
\bottomrule
\end{tabular}
\caption{Average input length for each validation set. Reported input lengths include the prompt length (consistent across all inputs); prompts can be found in \autoref{app:prompts}. \label{table:avg_len_num_ex}}
\end{table*}
\begin{table*}[h!]
\centering
\small
\begin{tabular}{llllll}
  \toprule
  \multirow{2}{*}{\textbf{Dataset}} & \multirow{2}{*}{\textbf{Condition}} & \multirow{2}{*}{\textbf{Class}} & \multicolumn{3}{c}{\textbf{\% Sampled}}\\
  \cmidrule{4-6}
  & & & 0.25 & 0.50 & 0.75 \\
  \midrule
  \multirow{4}{*}{QNLI} & \multirow{2}{*}{False-L} & True & 31.35 & 36.27 & 40.91\\
  \cmidrule{3-6}
  & & False & 65.87 & 57.55 & 52.05\\
  \cmidrule{2-6}
  & \multirow{2}{*}{True-L} & True & 69.98 & 60.16 & 53.93\\
  \cmidrule{3-6}
  & & False & 31.56 & 36.33 & 40.61\\  
  \midrule
  \multirow{4}{*}{RTE} & \multirow{2}{*}{False-L} & True & 33.05 & 40.03 & 46.23\\
  \cmidrule{3-6}
  & & False & 122.98 & 90.66 & 76.35\\
  \cmidrule{2-6}
  & \multirow{2}{*}{True-L} & True & 124.51 & 91.02 & 77.78\\
  \cmidrule{3-6}
  & & False & 33.44 & 39.98 & 47.21\\  
  \midrule
  \multirow{4}{*}{WNLI} & \multirow{2}{*}{False-L} & True & 25.23 & 28.01 & 31.70\\
  \cmidrule{3-6}
  & & False & 56.01 & 47.21 & 40.60\\
  \cmidrule{2-6}
  & \multirow{2}{*}{True-L} & True & 60.08 & 48.73 & 43.12\\
  \cmidrule{3-6}
  & & False & 24.72 & 28.12 & 31.05\\  
  \midrule
  \multirow{4}{*}{HANS} & \multirow{2}{*}{False-L} & True & 16.92 & 18.72 & 19.65\\
  \cmidrule{3-6}
  & & False & 23.59 & 22.65 & 21.97\\
  \cmidrule{2-6}
  & \multirow{2}{*}{True-L} & True & 23.29 & 22.39 & 21.77\\
  \cmidrule{3-6}
  & & False & 18.17 & 19.40 & 20.16\\  
  \midrule
  \multirow{4}{*}{SST2} & \multirow{2}{*}{False-L} & True & 11.07 & 12.66 & 14.92\\
  \cmidrule{3-6}
  & & False & 29.57 & 23.90 & 20.36\\
  \cmidrule{2-6}
  & \multirow{2}{*}{True-L} & True & 30.55 & 25.14 & 21.37\\
  \cmidrule{3-6}
  & & False & 11.06 & 12.26 & 14.27\\  
  \midrule
  \multirow{4}{*}{MRPC} & \multirow{2}{*}{False-L} & True & 44.89 & 49.24 & 53.12\\
  \cmidrule{3-6}
  & & False & 74.31 & 69.98 & 65.67\\
  \cmidrule{2-6}
  & \multirow{2}{*}{True-L} & True & 70.53 & 65.61 & 61.50\\
  \cmidrule{3-6}
  & & False & 47.12 & 52.35 & 56.78\\  
  \midrule
  \multirow{4}{*}{PAWS-X$_{\textsc{EN}}$} & \multirow{2}{*}{False-L} & True & 44.61 & 49.51 & 53.98\\
  \cmidrule{3-6}
  & & False & 72.44 & 67.80 & 63.31\\
  \cmidrule{2-6}
  & \multirow{2}{*}{True-L} & True & 72.58 & 67.93 & 63.46\\
  \cmidrule{3-6}
  & & False & 44.56 & 49.53 & 54.04\\  

\bottomrule
\end{tabular}
\caption{Average input length across different sampling bins (by percentage of data sampled from). Reported input lengths include the prompt length (consistent across all inputs); prompts can be found in \autoref{app:prompts}. \label{table:avg_len_diff}}
\end{table*}

\clearpage
\subsection{Additional Length Bias Results}\label{app:q1-results}
\begin{figure*}[h!]
\centering
\begin{minipage}[t]{\linewidth}
\begin{subfigure}{\linewidth}
    \centering
    \includegraphics[width=0.19\textwidth]            {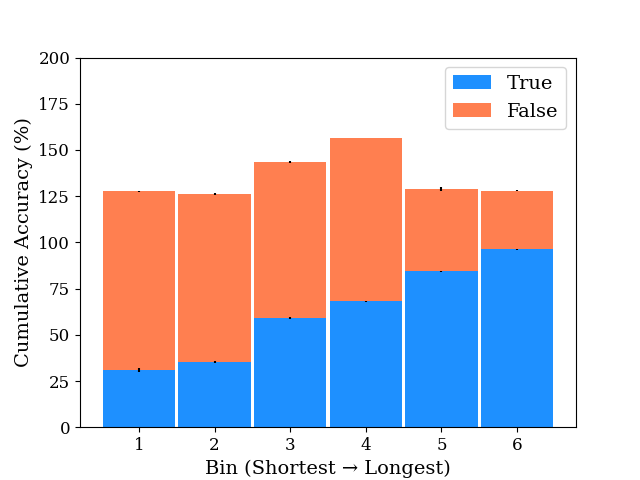}
    \includegraphics[width=0.19\textwidth]            {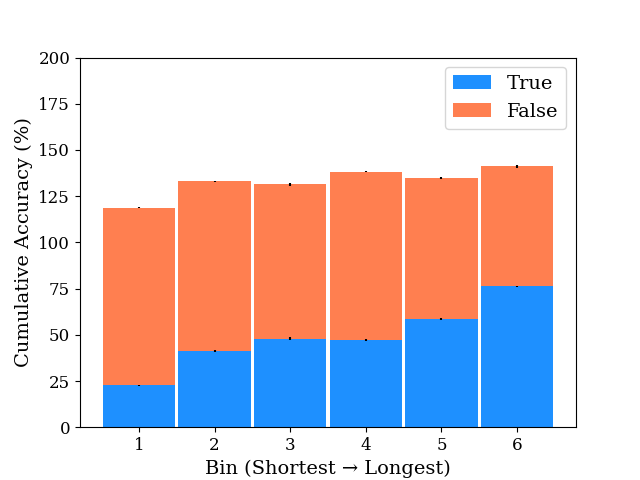}
    \includegraphics[width=0.19\textwidth]            {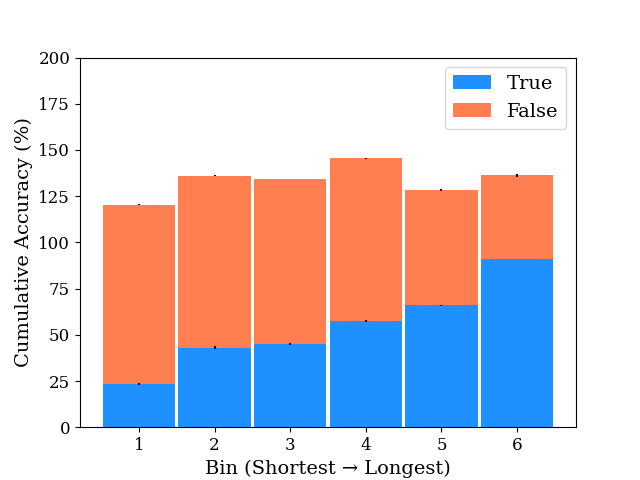}
    \includegraphics[width=0.19\textwidth]            {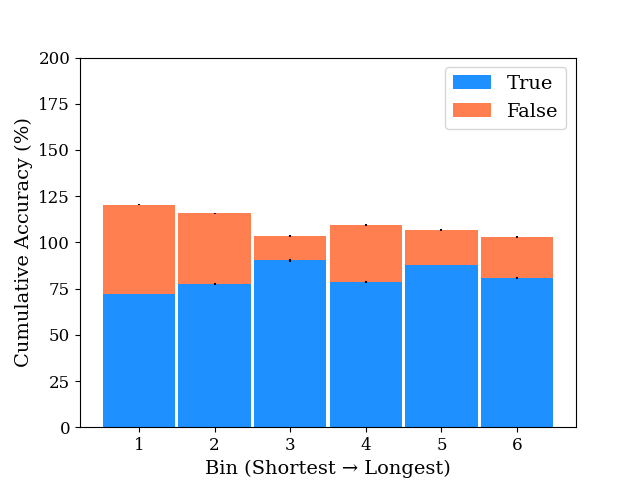}
    \includegraphics[width=0.19\textwidth]            {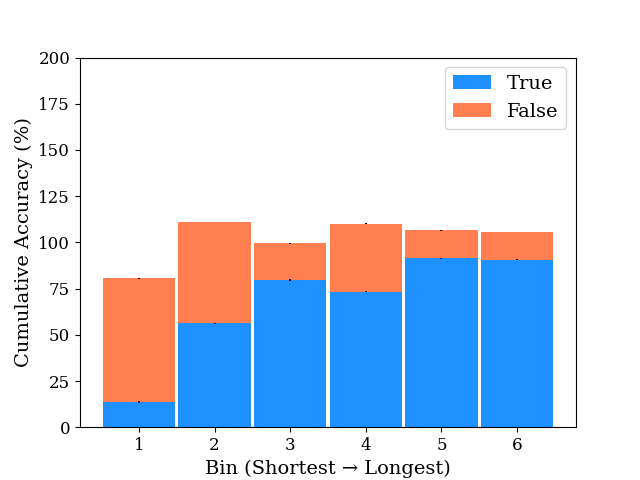}
    \caption{Hans}
\end{subfigure}
\begin{subfigure}{\linewidth}
    \centering
    \includegraphics[width=0.19\textwidth]            {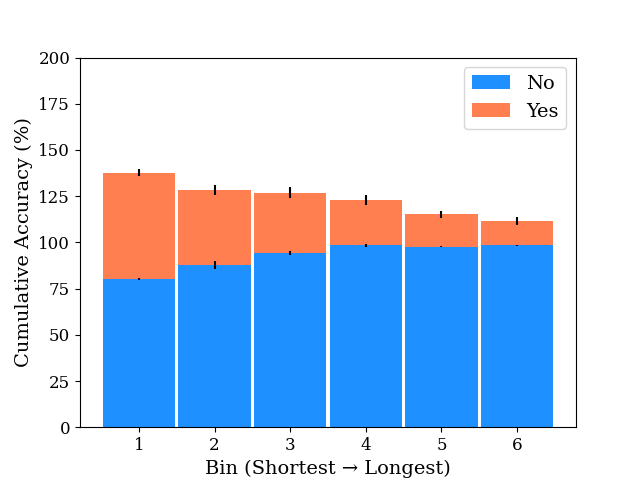}
    \includegraphics[width=0.19\textwidth]            {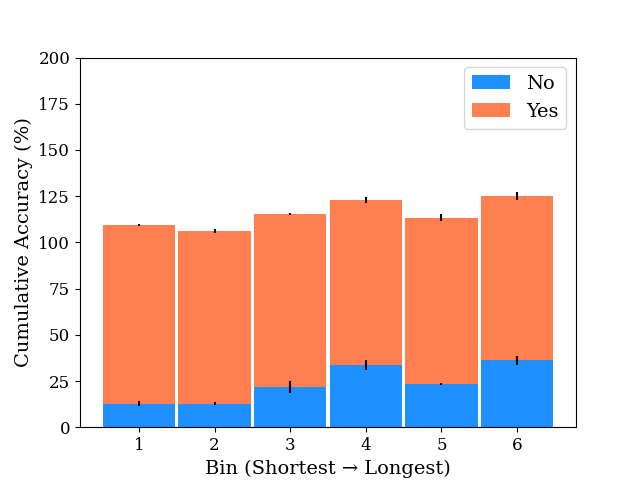}
    \includegraphics[width=0.19\textwidth]            {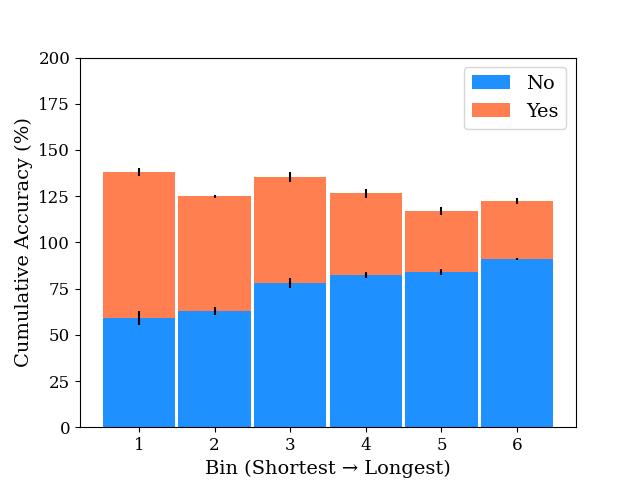}
    \includegraphics[width=0.19\textwidth]            {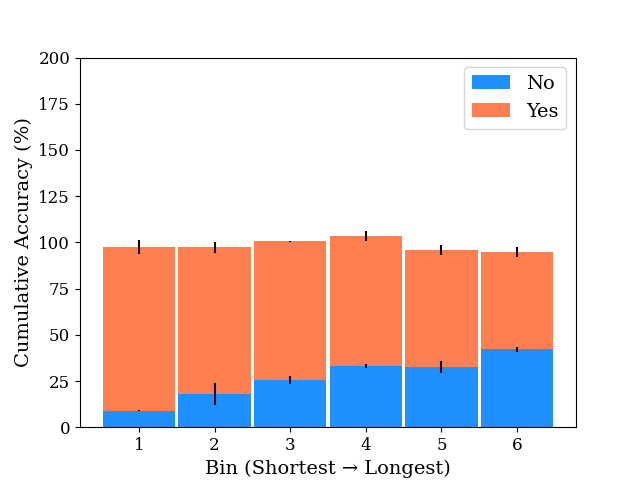}
    \includegraphics[width=0.19\textwidth]            {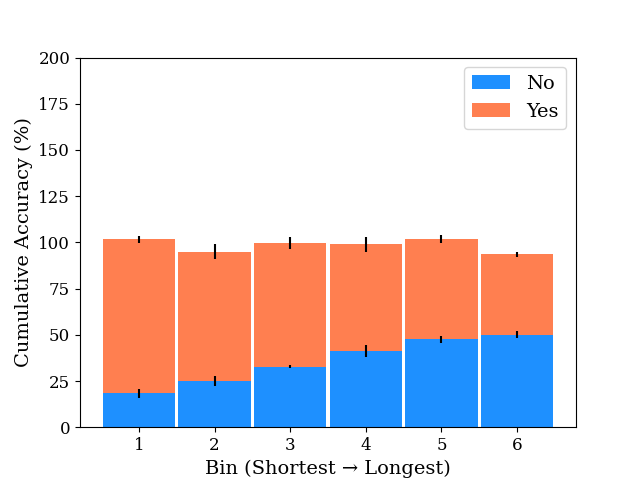}
    \caption{PAWS-X$_{\textsc{EN}}$}
\end{subfigure}
\begin{subfigure}{\linewidth}
    \centering
    \includegraphics[width=0.19\textwidth]            {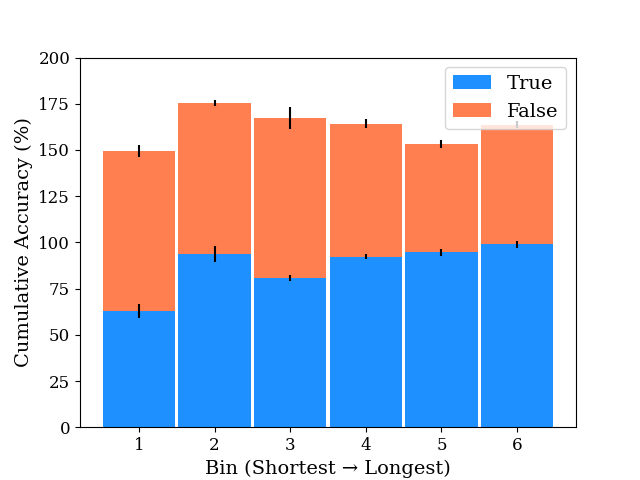}
    \includegraphics[width=0.19\textwidth]            {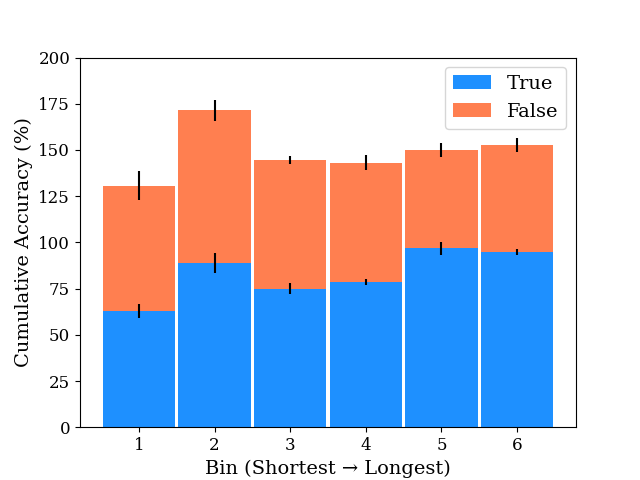}
    \includegraphics[width=0.19\textwidth]            {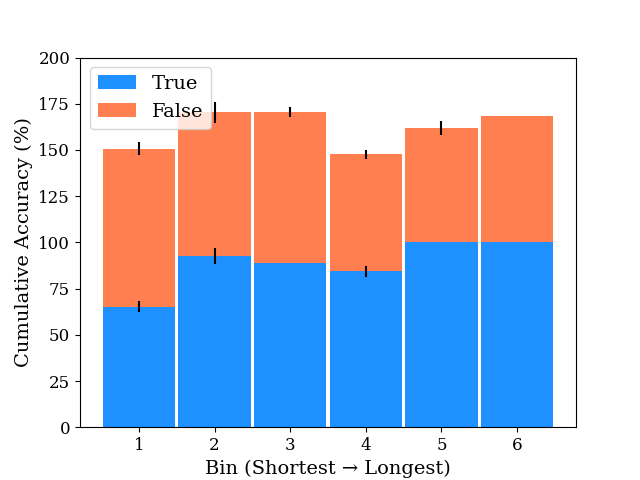}
    \includegraphics[width=0.19\textwidth]            {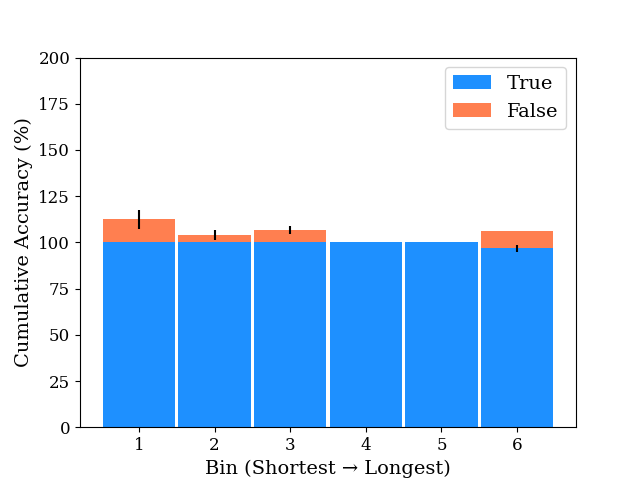}
    \includegraphics[width=0.19\textwidth]            {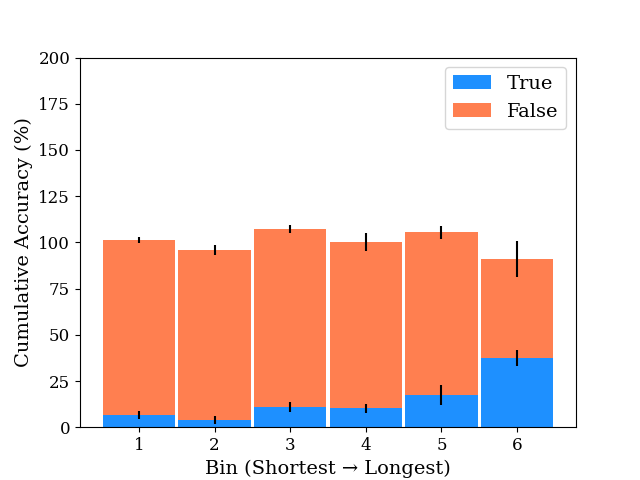}
    \caption{RTE}
\end{subfigure}
\begin{subfigure}{\linewidth}
    \centering
    \includegraphics[width=0.19\textwidth]            {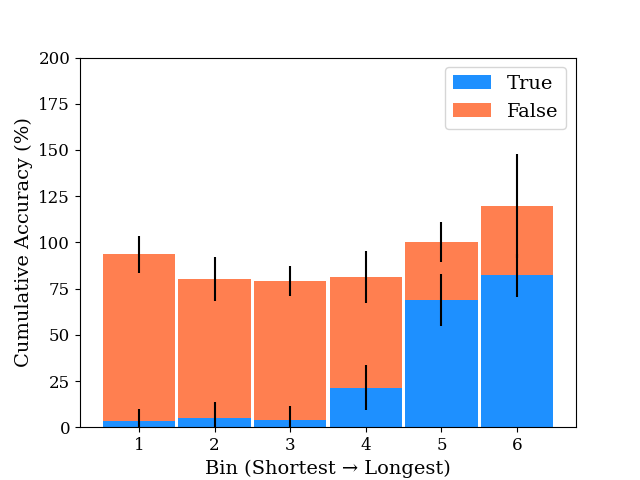}
    \includegraphics[width=0.19\textwidth]            {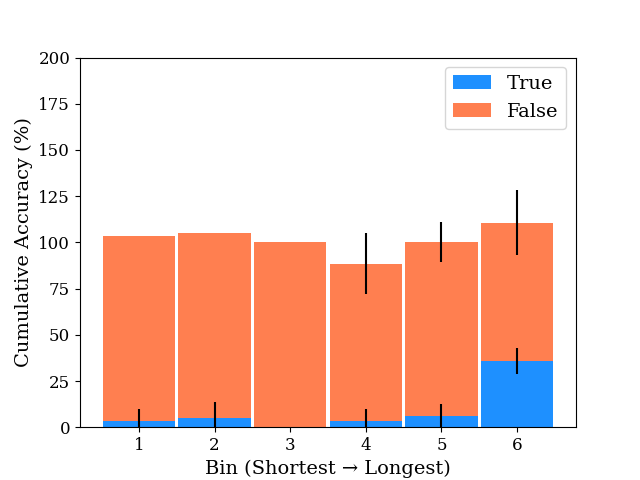}
    \includegraphics[width=0.19\textwidth]            {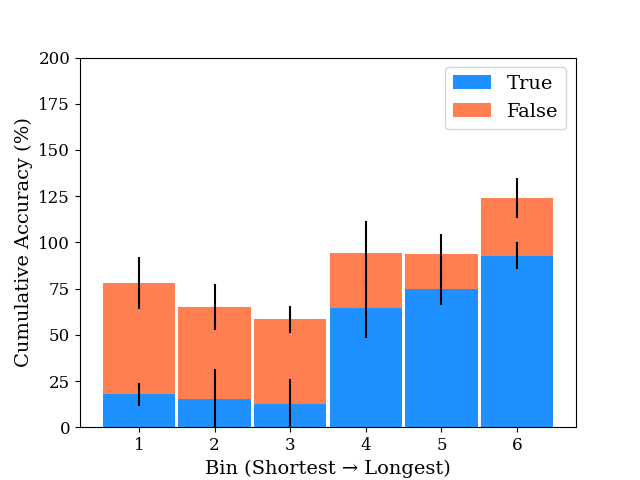}
    \includegraphics[width=0.19\textwidth]            {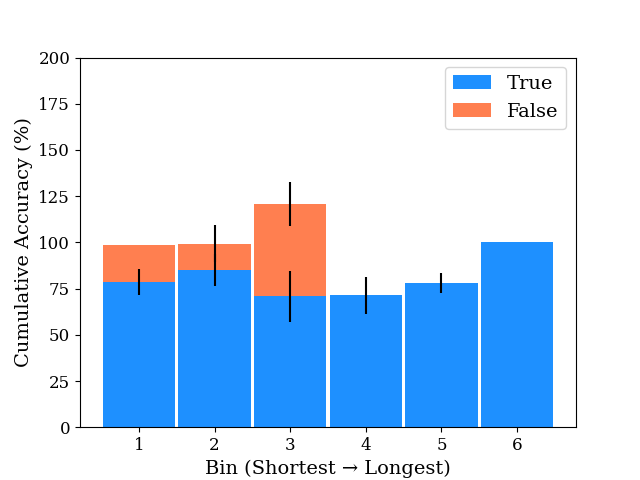}
    \includegraphics[width=0.19\textwidth]            {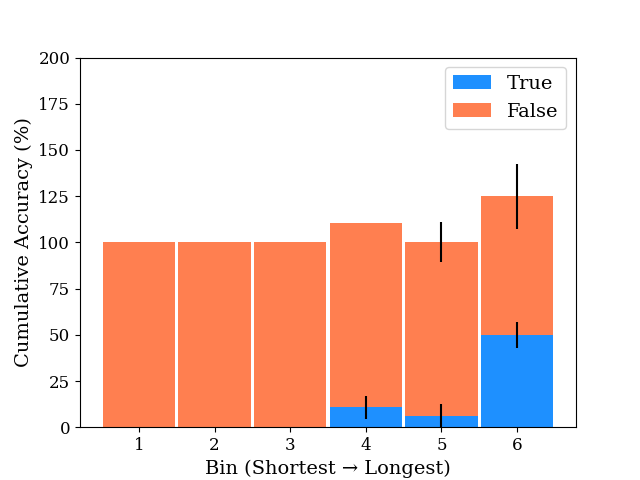}
    \caption{WNLI}
\end{subfigure}
\begin{subfigure}{\linewidth}
    \centering
    \includegraphics[width=0.19\textwidth]            {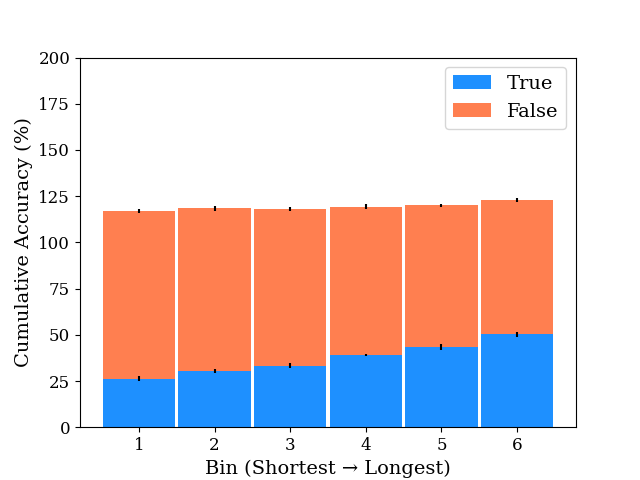}
    \includegraphics[width=0.19\textwidth]            {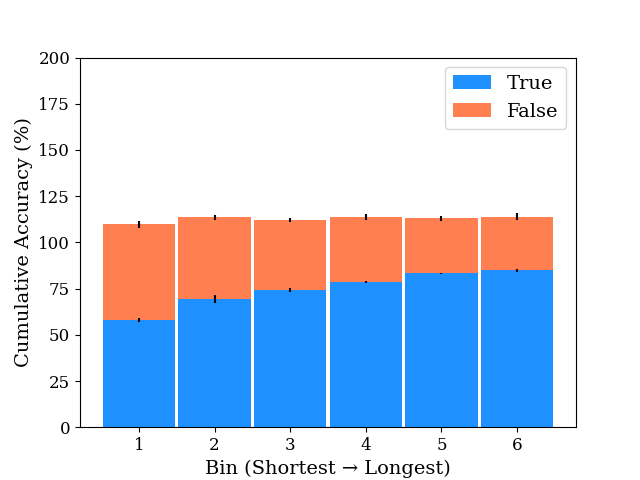}
    \includegraphics[width=0.19\textwidth]            {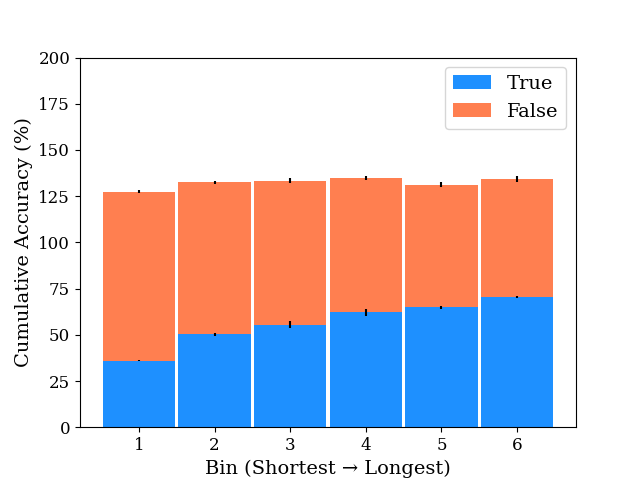}
    \includegraphics[width=0.19\textwidth]            {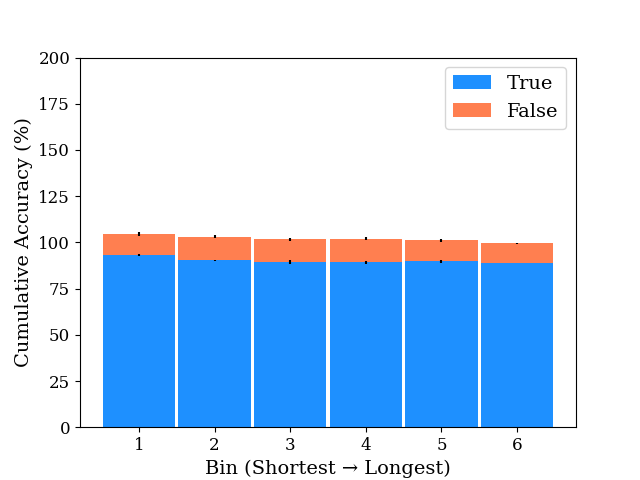}
    \includegraphics[width=0.19\textwidth]            {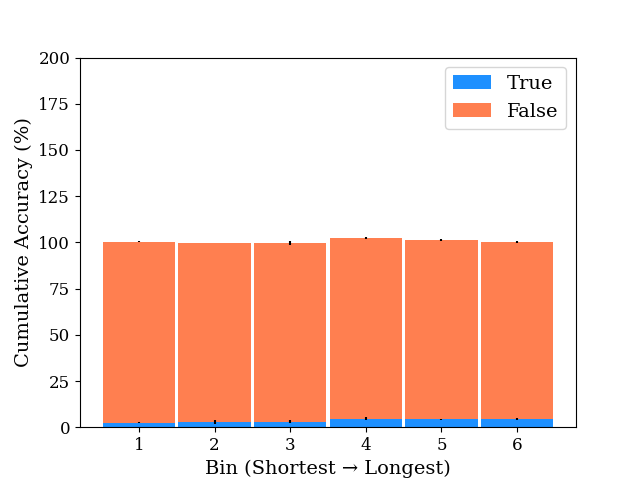}
    \caption{QNLI}
\end{subfigure}
\begin{subfigure}{\linewidth}
    \centering
    \includegraphics[width=0.19\textwidth]            {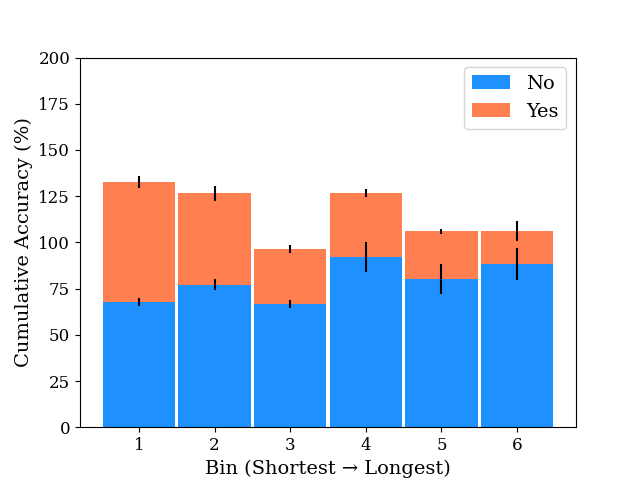}
    \includegraphics[width=0.19\textwidth]            {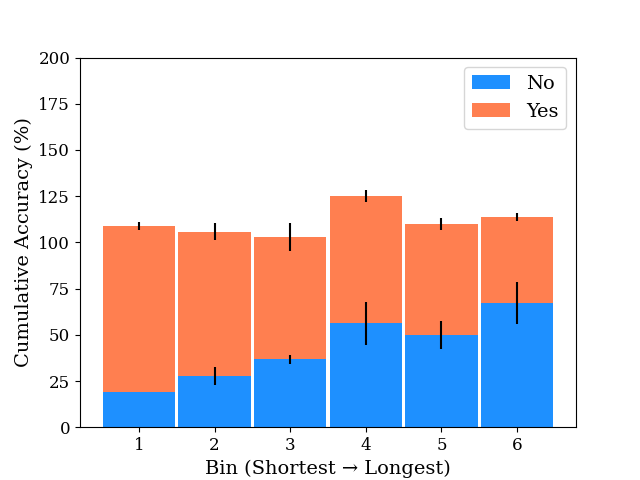}
    \includegraphics[width=0.19\textwidth]            {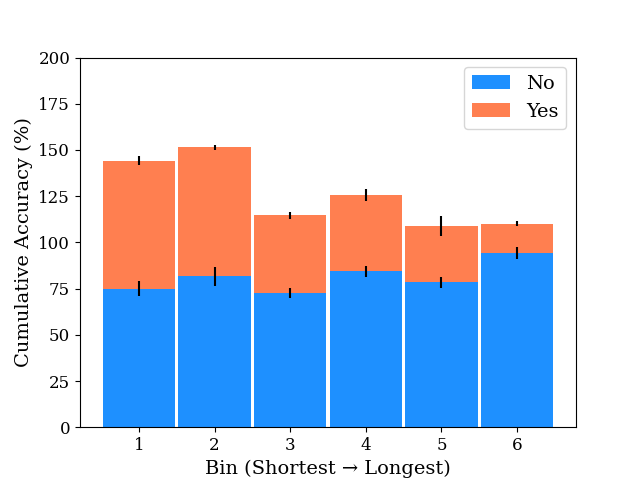}
    \includegraphics[width=0.19\textwidth]            {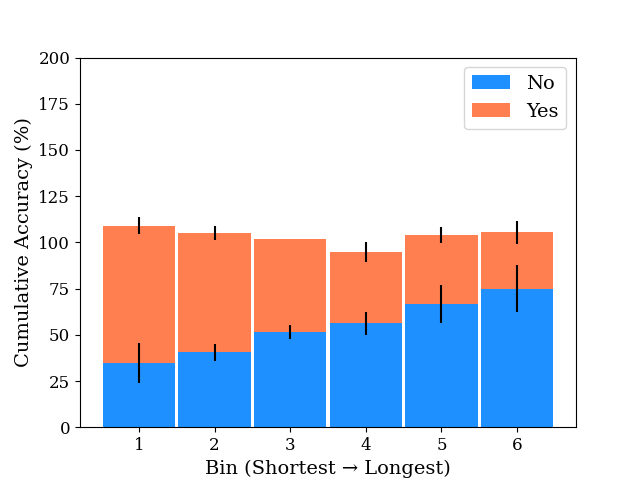}
    \includegraphics[width=0.19\textwidth]            {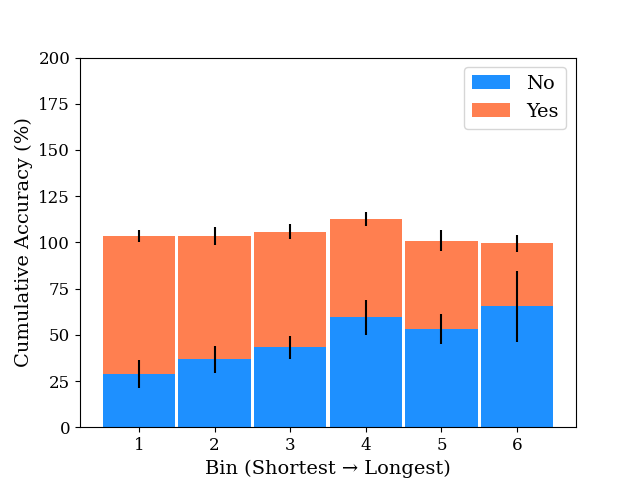}
    \caption{MRPC}
\end{subfigure}
\begin{subfigure}{\linewidth}
    \centering
    \includegraphics[width=0.19\textwidth]            {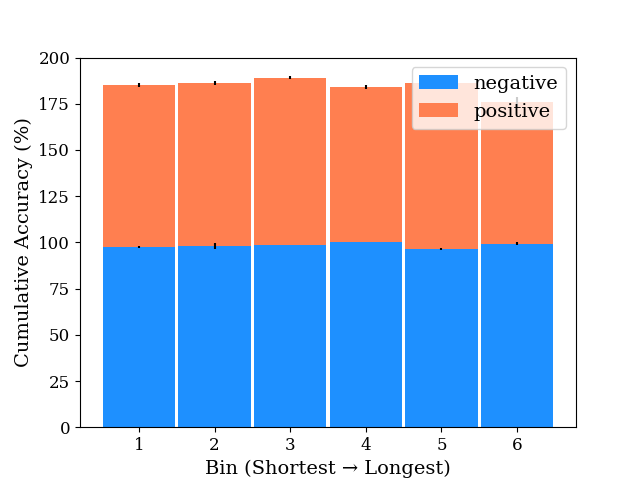}
    \includegraphics[width=0.19\textwidth]            {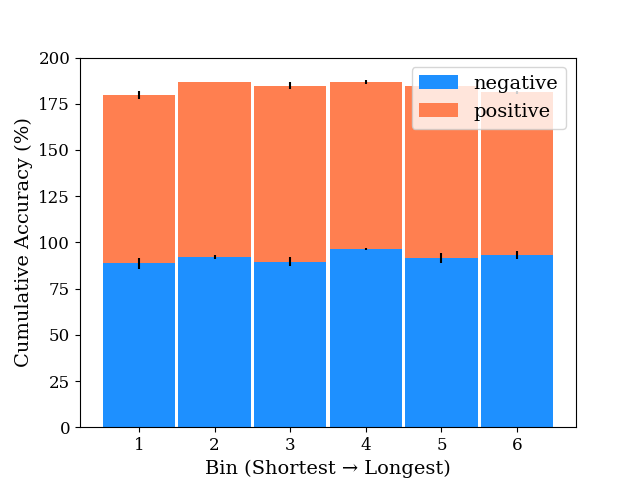}
    \includegraphics[width=0.19\textwidth]            {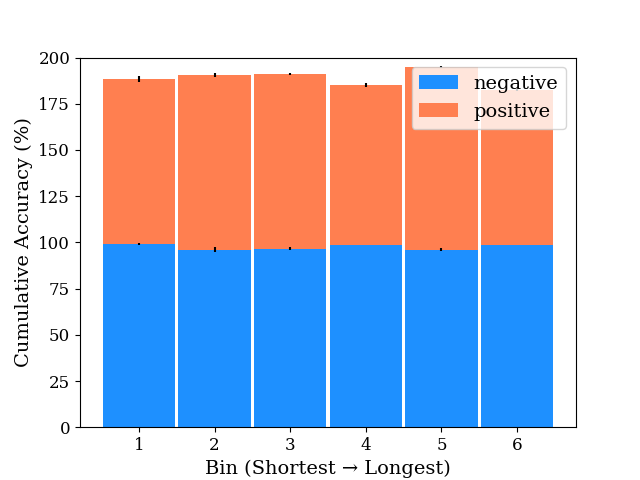}
    \includegraphics[width=0.19\textwidth]            {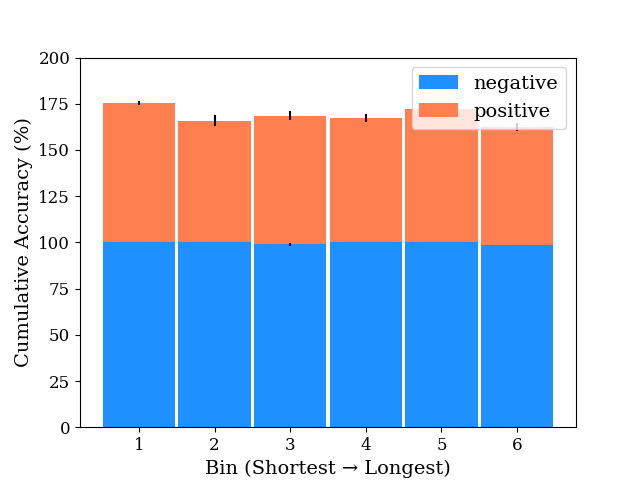}
    \includegraphics[width=0.19\textwidth]            {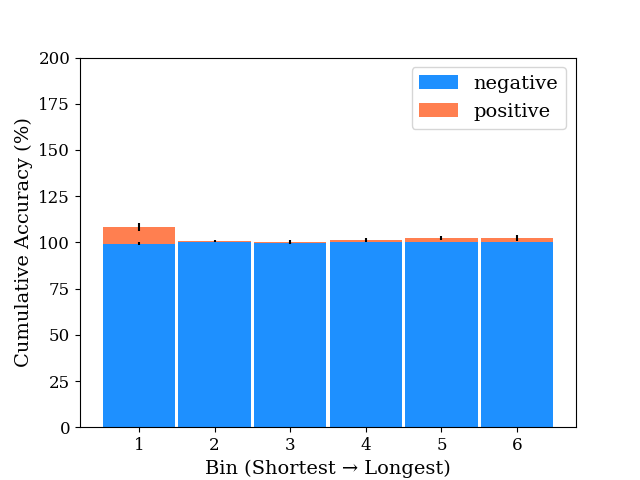}
    \caption{SST-2}
\end{subfigure}
\end{minipage}
\hfill
\begin{minipage}[c]{\linewidth}
    \caption{ICL performance of Llama3 8B, Llama2 7B, Mistral 7B, OPT 6.7B, and GPT Neo 2.7B (from left to right) where $y_1$ (Blue) samples long demonstrations and $y_2$ (Orange) samples short demonstrations.}
\end{minipage}
\end{figure*}

\begin{figure*}[t!]
\centering
\begin{minipage}[t]{\linewidth}
\begin{subfigure}{\linewidth}
    \centering
    \includegraphics[width=0.19\textwidth]            {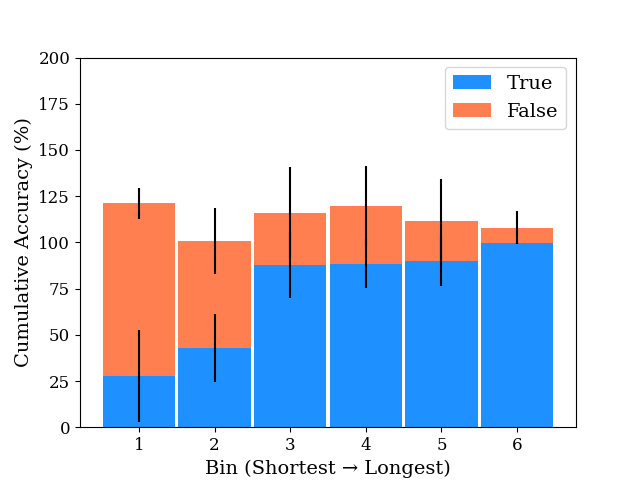}
    \includegraphics[width=0.19\textwidth]            {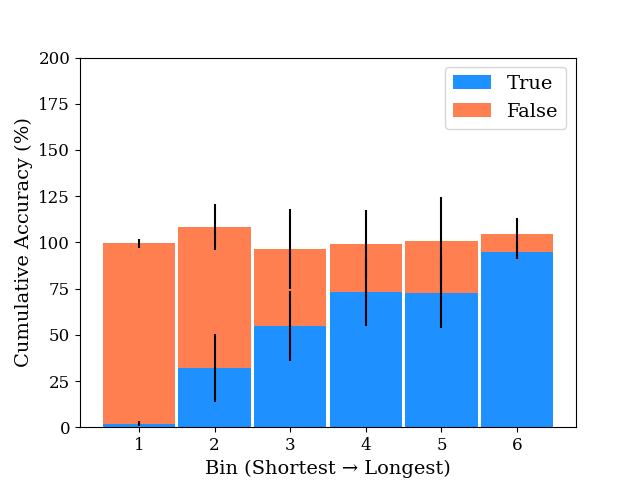}
    \includegraphics[width=0.19\textwidth]            {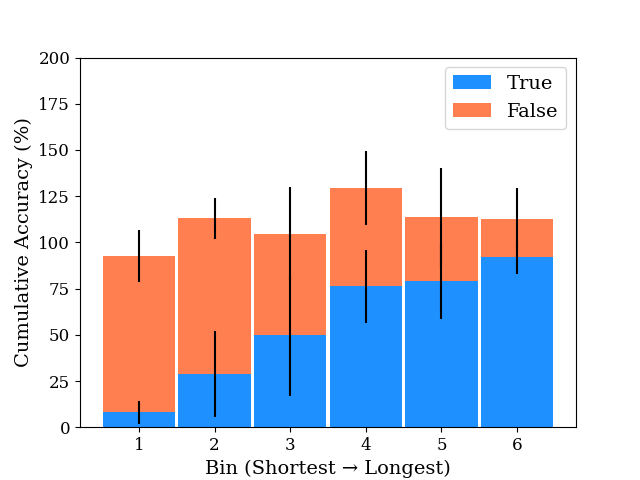}
    \includegraphics[width=0.19\textwidth]            {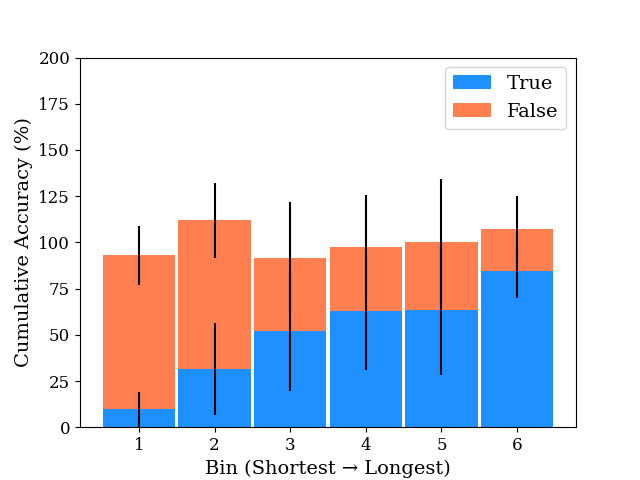}
    \includegraphics[width=0.19\textwidth]            {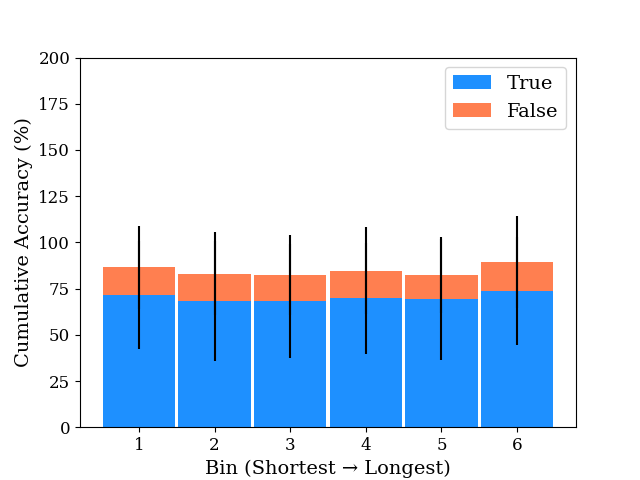}
    \caption{Hans}
\end{subfigure}
\begin{subfigure}{\linewidth}
    \centering
    \includegraphics[width=0.19\textwidth]            {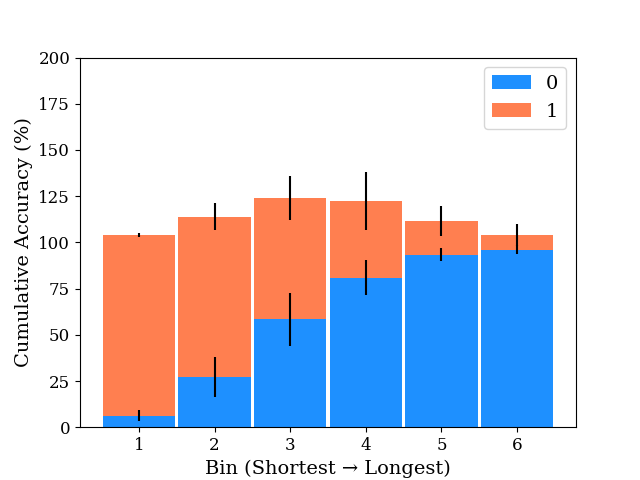}
    \includegraphics[width=0.19\textwidth]            {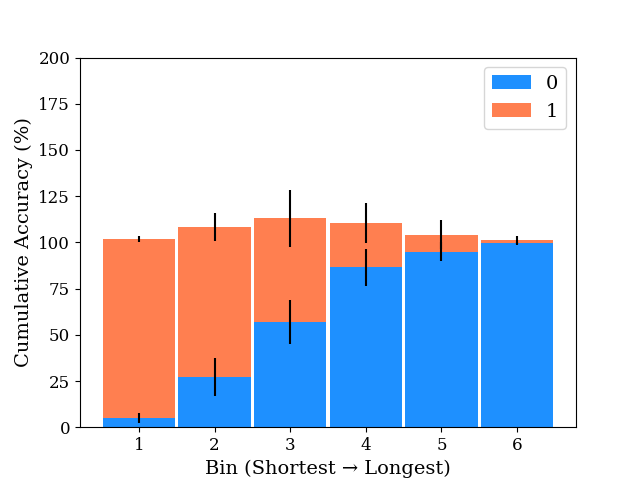}
    \includegraphics[width=0.19\textwidth]            {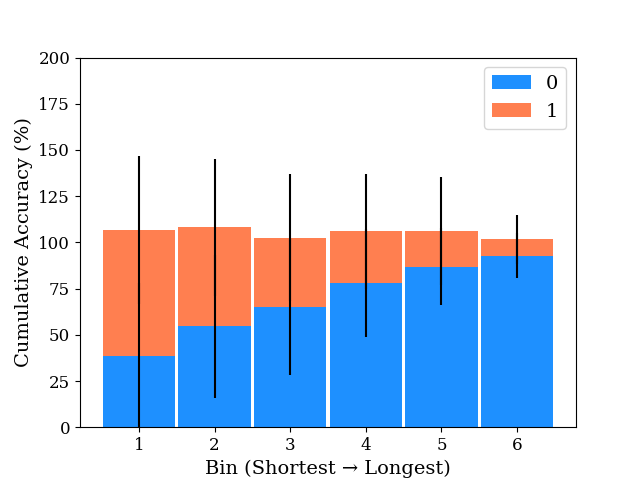}
    \includegraphics[width=0.19\textwidth]            {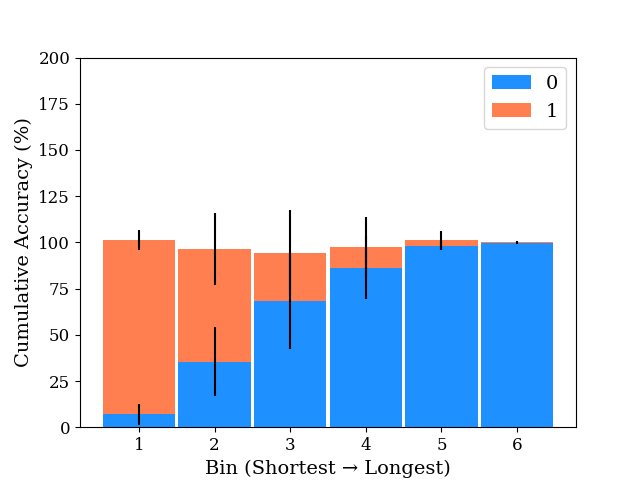}
    \includegraphics[width=0.19\textwidth]            {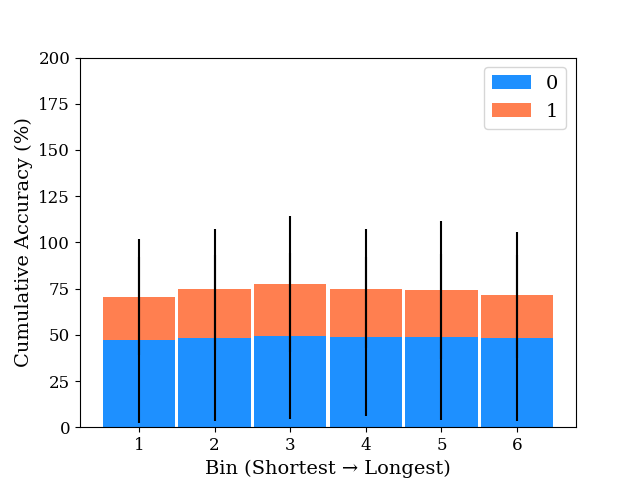}
    \caption{PAWS-X$_{\textsc{EN}}$}
\end{subfigure}
\begin{subfigure}{\linewidth}
    \centering
    \includegraphics[width=0.19\textwidth]            {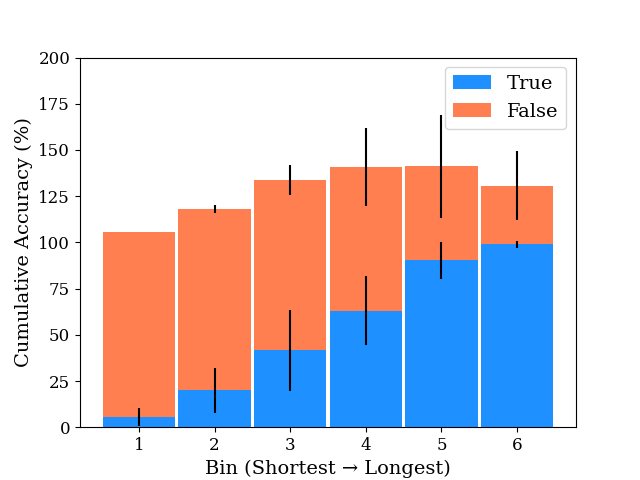}
    \includegraphics[width=0.19\textwidth]            {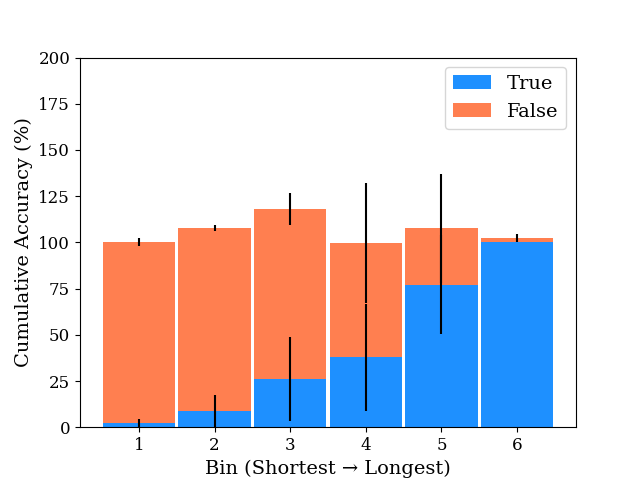}
    \includegraphics[width=0.19\textwidth]            {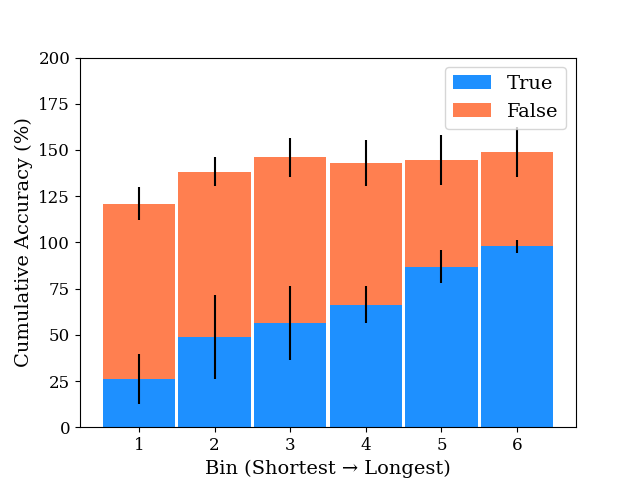}
    \includegraphics[width=0.19\textwidth]            {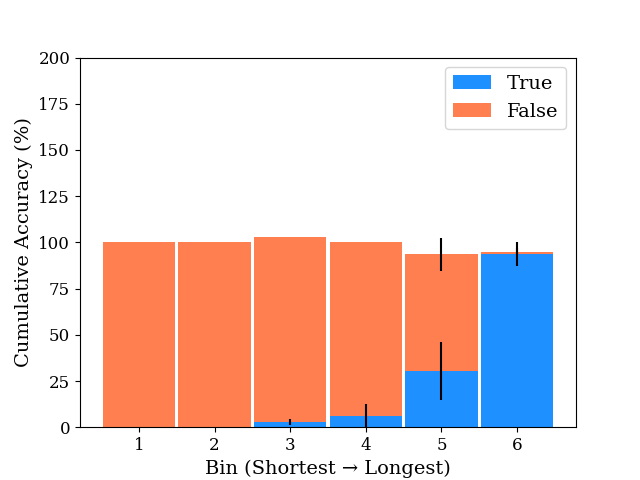}
    \includegraphics[width=0.19\textwidth]            {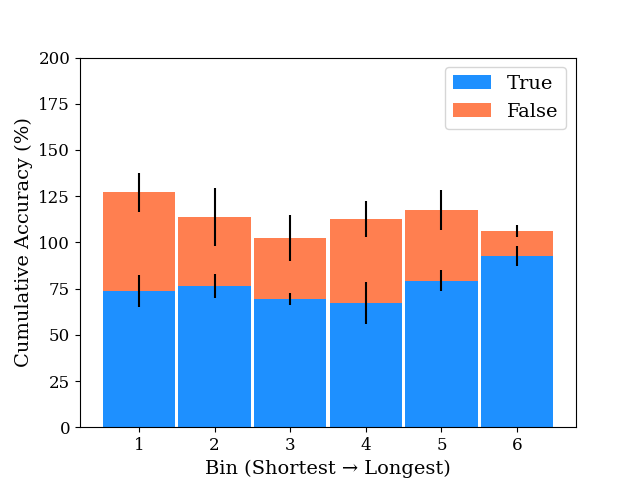}
    \caption{RTE}
\end{subfigure}
\begin{subfigure}{\linewidth}
    \centering
    \includegraphics[width=0.19\textwidth]            {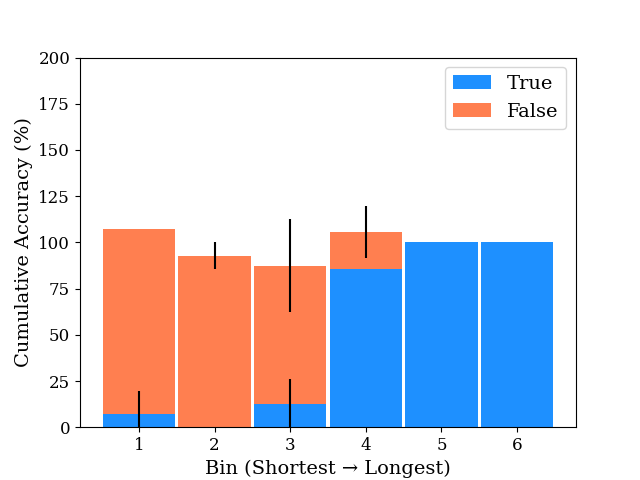}
    \includegraphics[width=0.19\textwidth]            {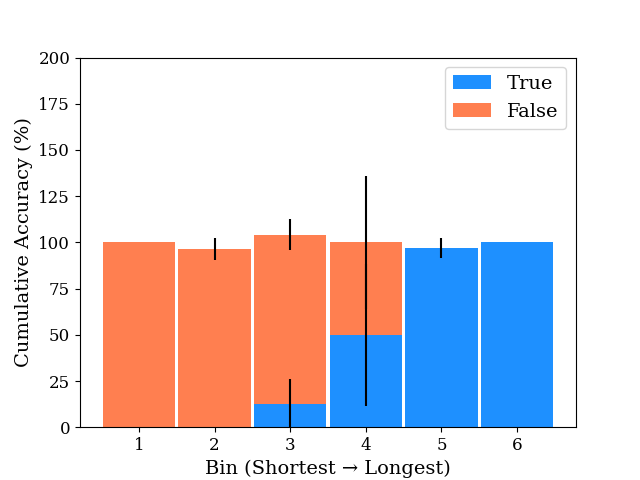}
    \includegraphics[width=0.19\textwidth]            {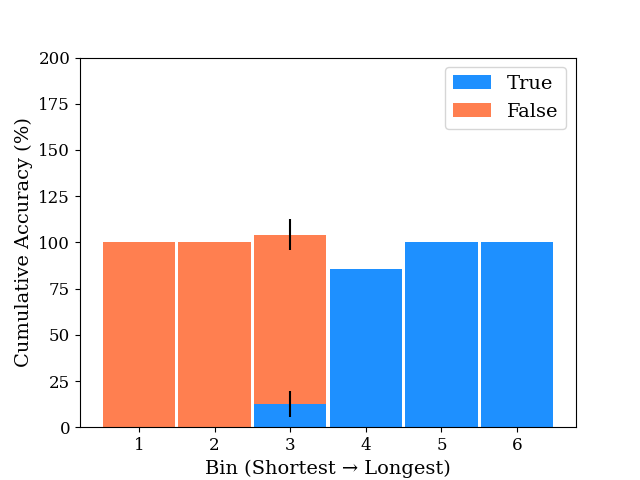}
    \includegraphics[width=0.19\textwidth]            {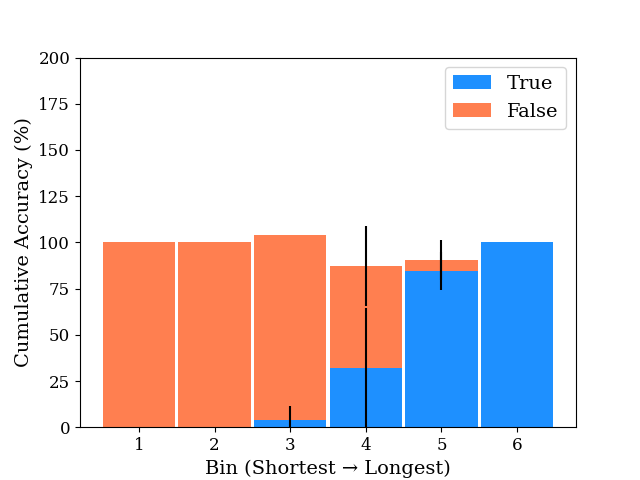}
    \includegraphics[width=0.19\textwidth]            {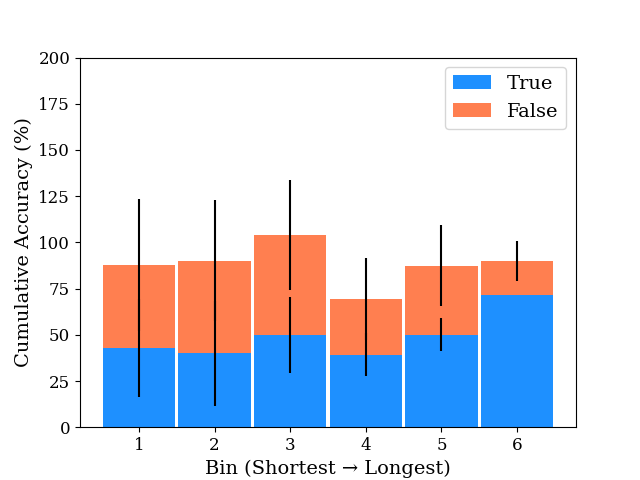}
    \caption{WNLI}
\end{subfigure}
\begin{subfigure}{\linewidth}
    \centering
    \includegraphics[width=0.19\textwidth]            {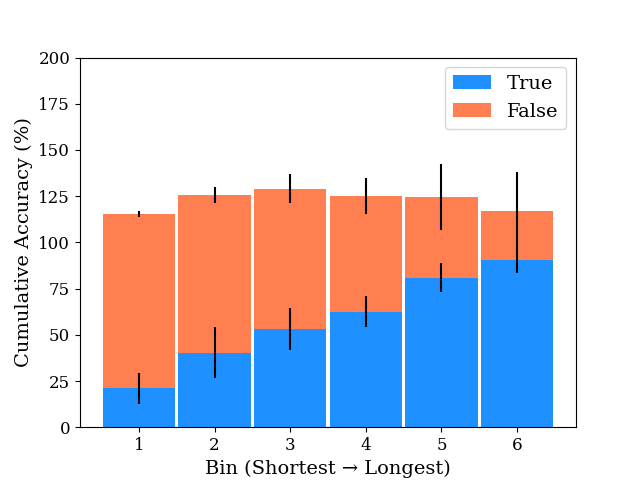}
    \includegraphics[width=0.19\textwidth]            {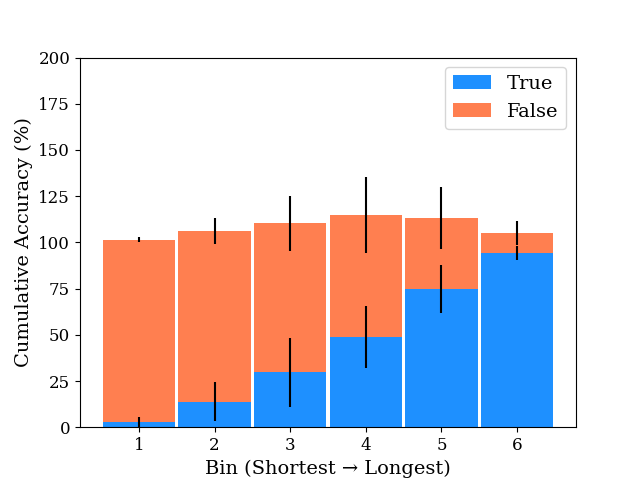}
    \includegraphics[width=0.19\textwidth]            {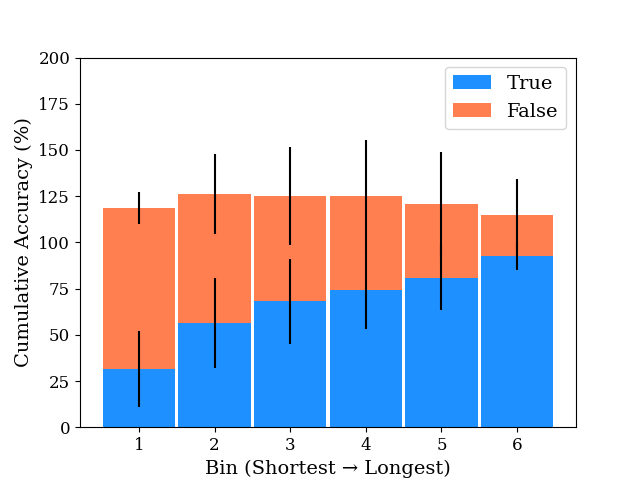}
    \includegraphics[width=0.19\textwidth]            {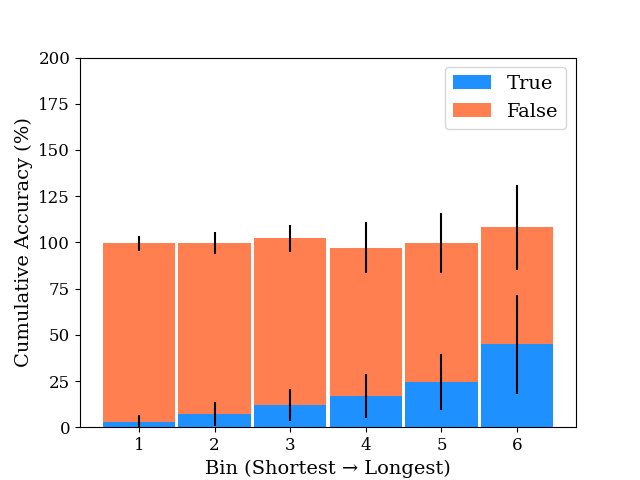}
    \includegraphics[width=0.19\textwidth]            {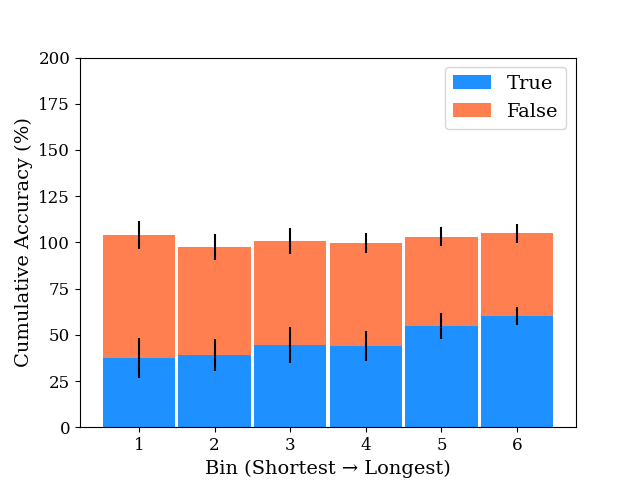}
    \caption{QNLI}
\end{subfigure}
\begin{subfigure}{\linewidth}
    \centering
    \includegraphics[width=0.19\textwidth]            {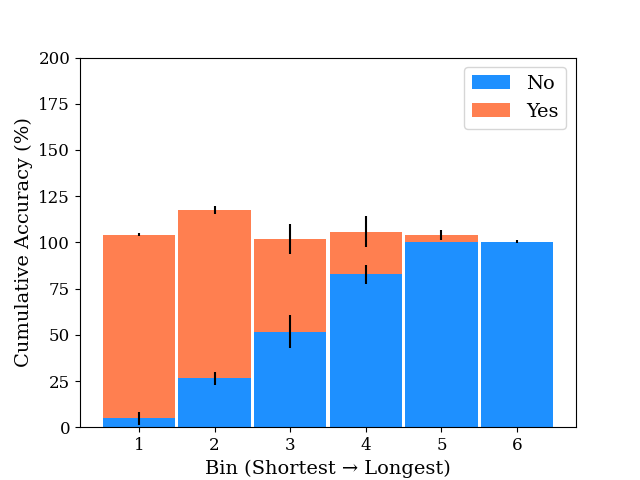}
    \includegraphics[width=0.19\textwidth]            {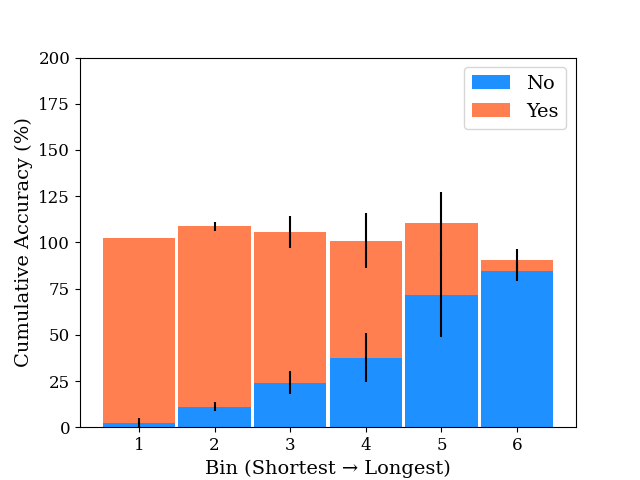}
    \includegraphics[width=0.19\textwidth]            {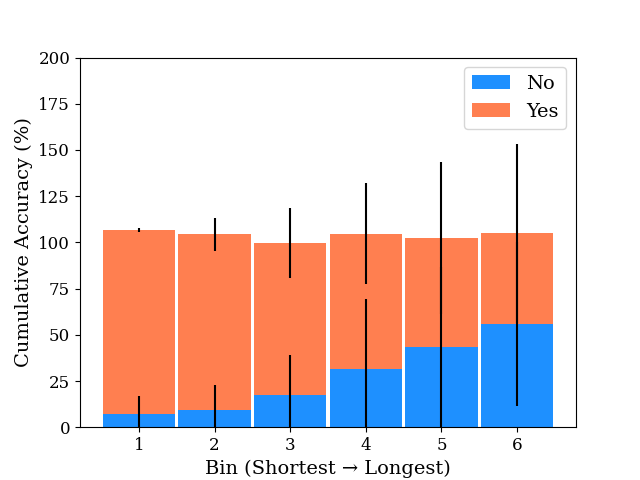}
    \includegraphics[width=0.19\textwidth]            {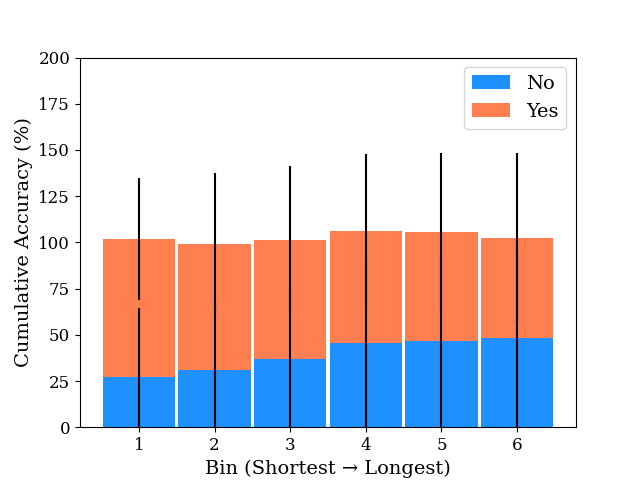}
    \includegraphics[width=0.19\textwidth]            {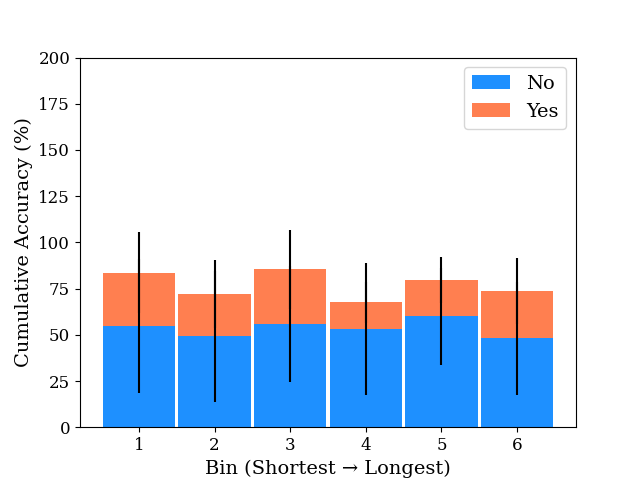}
    \caption{MRPC}
\end{subfigure}
\begin{subfigure}{\linewidth}
    \centering
    \includegraphics[width=0.19\textwidth]            {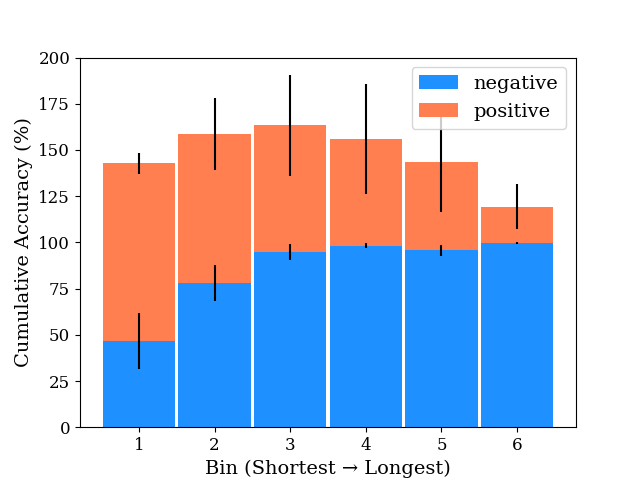}
    \includegraphics[width=0.19\textwidth]            {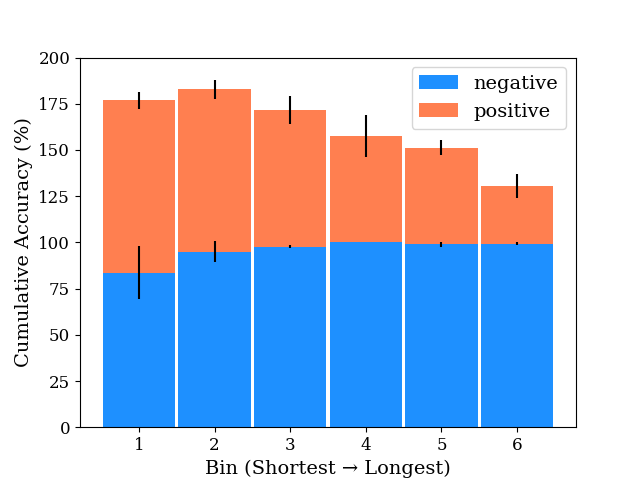}
    \includegraphics[width=0.19\textwidth]            {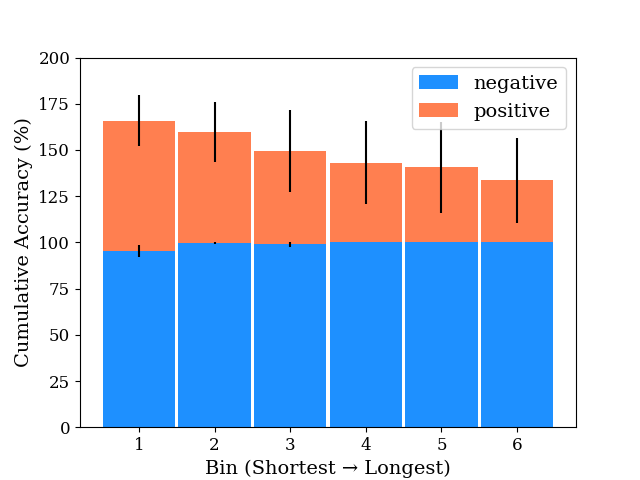}
    \includegraphics[width=0.19\textwidth]            {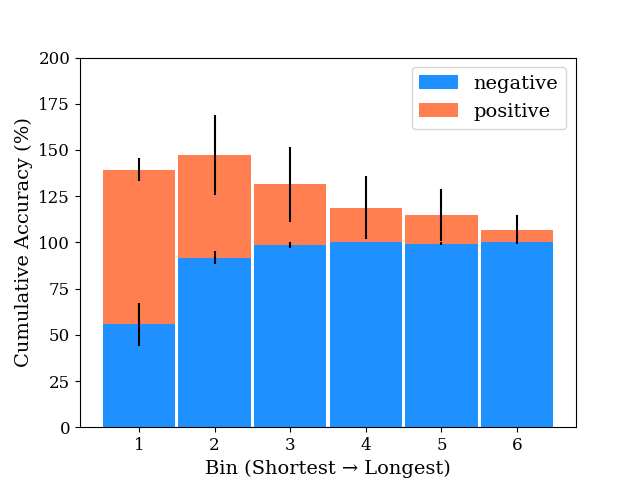}
    \includegraphics[width=0.19\textwidth]            {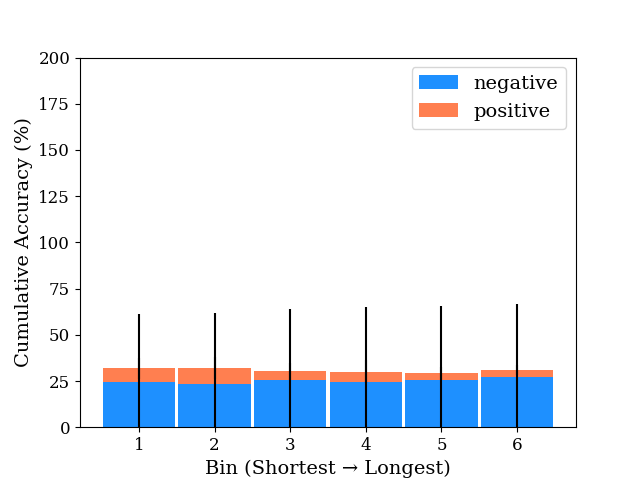}
    \caption{SST-2}
\end{subfigure}
\end{minipage}
\hfill
\begin{minipage}[c]{\linewidth}
    \caption{Finetuning performance of Llama3 8B, Llama2 7B, Mistral 7B, OPT 6.7B, and GPT Neo 2.7B (from left to right) where $y_1$ (Blue) samples long demonstrations and $y_2$ (Orange) samples short demonstrations.}
\end{minipage}
\end{figure*}
\begin{figure*}[t!]
\centering
\begin{minipage}[t]{\linewidth}
\begin{subfigure}{\linewidth}
    \centering
    \includegraphics[width=0.19\textwidth]            {latex/figures/q1_icl/hans_llama3_8b_class2_6bins.png}
    \includegraphics[width=0.19\textwidth]            {latex/figures/q1_icl/hans_llama2_7b_class2_6bins.png}
    \includegraphics[width=0.19\textwidth]            {latex/figures/q1_icl/hans_mistral_7b_class2_6bins.png}
    \includegraphics[width=0.19\textwidth]            {latex/figures/q1_icl/hans_opt_6dot7b_class2_6bins.png}
    \includegraphics[width=0.19\textwidth]            {latex/figures/q1_icl/hans_gpt_2dot7b_class2_6bins.png}
    \caption{Hans}
\end{subfigure}
\begin{subfigure}{\linewidth}
    \centering
    \includegraphics[width=0.19\textwidth]            {latex/figures/q1_icl/en_llama3_8b_class2_6bins.png}
    \includegraphics[width=0.19\textwidth]            {latex/figures/q1_icl/en_llama2_7b_class2_6bins.png}
    \includegraphics[width=0.19\textwidth]            {latex/figures/q1_icl/en_mistral_7b_class2_6bins.png}
    \includegraphics[width=0.19\textwidth]            {latex/figures/q1_icl/en_opt_6dot7b_class2_6bins.png}
    \includegraphics[width=0.19\textwidth]            {latex/figures/q1_icl/en_gpt_2dot7b_class2_6bins.png}
    \caption{PAWS-X$_{\textsc{EN}}$}
\end{subfigure}
\begin{subfigure}{\linewidth}
    \centering
    \includegraphics[width=0.19\textwidth]            {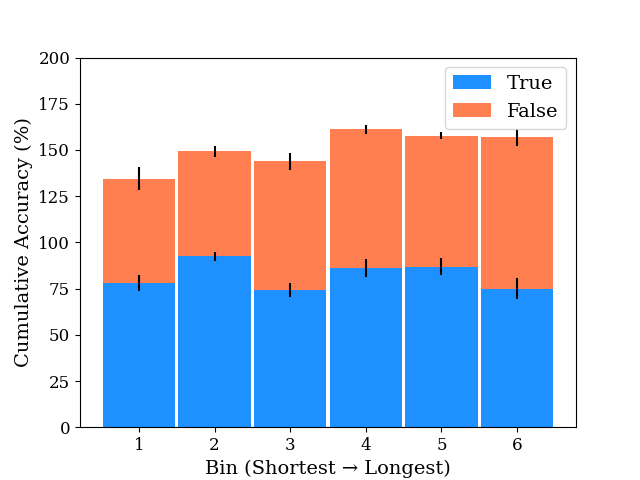}
    \includegraphics[width=0.19\textwidth]            {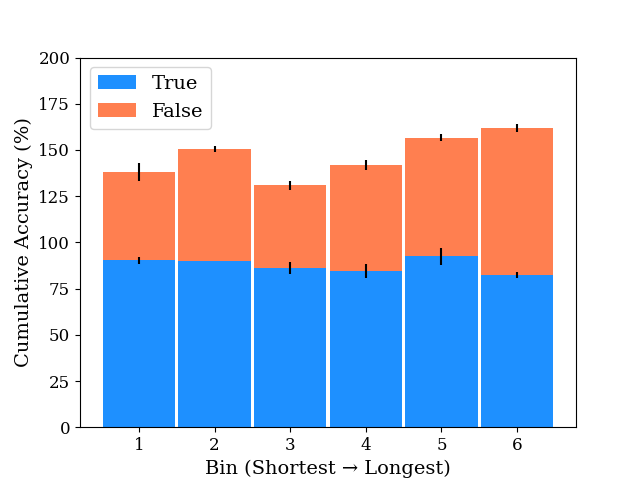}
    \includegraphics[width=0.19\textwidth]            {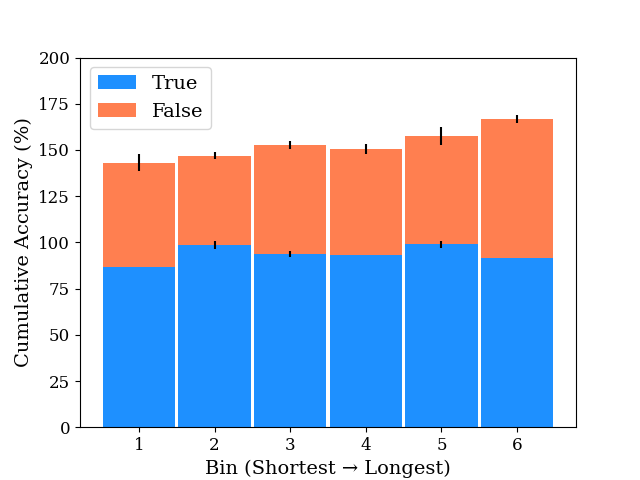}
    \includegraphics[width=0.19\textwidth]            {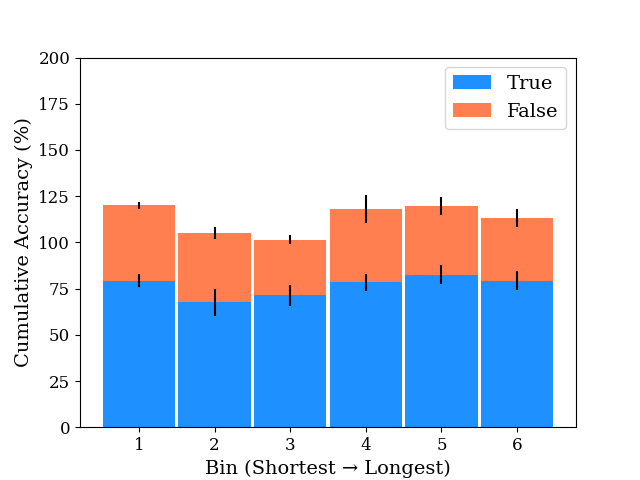}
    \includegraphics[width=0.19\textwidth]            {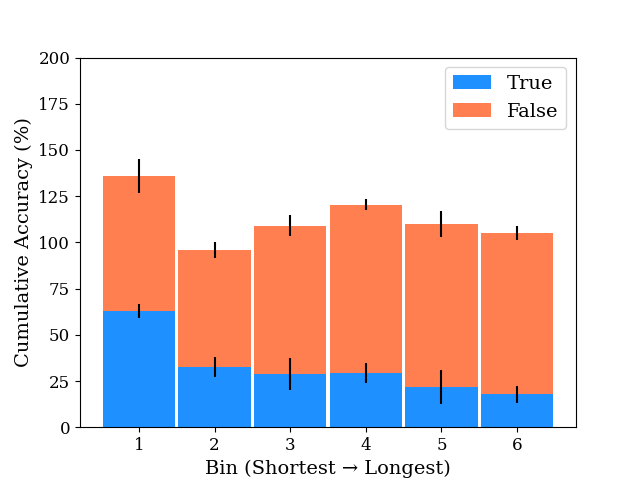}
    \caption{RTE}
\end{subfigure}
\begin{subfigure}{\linewidth}
    \centering
    \includegraphics[width=0.19\textwidth]            {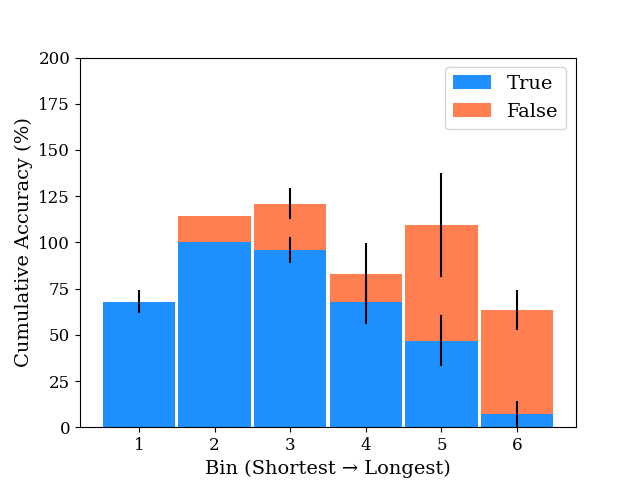}
    \includegraphics[width=0.19\textwidth]            {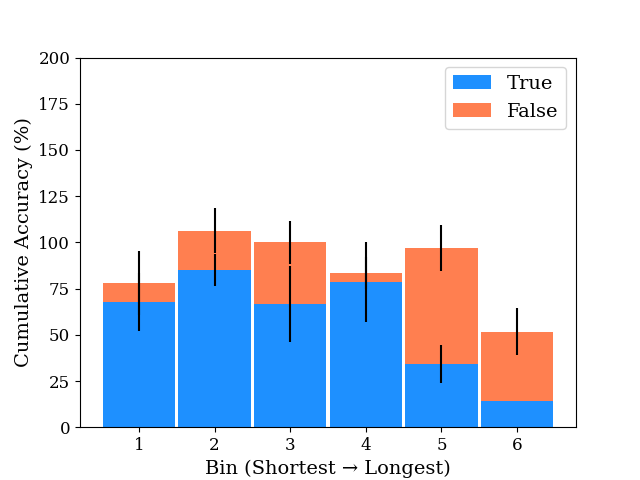}
    \includegraphics[width=0.19\textwidth]            {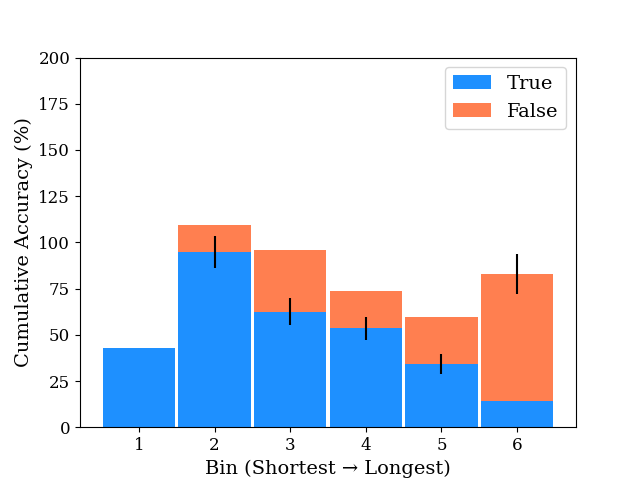}
    \includegraphics[width=0.19\textwidth]            {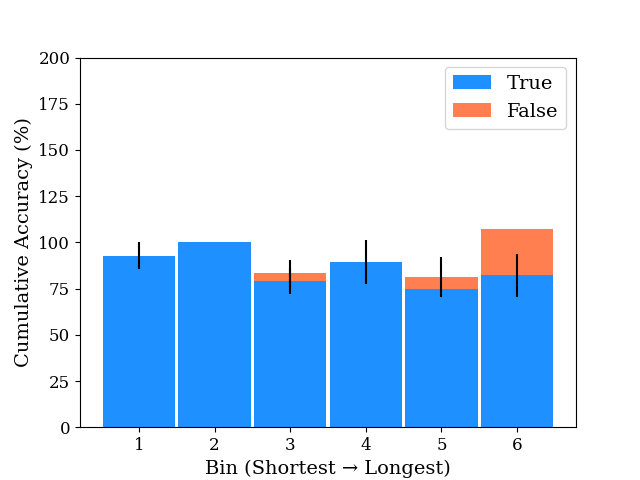}
    \includegraphics[width=0.19\textwidth]            {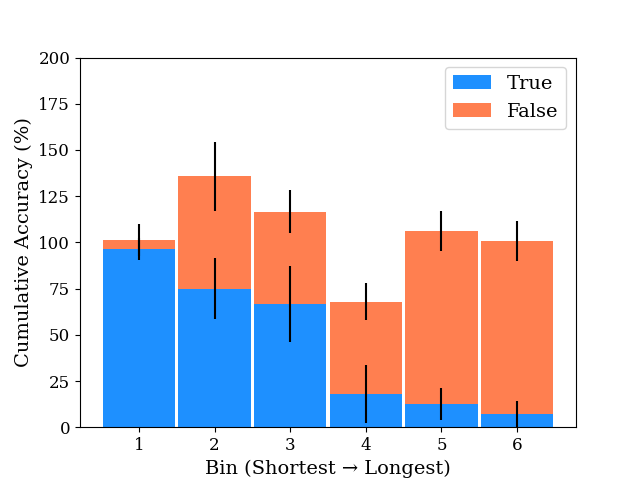}
    \caption{WNLI}
\end{subfigure}
\begin{subfigure}{\linewidth}
    \centering
    \includegraphics[width=0.19\textwidth]            {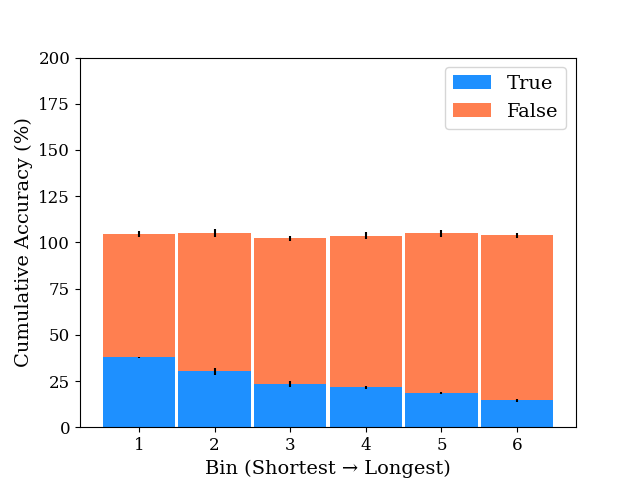}
    \includegraphics[width=0.19\textwidth]            {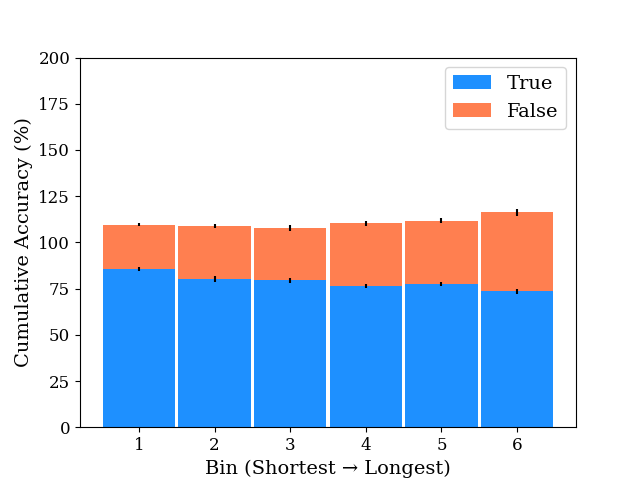}
    \includegraphics[width=0.19\textwidth]            {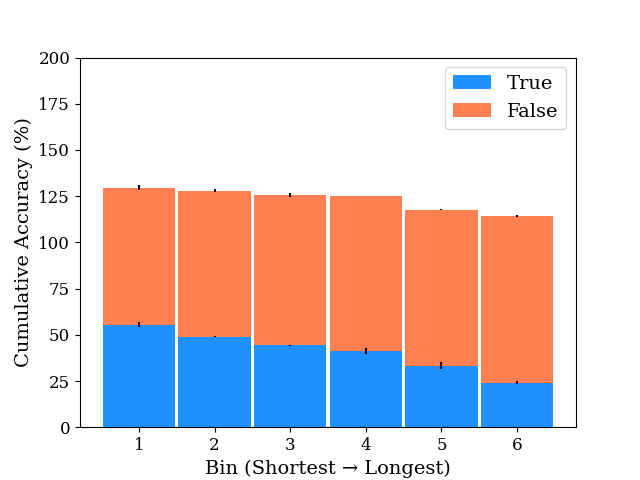}
    \includegraphics[width=0.19\textwidth]            {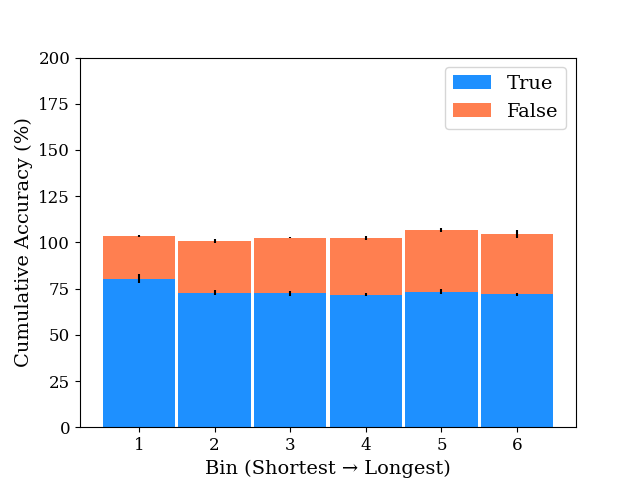}
    \includegraphics[width=0.19\textwidth]            {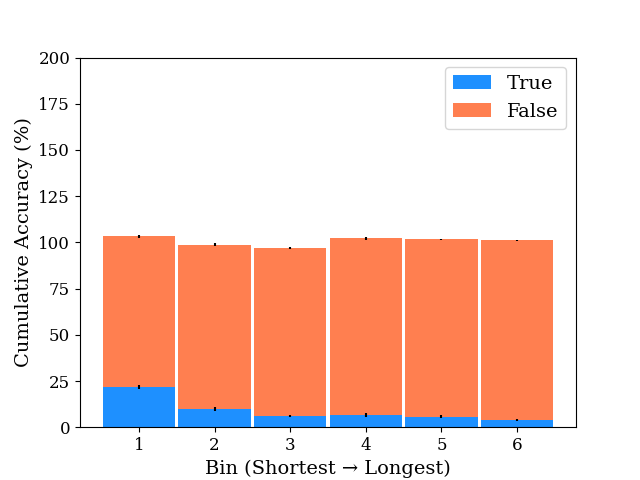}
    \caption{QNLI}
\end{subfigure}
\begin{subfigure}{\linewidth}
    \centering
    \includegraphics[width=0.19\textwidth]            {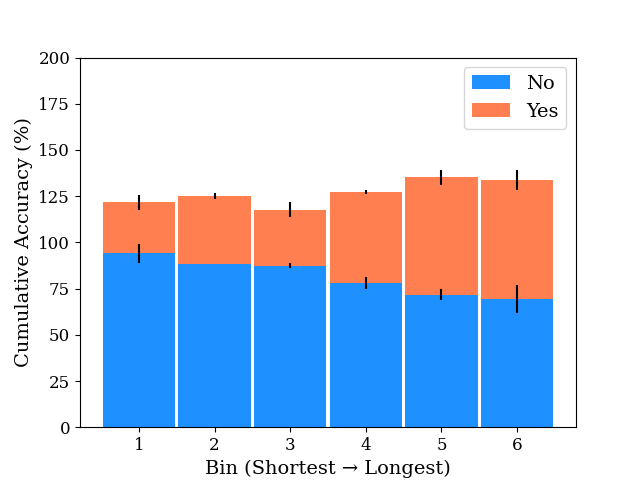}
    \includegraphics[width=0.19\textwidth]            {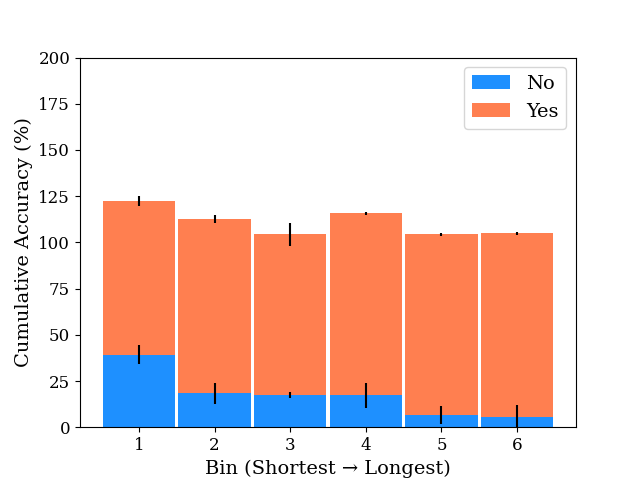}
    \includegraphics[width=0.19\textwidth]            {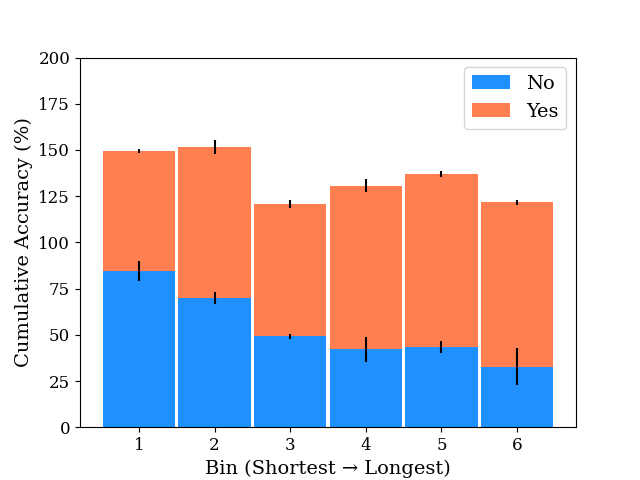}
    \includegraphics[width=0.19\textwidth]            {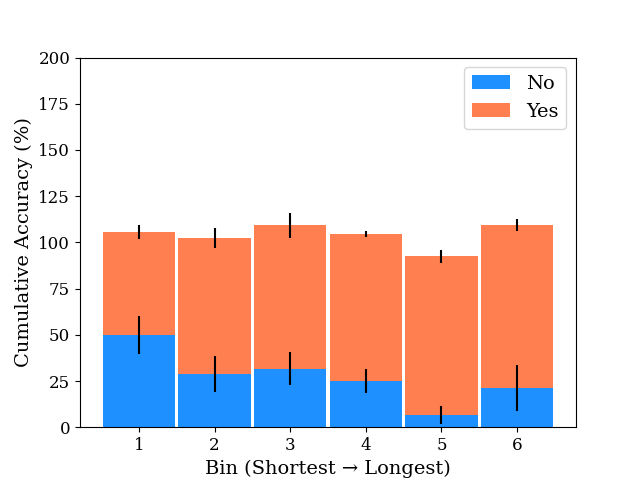}
    \includegraphics[width=0.19\textwidth]            {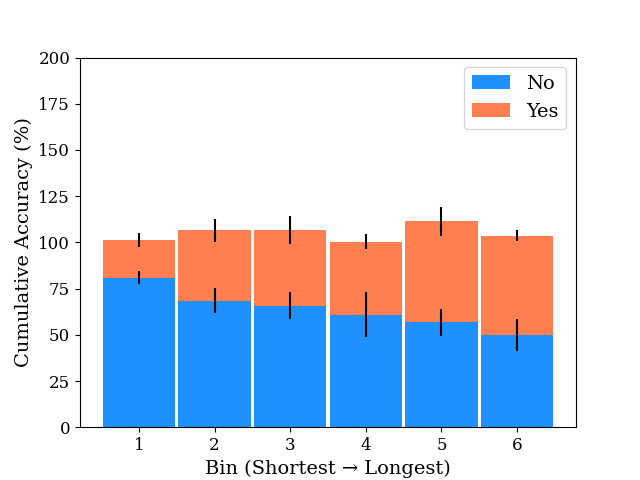}
    \caption{MRPC}
\end{subfigure}
\begin{subfigure}{\linewidth}
    \centering
    \includegraphics[width=0.19\textwidth]            {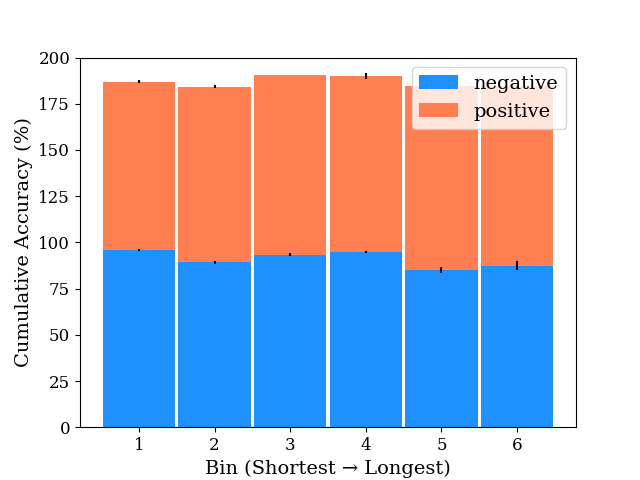}
    \includegraphics[width=0.19\textwidth]            {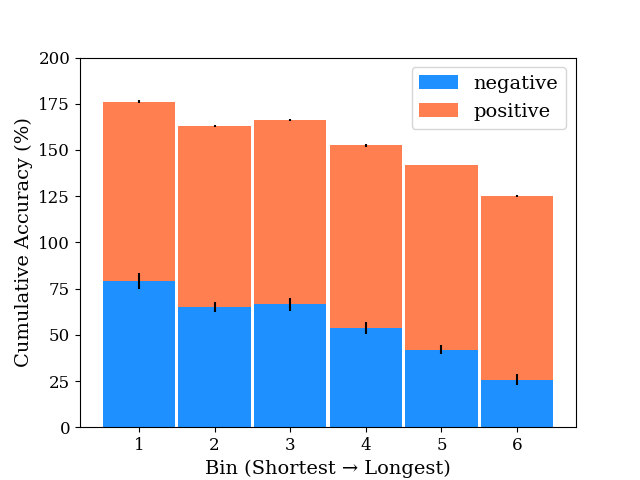}
    \includegraphics[width=0.19\textwidth]            {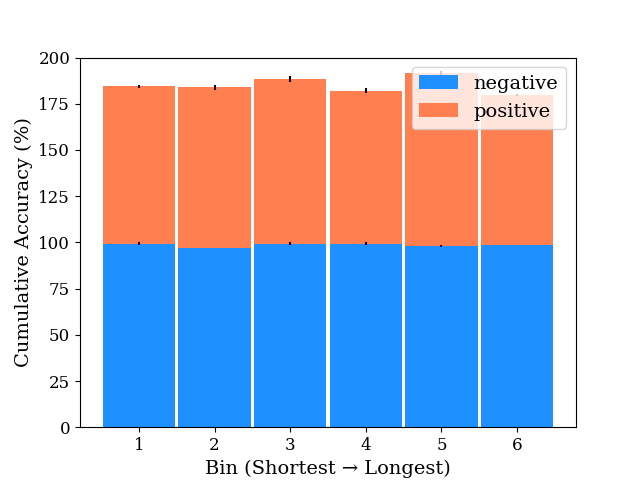}
    \includegraphics[width=0.19\textwidth]            {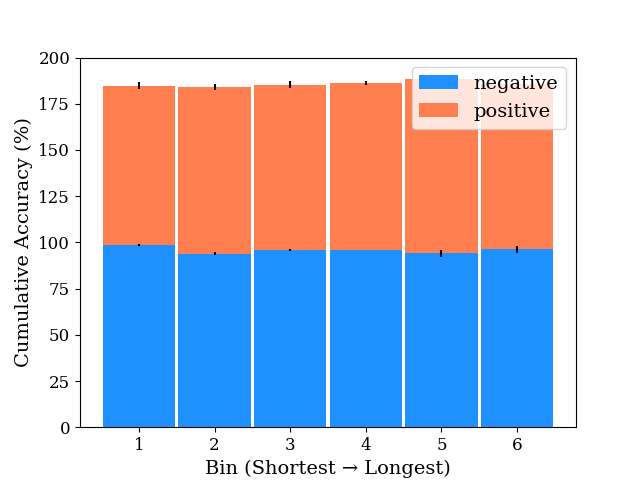}
    \includegraphics[width=0.19\textwidth]            {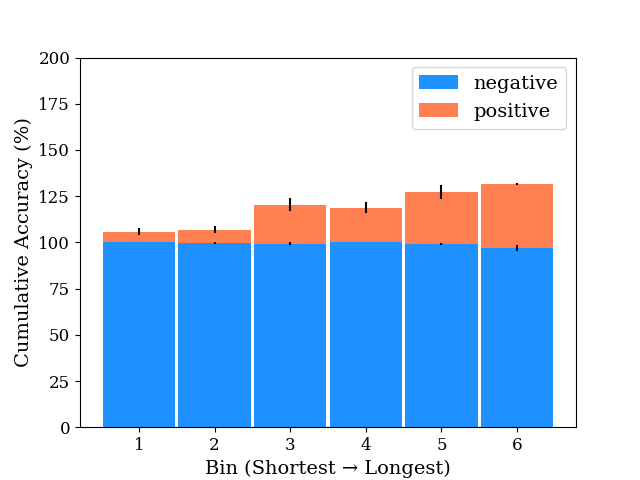}
    \caption{SST-2}
\end{subfigure}
\end{minipage}
\hfill
\begin{minipage}[c]{\linewidth}
    \caption{ICL performance of Llama3 8B, Llama2 7B, Mistral 7B, OPT 6.7B, and GPT Neo 2.7B (from left to right) where $y_1$ (Blue) samples short demonstrations and $y_2$ (Orange) samples long demonstrations.}
\end{minipage}
\end{figure*}

\begin{figure*}[t!]
\centering
\begin{minipage}[t]{\linewidth}
\begin{subfigure}{\linewidth}
    \centering
    \includegraphics[width=0.19\textwidth]            {latex/figures/ft/hans_200_llama3_8b_class2_6bins.png}
    \includegraphics[width=0.19\textwidth]            {latex/figures/ft/hans_200_llama2_7b_class2_6bins.png}
    \includegraphics[width=0.19\textwidth]            {latex/figures/ft/hans_200_mistral_7b_class2_6bins.png}
    \includegraphics[width=0.19\textwidth]            {latex/figures/ft/hans_200_opt_6dot7b_class2_6bins.png}
    \includegraphics[width=0.19\textwidth]            {latex/figures/ft/hans_200_gpt_2dot7b_class2_6bins.png}
    \caption{Hans}
\end{subfigure}
\begin{subfigure}{\linewidth}
    \centering
    \includegraphics[width=0.19\textwidth]            {latex/figures/ft/paws_200_llama3_8b_class2_6bins.png}
    \includegraphics[width=0.19\textwidth]            {latex/figures/ft/paws_200_llama2_7b_class2_6bins.png}
    \includegraphics[width=0.19\textwidth]            {latex/figures/ft/paws_200_mistral_7b_class2_6bins.png}
    \includegraphics[width=0.19\textwidth]            {latex/figures/ft/paws_200_opt_6dot7b_class2_6bins.png}
    \includegraphics[width=0.19\textwidth]            {latex/figures/ft/paws_200_gpt_2dot7b_class2_6bins.png}
    \caption{PAWS-X$_{\textsc{EN}}$}
\end{subfigure}
\begin{subfigure}{\linewidth}
    \centering
    \includegraphics[width=0.19\textwidth]            {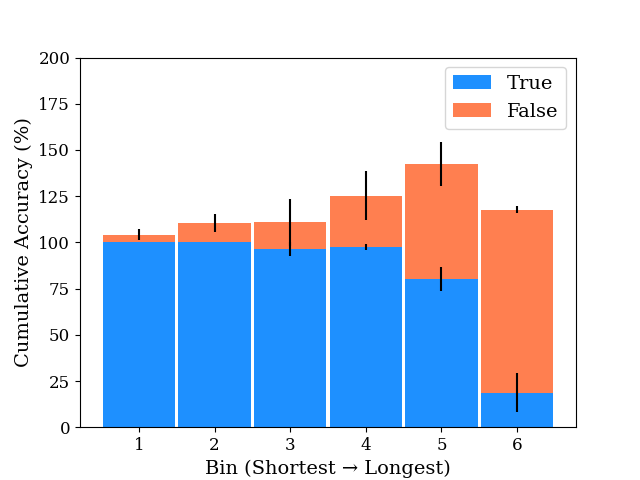}
    \includegraphics[width=0.19\textwidth]            {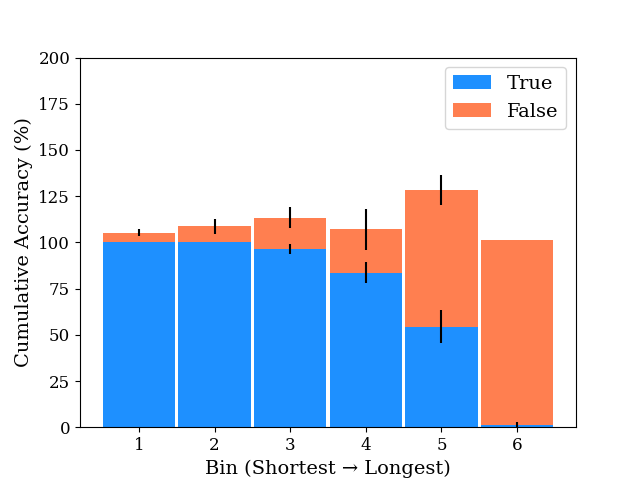}
    \includegraphics[width=0.19\textwidth]            {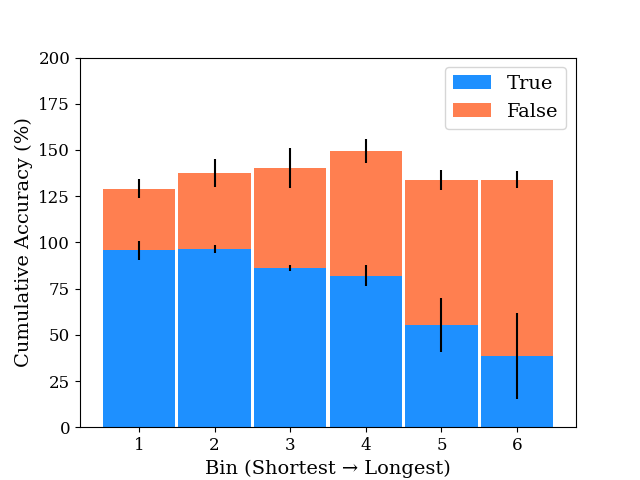}
    \includegraphics[width=0.19\textwidth]            {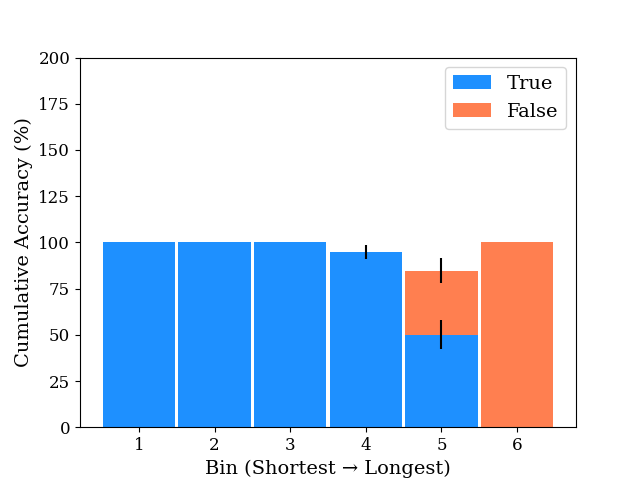}
    \includegraphics[width=0.19\textwidth]            {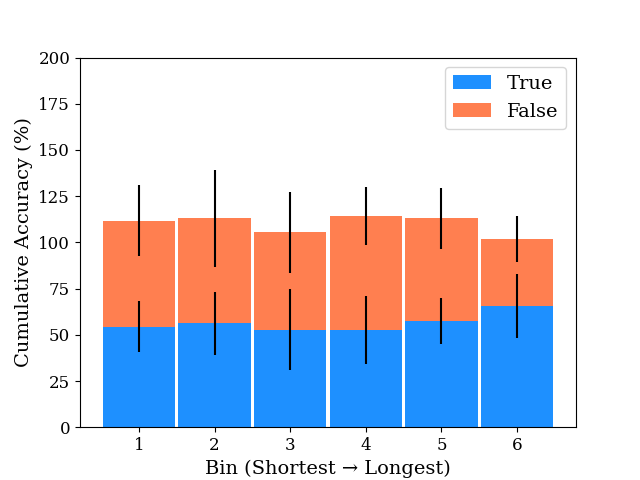}
    \caption{RTE}
\end{subfigure}
\begin{subfigure}{\linewidth}
    \centering
    \includegraphics[width=0.19\textwidth]            {latex/figures/ft/wnli_200_llama3_8b_class2_6bins.png}
    \includegraphics[width=0.19\textwidth]            {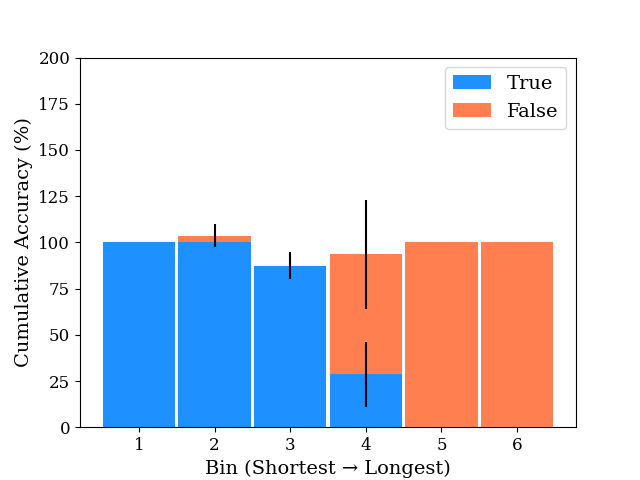}
    \includegraphics[width=0.19\textwidth]            {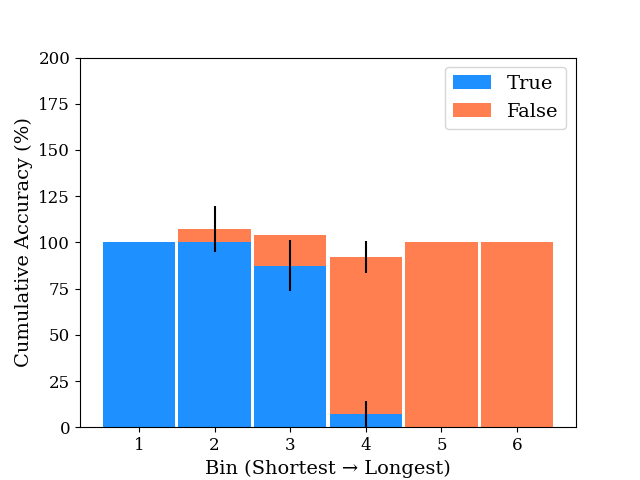}
    \includegraphics[width=0.19\textwidth]            {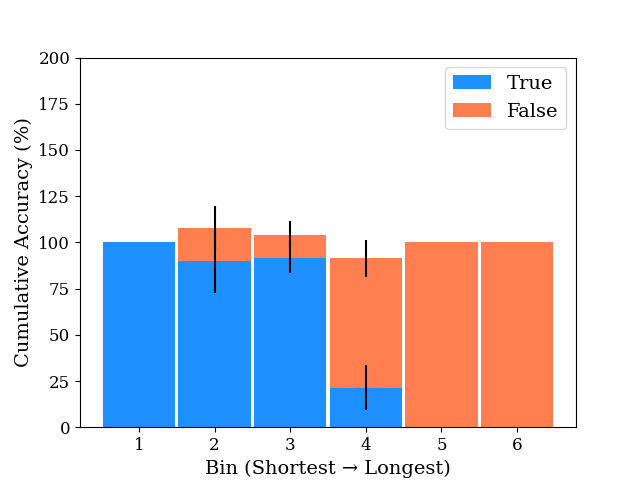}
    \includegraphics[width=0.19\textwidth]            {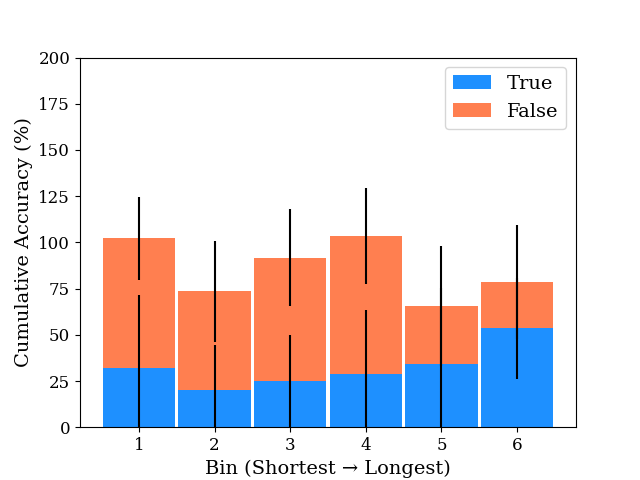}
    \caption{WNLI}
\end{subfigure}
\begin{subfigure}{\linewidth}
    \centering
    \includegraphics[width=0.19\textwidth]            {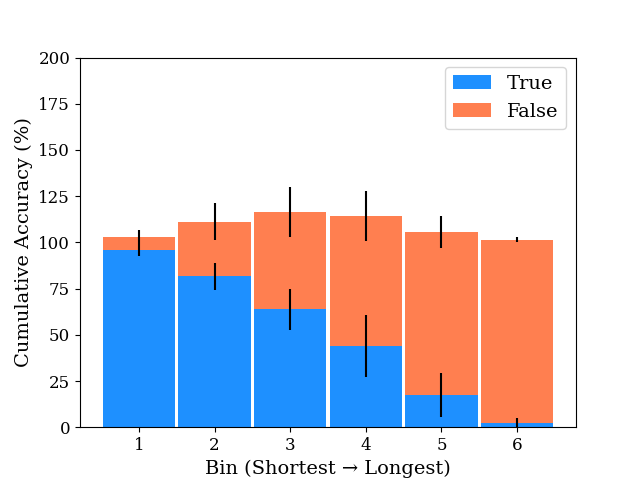}
    \includegraphics[width=0.19\textwidth]            {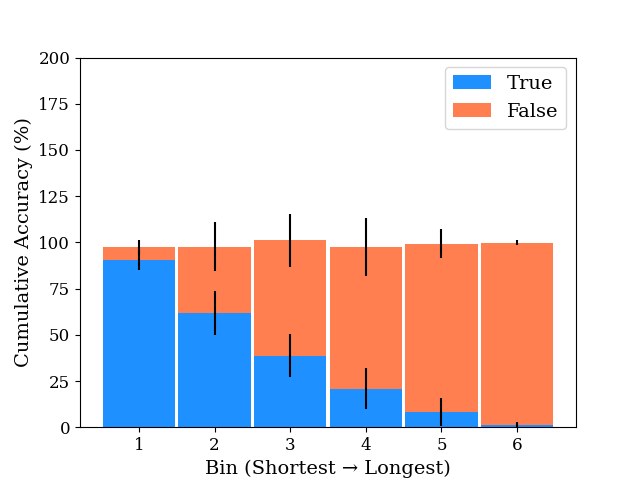}
    \includegraphics[width=0.19\textwidth]            {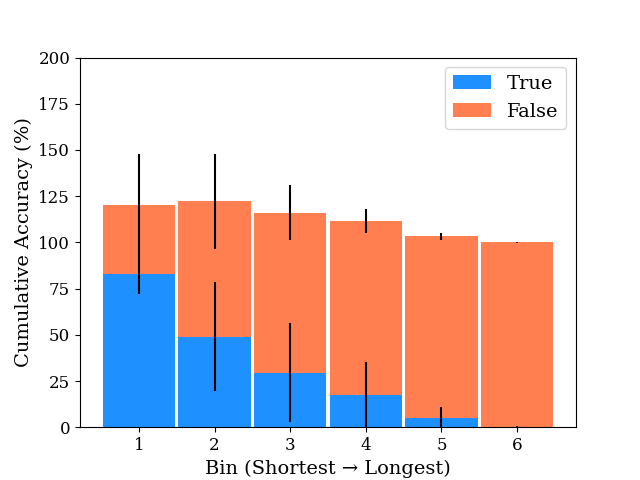}
    \includegraphics[width=0.19\textwidth]            {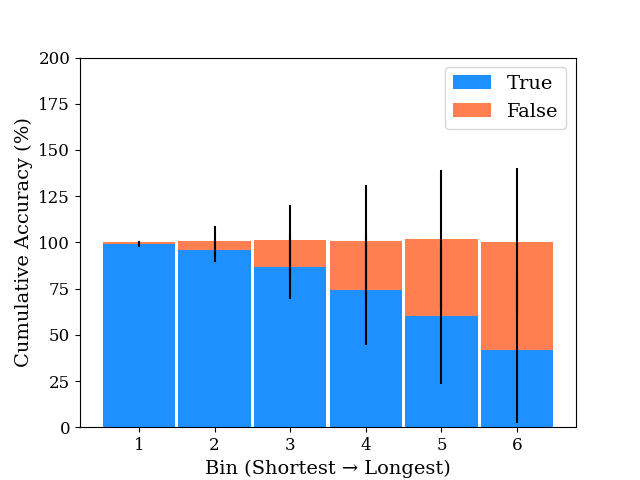}
    \includegraphics[width=0.19\textwidth]            {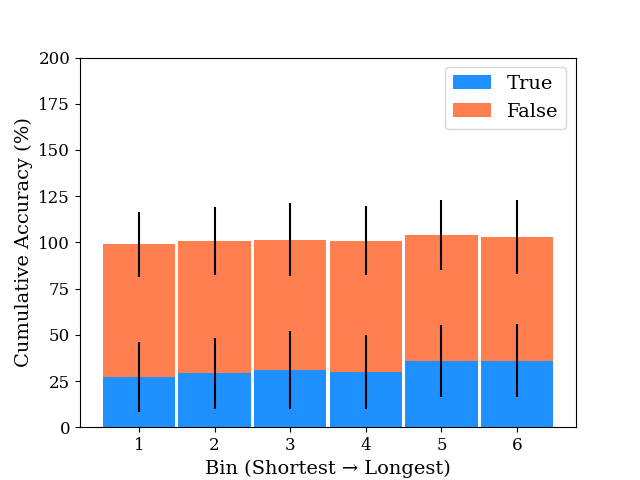}
    \caption{QNLI}
\end{subfigure}
\begin{subfigure}{\linewidth}
    \centering
    \includegraphics[width=0.19\textwidth]            {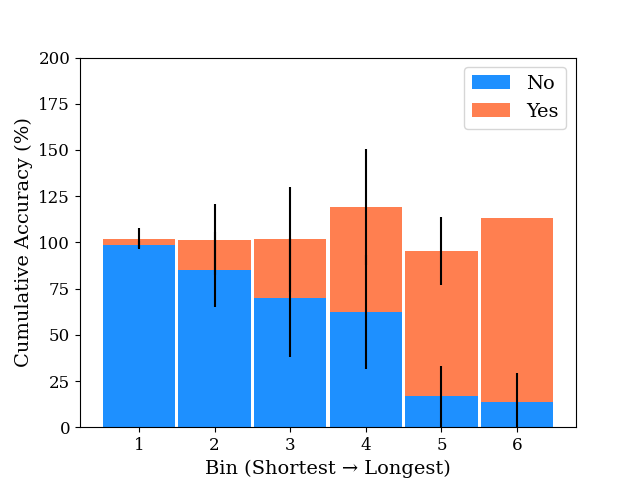}
    \includegraphics[width=0.19\textwidth]            {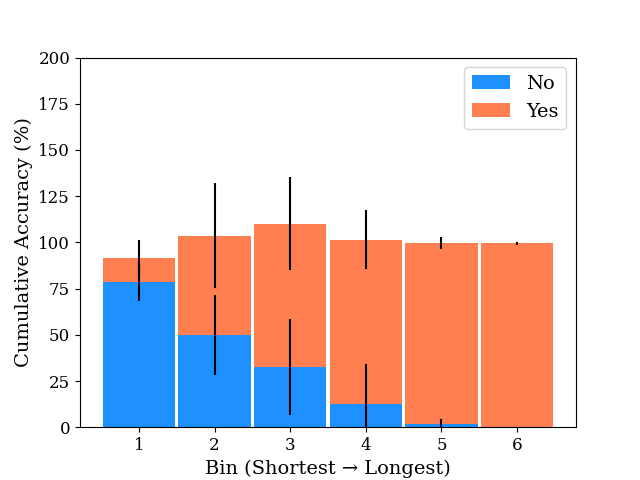}
    \includegraphics[width=0.19\textwidth]            {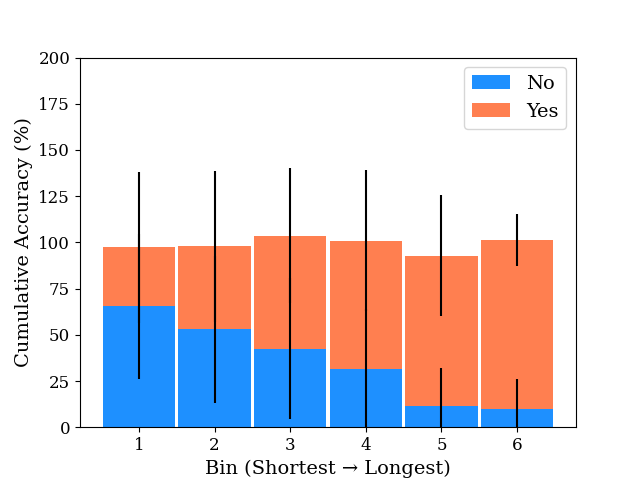}
    \includegraphics[width=0.19\textwidth]            {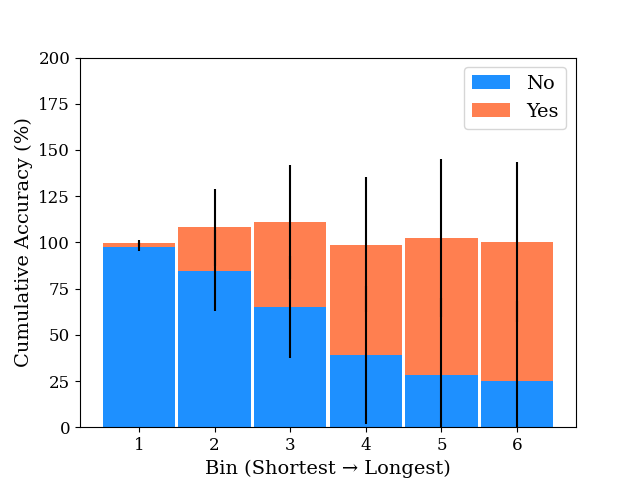}
    \includegraphics[width=0.19\textwidth]            {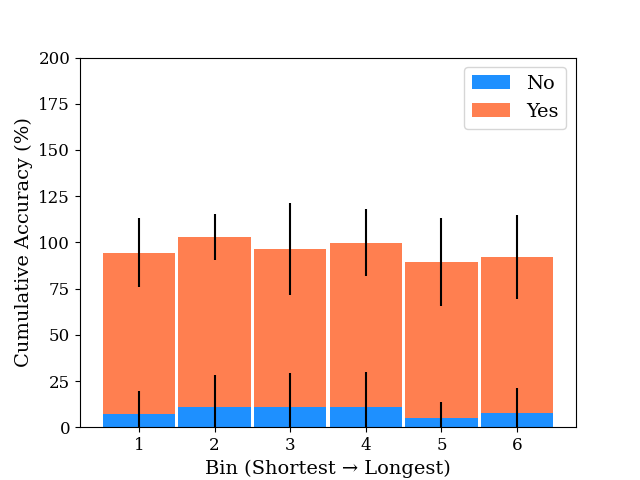}
    \caption{MRPC}
\end{subfigure}
\begin{subfigure}{\linewidth}
    \centering
    \includegraphics[width=0.19\textwidth]            {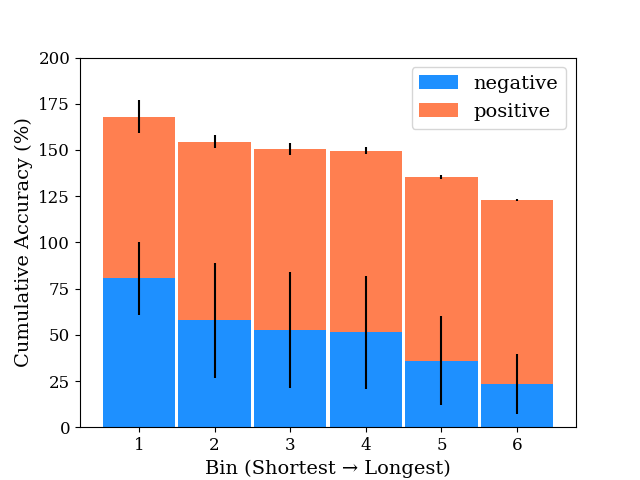}
    \includegraphics[width=0.19\textwidth]            {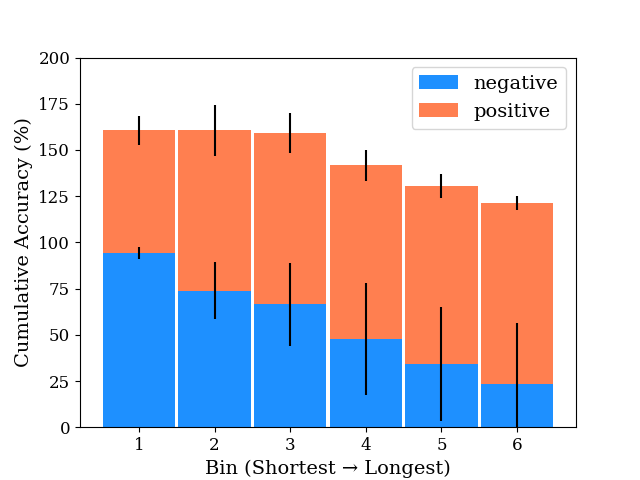}
    \includegraphics[width=0.19\textwidth]            {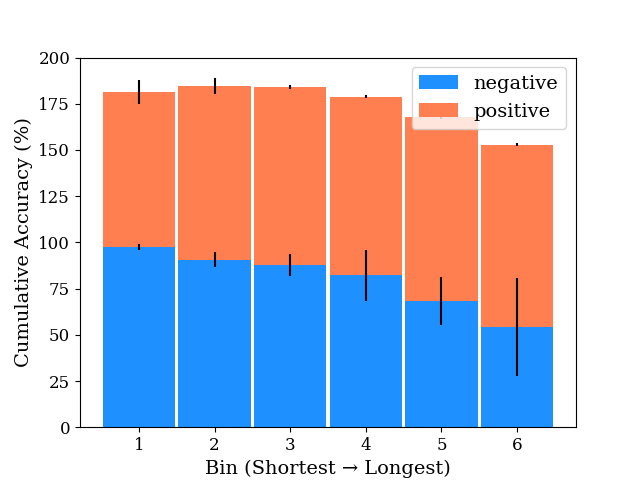}
    \includegraphics[width=0.19\textwidth]            {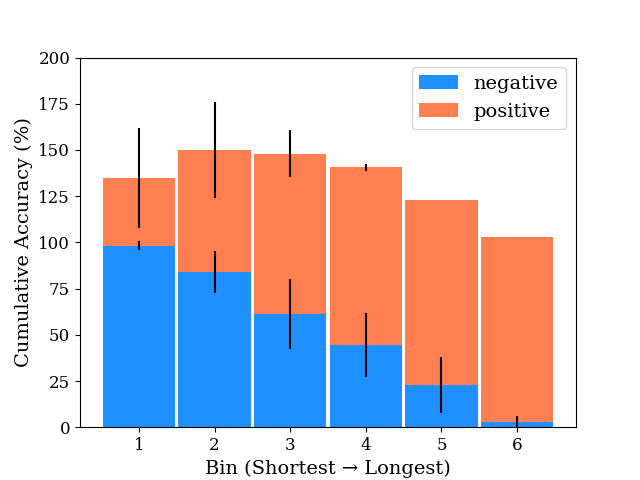}
    \includegraphics[width=0.19\textwidth]            {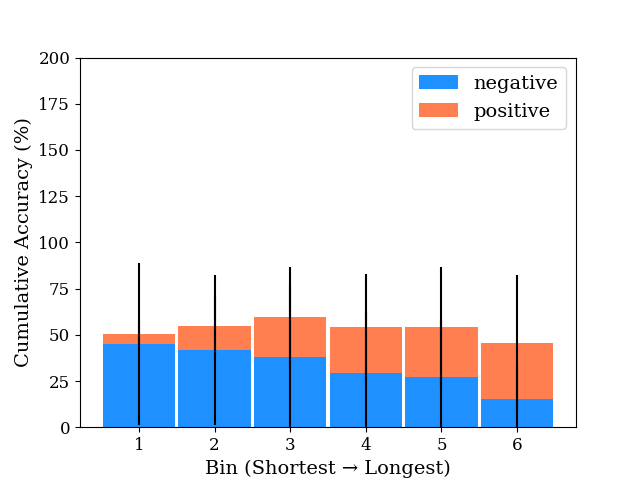}
    \caption{SST-2}
\end{subfigure}
\end{minipage}
\hfill
\begin{minipage}[c]{\linewidth}
    \caption{Finetuning performance of Llama3 8B, Llama2 7B, Mistral 7B, OPT 6.7B, and GPT Neo 2.7B (from left to right) where $y_1$ (Blue) samples short demonstrations and $y_2$ (Orange) samples long demonstrations.}
\end{minipage}
\end{figure*}
\begin{figure*}[t!]
\centering
\begin{minipage}[t]{\linewidth}
\begin{subfigure}{\linewidth}
    \centering
    \includegraphics[width=0.19\textwidth]            {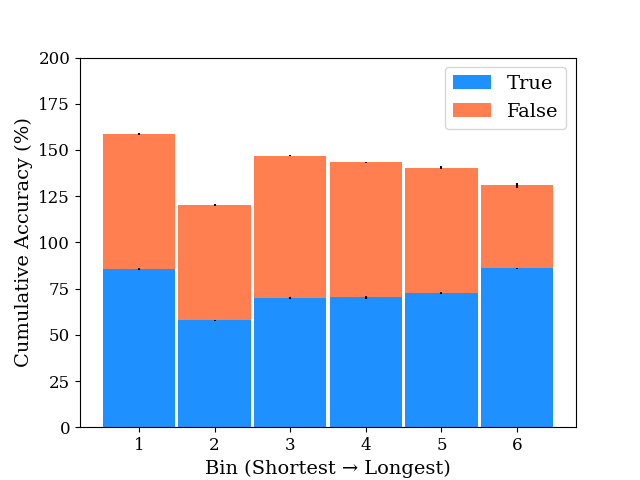}
    \includegraphics[width=0.19\textwidth]            {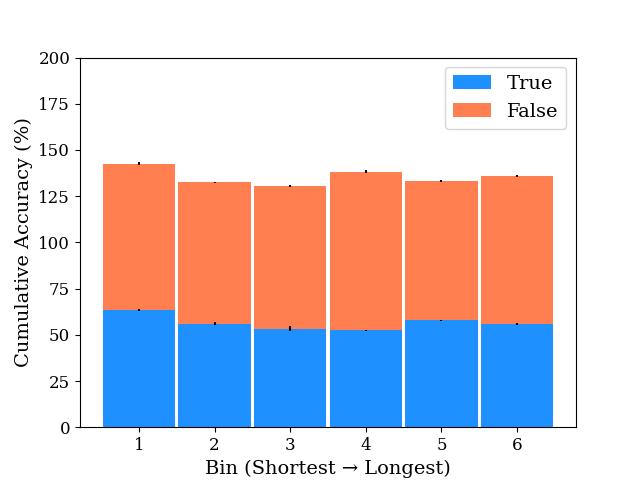}
    \includegraphics[width=0.19\textwidth]            {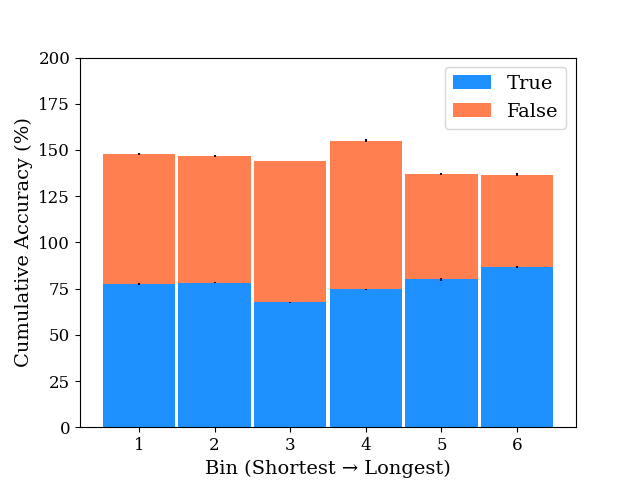}
    \includegraphics[width=0.19\textwidth]            {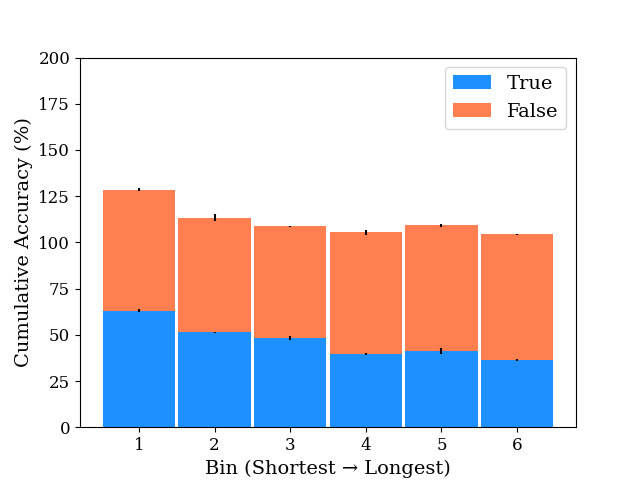}
    \includegraphics[width=0.19\textwidth]            {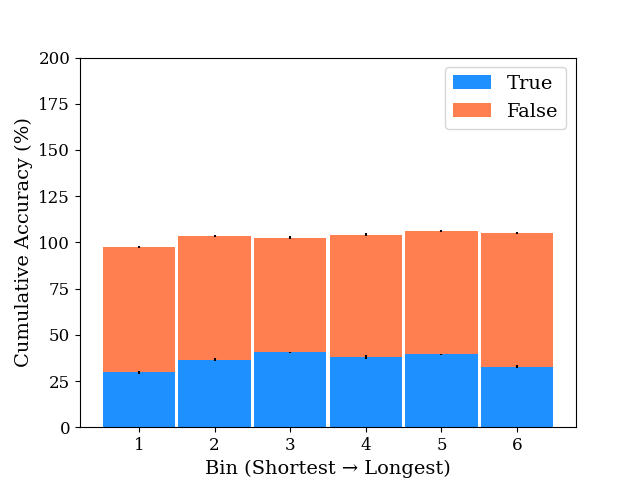}
    \caption{Hans}
\end{subfigure}
\begin{subfigure}{\linewidth}
    \centering
    \includegraphics[width=0.19\textwidth]            {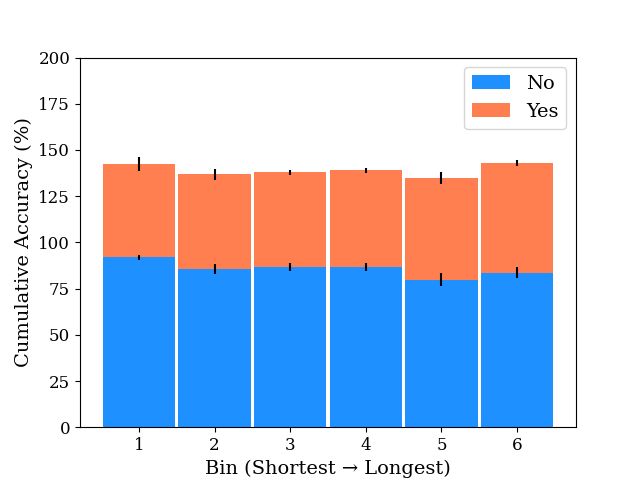}
    \includegraphics[width=0.19\textwidth]            {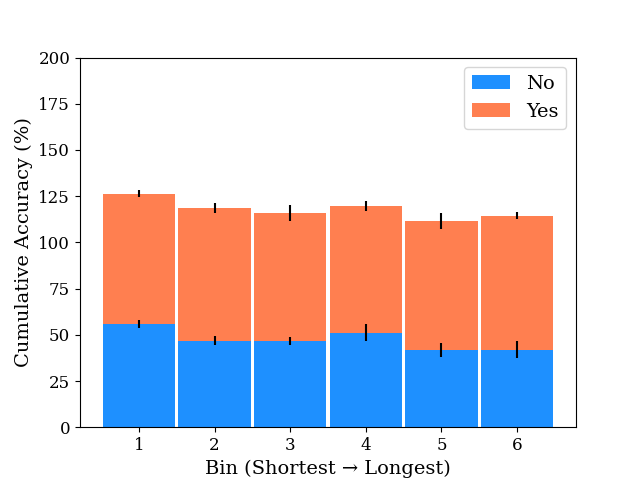}
    \includegraphics[width=0.19\textwidth]            {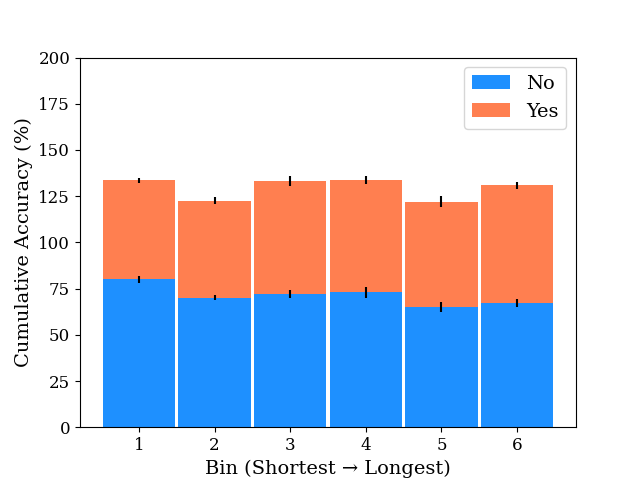}
    \includegraphics[width=0.19\textwidth]            {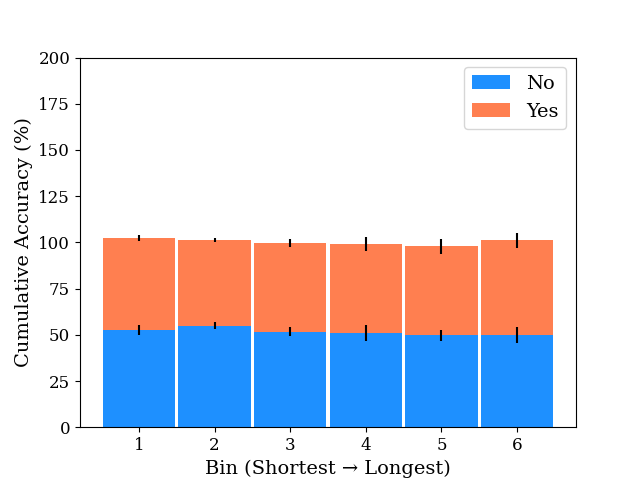}
    \includegraphics[width=0.19\textwidth]            {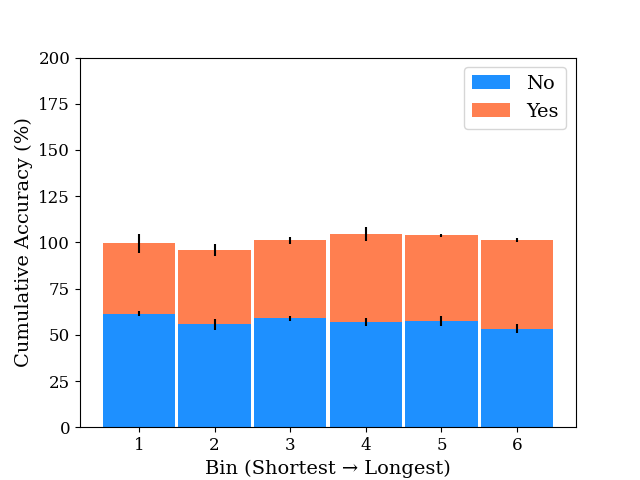}
    \caption{PAWS-X$_{\textsc{EN}}$}
\end{subfigure}
\begin{subfigure}{\linewidth}
    \centering
    \includegraphics[width=0.19\textwidth]            {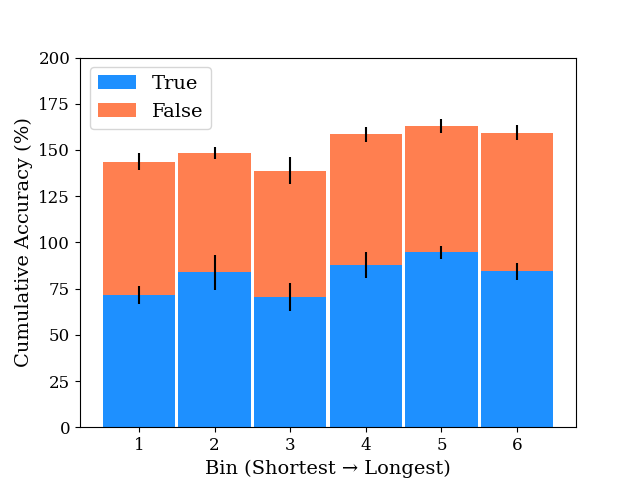}
    \includegraphics[width=0.19\textwidth]            {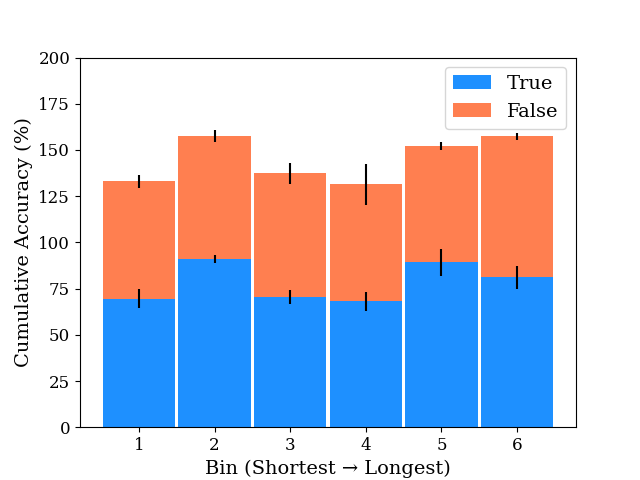}
    \includegraphics[width=0.19\textwidth]            {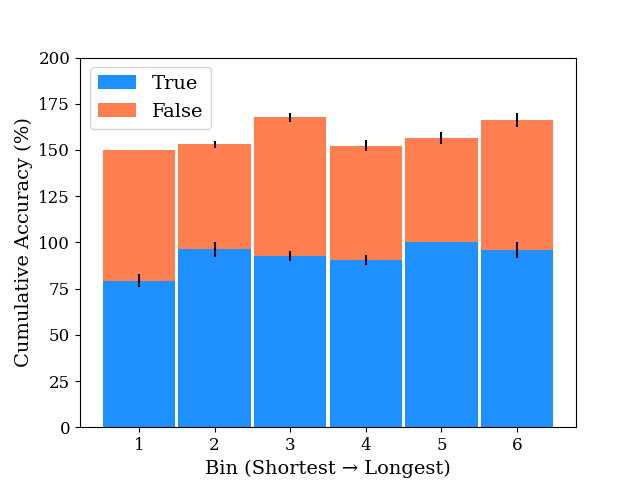}
    \includegraphics[width=0.19\textwidth]            {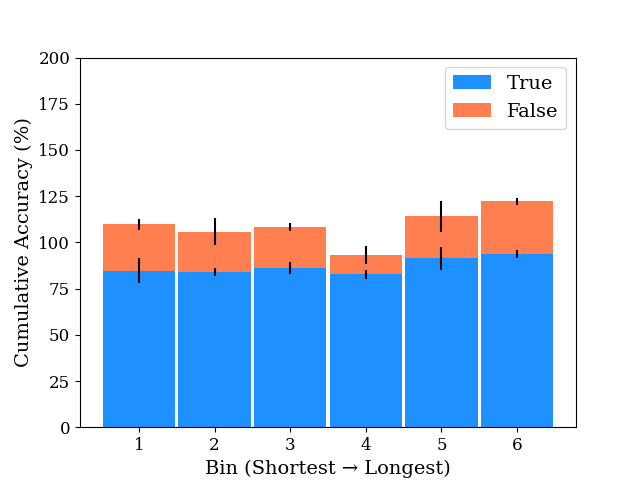}
    \includegraphics[width=0.19\textwidth]            {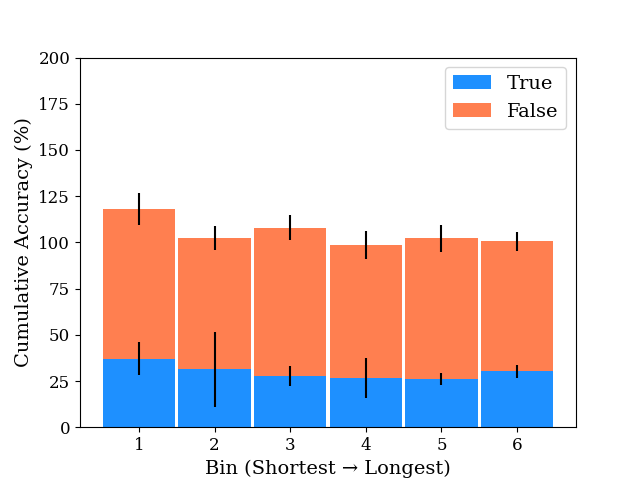}
    \caption{RTE}
\end{subfigure}
\begin{subfigure}{\linewidth}
    \centering
    \includegraphics[width=0.19\textwidth]            {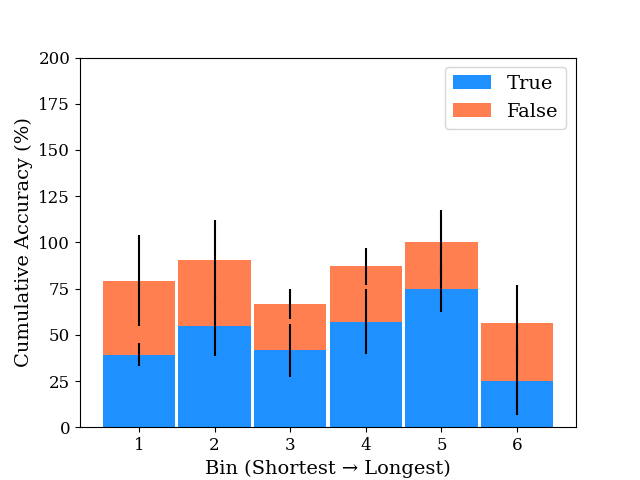}
    \includegraphics[width=0.19\textwidth]            {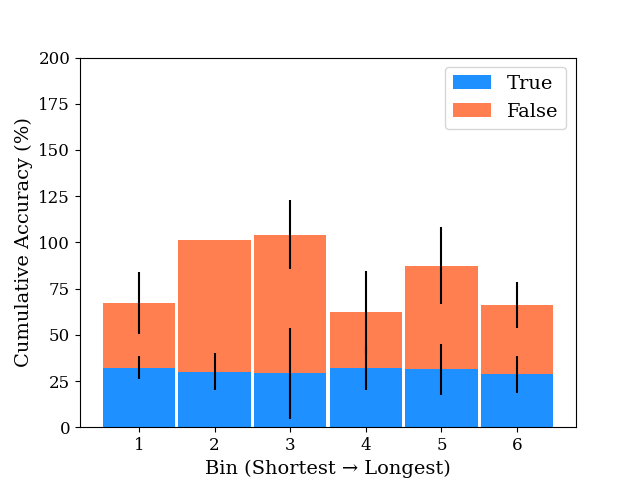}
    \includegraphics[width=0.19\textwidth]            {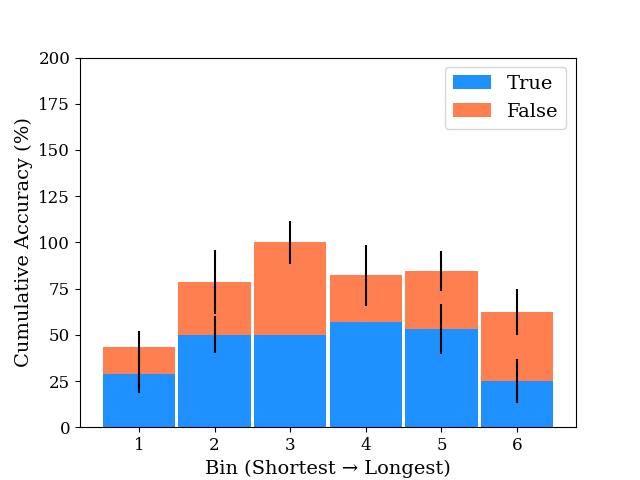}
    \includegraphics[width=0.19\textwidth]            {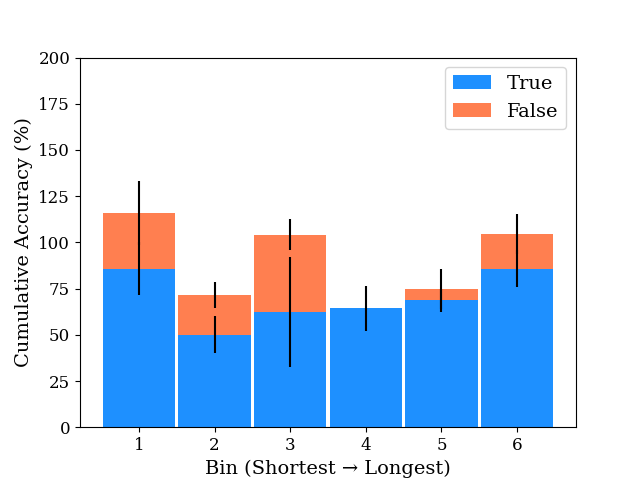}
    \includegraphics[width=0.19\textwidth]            {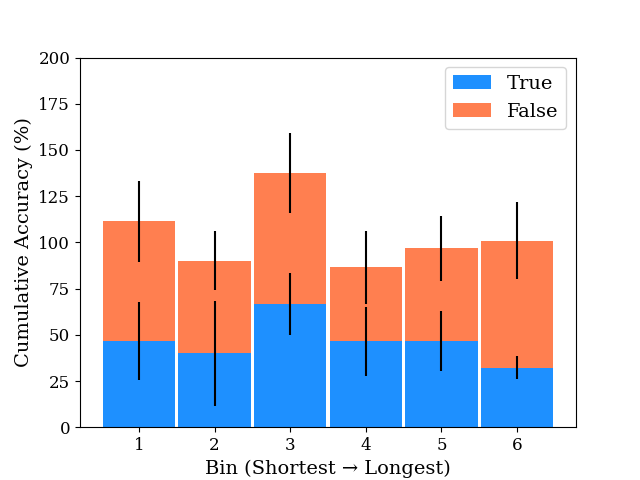}
    \caption{WNLI}
\end{subfigure}
\begin{subfigure}{\linewidth}
    \centering
    \includegraphics[width=0.19\textwidth]            {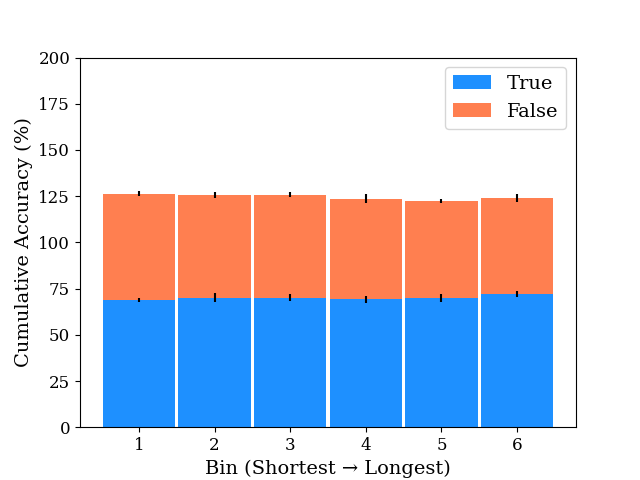}
    \includegraphics[width=0.19\textwidth]            {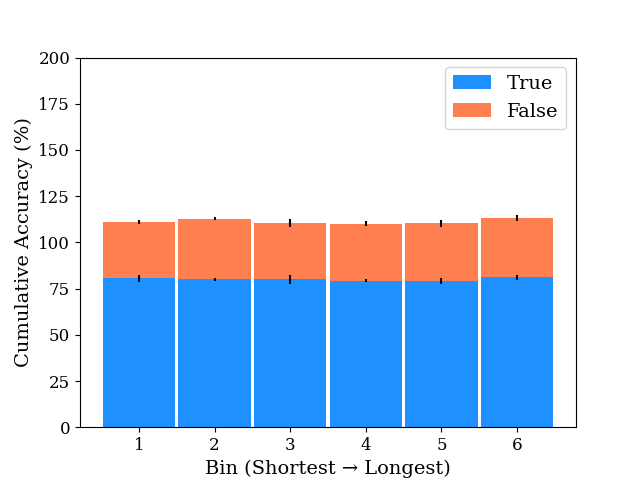}
    \includegraphics[width=0.19\textwidth]            {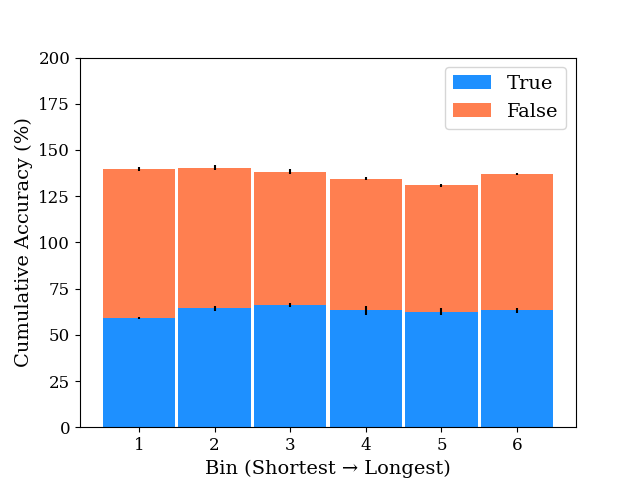}
    \includegraphics[width=0.19\textwidth]            {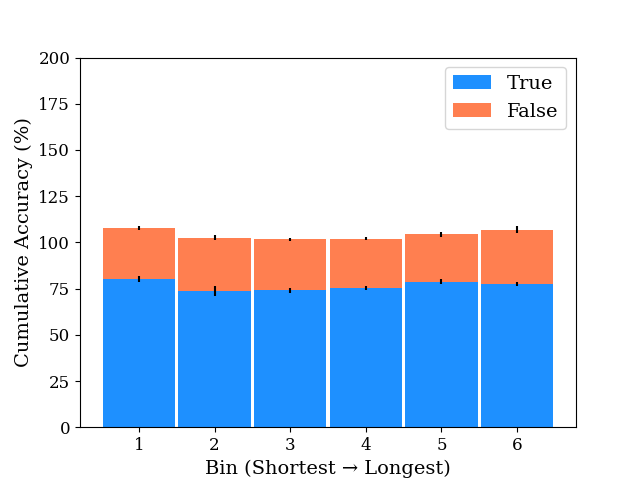}
    \includegraphics[width=0.19\textwidth]            {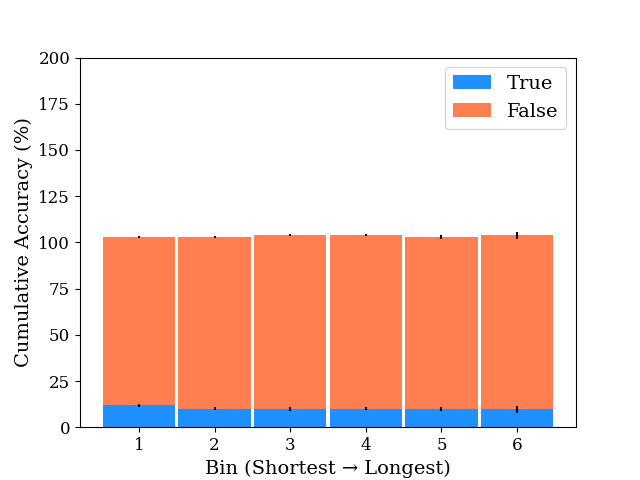}
    \caption{QNLI}
\end{subfigure}
\begin{subfigure}{\linewidth}
    \centering
    \includegraphics[width=0.19\textwidth]            {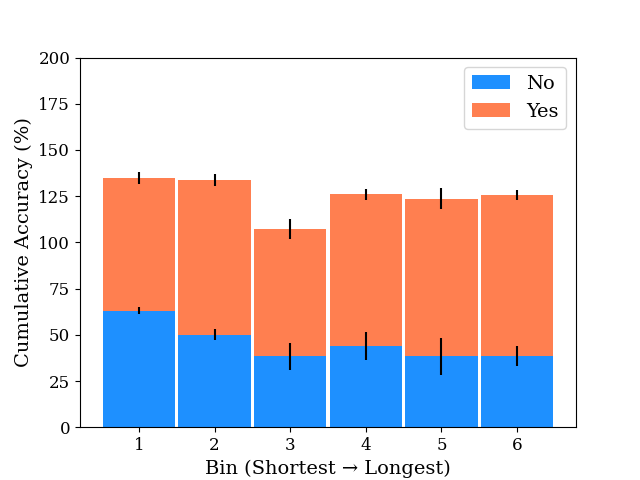}
    \includegraphics[width=0.19\textwidth]            {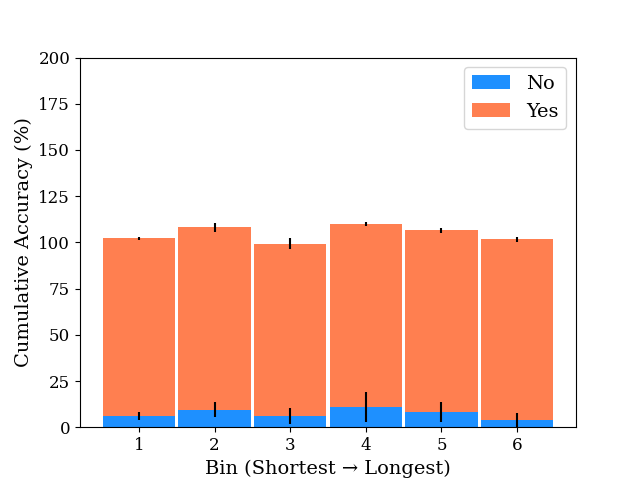}
    \includegraphics[width=0.19\textwidth]            {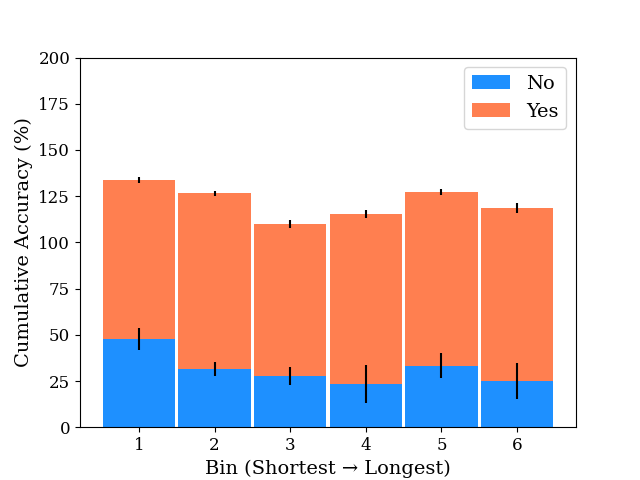}
    \includegraphics[width=0.19\textwidth]            {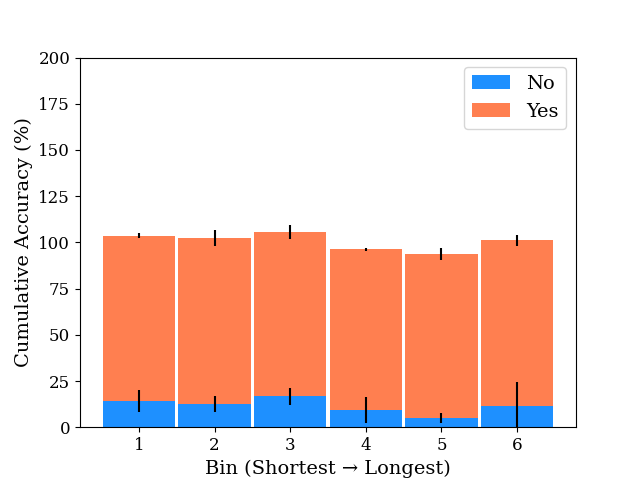}
    \includegraphics[width=0.19\textwidth]            {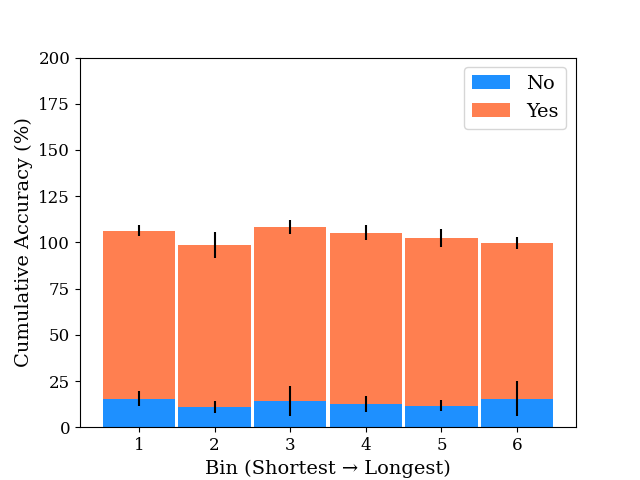}
    \caption{MRPC}
\end{subfigure}
\begin{subfigure}{\linewidth}
    \centering
    \includegraphics[width=0.19\textwidth]            {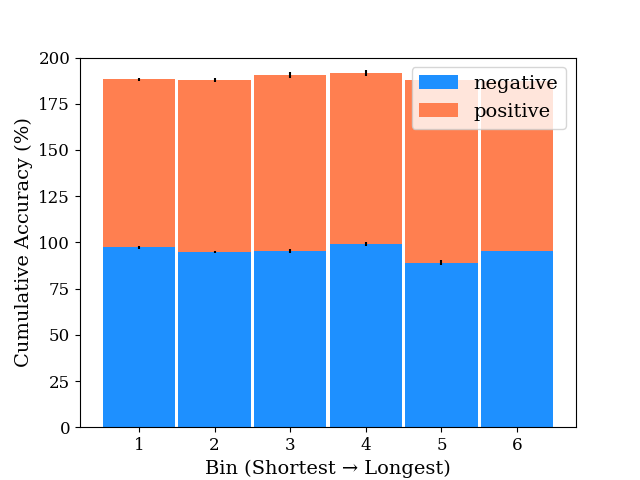}
    \includegraphics[width=0.19\textwidth]            {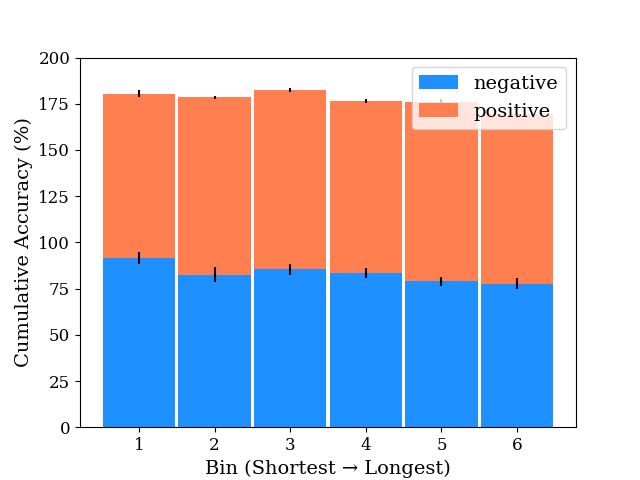}
    \includegraphics[width=0.19\textwidth]            {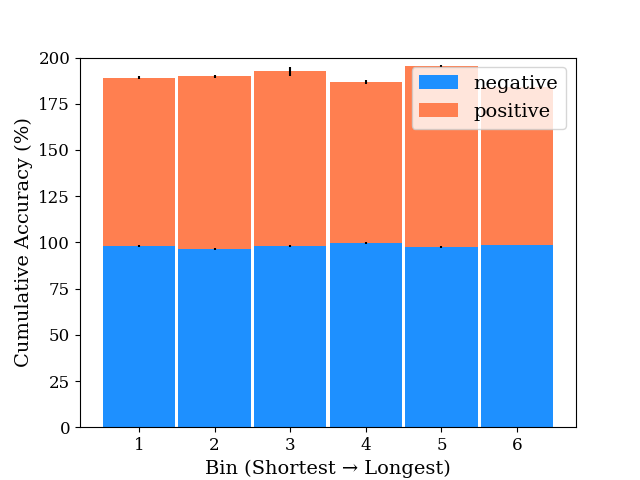}
    \includegraphics[width=0.19\textwidth]            {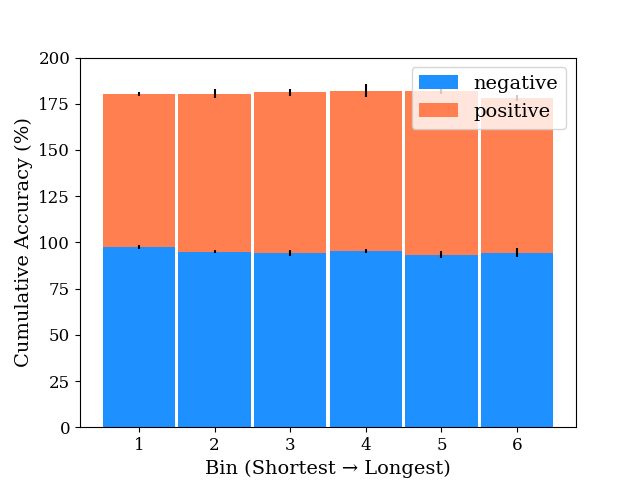}
    \includegraphics[width=0.19\textwidth]            {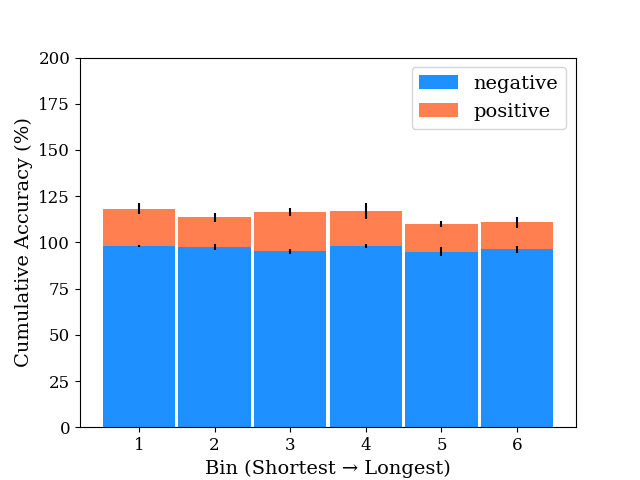}
    \caption{SST-2}
\end{subfigure}
\end{minipage}
\hfill
\begin{minipage}[c]{\linewidth}
    \caption{ICL performance of Llama3 8B, Llama2 7B, Mistral 7B, OPT 6.7B, and GPT Neo 2.7B (from left to right) where $y_1$ (Blue) and $y_2$ (Orange) are both randomly sampled.}
\end{minipage}
\end{figure*}

\begin{figure*}[t!]
\centering
\begin{minipage}[t]{\linewidth}
\begin{subfigure}{\linewidth}
    \centering
    \includegraphics[width=0.19\textwidth]            {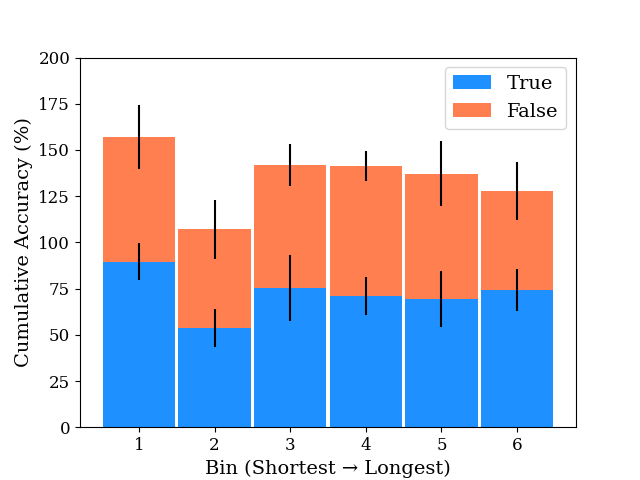}
    \includegraphics[width=0.19\textwidth]            {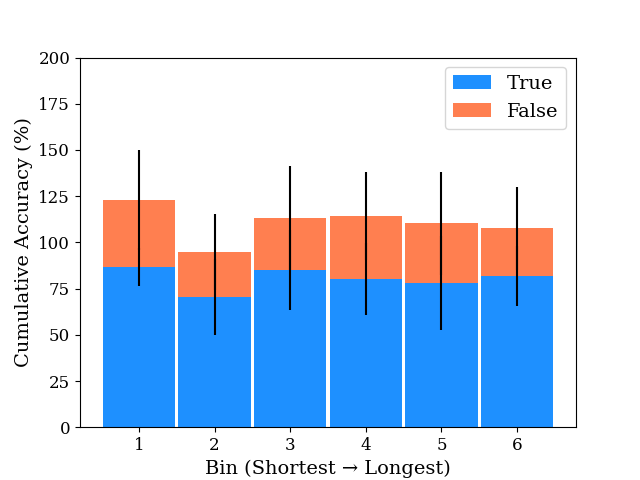}
    \includegraphics[width=0.19\textwidth]            {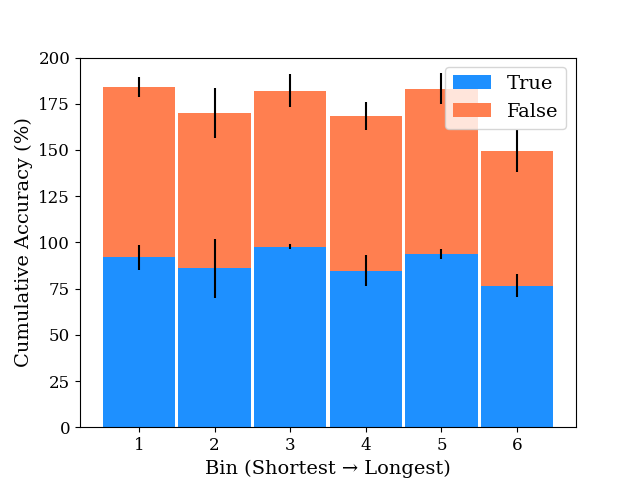}
    \includegraphics[width=0.19\textwidth]            {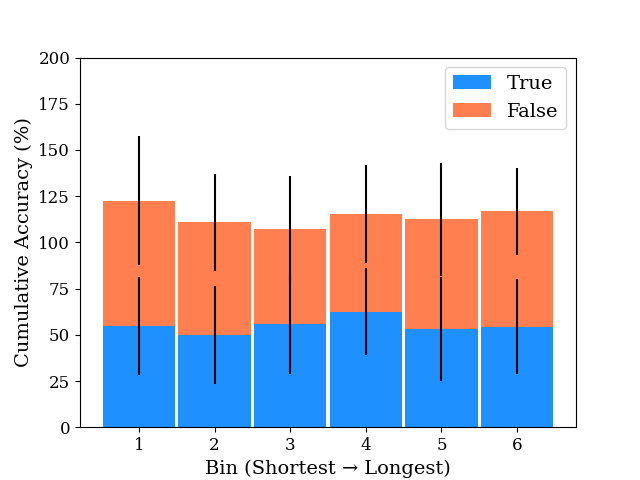}
    \includegraphics[width=0.19\textwidth]            {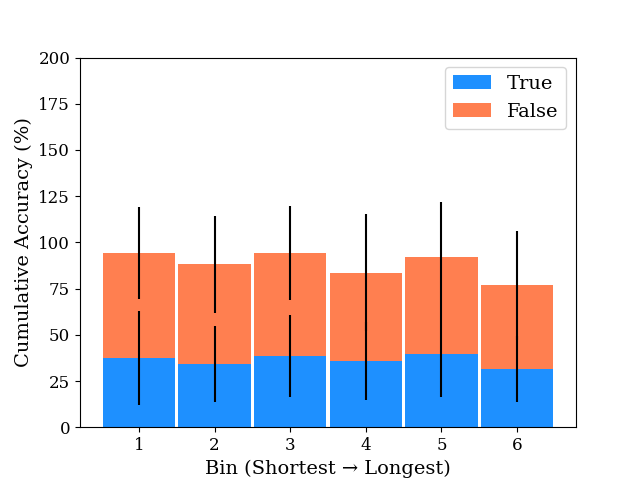}
    \caption{Hans}
\end{subfigure}
\begin{subfigure}{\linewidth}
    \centering
    \includegraphics[width=0.19\textwidth]            {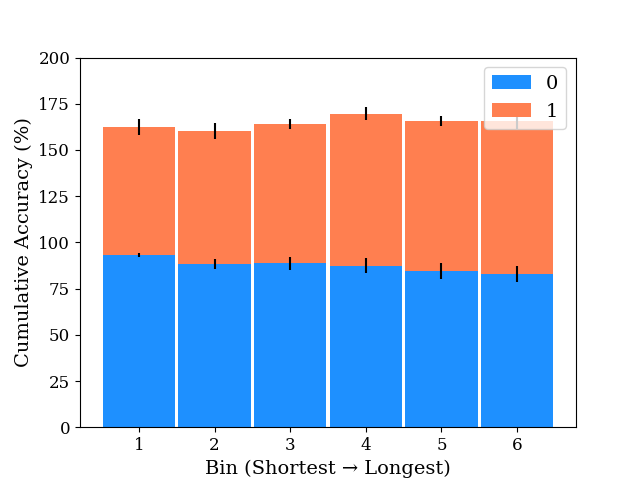}
    \includegraphics[width=0.19\textwidth]            {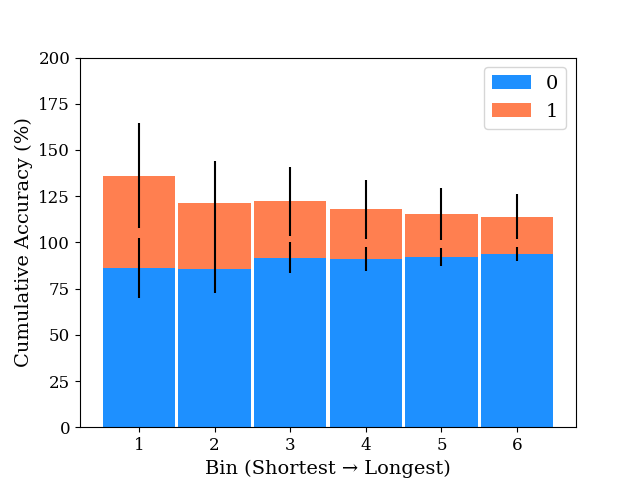}
    \includegraphics[width=0.19\textwidth]            {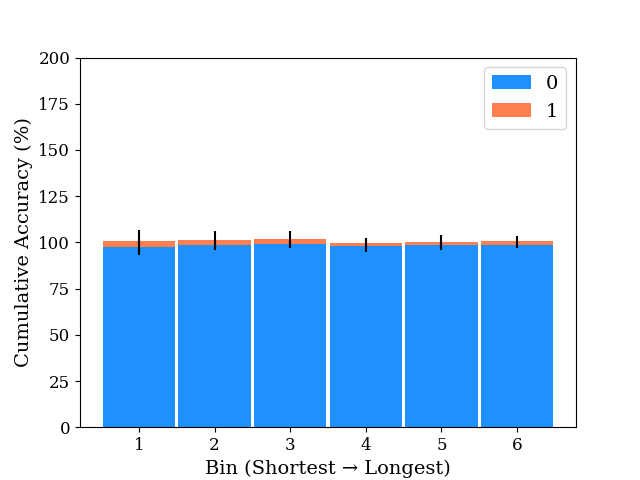}
    \includegraphics[width=0.19\textwidth]            {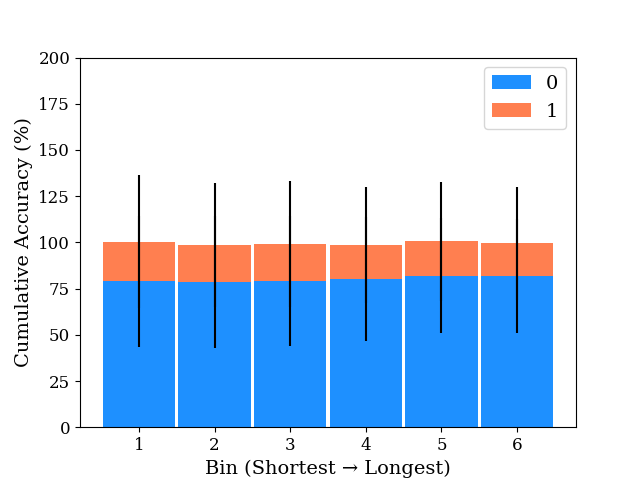}
    \includegraphics[width=0.19\textwidth]            {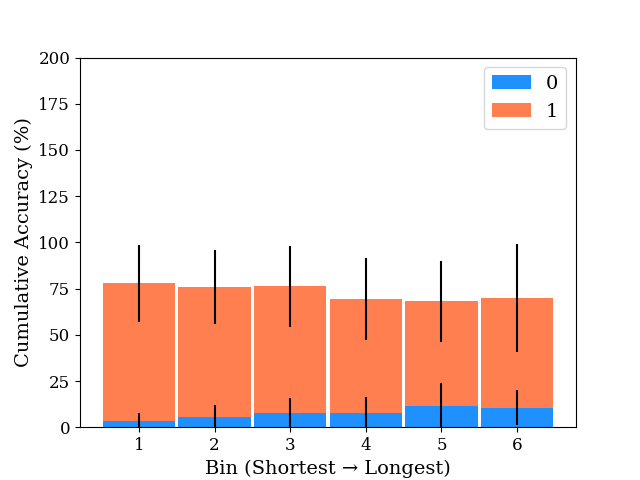}
    \caption{PAWS-X$_{\textsc{EN}}$}
\end{subfigure}
\begin{subfigure}{\linewidth}
    \centering
    \includegraphics[width=0.19\textwidth]            {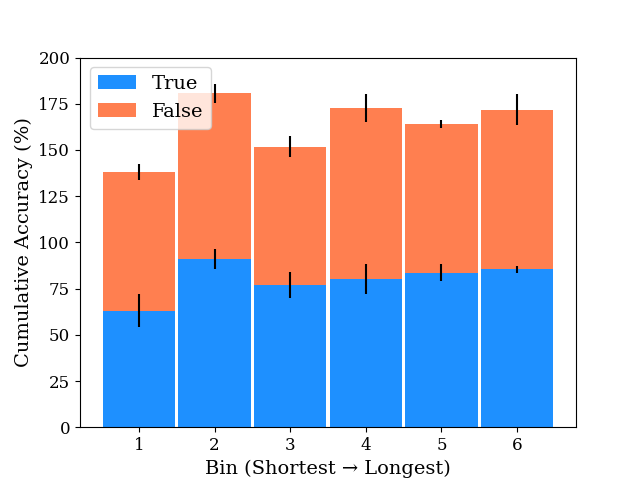}
    \includegraphics[width=0.19\textwidth]            {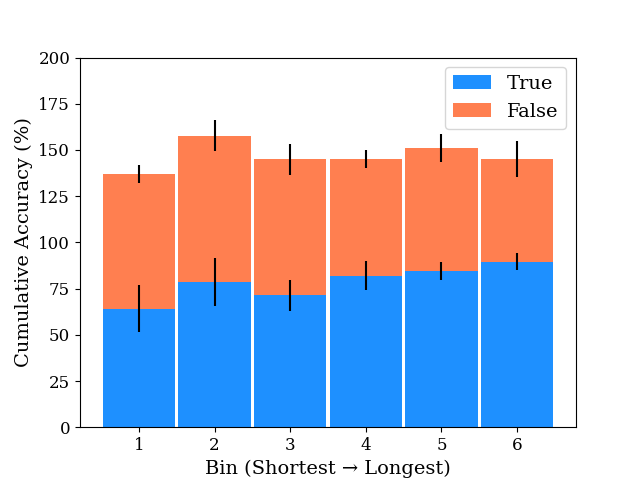}
    \includegraphics[width=0.19\textwidth]            {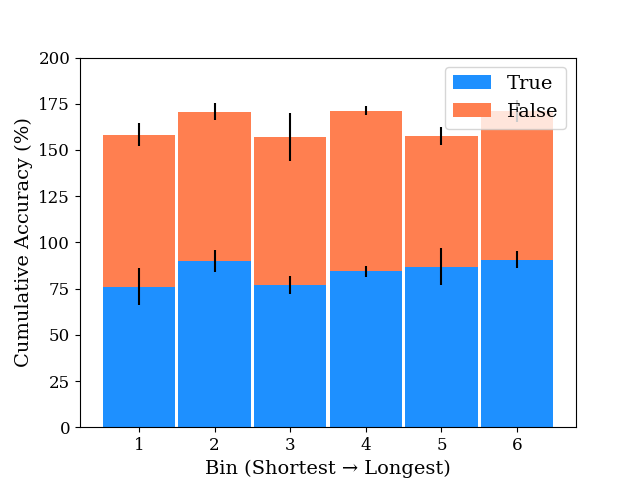}
    \includegraphics[width=0.19\textwidth]            {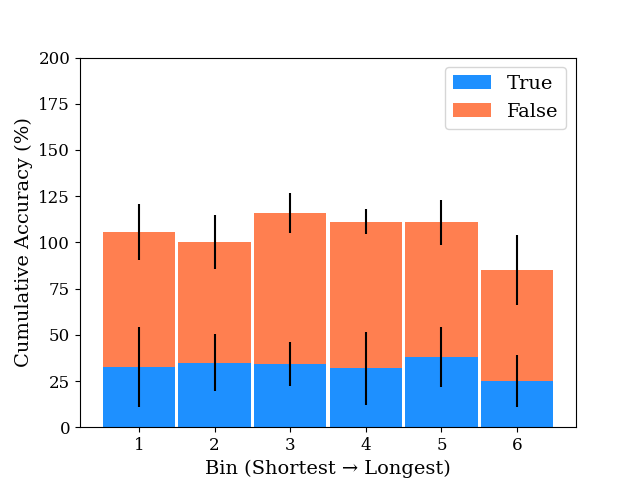}
    \includegraphics[width=0.19\textwidth]            {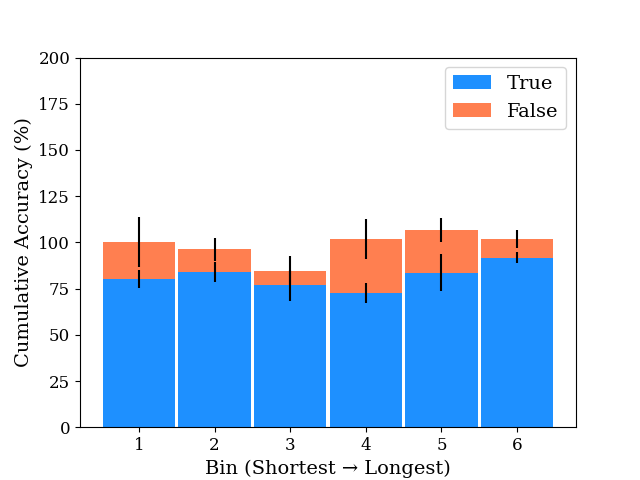}
    \caption{RTE}
\end{subfigure}
\begin{subfigure}{\linewidth}
    \centering
    \includegraphics[width=0.19\textwidth]            {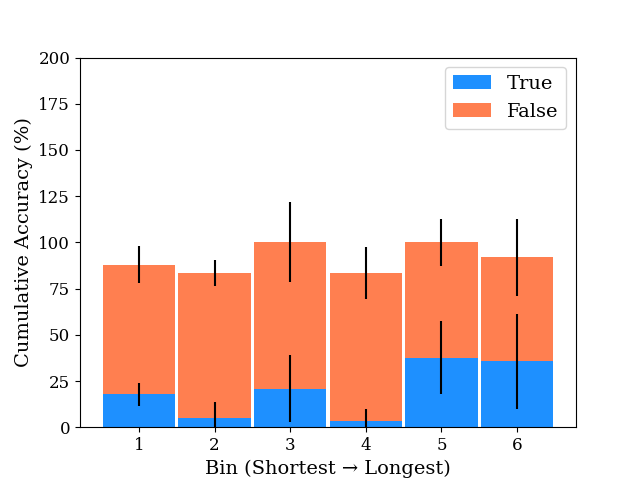}
    \includegraphics[width=0.19\textwidth]            {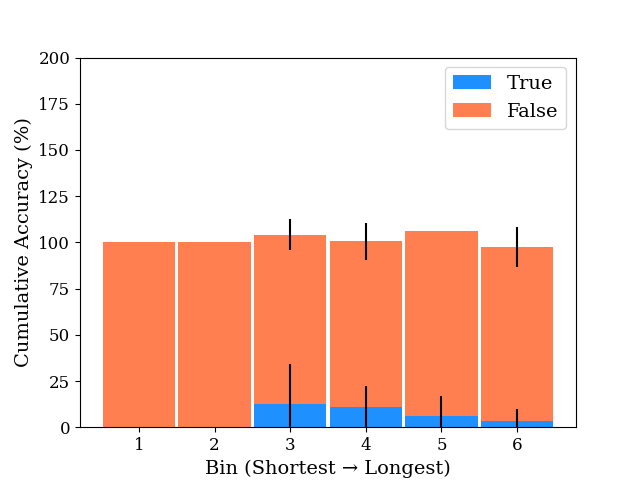}
    \includegraphics[width=0.19\textwidth]            {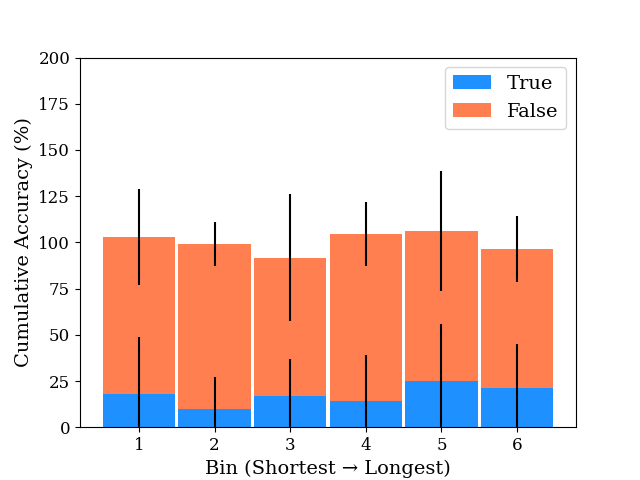}
    \includegraphics[width=0.19\textwidth]            {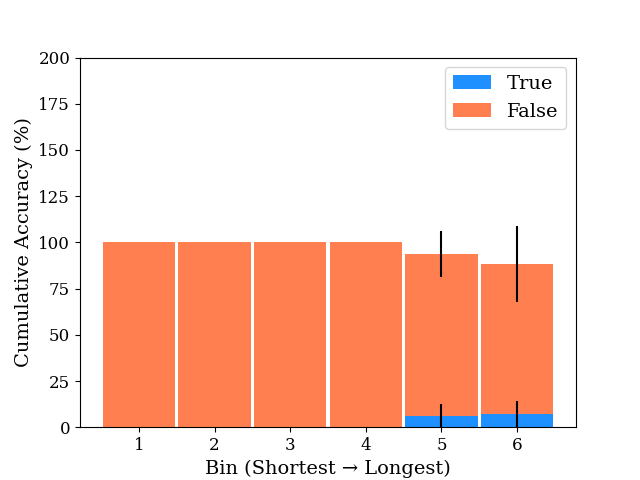}
    \includegraphics[width=0.19\textwidth]            {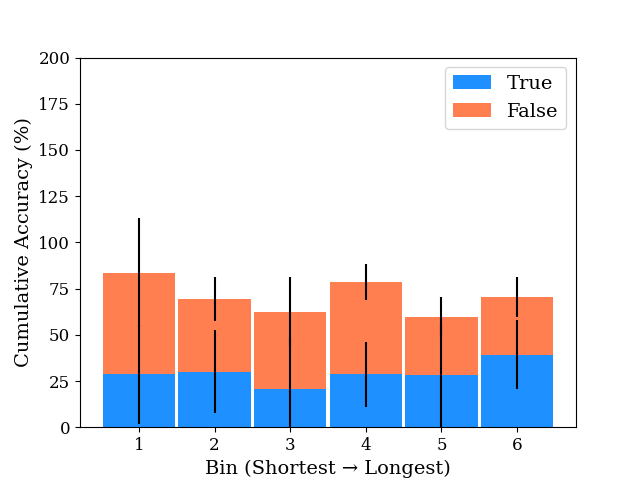}
    \caption{WNLI}
\end{subfigure}
\begin{subfigure}{\linewidth}
    \centering
    \includegraphics[width=0.19\textwidth]            {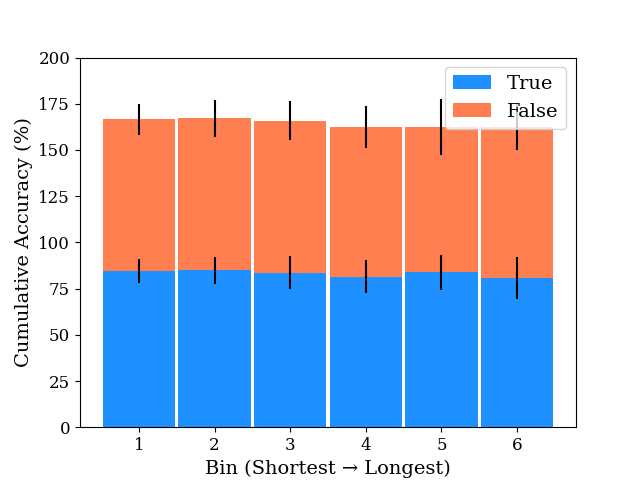}
    \includegraphics[width=0.19\textwidth]            {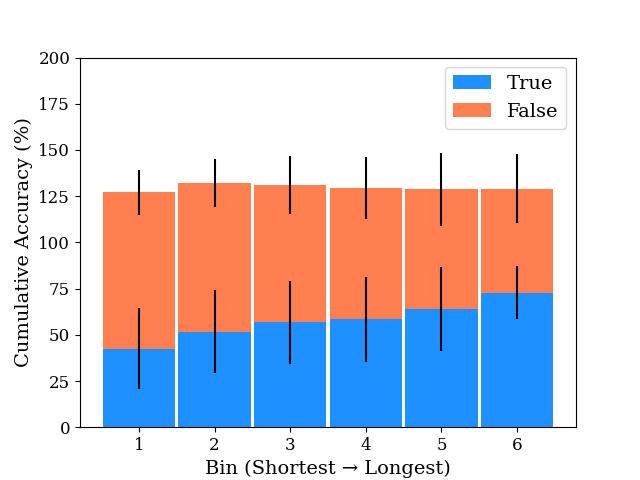}
    \includegraphics[width=0.19\textwidth]            {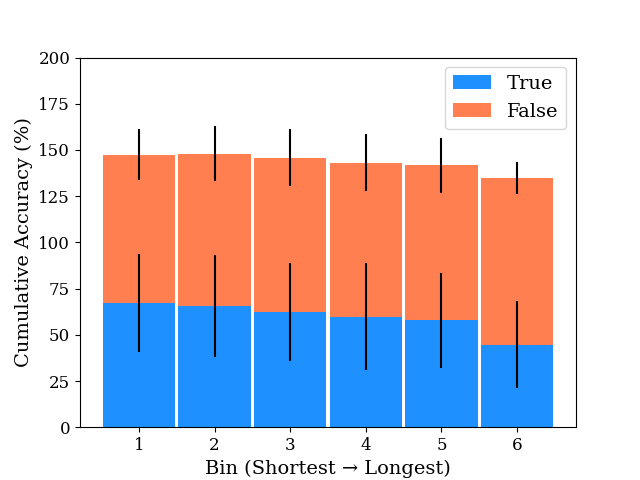}
    \includegraphics[width=0.19\textwidth]            {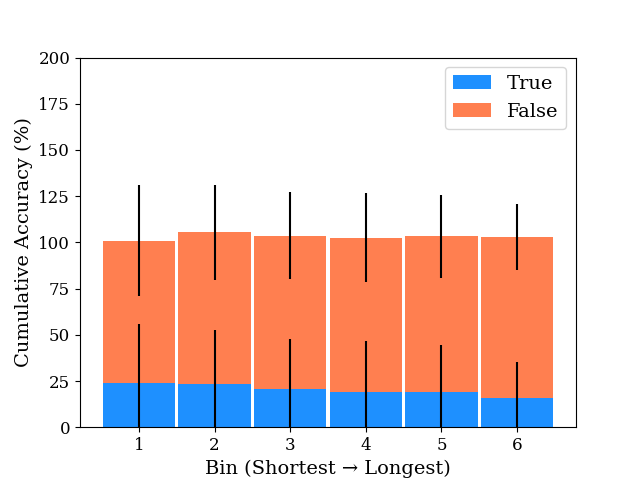}
    \includegraphics[width=0.19\textwidth]            {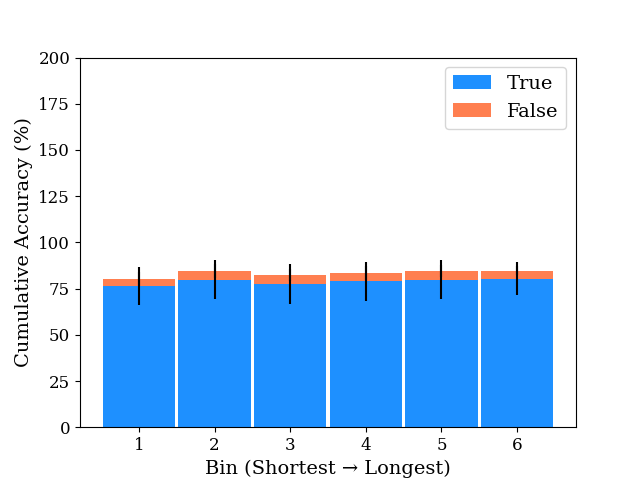}
    \caption{QNLI}
\end{subfigure}
\begin{subfigure}{\linewidth}
    \centering
    \includegraphics[width=0.19\textwidth]            {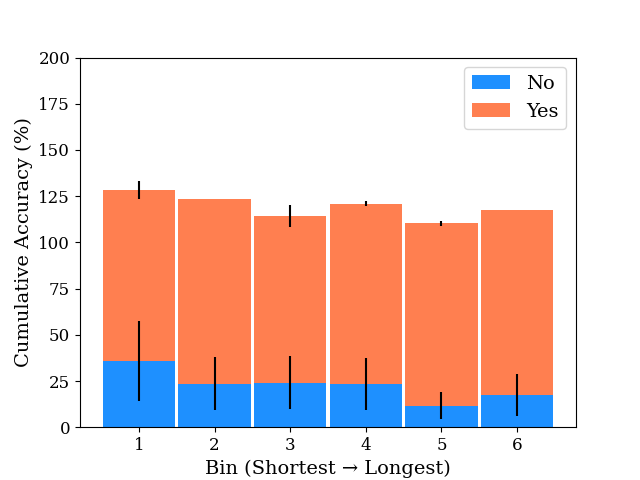}
    \includegraphics[width=0.19\textwidth]            {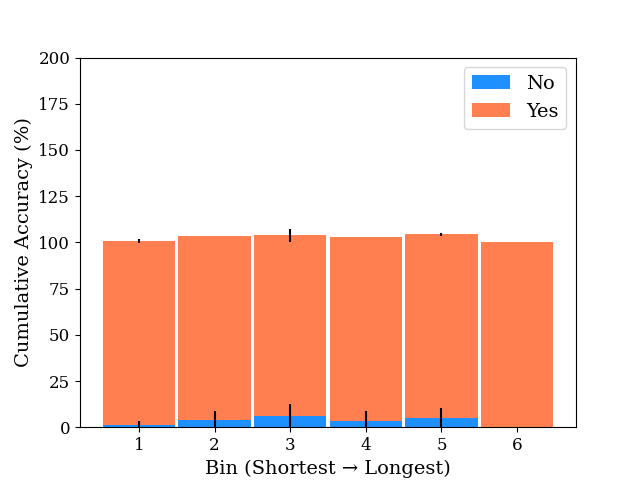}
    \includegraphics[width=0.19\textwidth]            {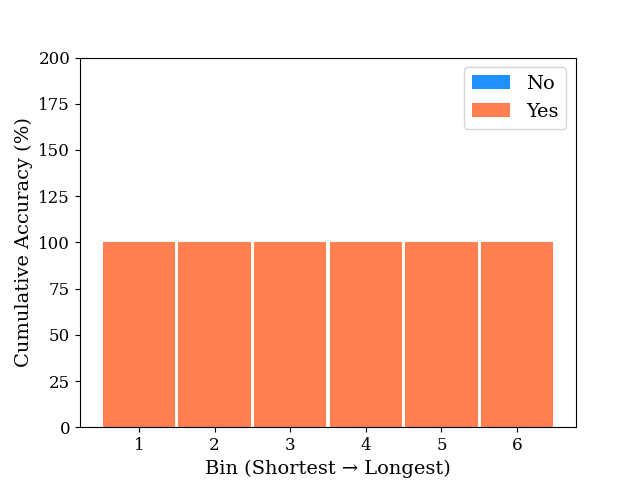}
    \includegraphics[width=0.19\textwidth]            {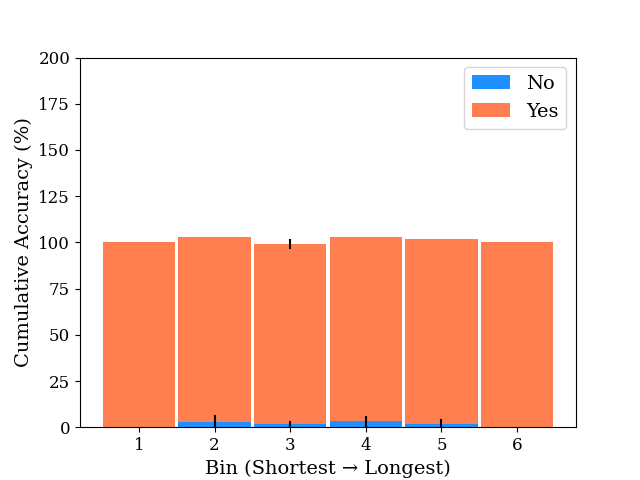}
    \includegraphics[width=0.19\textwidth]            {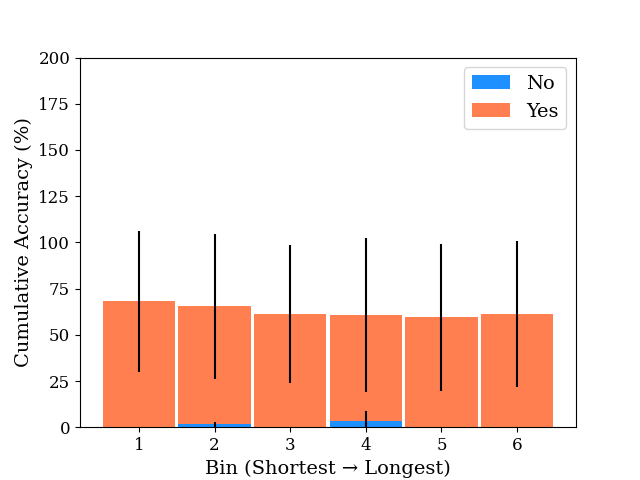}
    \caption{MRPC}
\end{subfigure}
\begin{subfigure}{\linewidth}
    \centering
    \includegraphics[width=0.19\textwidth]            {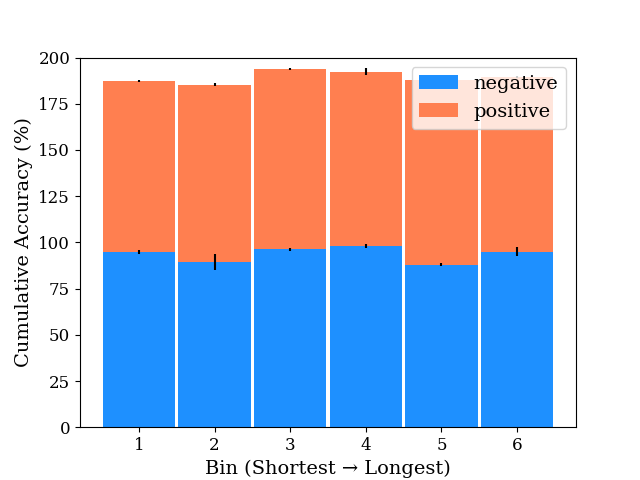}
    \includegraphics[width=0.19\textwidth]            {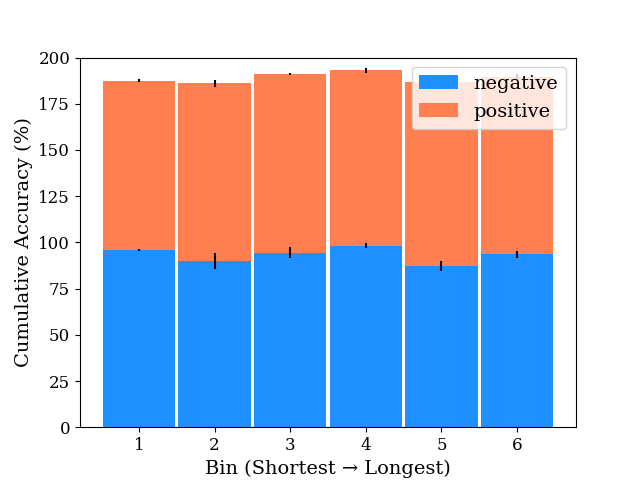}
    \includegraphics[width=0.19\textwidth]            {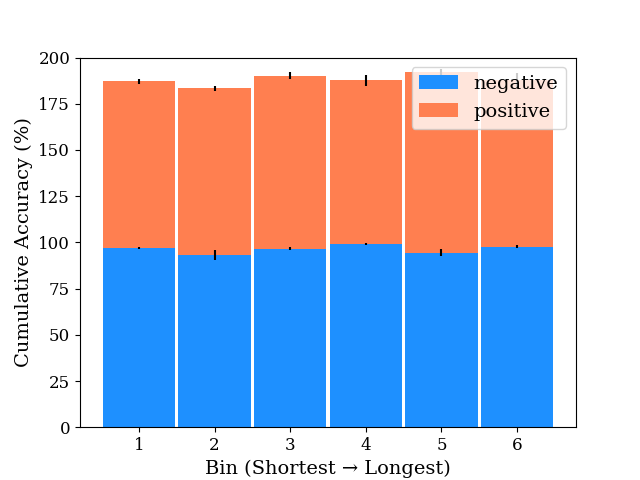}
    \includegraphics[width=0.19\textwidth]            {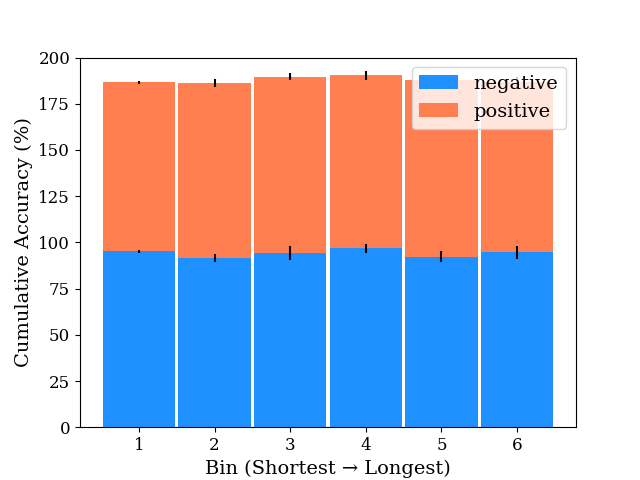}
    \includegraphics[width=0.19\textwidth]            {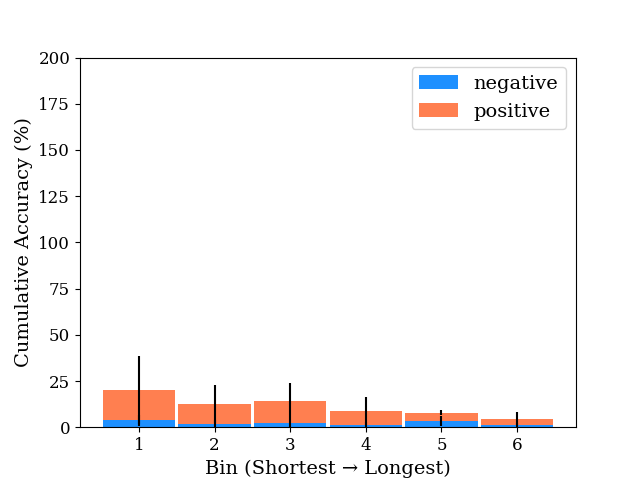}
    \caption{SST-2}
\end{subfigure}
\end{minipage}
\hfill
\begin{minipage}[c]{\linewidth}
    \caption{Finetuning performance of Llama3 8B, Llama2 7B, Mistral 7B, OPT 6.7B, and GPT Neo 2.7B (from left to right) where $y_1$ (Blue) and $y_2$ (Orange) are both randomly sampled.}
\end{minipage}
\end{figure*}

\clearpage
\subsection{Additional Model Parameter (OPT) Results}\label{app:opt-results}
Each of the following figures shows validation performance when varying the number of model parameters using the OPT model family. Bin 0 contains the shortest demonstrations and Bin 5 contains the longest demonstrations. Each subfigure shows the validation accuracy on a single class when in-context instances belonging to the respective class were sampled from long instances, short instances, and randomly sampled (left to right).
\begin{figure}[h!]
\vspace{-2.5mm}
\centering
\begin{minipage}[t]{\linewidth}
\begin{subfigure}{\linewidth}
    \centering
    \includegraphics[width=0.32\textwidth]            {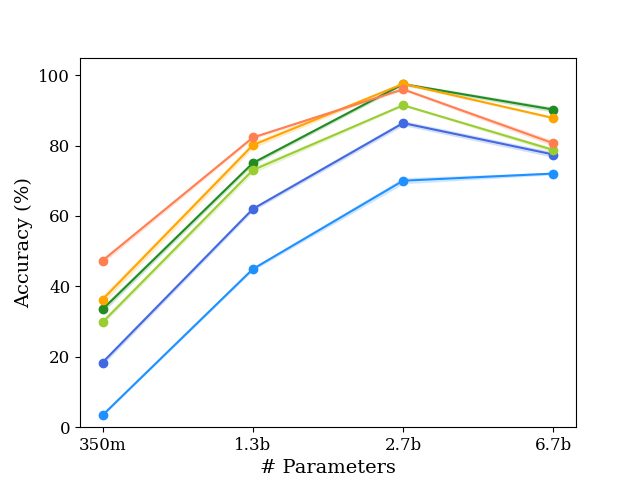}
    \includegraphics[width=0.32\textwidth]            {latex/figures/lineplots/parameters/hans_opt_class2_cls1.png}
    \includegraphics[width=0.32\textwidth]            {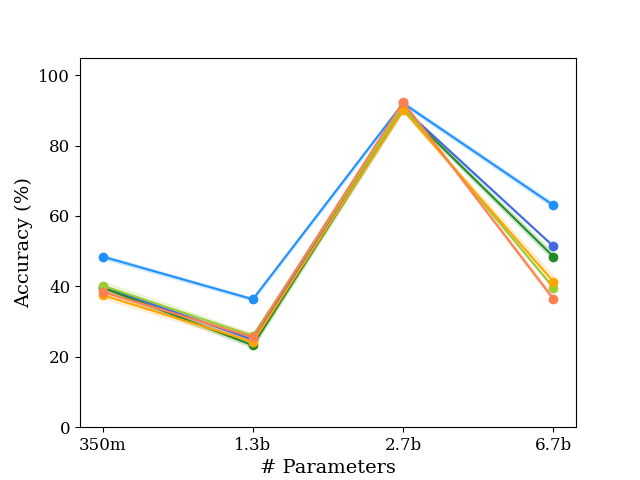}
    \caption{$y_1$}
\end{subfigure}
    \vspace{-2.5mm}
\begin{subfigure}{\linewidth}
    \centering
    \includegraphics[width=0.32\textwidth]            {latex/figures/lineplots/parameters/hans_opt_class2_cls2.png}
    \includegraphics[width=0.32\textwidth]            {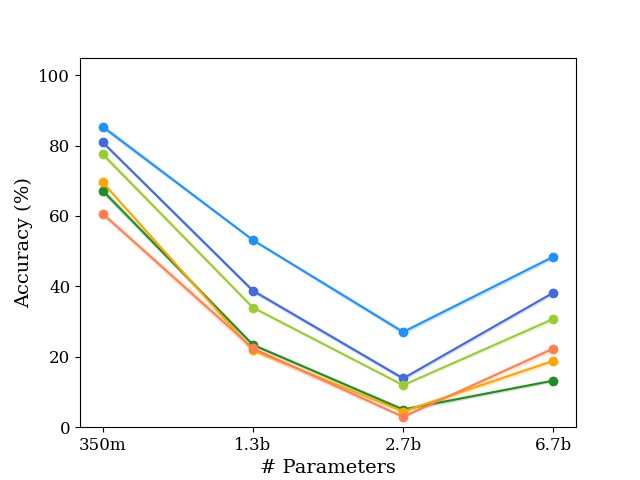}
    \includegraphics[width=0.32\textwidth]            {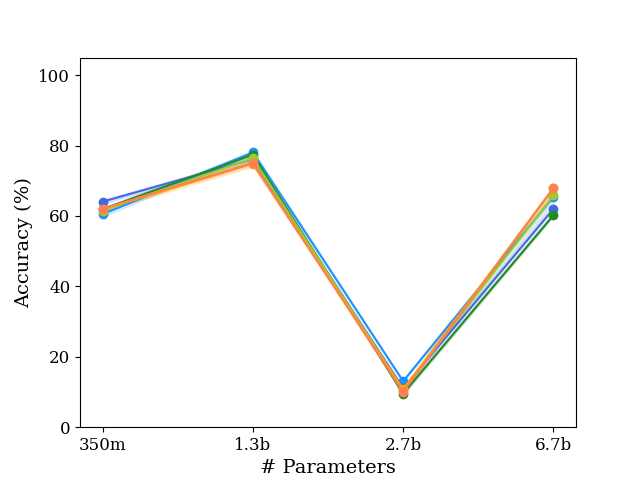}
    \caption{$y_2$}
\end{subfigure}
\end{minipage}
    \hfill
    \begin{minipage}[c]{\linewidth}
        \centering
            \begin{subfigure}{0.4 \linewidth}
            \centering
                        \vspace{2mm}
            \includegraphics[width=\textwidth]{latex/figures/lineplots/legend.pdf}
        \end{subfigure}%
    \end{minipage}
    \hfill
    \vspace{-2.5mm}
\begin{minipage}[c]{\linewidth}
    \caption{Hans dataset (OPT)}
\end{minipage}
\end{figure}
\vspace{-5mm}

\begin{figure}[h!]
\vspace{-2.5mm}
\centering
\begin{minipage}[t]{\linewidth}
\begin{subfigure}{\linewidth}
    \centering
    \includegraphics[width=0.32\textwidth]            {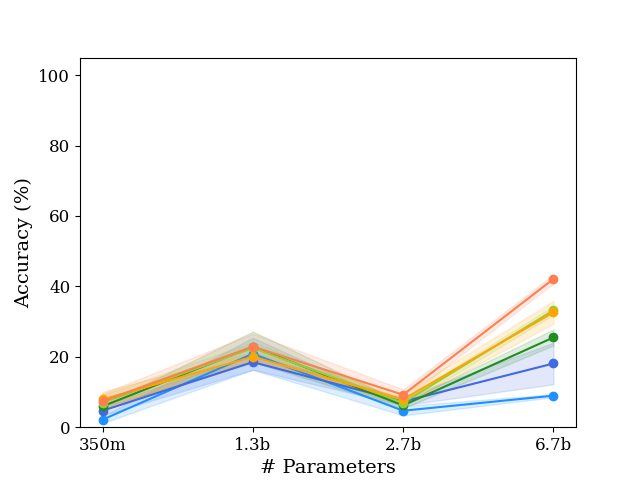}
    \includegraphics[width=0.32\textwidth]            {latex/figures/lineplots/parameters/en_opt_class2_cls1.png}
    \includegraphics[width=0.32\textwidth]            {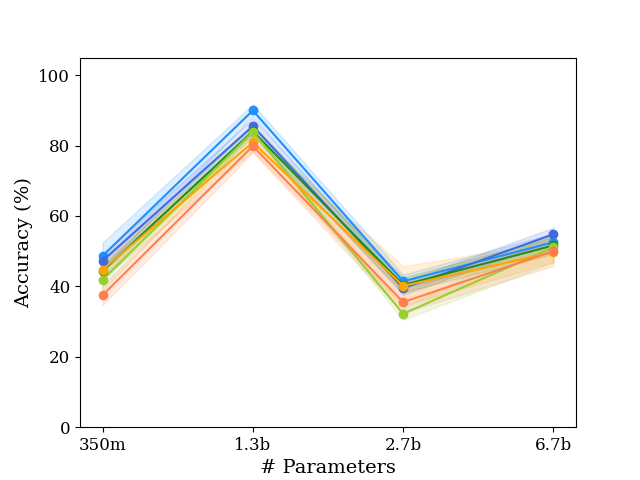}
    \caption{$y_1$}
\end{subfigure}
    \vspace{-2.5mm}
\begin{subfigure}{\linewidth}
    \centering
    \includegraphics[width=0.32\textwidth]            {latex/figures/lineplots/parameters/en_opt_class2_cls2.png}
    \includegraphics[width=0.32\textwidth]            {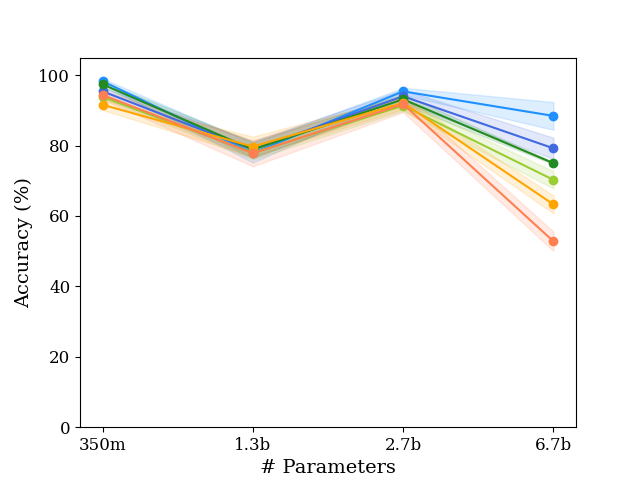}
    \includegraphics[width=0.32\textwidth]            {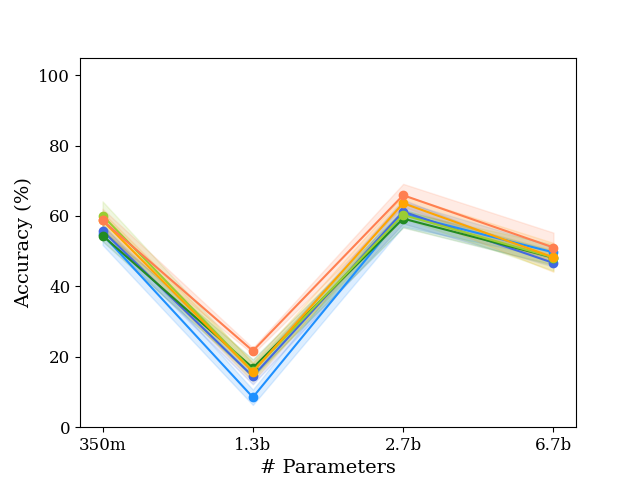}
    \caption{$y_2$}
\end{subfigure}
\end{minipage}
    \hfill
    \begin{minipage}[c]{\linewidth}
        \centering
            \begin{subfigure}{0.4 \linewidth}
            \centering
                        \vspace{2mm}
            \includegraphics[width=\textwidth]{latex/figures/lineplots/legend.pdf}
        \end{subfigure}%
    \end{minipage}
    \hfill
    \vspace{-2.5mm}
\begin{minipage}[c]{\linewidth}
    \caption{PAWS-X$_{\textsc{EN}}$ dataset (OPT)}
\end{minipage}
\end{figure}
\vspace{-5mm}

\begin{figure}[h!]
\vspace{-2.5mm}
\centering
\begin{minipage}[t]{\linewidth}
\begin{subfigure}{\linewidth}
    \centering
    \includegraphics[width=0.32\textwidth]            {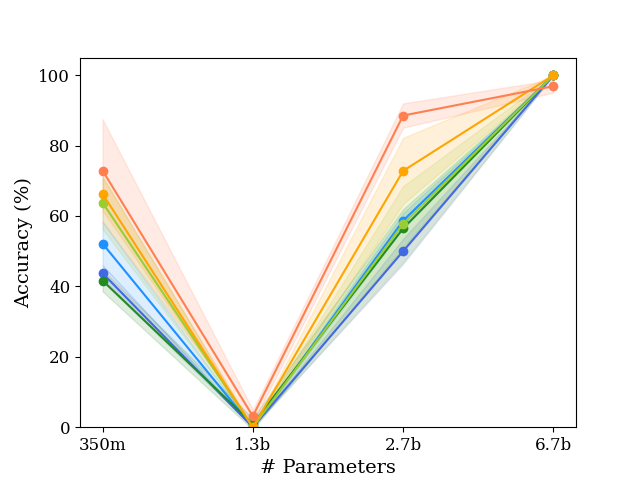}
    \includegraphics[width=0.32\textwidth]            {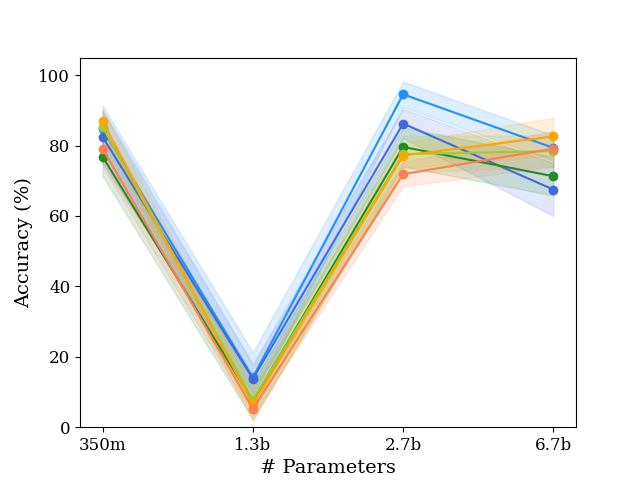}
    \includegraphics[width=0.32\textwidth]            {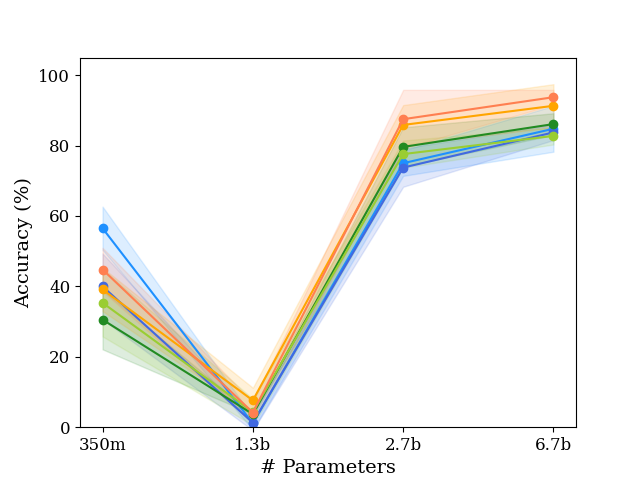}
    \caption{$y_1$}
\end{subfigure}
    \vspace{-2.5mm}
\begin{subfigure}{\linewidth}
    \centering
    \includegraphics[width=0.32\textwidth]            {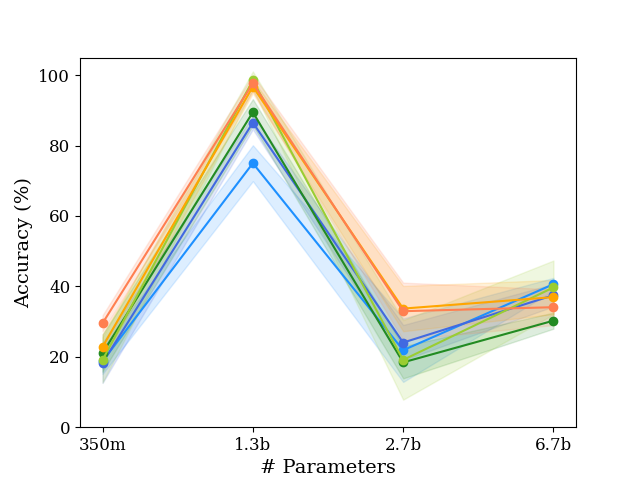}
    \includegraphics[width=0.32\textwidth]            {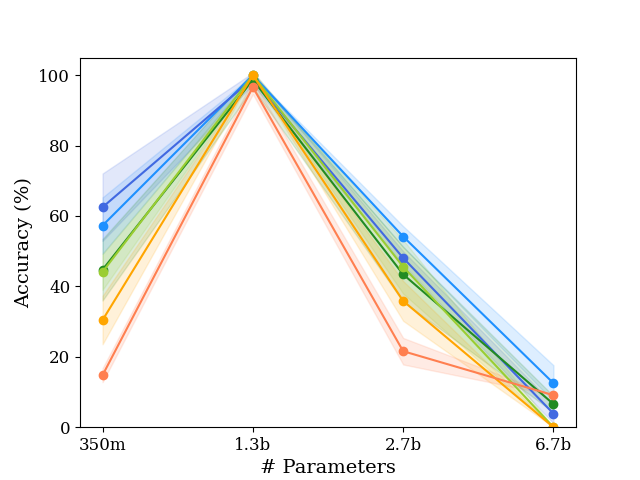}
    \includegraphics[width=0.32\textwidth]            {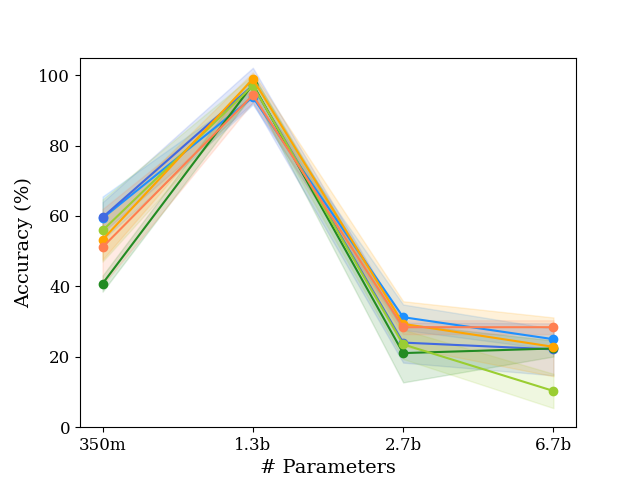}
    \caption{$y_2$}
\end{subfigure}
\end{minipage}
    \hfill
    \begin{minipage}[c]{\linewidth}
        \centering
            \begin{subfigure}{0.4 \linewidth}
            \centering
                        \vspace{2mm}
            \includegraphics[width=\textwidth]{latex/figures/lineplots/legend.pdf}
        \end{subfigure}%
    \end{minipage}
    \hfill
    \vspace{-2.5mm}
\begin{minipage}[c]{\linewidth}
    \caption{RTE dataset (OPT)}
\end{minipage}
\end{figure}
\vspace{-7mm}

\begin{figure}[h!]
\vspace{-2.5mm}
\centering
\begin{minipage}[t]{\linewidth}
\begin{subfigure}{\linewidth}
    \centering
    \includegraphics[width=0.32\textwidth]            {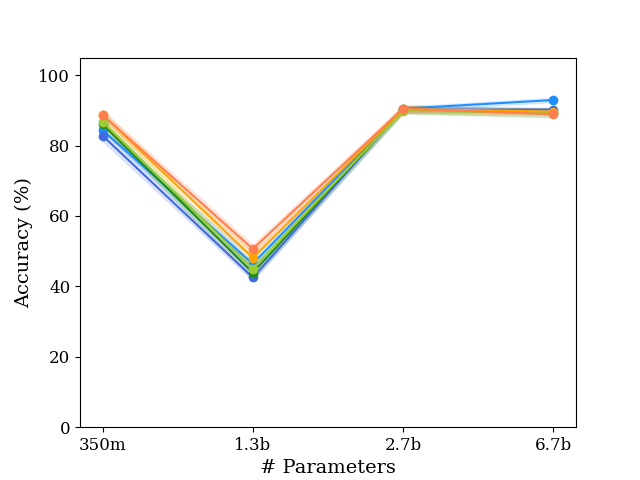}
    \includegraphics[width=0.32\textwidth]            {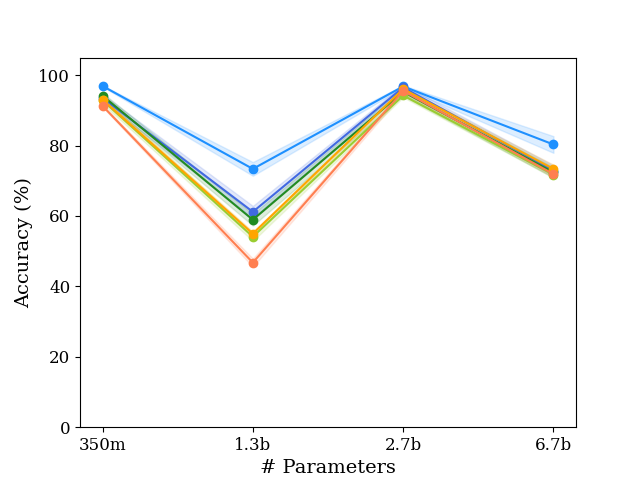}
    \includegraphics[width=0.32\textwidth]            {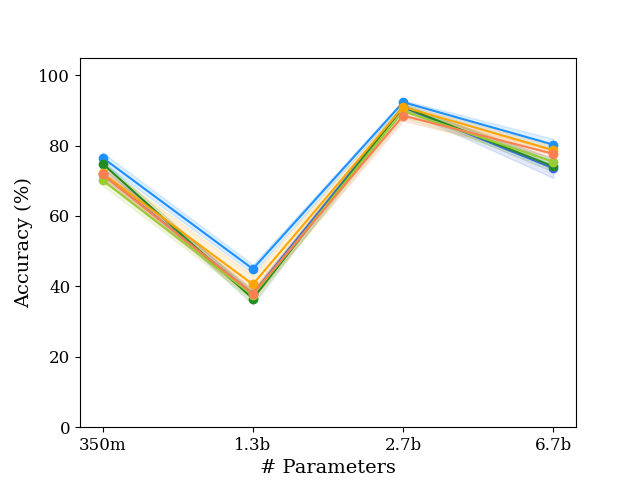}
    \caption{$y_1$}
\end{subfigure}
    \vspace{-2.5mm}
\begin{subfigure}{\linewidth}
    \centering
    \includegraphics[width=0.32\textwidth]            {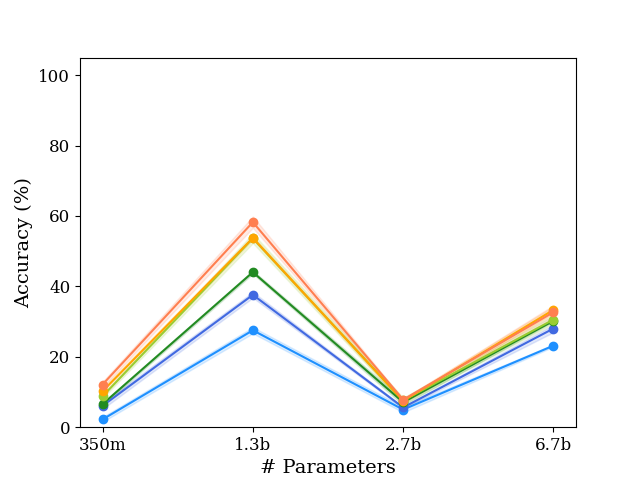}
    \includegraphics[width=0.32\textwidth]            {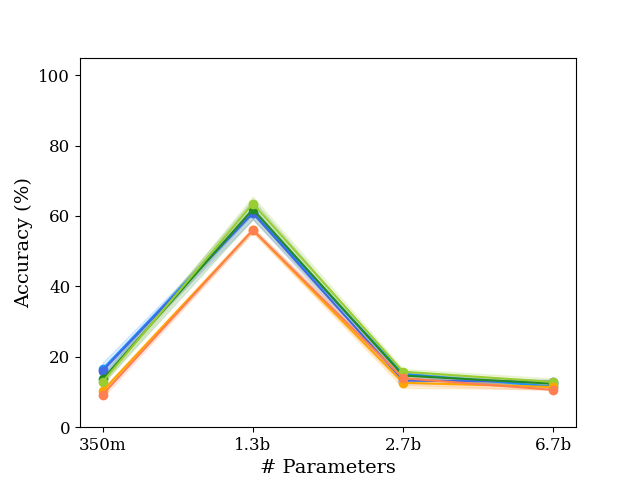}
    \includegraphics[width=0.32\textwidth]            {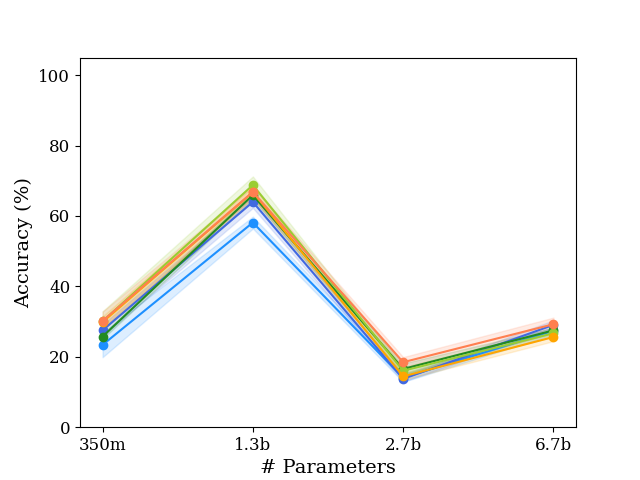}
    \caption{$y_2$}
\end{subfigure}
\end{minipage}
    \hfill
    \begin{minipage}[c]{\linewidth}
        \centering
            \begin{subfigure}{0.4 \linewidth}
            \centering
                        \vspace{2mm}
            \includegraphics[width=\textwidth]{latex/figures/lineplots/legend.pdf}
        \end{subfigure}%
    \end{minipage}
    \hfill
\vspace{-2.5mm}
\begin{minipage}[c]{\linewidth}
    \caption{QNLI dataset (OPT)}
\end{minipage}
\end{figure}
\vspace{-7mm}

\begin{figure}[h!]
\vspace{-2.5mm}
\centering
\begin{minipage}[t]{\linewidth}
\begin{subfigure}{\linewidth}
    \centering
    \includegraphics[width=0.32\textwidth]            {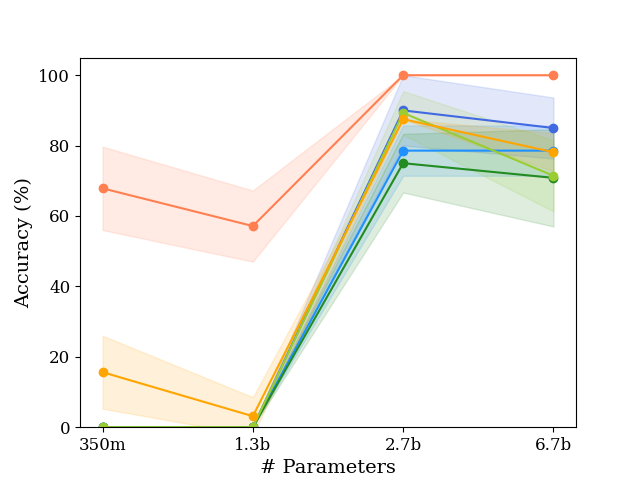}
    \includegraphics[width=0.32\textwidth]            {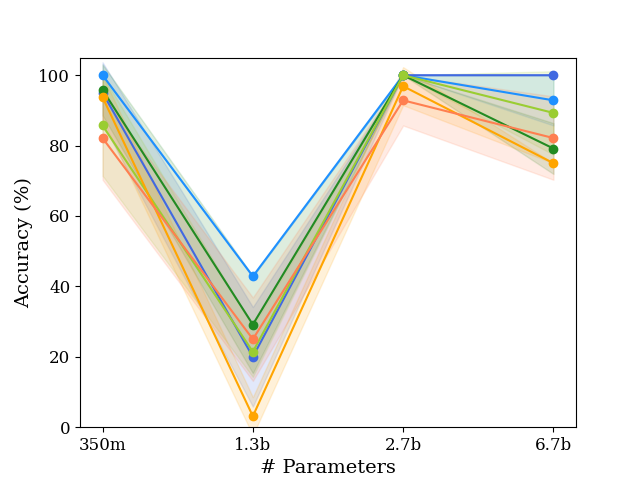}
    \includegraphics[width=0.32\textwidth]            {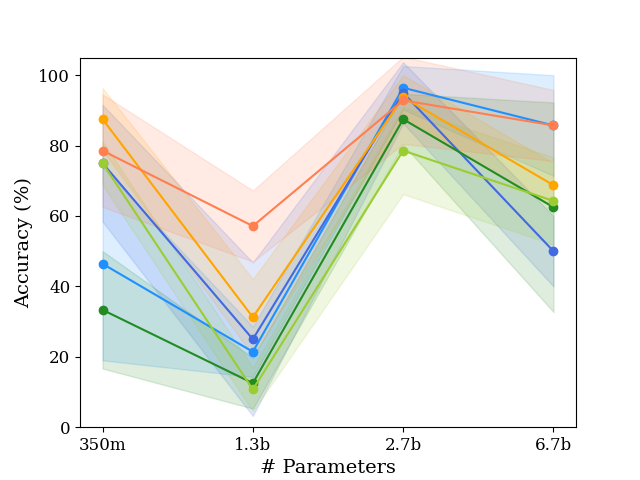}
    \caption{$y_1$}
\end{subfigure}
    \vspace{-2.5mm}
\begin{subfigure}{\linewidth}
    \centering
    \includegraphics[width=0.32\textwidth]            {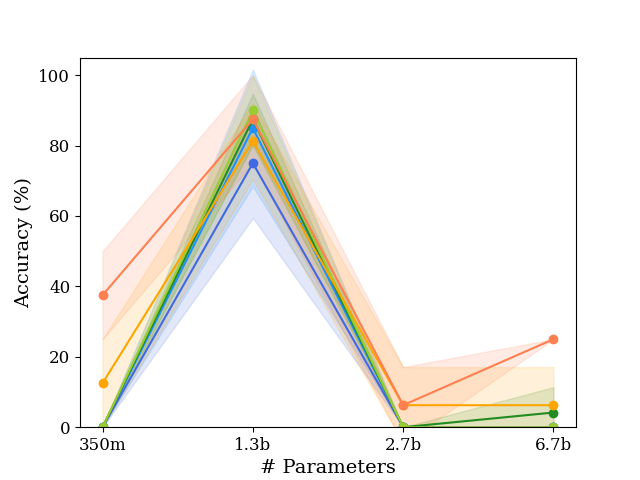}
    \includegraphics[width=0.32\textwidth]            {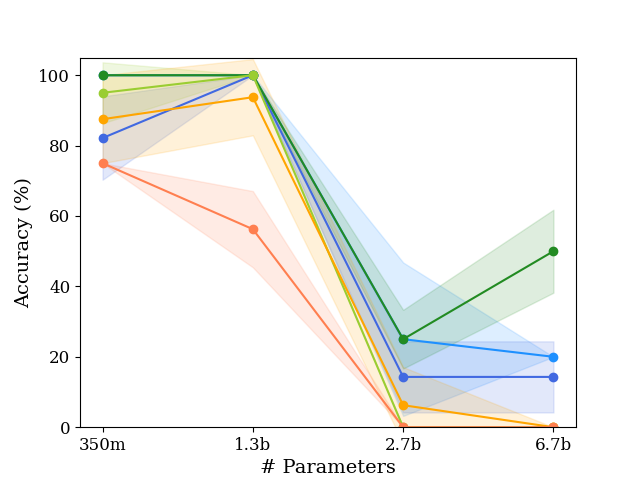}
    \includegraphics[width=0.32\textwidth]            {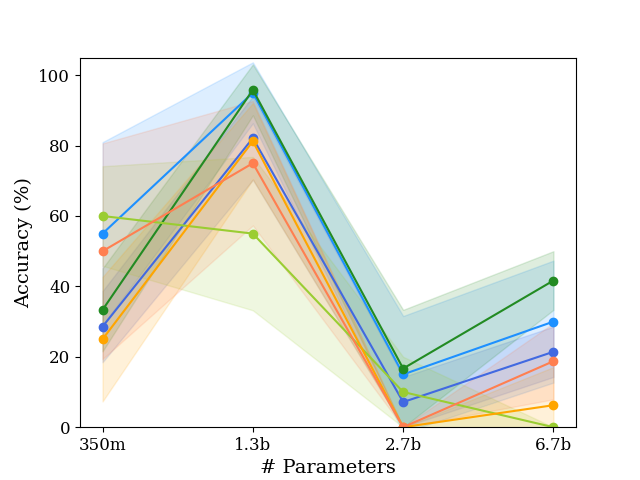}
    \caption{$y_2$}
\end{subfigure}
\end{minipage}
    \hfill
    \begin{minipage}[c]{\linewidth}
        \centering
            \begin{subfigure}{0.4 \linewidth}
            \centering
                        \vspace{2mm}
            \includegraphics[width=\textwidth]{latex/figures/lineplots/legend.pdf}
        \end{subfigure}%
    \end{minipage}
    \hfill
    \vspace{-2.5mm}
\begin{minipage}[c]{\linewidth}
    \caption{WNLI dataset (OPT)}
\end{minipage}
\end{figure}
\vspace{-7mm}

\begin{figure}[h!]
\vspace{-2.5mm}
\centering
\begin{minipage}[t]{\linewidth}
\begin{subfigure}{\linewidth}
    \centering
    \includegraphics[width=0.32\textwidth]            {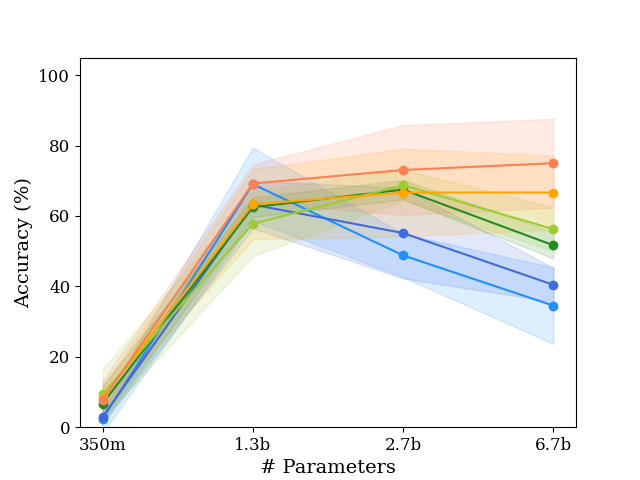}
    \includegraphics[width=0.32\textwidth]            {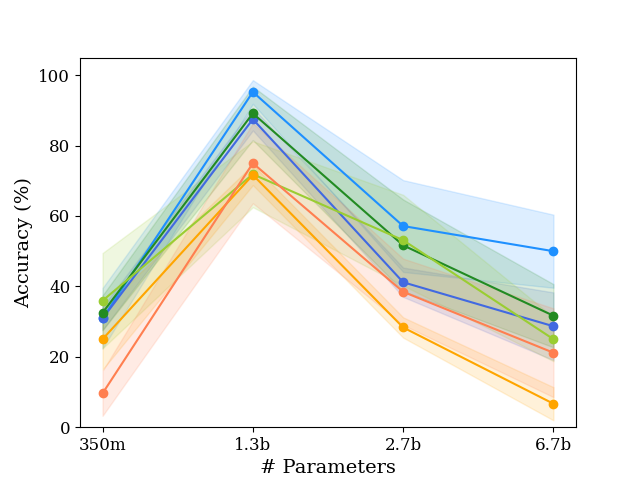}
    \includegraphics[width=0.32\textwidth]            {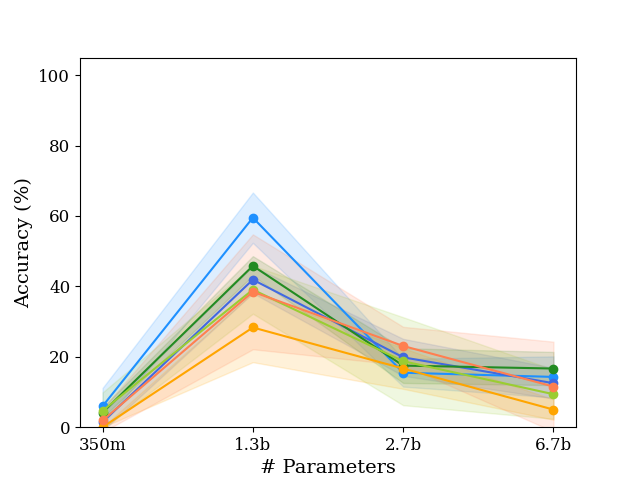}
    \caption{$y_1$}
\end{subfigure}
    \vspace{-2.5mm}
\begin{subfigure}{\linewidth}
    \centering
    \includegraphics[width=0.32\textwidth]            {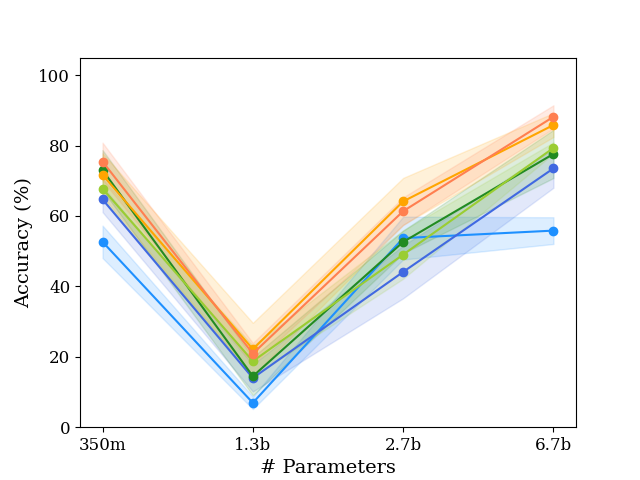}
    \includegraphics[width=0.32\textwidth]            {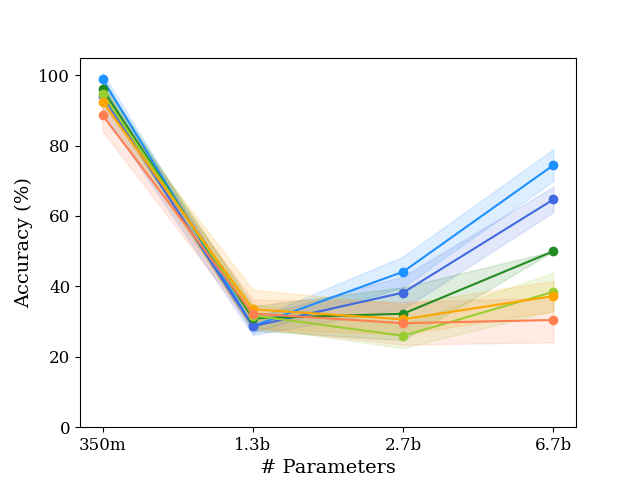}
    \includegraphics[width=0.32\textwidth]            {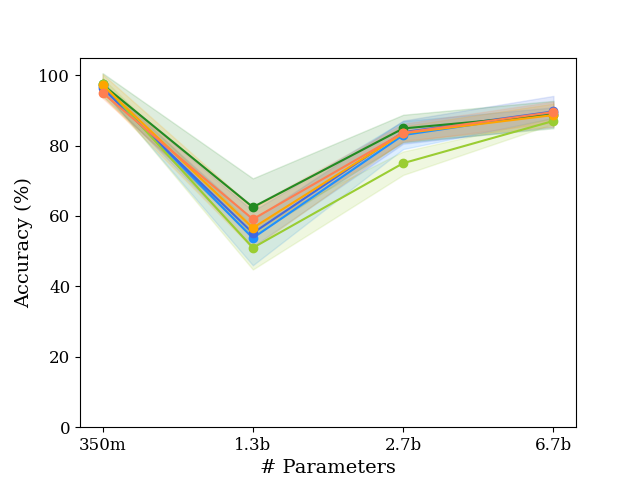}
    \caption{$y_2$}
\end{subfigure}
\end{minipage}
    \hfill
    \begin{minipage}[c]{\linewidth}
        \centering
            \begin{subfigure}{0.4 \linewidth}
            \centering
            \vspace{2mm}
            \includegraphics[width=\textwidth]{latex/figures/lineplots/legend.pdf}
        \end{subfigure}%
    \end{minipage}
    \hfill
    \vspace{-2.5mm}
\begin{minipage}[c]{\linewidth}
    \caption{MPRC dataset (OPT)}
\end{minipage}
\end{figure}
\vspace{-7mm}

\begin{figure}[h!]
\vspace{-2.5mm}
\centering
\begin{minipage}[t]{\linewidth}
\begin{subfigure}{\linewidth}
    \centering
    \includegraphics[width=0.32\textwidth]            {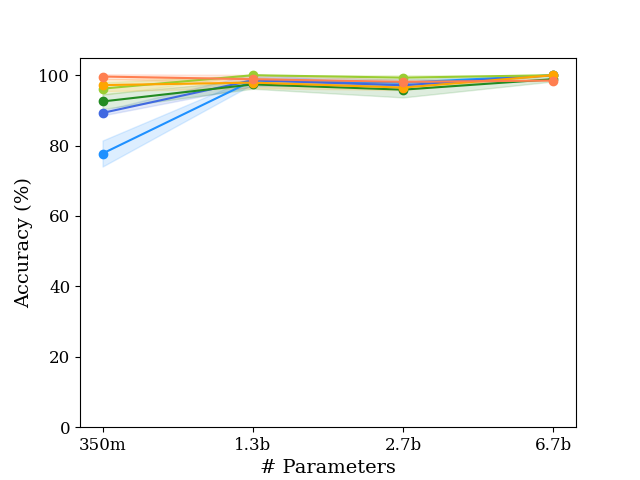}
    \includegraphics[width=0.32\textwidth]            {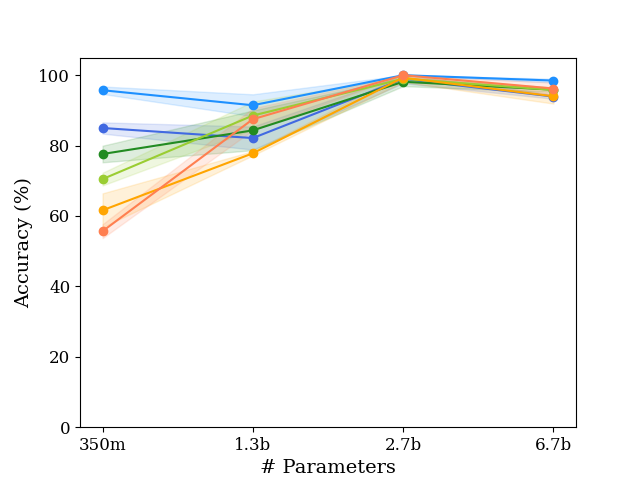}
    \includegraphics[width=0.32\textwidth]            {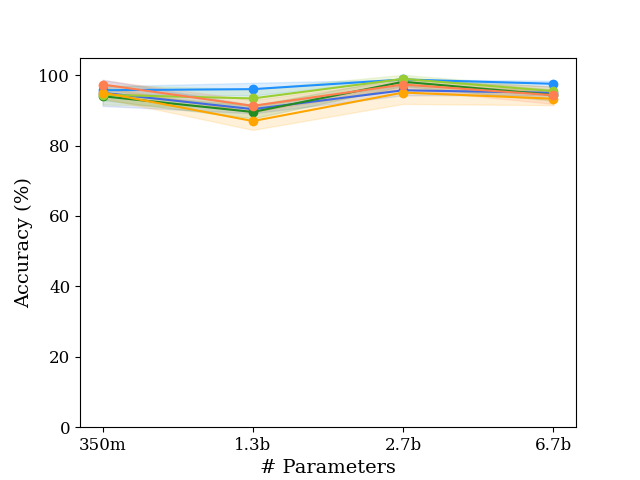}
    \caption{$y_1$}
\end{subfigure}
    \vspace{-2.5mm}
\begin{subfigure}{\linewidth}
    \centering
    \includegraphics[width=0.32\textwidth]            {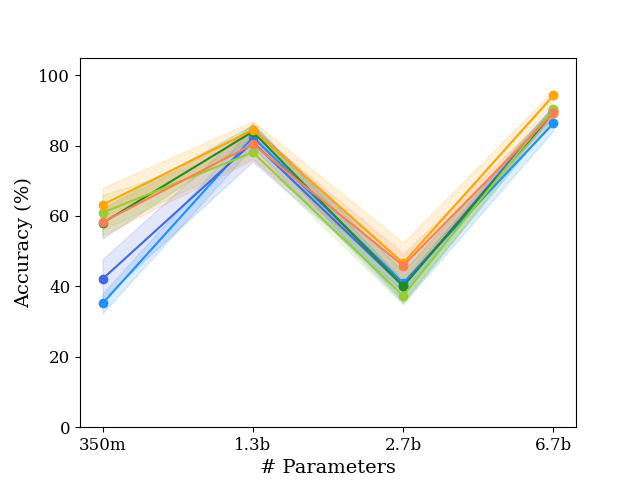}
    \includegraphics[width=0.32\textwidth]            {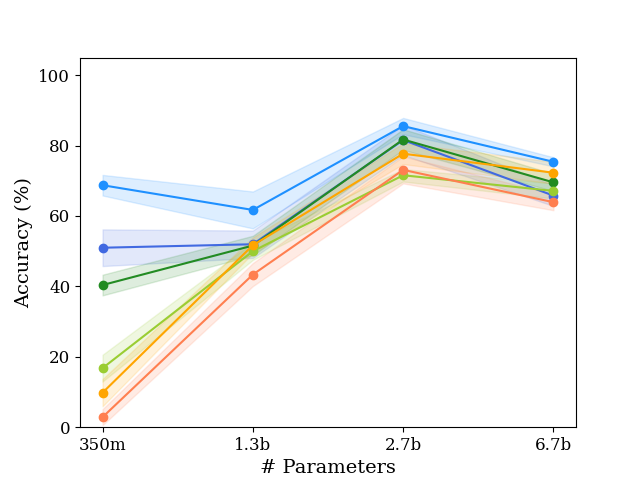}
    \includegraphics[width=0.32\textwidth]            {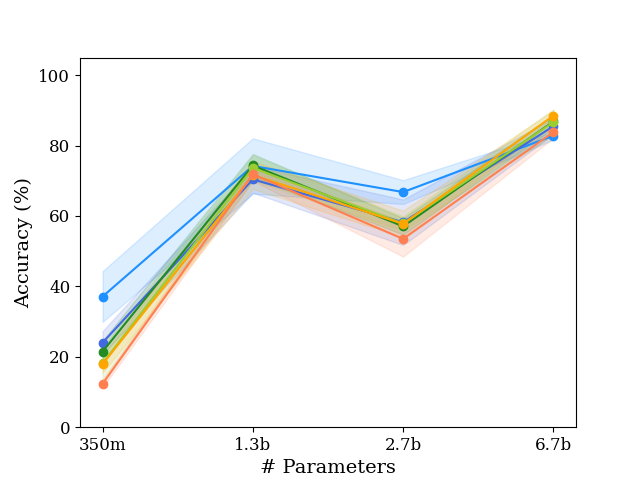}
    \caption{$y_2$}
\end{subfigure}
\end{minipage}
    \hfill
    \begin{minipage}[c]{\linewidth}
        \centering
            \begin{subfigure}{0.4 \linewidth}
            \centering
            \vspace{2mm}
            \includegraphics[width=\textwidth]{latex/figures/lineplots/legend.pdf}
        \end{subfigure}%
    \end{minipage}
    \hfill
    \vspace{-2.5mm}
\begin{minipage}[c]{\linewidth}
    \caption{SST-2 dataset (OPT)}
\end{minipage}
\end{figure}
\vspace{-7mm}

\clearpage
\subsection{Additional Number of Examples Results}\label{app:num-ex-results}
Each of the following figures shows validation performance when varying the number of examples using Llama3 8B and GPT Neo 2.7B. Bin 0 contains the shortest demonstrations and Bin 5 contains the longest demonstrations. Each subfigure shows the validation accuracy on a single class when in-context instances belonging to the respective class were sampled from long instances, short instances, and randomly sampled (left to right).

\begin{figure}[h!]
\vspace{-2.5mm}
\centering
\begin{minipage}[t]{\linewidth}
\begin{subfigure}{\linewidth}
    \centering
    \includegraphics[width=0.32\textwidth]            {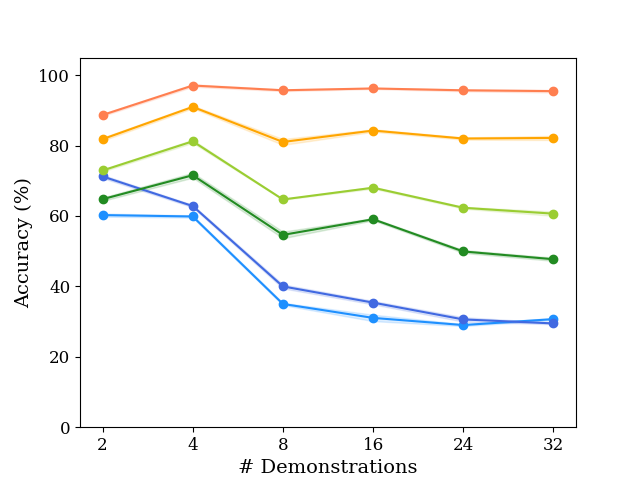}
    \includegraphics[width=0.32\textwidth]            {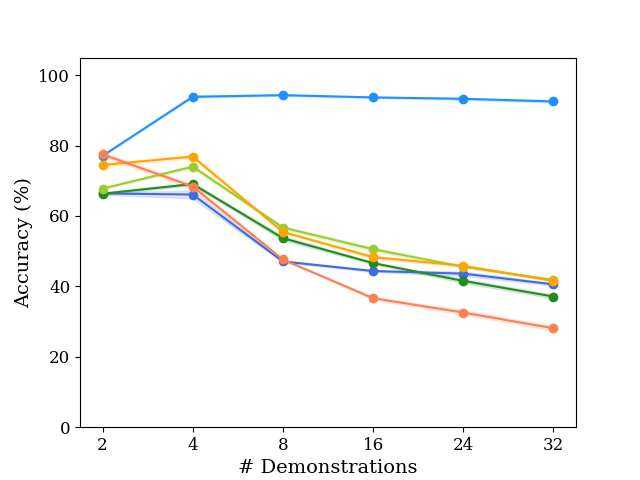}
    \includegraphics[width=0.32\textwidth]            {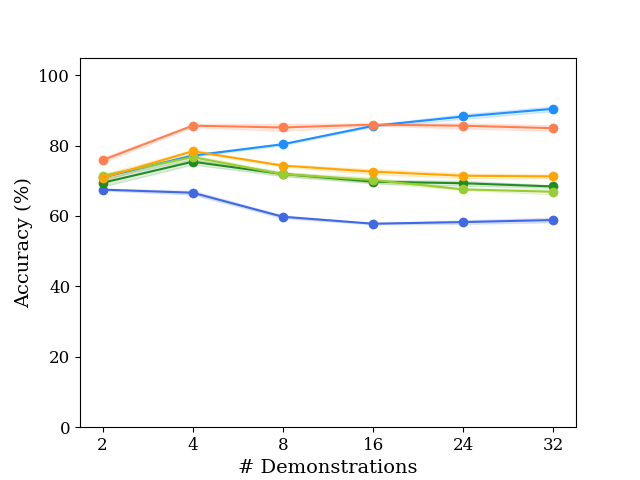}
    \caption{$y_1$}
\end{subfigure}
    \vspace{-2.5mm}
\begin{subfigure}{\linewidth}
    \centering
    \includegraphics[width=0.32\textwidth]            {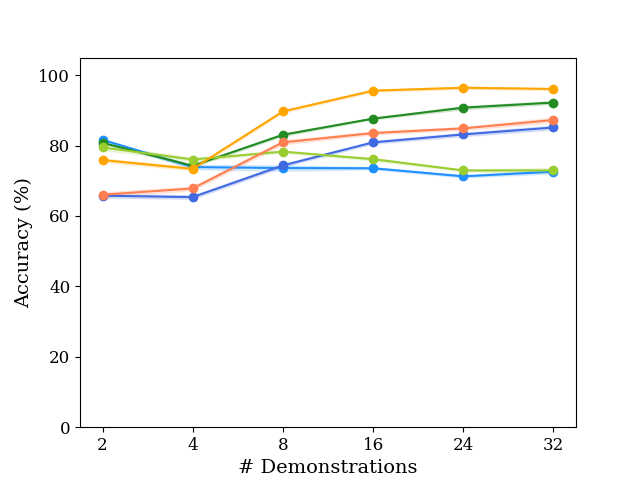}
    \includegraphics[width=0.32\textwidth]            {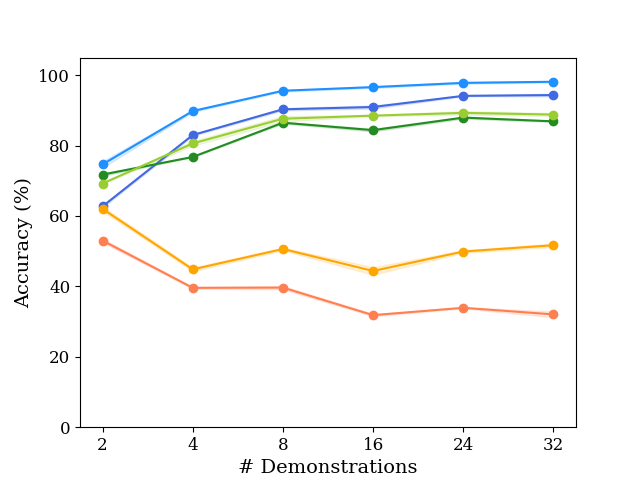}
    \includegraphics[width=0.32\textwidth]            {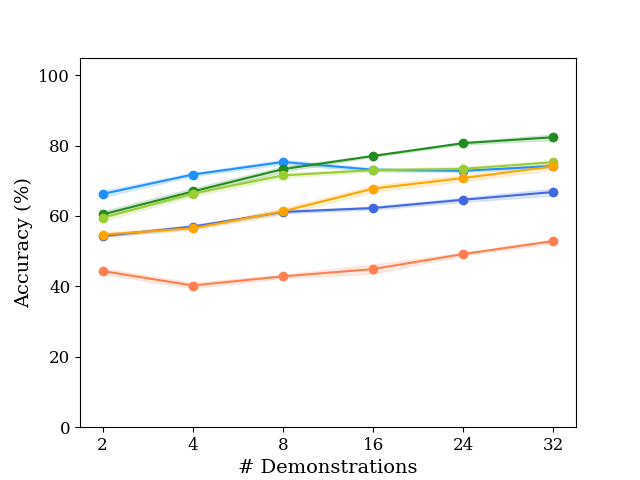}
    \caption{$y_2$}
\end{subfigure}
\end{minipage}
    \hfill
    \begin{minipage}[c]{\linewidth}
        \centering
            \begin{subfigure}{0.4 \linewidth}
            \centering
                        \vspace{2mm}
            \includegraphics[width=\textwidth]{latex/figures/lineplots/legend.pdf}
        \end{subfigure}%
    \end{minipage}
    \hfill
    \vspace{-2.5mm}
\begin{minipage}[c]{\linewidth}
    \caption{Hans dataset (Llama 3 8B)}
\end{minipage}
\end{figure}
\vspace{-5mm}

\begin{figure}[h!]
\vspace{-2.5mm}
\centering
\begin{minipage}[t]{\linewidth}
\begin{subfigure}{\linewidth}
    \centering
    \includegraphics[width=0.32\textwidth]            {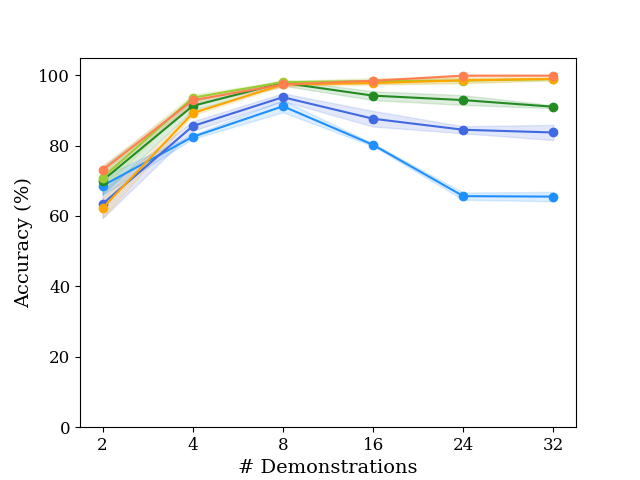}
    \includegraphics[width=0.32\textwidth]            {latex/figures/lineplots/examples/en_llama3_8b_class2_cls1.png}
    \includegraphics[width=0.32\textwidth]            {latex/figures/lineplots/examples/en_llama3_8b_random_cls1.png}
    \caption{$y_1$}
\end{subfigure}
    \vspace{-2.5mm}
\begin{subfigure}{\linewidth}
    \centering
    \includegraphics[width=0.32\textwidth]            {latex/figures/lineplots/examples/en_llama3_8b_class2_cls2.png}
    \includegraphics[width=0.32\textwidth]            {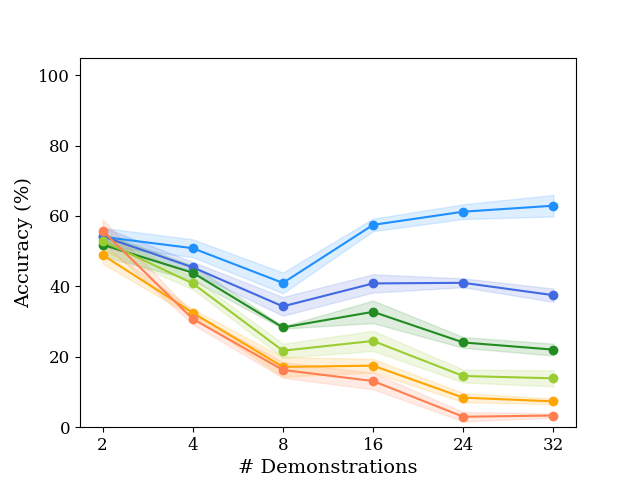}
    \includegraphics[width=0.32\textwidth]            {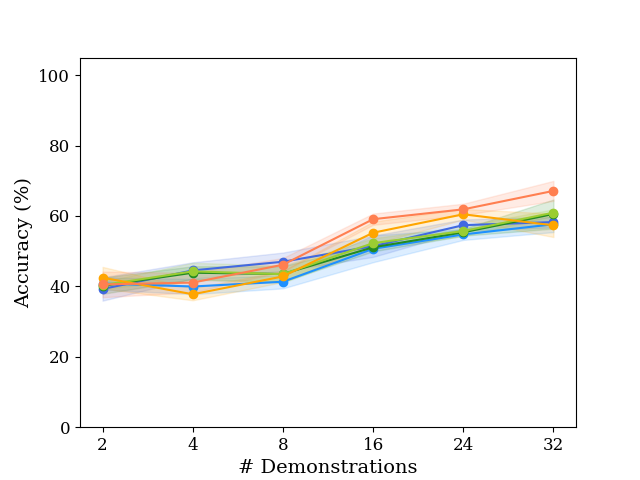}
    \caption{$y_2$}
\end{subfigure}
\end{minipage}
    \hfill
    \begin{minipage}[c]{\linewidth}
        \centering
            \begin{subfigure}{0.4 \linewidth}
            \centering
                        \vspace{2mm}
            \includegraphics[width=\textwidth]{latex/figures/lineplots/legend.pdf}
        \end{subfigure}%
    \end{minipage}
    \hfill
    \vspace{-2.5mm}
\begin{minipage}[c]{\linewidth}
    \caption{PAWS-X$_{\textsc{EN}}$ dataset (Llama 3 8B)}
\end{minipage}
\end{figure}
\vspace{-5mm}

\begin{figure}[h!]
\vspace{-2.5mm}
\centering
\begin{minipage}[t]{\linewidth}
\begin{subfigure}{\linewidth}
    \centering
    \includegraphics[width=0.32\textwidth]            {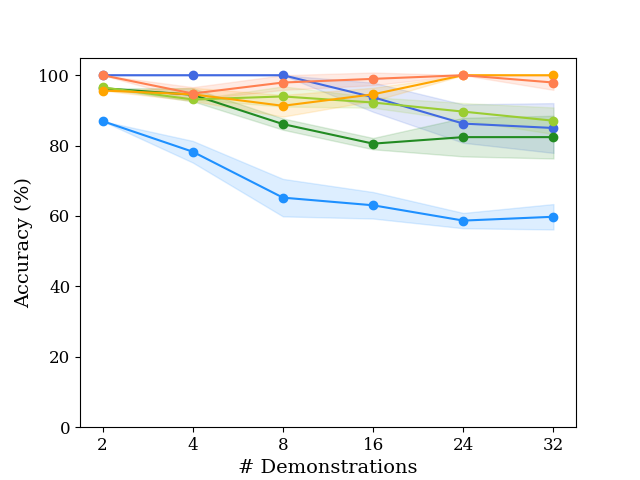}
    \includegraphics[width=0.32\textwidth]            {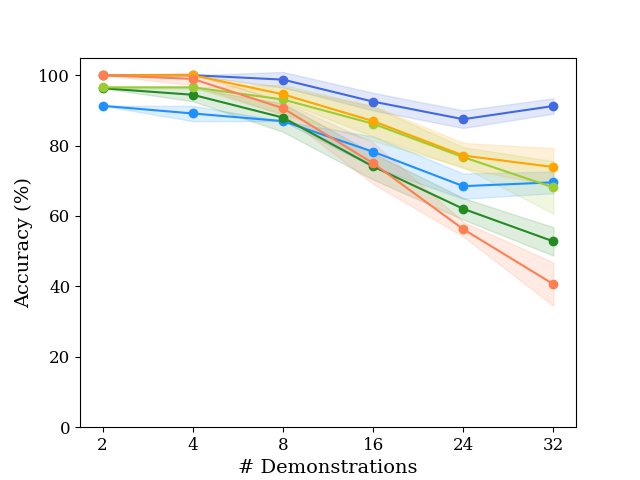}
    \includegraphics[width=0.32\textwidth]            {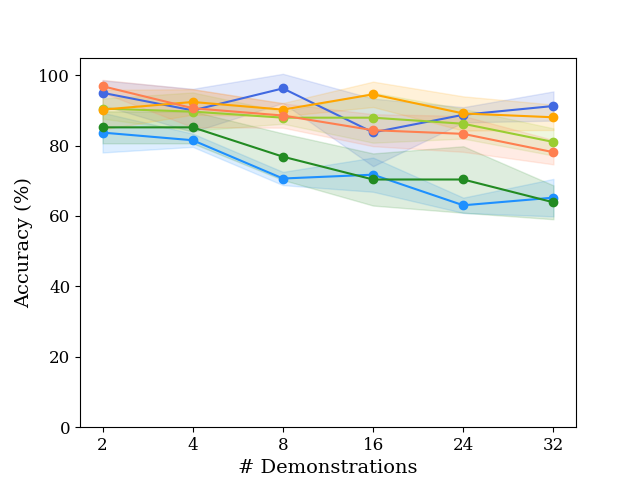}
    \caption{$y_1$}
\end{subfigure}
    \vspace{-2.5mm}
\begin{subfigure}{\linewidth}
    \centering
    \includegraphics[width=0.32\textwidth]            {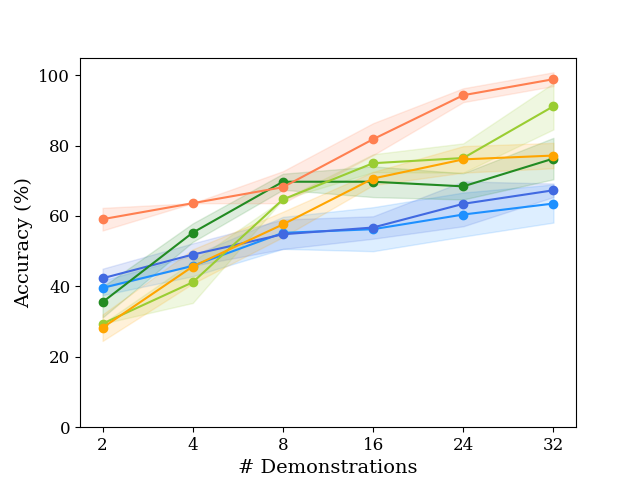}
    \includegraphics[width=0.32\textwidth]            {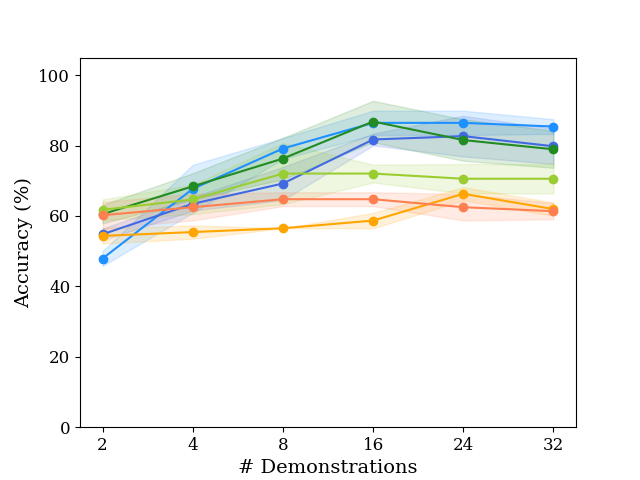}
    \includegraphics[width=0.32\textwidth]            {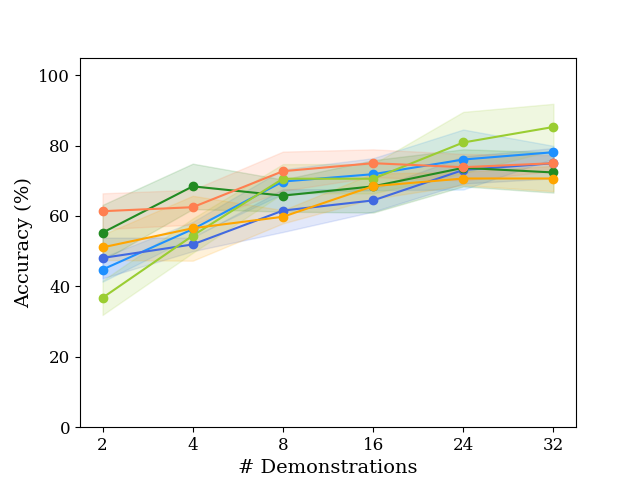}
    \caption{$y_2$}
\end{subfigure}
\end{minipage}
    \hfill
    \begin{minipage}[c]{\linewidth}
        \centering
            \begin{subfigure}{0.4 \linewidth}
            \centering
                        \vspace{2mm}
            \includegraphics[width=\textwidth]{latex/figures/lineplots/legend.pdf}
        \end{subfigure}%
    \end{minipage}
    \hfill
    \vspace{-2.5mm}
\begin{minipage}[c]{\linewidth}
    \caption{RTE dataset (Llama 3 8B)}
\end{minipage}
\end{figure}
\vspace{-7mm}

\begin{figure}[h!]
\vspace{-2.5mm}
\centering
\begin{minipage}[t]{\linewidth}
\begin{subfigure}{\linewidth}
    \centering
    \includegraphics[width=0.32\textwidth]            {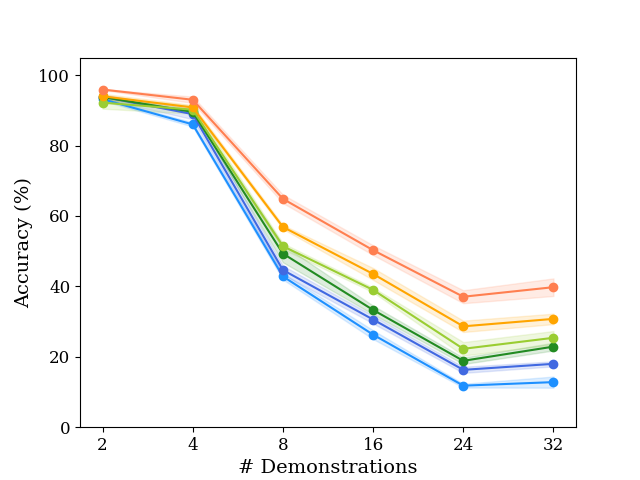}
    \includegraphics[width=0.32\textwidth]            {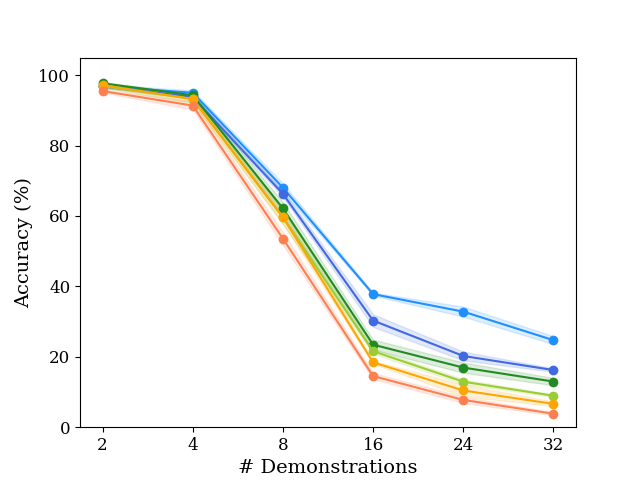}
    \includegraphics[width=0.32\textwidth]            {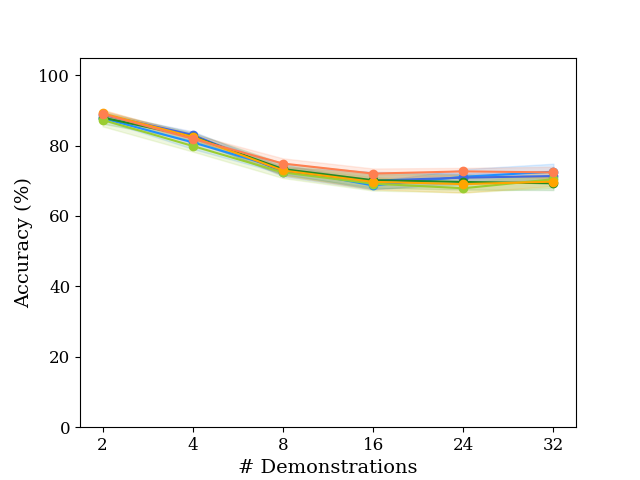}
    \caption{$y_1$}
\end{subfigure}
    \vspace{-2.5mm}
\begin{subfigure}{\linewidth}
    \centering
    \includegraphics[width=0.32\textwidth]            {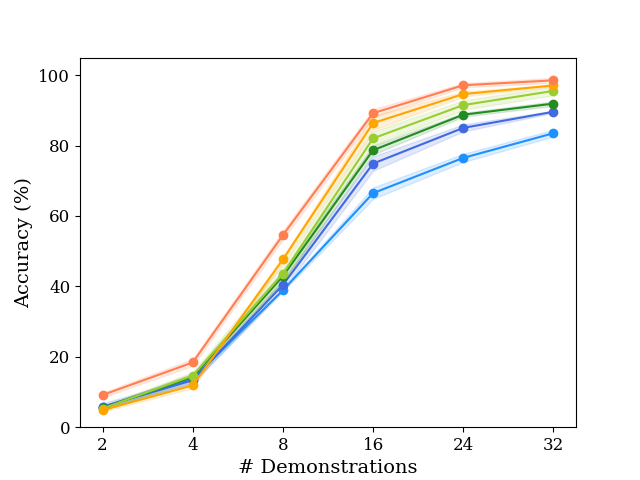}
    \includegraphics[width=0.32\textwidth]            {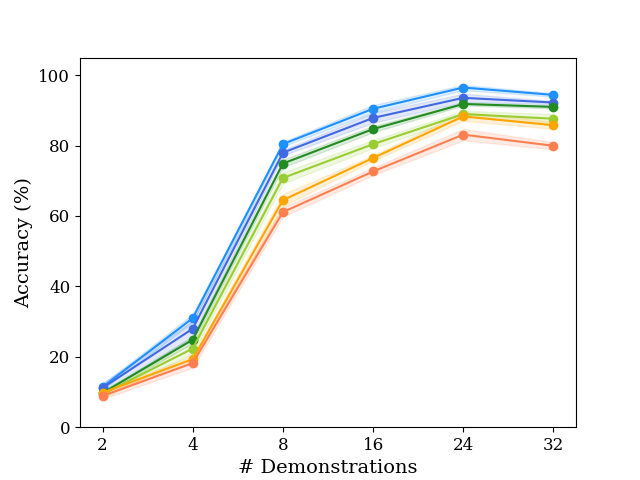}
    \includegraphics[width=0.32\textwidth]            {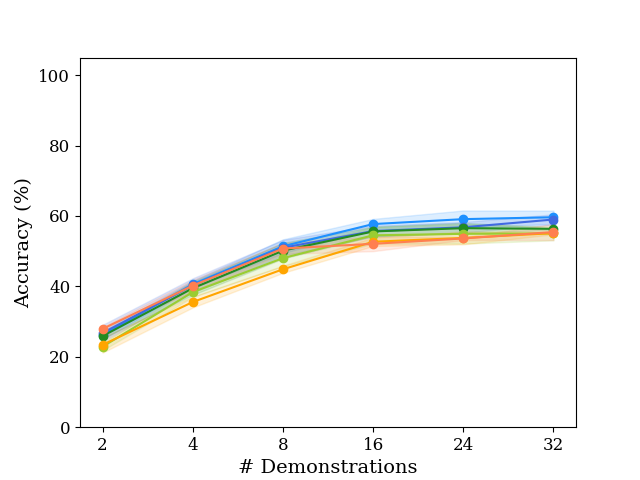}
    \caption{$y_2$}
\end{subfigure}
\end{minipage}
    \hfill
    \begin{minipage}[c]{\linewidth}
        \centering
            \begin{subfigure}{0.4 \linewidth}
            \centering
                        \vspace{2mm}
            \includegraphics[width=\textwidth]{latex/figures/lineplots/legend.pdf}
        \end{subfigure}%
    \end{minipage}
    \hfill
\vspace{-2.5mm}
\begin{minipage}[c]{\linewidth}
    \caption{QNLI dataset (Llama 3 8B)}
\end{minipage}
\end{figure}
\vspace{-7mm}

\begin{figure}[h!]
\vspace{-2.5mm}
\centering
\begin{minipage}[t]{\linewidth}
\begin{subfigure}{\linewidth}
    \centering
    \includegraphics[width=0.32\textwidth]            {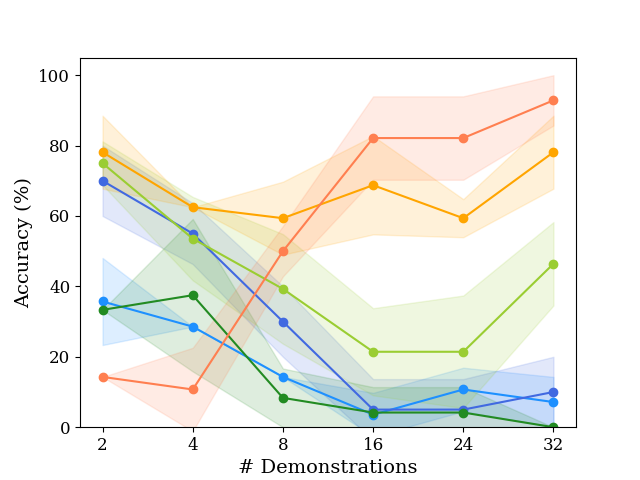}
    \includegraphics[width=0.32\textwidth]            {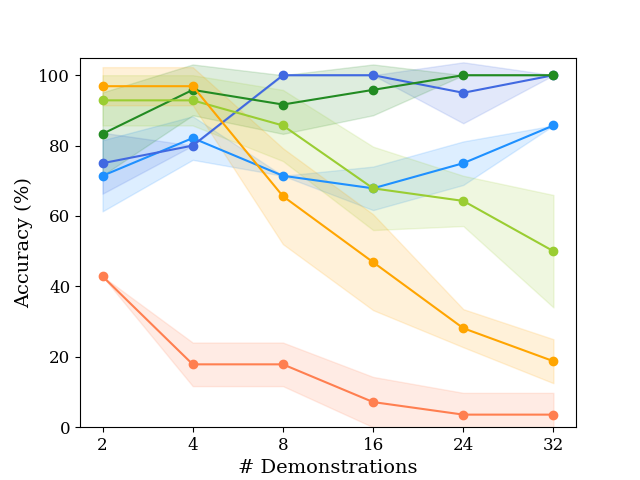}
    \includegraphics[width=0.32\textwidth]            {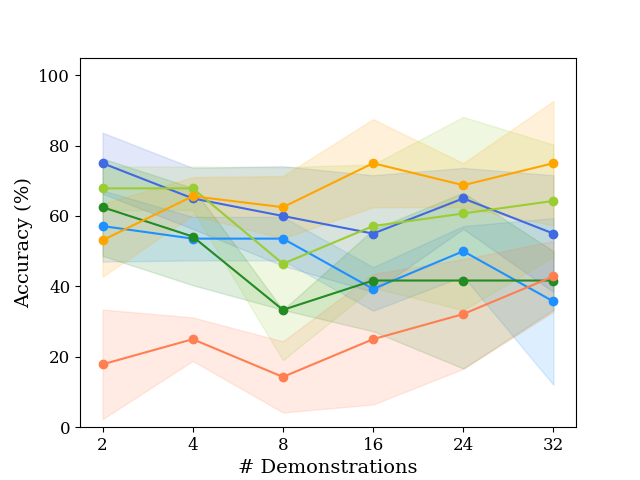}
    \caption{$y_1$}
\end{subfigure}
    \vspace{-2.5mm}
\begin{subfigure}{\linewidth}
    \centering
    \includegraphics[width=0.32\textwidth]            {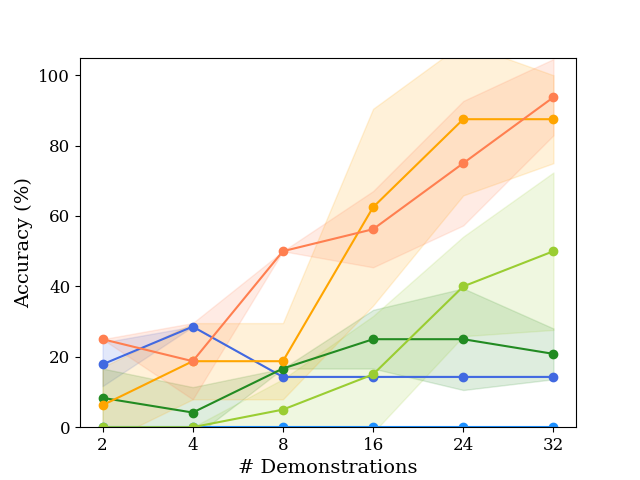}
    \includegraphics[width=0.32\textwidth]            {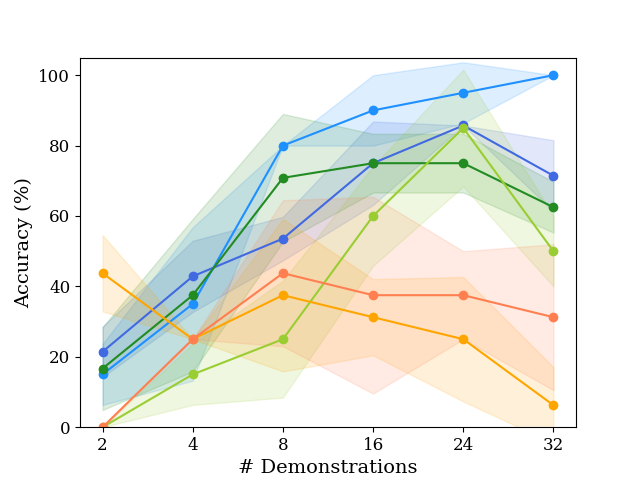}
    \includegraphics[width=0.32\textwidth]            {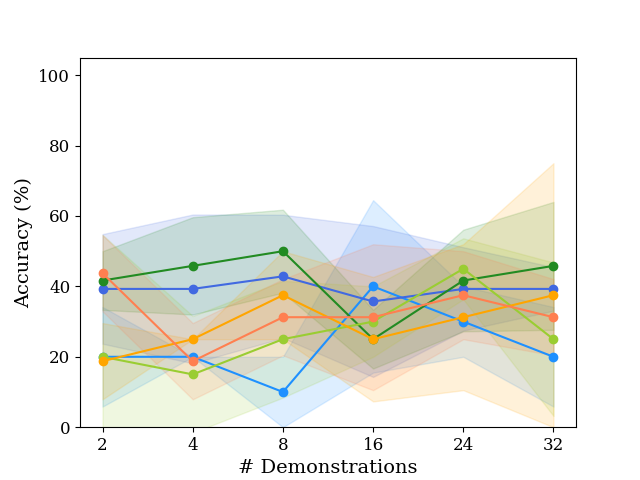}
    \caption{$y_2$}
\end{subfigure}
\end{minipage}
    \hfill
    \begin{minipage}[c]{\linewidth}
        \centering
            \begin{subfigure}{0.4 \linewidth}
            \centering
                        \vspace{2mm}
            \includegraphics[width=\textwidth]{latex/figures/lineplots/legend.pdf}
        \end{subfigure}%
    \end{minipage}
    \hfill
    \vspace{-2.5mm}
\begin{minipage}[c]{\linewidth}
    \caption{WNLI dataset (Llama 3 8B)}
\end{minipage}
\end{figure}
\vspace{-7mm}

\begin{figure}[h!]
\vspace{-2.5mm}
\centering
\begin{minipage}[t]{\linewidth}
\begin{subfigure}{\linewidth}
    \centering
    \includegraphics[width=0.32\textwidth]            {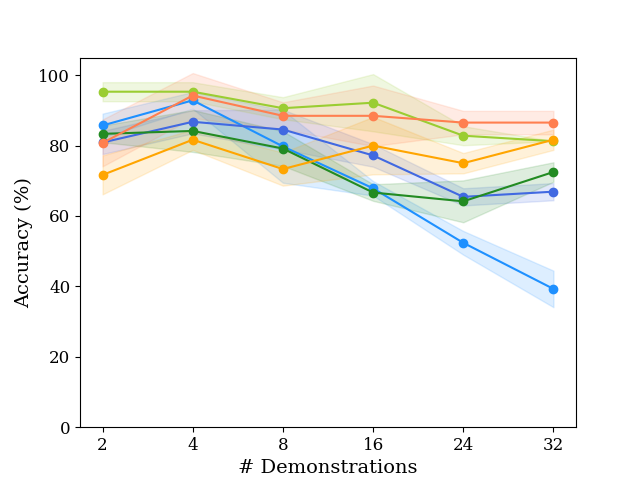}
    \includegraphics[width=0.32\textwidth]            {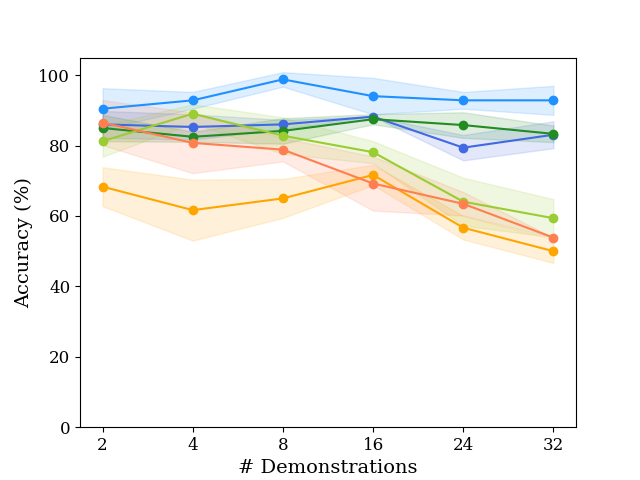}
    \includegraphics[width=0.32\textwidth]            {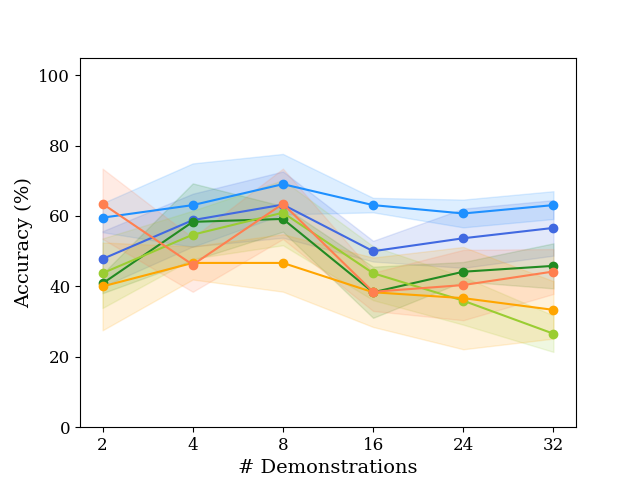}
    \caption{$y_1$}
\end{subfigure}
    \vspace{-2.5mm}
\begin{subfigure}{\linewidth}
    \centering
    \includegraphics[width=0.32\textwidth]            {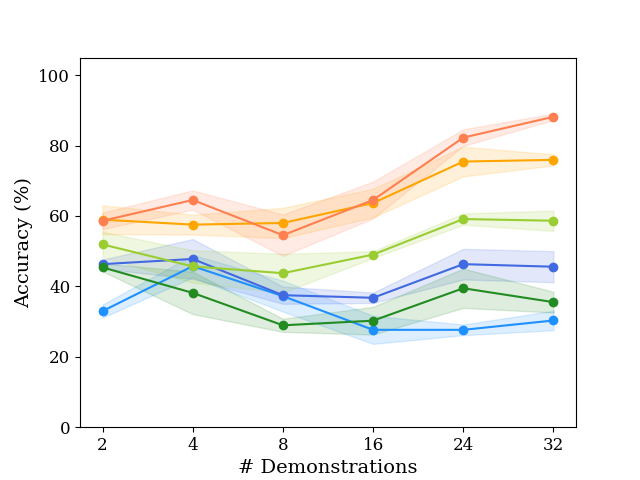}
    \includegraphics[width=0.32\textwidth]            {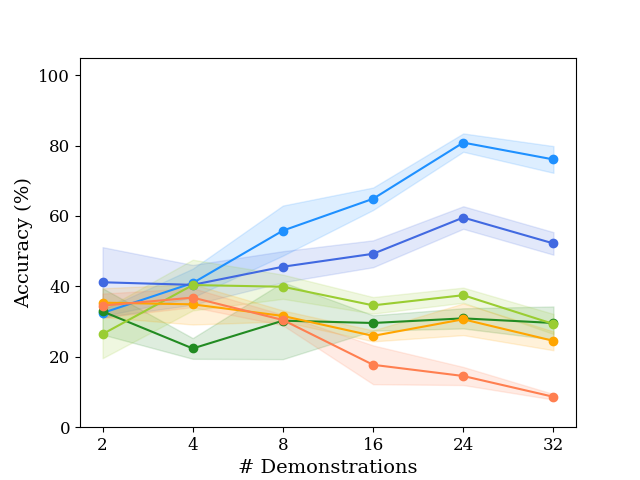}
    \includegraphics[width=0.32\textwidth]            {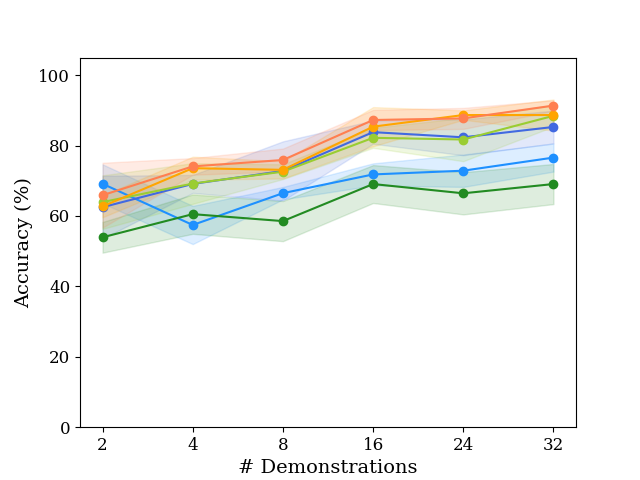}
    \caption{$y_2$}
\end{subfigure}
\end{minipage}
    \hfill
    \begin{minipage}[c]{\linewidth}
        \centering
            \begin{subfigure}{0.4 \linewidth}
            \centering
            \vspace{2mm}
            \includegraphics[width=\textwidth]{latex/figures/lineplots/legend.pdf}
        \end{subfigure}%
    \end{minipage}
    \hfill
    \vspace{-2.5mm}
\begin{minipage}[c]{\linewidth}
    \caption{MPRC dataset (Llama 3 8B)}
\end{minipage}
\end{figure}
\vspace{-7mm}

\begin{figure}[h!]
\vspace{-2.5mm}
\centering
\begin{minipage}[t]{\linewidth}
\begin{subfigure}{\linewidth}
    \centering
    \includegraphics[width=0.32\textwidth]            {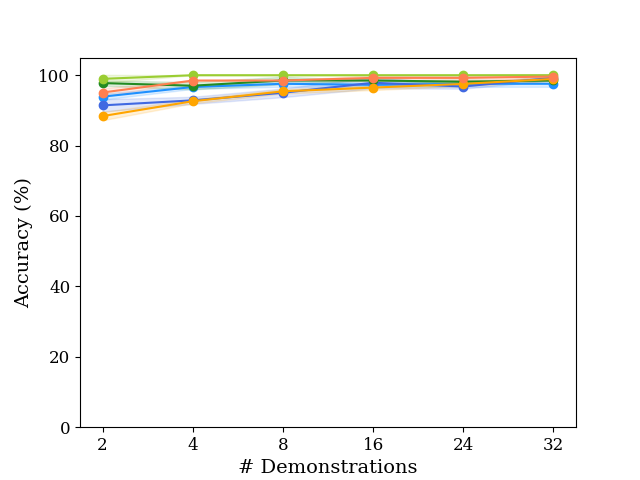}
    \includegraphics[width=0.32\textwidth]            {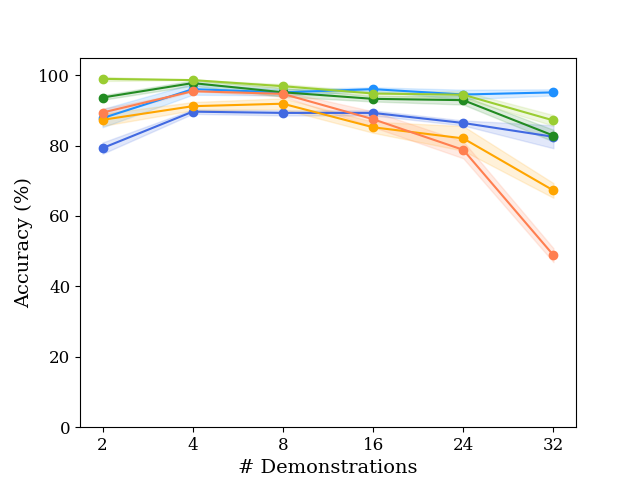}
    \includegraphics[width=0.32\textwidth]            {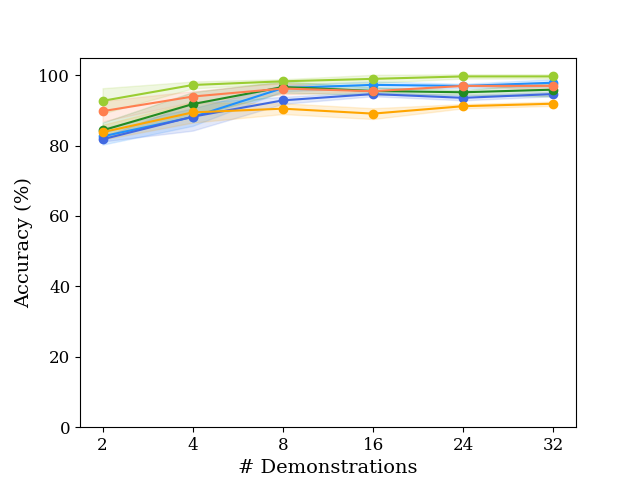}
    \caption{$y_1$}
\end{subfigure}
    \vspace{-2.5mm}
\begin{subfigure}{\linewidth}
    \centering
    \includegraphics[width=0.32\textwidth]            {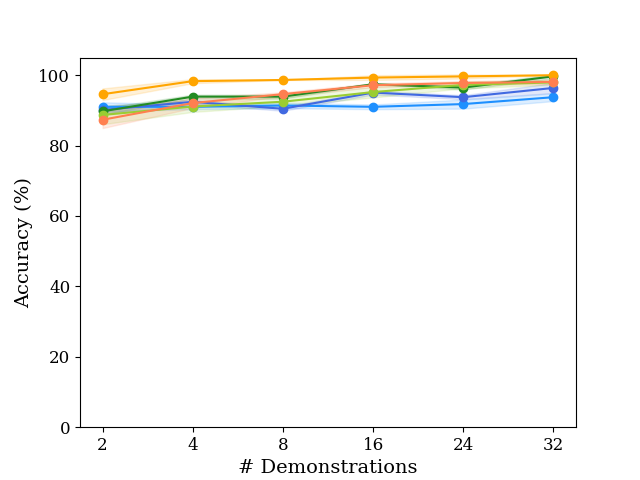}
    \includegraphics[width=0.32\textwidth]            {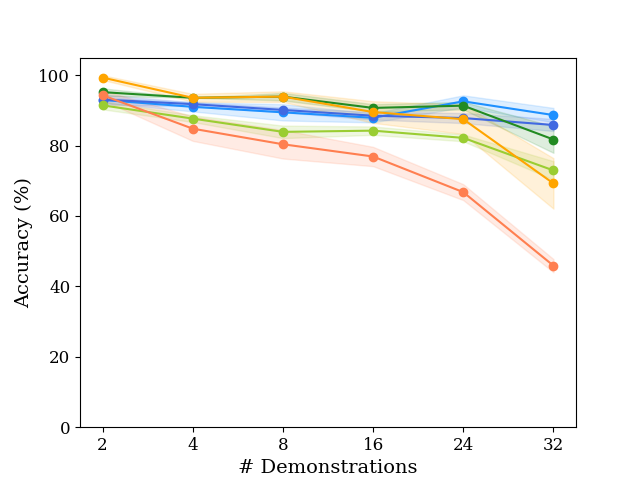}
    \includegraphics[width=0.32\textwidth]            {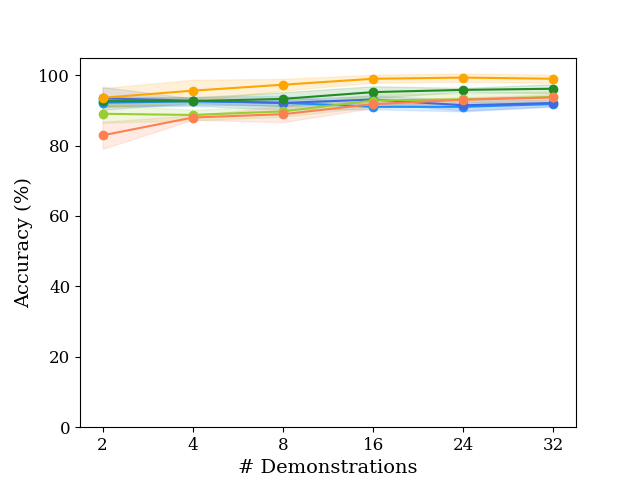}
    \caption{$y_2$}
\end{subfigure}
\end{minipage}
    \hfill
    \begin{minipage}[c]{\linewidth}
        \centering
            \begin{subfigure}{0.4 \linewidth}
            \centering
            \vspace{2mm}
            \includegraphics[width=\textwidth]{latex/figures/lineplots/legend.pdf}
        \end{subfigure}%
    \end{minipage}
    \hfill
    \vspace{-2.5mm}
\begin{minipage}[c]{\linewidth}
    \caption{SST-2 dataset (Llama 3 8B)}
\end{minipage}
\end{figure}
\vspace{-7mm}
\begin{figure}[h!]
\vspace{-2.5mm}
\centering
\begin{minipage}[t]{\linewidth}
\begin{subfigure}{\linewidth}
    \centering
    \includegraphics[width=0.32\textwidth]            {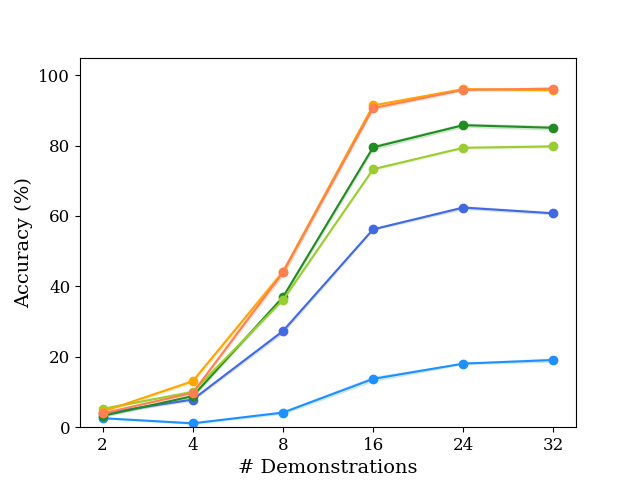}
    \includegraphics[width=0.32\textwidth]            {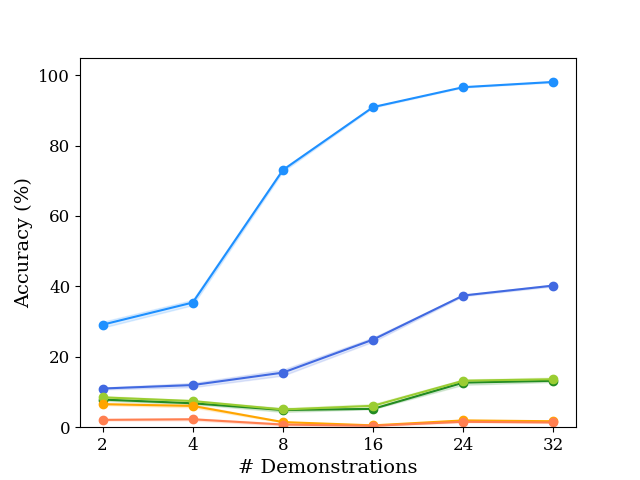}
    \includegraphics[width=0.32\textwidth]            {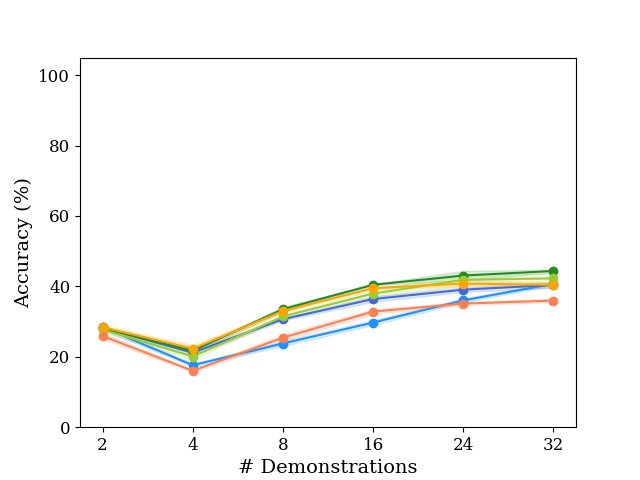}
    \caption{$y_1$}
\end{subfigure}
    \vspace{-2.5mm}
\begin{subfigure}{\linewidth}
    \centering
    \includegraphics[width=0.32\textwidth]            {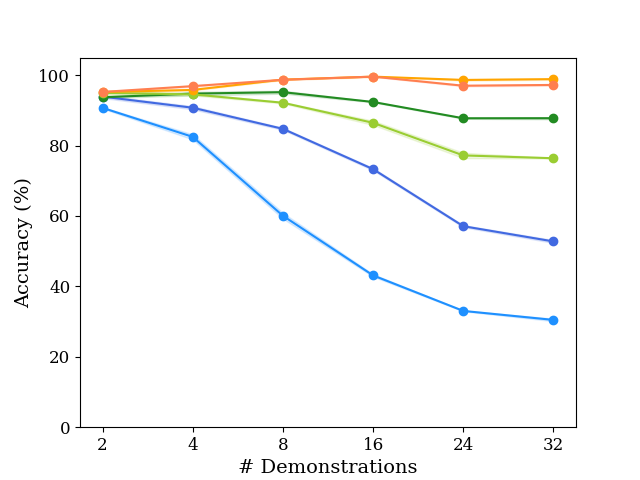}
    \includegraphics[width=0.32\textwidth]            {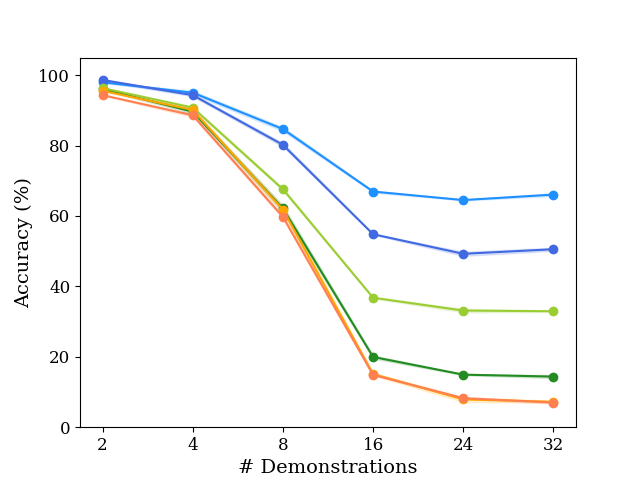}
    \includegraphics[width=0.32\textwidth]            {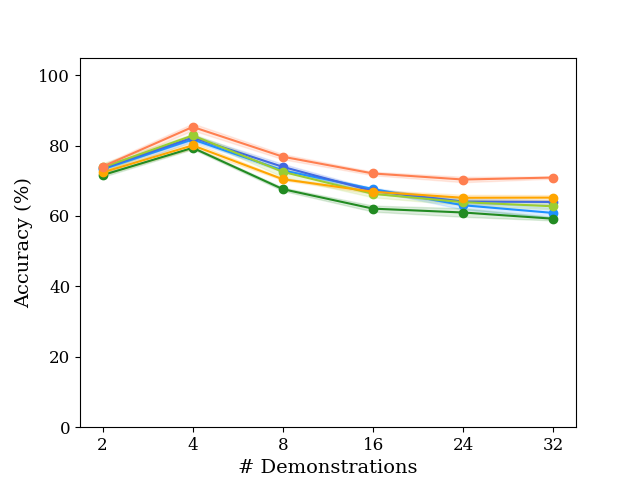}
    \caption{$y_2$}
\end{subfigure}
\end{minipage}
    \hfill
    \begin{minipage}[c]{\linewidth}
        \centering
            \begin{subfigure}{0.4 \linewidth}
            \centering
                        \vspace{2mm}
            \includegraphics[width=\textwidth]{latex/figures/lineplots/legend.pdf}
        \end{subfigure}%
    \end{minipage}
    \hfill
    \vspace{-2.5mm}
\begin{minipage}[c]{\linewidth}
    \caption{Hans dataset (GPT Neo 2.7B)}
\end{minipage}
\end{figure}
\vspace{-5mm}

\begin{figure}[h!]
\vspace{-2.5mm}
\centering
\begin{minipage}[t]{\linewidth}
\begin{subfigure}{\linewidth}
    \centering
    \includegraphics[width=0.32\textwidth]            {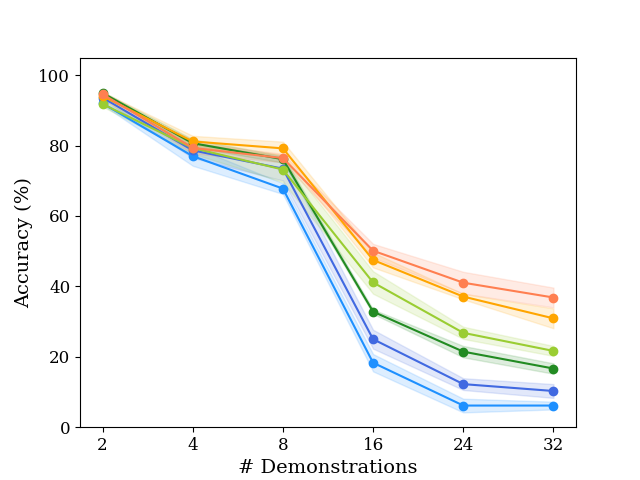}
    \includegraphics[width=0.32\textwidth]            {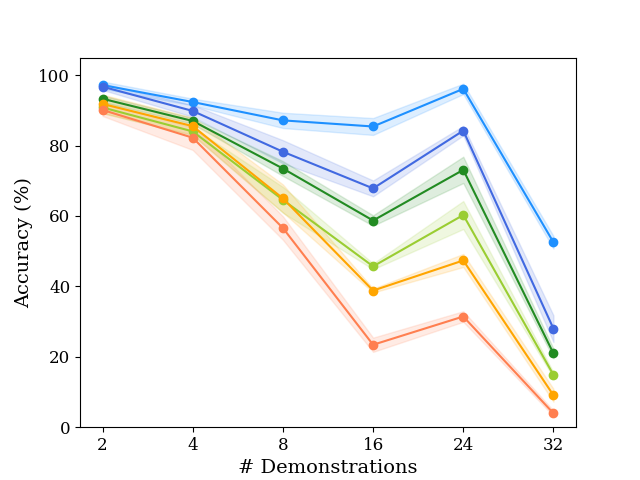}
    \includegraphics[width=0.32\textwidth]            {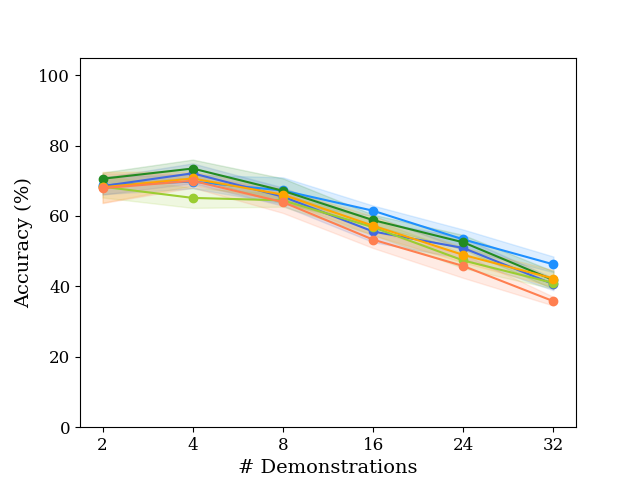}
    \caption{$y_1$}
\end{subfigure}
    \vspace{-2.5mm}
\begin{subfigure}{\linewidth}
    \centering
    \includegraphics[width=0.32\textwidth]            {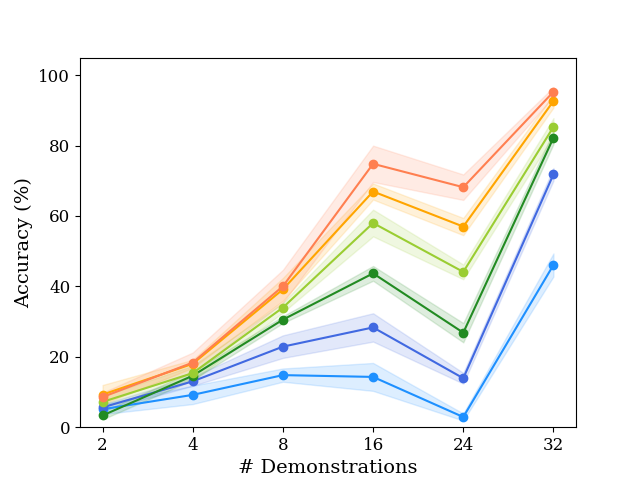}
    \includegraphics[width=0.32\textwidth]            {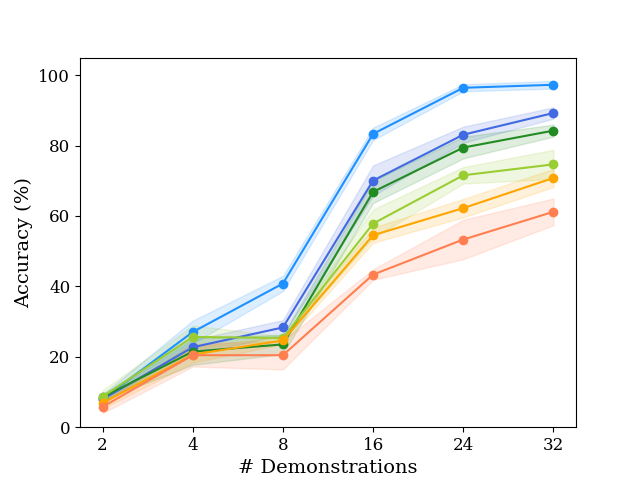}
    \includegraphics[width=0.32\textwidth]            {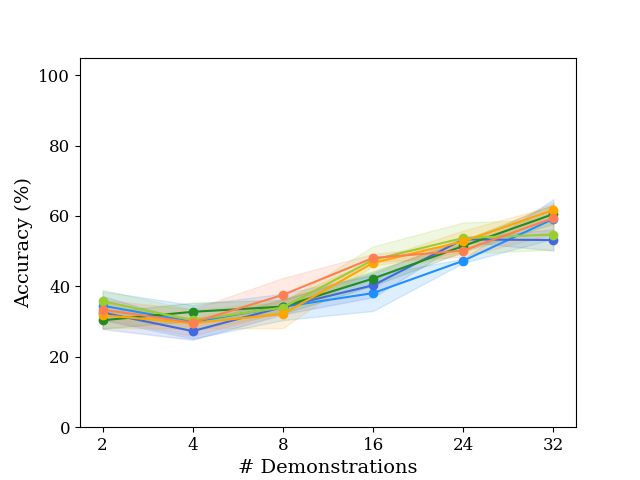}
    \caption{$y_2$}
\end{subfigure}
\end{minipage}
    \hfill
    \begin{minipage}[c]{\linewidth}
        \centering
            \begin{subfigure}{0.4 \linewidth}
            \centering
                        \vspace{2mm}
            \includegraphics[width=\textwidth]{latex/figures/lineplots/legend.pdf}
        \end{subfigure}%
    \end{minipage}
    \hfill
    \vspace{-2.5mm}
\begin{minipage}[c]{\linewidth}
    \caption{PAWS-X$_{\textsc{EN}}$ dataset (GPT Neo 2.7B)}
\end{minipage}
\end{figure}
\vspace{-5mm}

\begin{figure}[h!]
\vspace{-2.5mm}
\centering
\begin{minipage}[t]{\linewidth}
\begin{subfigure}{\linewidth}
    \centering
    \includegraphics[width=0.32\textwidth]            {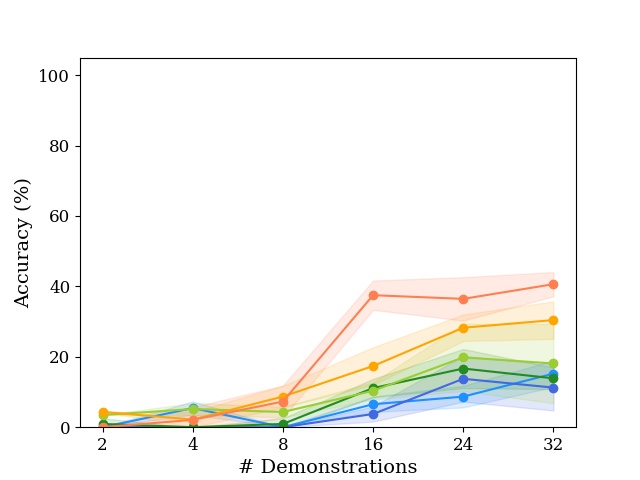}
    \includegraphics[width=0.32\textwidth]            {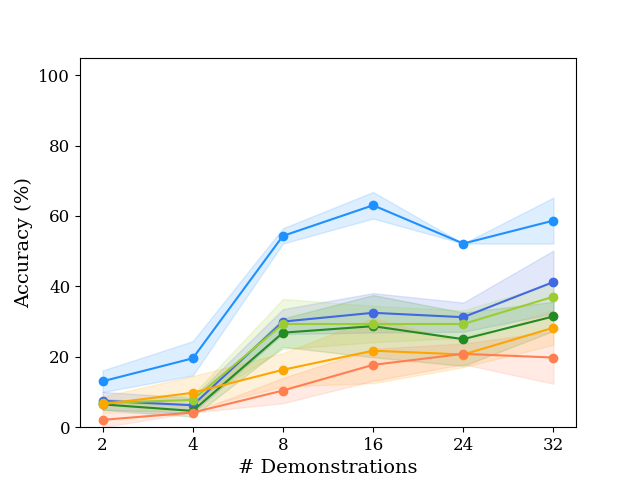}
    \includegraphics[width=0.32\textwidth]            {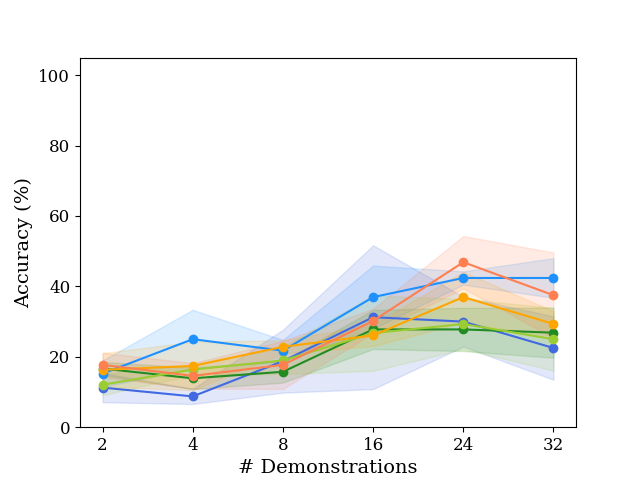}
    \caption{$y_1$}
\end{subfigure}
    \vspace{-2.5mm}
\begin{subfigure}{\linewidth}
    \centering
    \includegraphics[width=0.32\textwidth]            {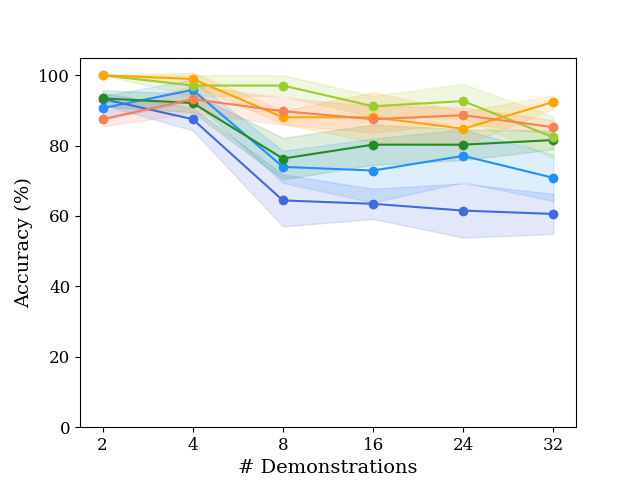}
    \includegraphics[width=0.32\textwidth]            {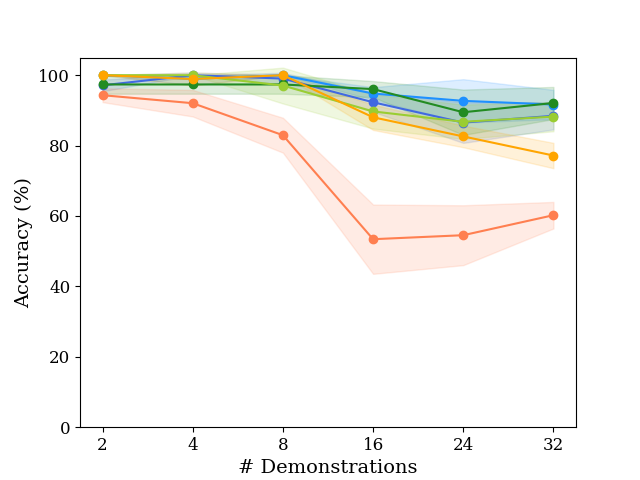}
    \includegraphics[width=0.32\textwidth]            {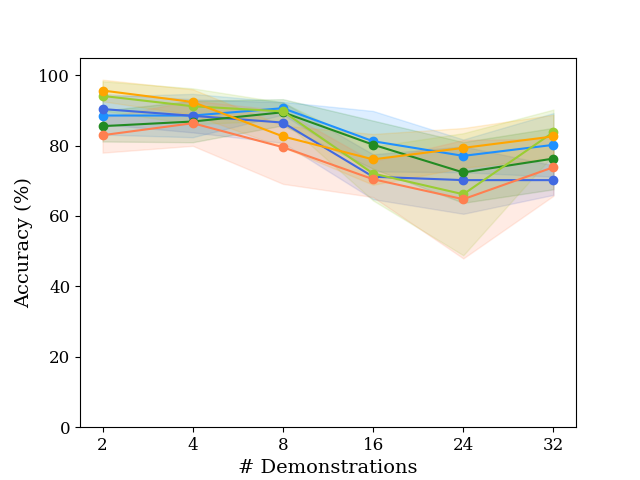}
    \caption{$y_2$}
\end{subfigure}
\end{minipage}
    \hfill
    \begin{minipage}[c]{\linewidth}
        \centering
            \begin{subfigure}{0.4 \linewidth}
            \centering
                        \vspace{2mm}
            \includegraphics[width=\textwidth]{latex/figures/lineplots/legend.pdf}
        \end{subfigure}%
    \end{minipage}
    \hfill
    \vspace{-2.5mm}
\begin{minipage}[c]{\linewidth}
    \caption{RTE dataset (GPT Neo 2.7B)}
\end{minipage}
\end{figure}
\vspace{-7mm}

\begin{figure}[h!]
\vspace{-2.5mm}
\centering
\begin{minipage}[t]{\linewidth}
\begin{subfigure}{\linewidth}
    \centering
    \includegraphics[width=0.32\textwidth]            {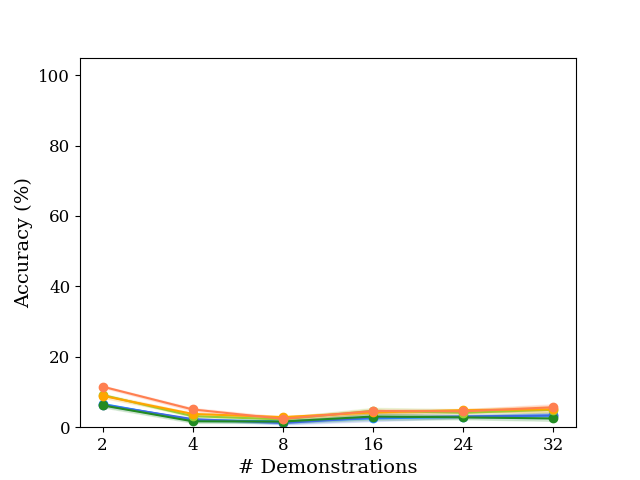}
    \includegraphics[width=0.32\textwidth]            {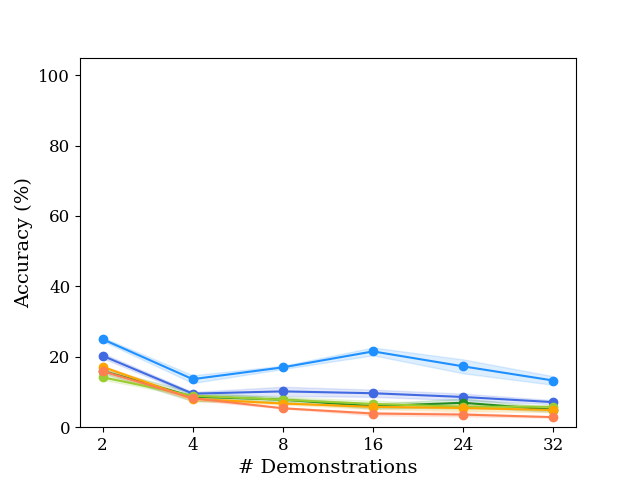}
    \includegraphics[width=0.32\textwidth]            {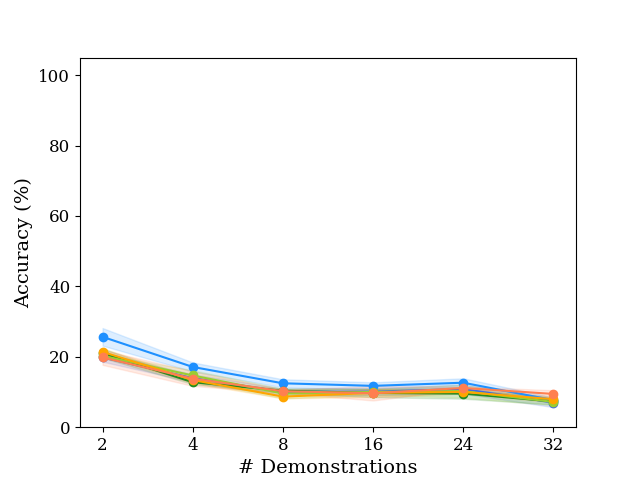}
    \caption{$y_1$}
\end{subfigure}
    \vspace{-2.5mm}
\begin{subfigure}{\linewidth}
    \centering
    \includegraphics[width=0.32\textwidth]            {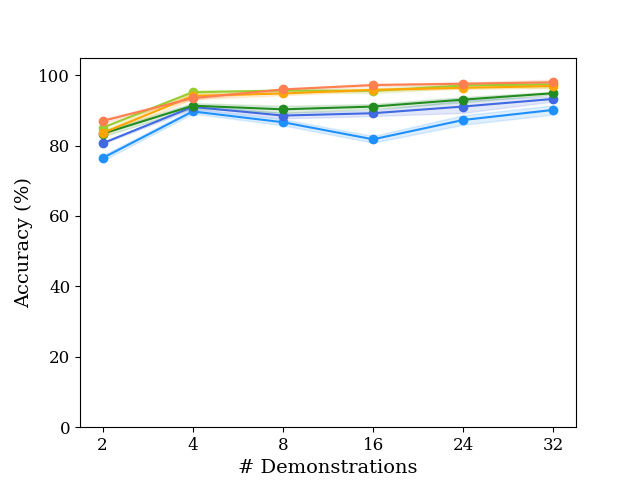}
    \includegraphics[width=0.32\textwidth]            {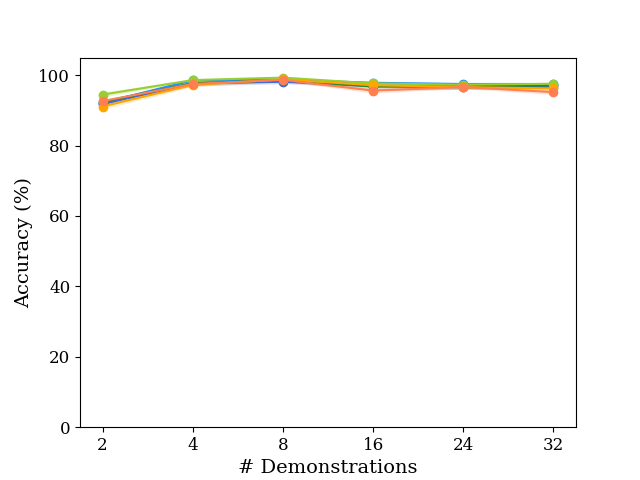}
    \includegraphics[width=0.32\textwidth]            {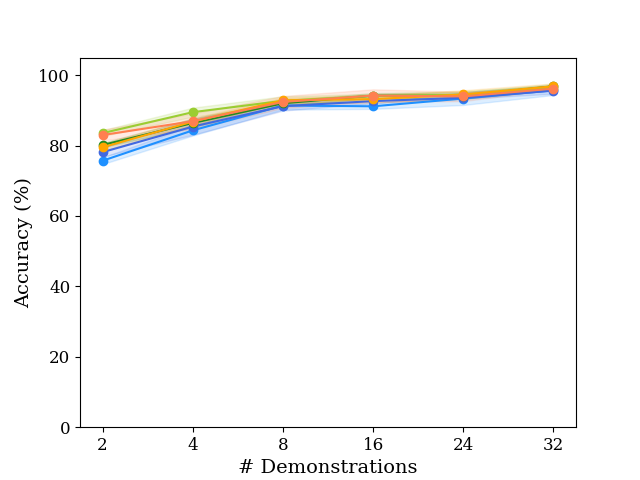}
    \caption{$y_2$}
\end{subfigure}
\end{minipage}
    \hfill
    \begin{minipage}[c]{\linewidth}
        \centering
            \begin{subfigure}{0.4 \linewidth}
            \centering
                        \vspace{2mm}
            \includegraphics[width=\textwidth]{latex/figures/lineplots/legend.pdf}
        \end{subfigure}%
    \end{minipage}
    \hfill
\vspace{-2.5mm}
\begin{minipage}[c]{\linewidth}
    \caption{QNLI dataset (GPT Neo 2.7B)}
\end{minipage}
\end{figure}
\vspace{-7mm}

\begin{figure}[h!]
\vspace{-2.5mm}
\centering
\begin{minipage}[t]{\linewidth}
\begin{subfigure}{\linewidth}
    \centering
    \includegraphics[width=0.32\textwidth]            {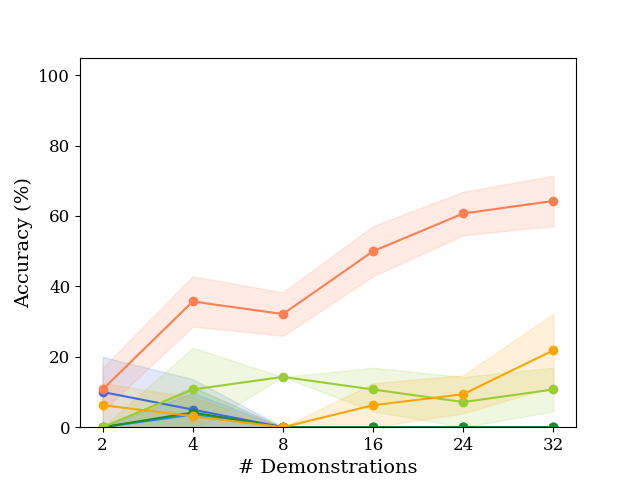}
    \includegraphics[width=0.32\textwidth]            {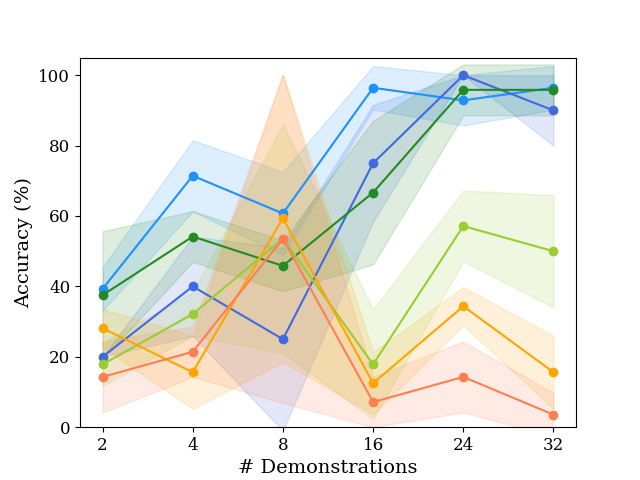}
    \includegraphics[width=0.32\textwidth]            {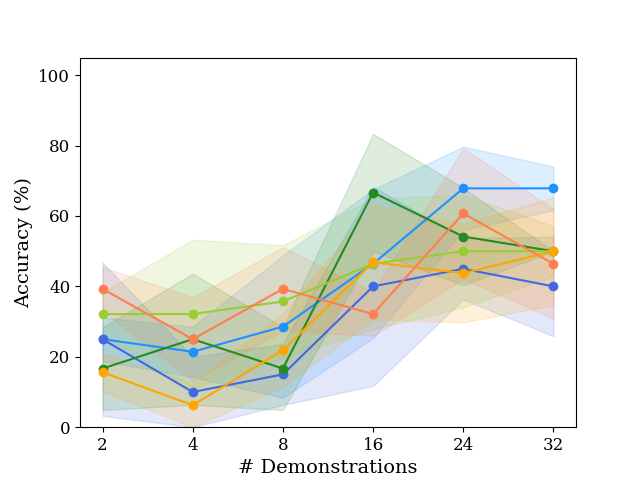}
    \caption{$y_1$}
\end{subfigure}
    \vspace{-2.5mm}
\begin{subfigure}{\linewidth}
    \centering
    \includegraphics[width=0.32\textwidth]            {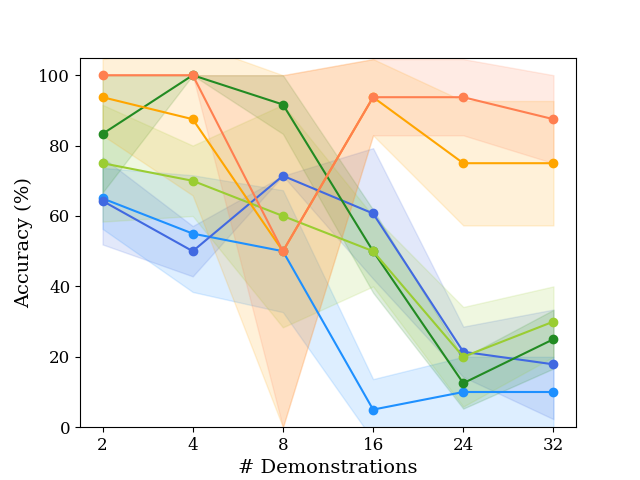}
    \includegraphics[width=0.32\textwidth]            {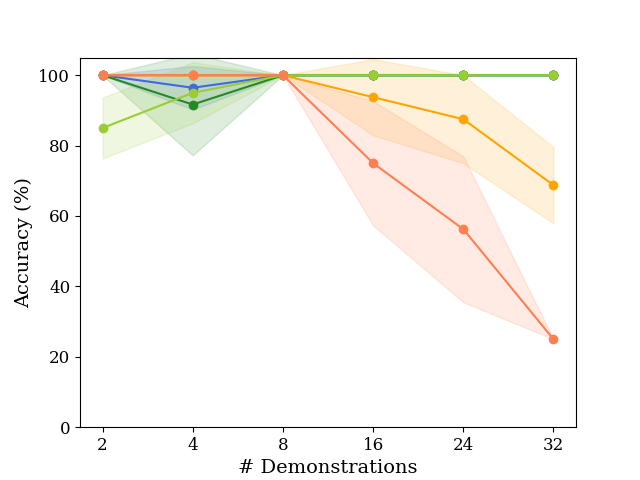}
    \includegraphics[width=0.32\textwidth]            {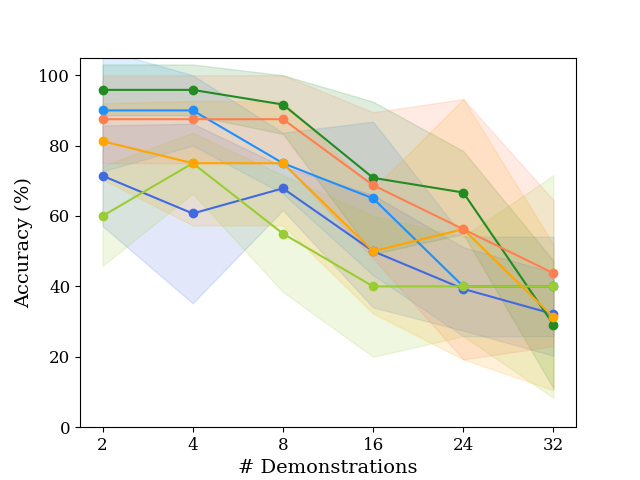}
    \caption{$y_2$}
\end{subfigure}
\end{minipage}
    \hfill
    \begin{minipage}[c]{\linewidth}
        \centering
            \begin{subfigure}{0.4 \linewidth}
            \centering
                        \vspace{2mm}
            \includegraphics[width=\textwidth]{latex/figures/lineplots/legend.pdf}
        \end{subfigure}%
    \end{minipage}
    \hfill
    \vspace{-2.5mm}
\begin{minipage}[c]{\linewidth}
    \caption{WNLI dataset (GPT Neo 2.7B)}
\end{minipage}
\end{figure}
\vspace{-7mm}

\begin{figure}[h!]
\vspace{-2.5mm}
\centering
\begin{minipage}[t]{\linewidth}
\begin{subfigure}{\linewidth}
    \centering
    \includegraphics[width=0.32\textwidth]            {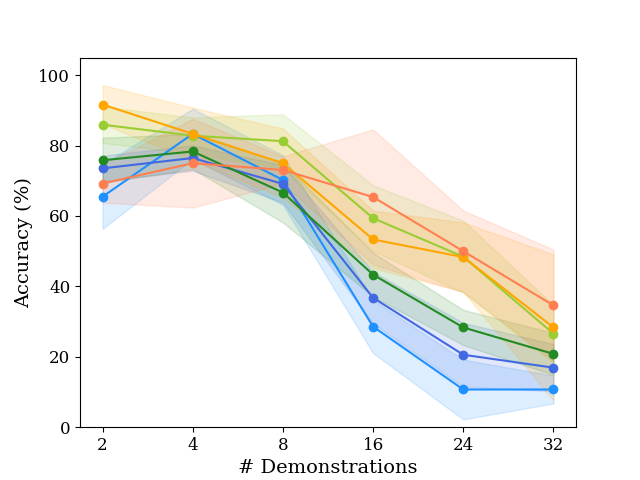}
    \includegraphics[width=0.32\textwidth]            {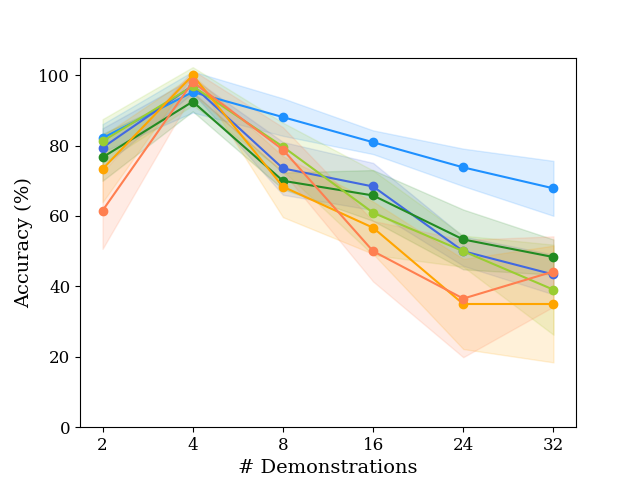}
    \includegraphics[width=0.32\textwidth]            {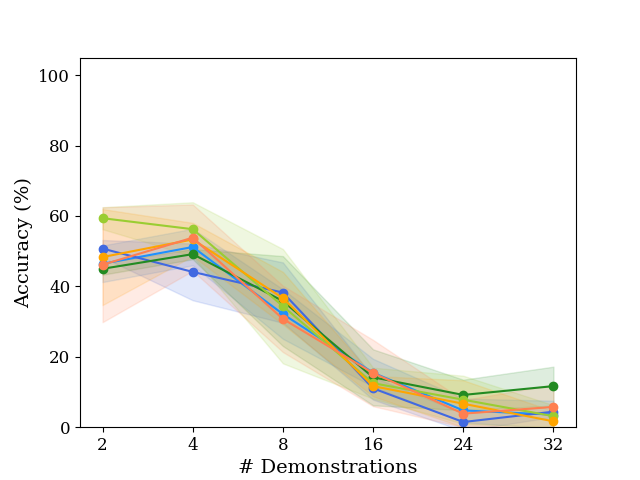}
    \caption{$y_1$}
\end{subfigure}
    \vspace{-2.5mm}
\begin{subfigure}{\linewidth}
    \centering
    \includegraphics[width=0.32\textwidth]            {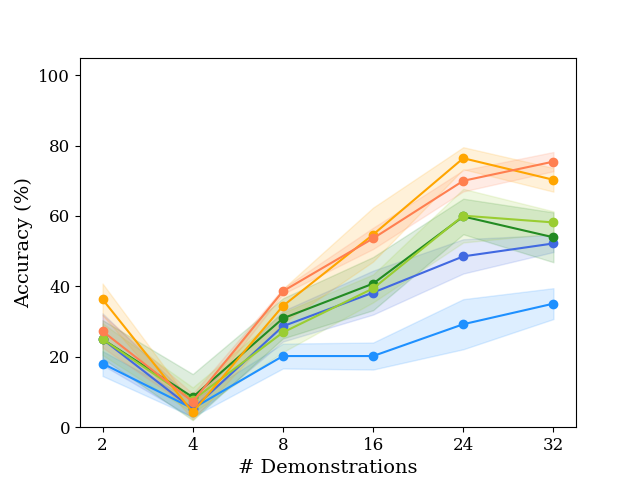}
    \includegraphics[width=0.32\textwidth]            {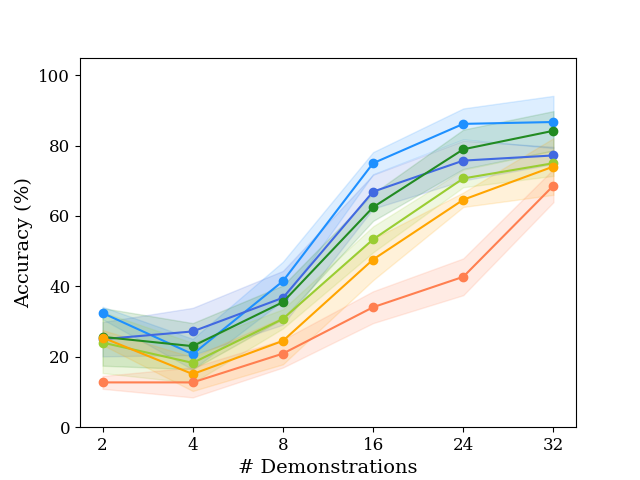}
    \includegraphics[width=0.32\textwidth]            {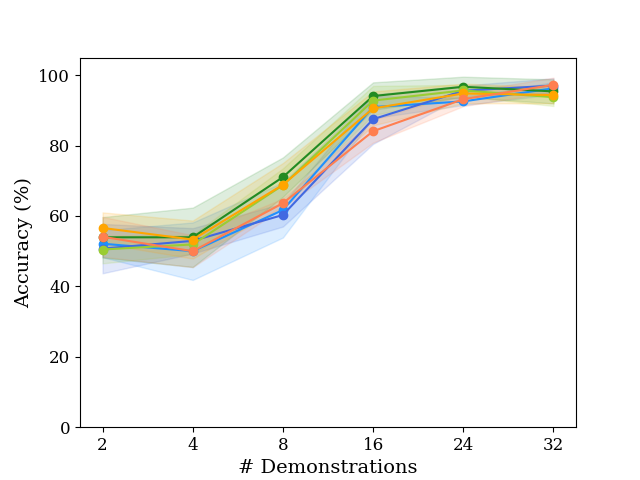}
    \caption{$y_2$}
\end{subfigure}
\end{minipage}
    \hfill
    \begin{minipage}[c]{\linewidth}
        \centering
            \begin{subfigure}{0.4 \linewidth}
            \centering
            \vspace{2mm}
            \includegraphics[width=\textwidth]{latex/figures/lineplots/legend.pdf}
        \end{subfigure}%
    \end{minipage}
    \hfill
    \vspace{-2.5mm}
\begin{minipage}[c]{\linewidth}
    \caption{MPRC dataset (GPT Neo 2.7B)}
\end{minipage}
\end{figure}
\vspace{-7mm}

\begin{figure}[h!]
\vspace{-2.5mm}
\centering
\begin{minipage}[t]{\linewidth}
\begin{subfigure}{\linewidth}
    \centering
    \includegraphics[width=0.32\textwidth]            {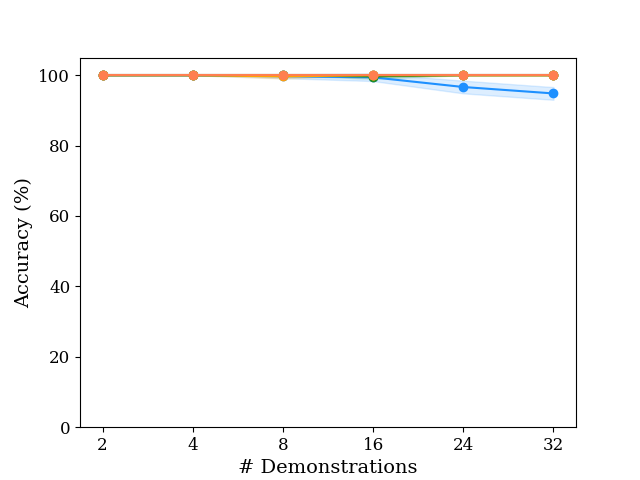}
    \includegraphics[width=0.32\textwidth]            {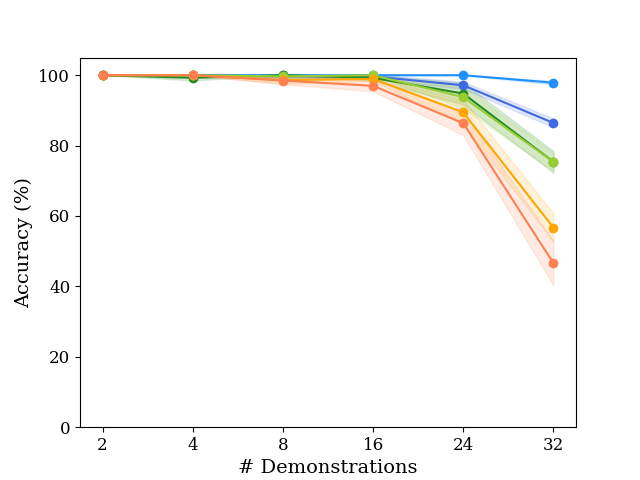}
    \includegraphics[width=0.32\textwidth]            {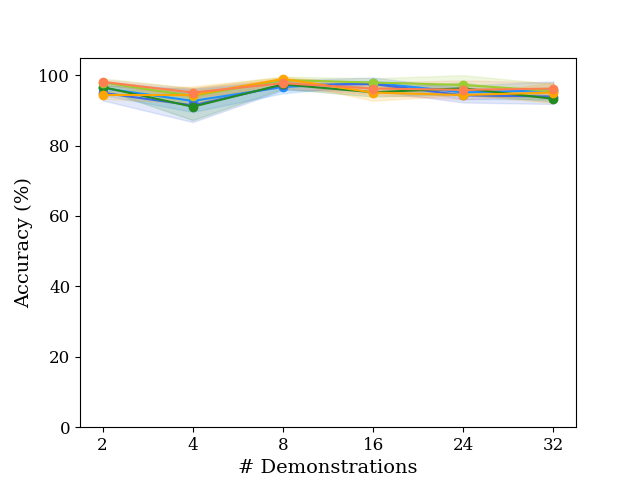}
    \caption{$y_1$}
\end{subfigure}
    \vspace{-2.5mm}
\begin{subfigure}{\linewidth}
    \centering
    \includegraphics[width=0.32\textwidth]            {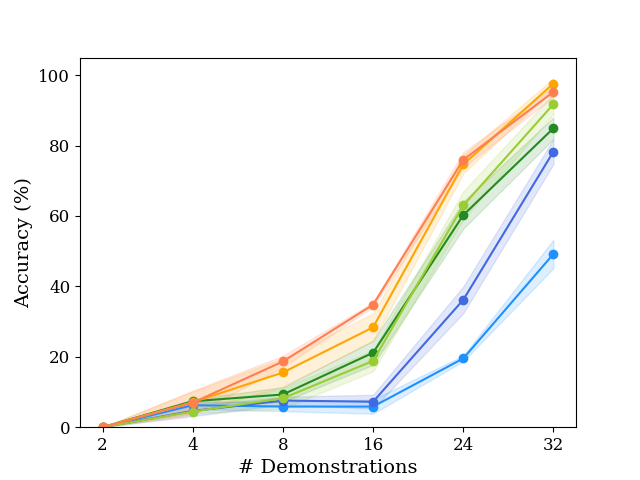}
    \includegraphics[width=0.32\textwidth]            {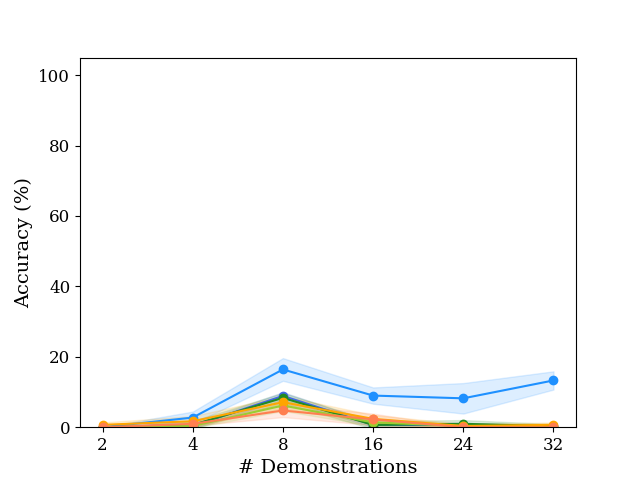}
    \includegraphics[width=0.32\textwidth]            {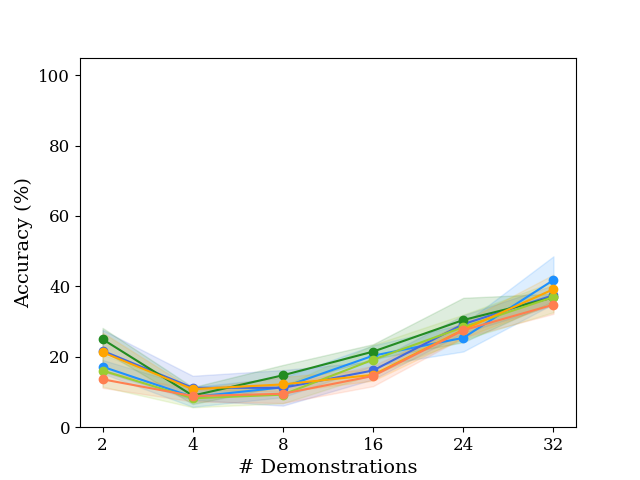}
    \caption{$y_2$}
\end{subfigure}
\end{minipage}
    \hfill
    \begin{minipage}[c]{\linewidth}
        \centering
            \begin{subfigure}{0.4 \linewidth}
            \centering
            \vspace{2mm}
            \includegraphics[width=\textwidth]{latex/figures/lineplots/legend.pdf}
        \end{subfigure}%
    \end{minipage}
    \hfill
    \vspace{-2.5mm}
\begin{minipage}[c]{\linewidth}
    \caption{SST-2 dataset (GPT Neo 2.7B)}
\end{minipage}
\end{figure}
\vspace{-7mm}

\clearpage

\subsection{Additional Length Difference Results}\label{app:len-diff-results}
Each of the following figures shows validation performance when varying the sampling percentage from each class using Llama3 8B and GPT Neo 2.7B. Bin 0 contains the shortest demonstrations and Bin 5 contains the longest demonstrations. Each subfigure shows the validation accuracy on a single class when in-context instances belonging to the respective class were sampled from long instances (left) and short instances (right).
\begin{figure}[h!]
\vspace{-2.5mm}
\centering
\begin{minipage}[t]{\linewidth}
\begin{subfigure}{\linewidth}
    \centering
    \includegraphics[width=0.48\textwidth]            {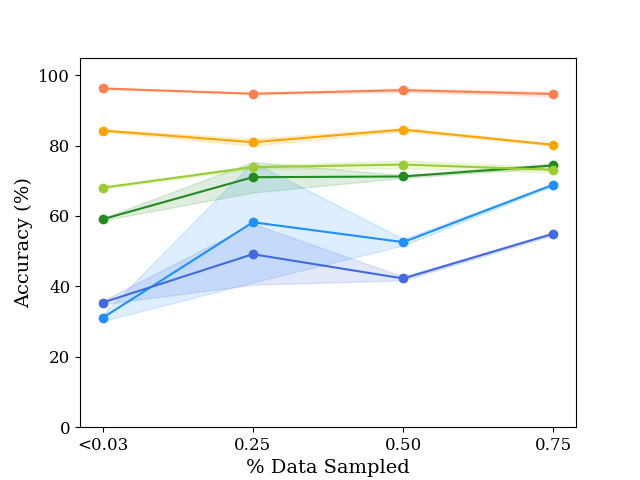}
    \includegraphics[width=0.48\textwidth]            {latex/figures/lineplots/length/hans_llama3_8b_class2_cls1.png}
    \caption{$y_1$}
\end{subfigure}
    \vspace{-2.5mm}
\begin{subfigure}{\linewidth}
    \centering
    \includegraphics[width=0.48\textwidth]            {latex/figures/lineplots/length/hans_llama3_8b_class2_cls2.png}
    \includegraphics[width=0.48\textwidth]            {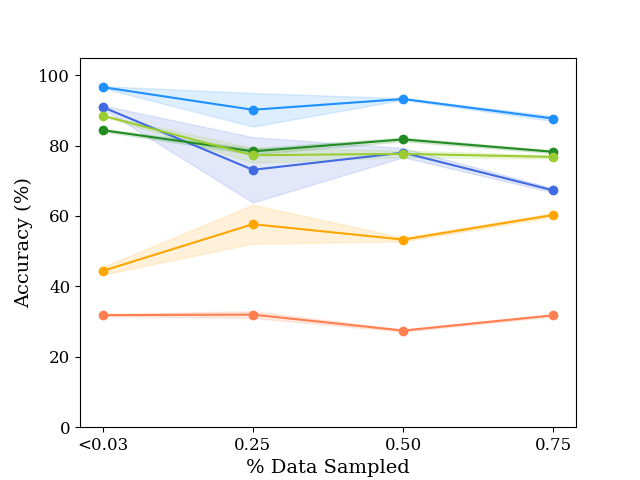}
    \caption{$y_2$}
\end{subfigure}
\end{minipage}
    \hfill
    \begin{minipage}[c]{\linewidth}
        \centering
            \begin{subfigure}{0.4 \linewidth}
            \centering
                        \vspace{2mm}
            \includegraphics[width=\textwidth]{latex/figures/lineplots/legend.pdf}
        \end{subfigure}%
    \end{minipage}
    \hfill
    \vspace{-2.5mm}
\begin{minipage}[c]{\linewidth}
    \caption{Hans dataset (Llama 3 8B)}
\end{minipage}
\end{figure}
\vspace{-5mm}

\begin{figure}[h!]
\vspace{-2.5mm}
\centering
\begin{minipage}[t]{\linewidth}
\begin{subfigure}{\linewidth}
    \centering
    \includegraphics[width=0.48\textwidth]            {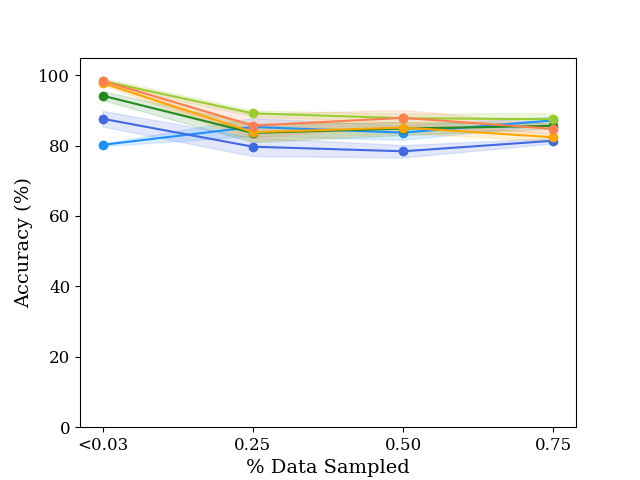}
    \includegraphics[width=0.48\textwidth]            {latex/figures/lineplots/length/en_llama3_8b_class2_cls1.png}
    \caption{$y_1$}
\end{subfigure}
    \vspace{-2.5mm}
\begin{subfigure}{\linewidth}
    \centering
    \includegraphics[width=0.48\textwidth]            {latex/figures/lineplots/length/en_llama3_8b_class2_cls2.png}
    \includegraphics[width=0.48\textwidth]            {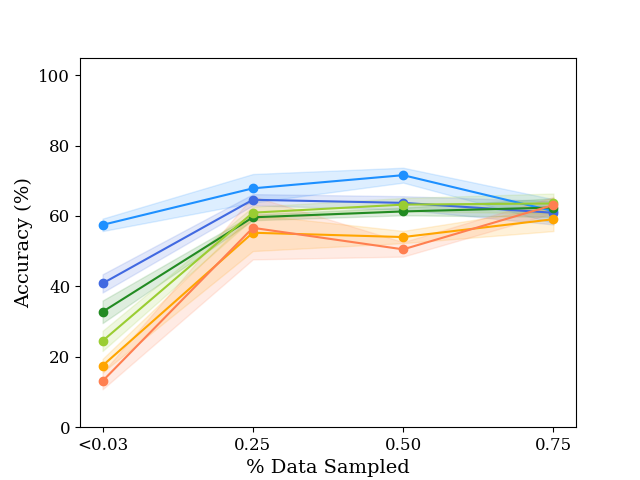}
    \caption{$y_2$}
\end{subfigure}
\end{minipage}
    \hfill
    \begin{minipage}[c]{\linewidth}
        \centering
            \begin{subfigure}{0.4 \linewidth}
            \centering
                        \vspace{2mm}
            \includegraphics[width=\textwidth]{latex/figures/lineplots/legend.pdf}
        \end{subfigure}%
    \end{minipage}
    \hfill
    \vspace{-2.5mm}
\begin{minipage}[c]{\linewidth}
    \caption{PAWS-X$_{\textsc{EN}}$ dataset (Llama 3 8B)}
\end{minipage}
\end{figure}
\vspace{-5mm}

\begin{figure}[h!]
\vspace{-2.5mm}
\centering
\begin{minipage}[t]{\linewidth}
\begin{subfigure}{\linewidth}
    \centering
    \includegraphics[width=0.48\textwidth]            {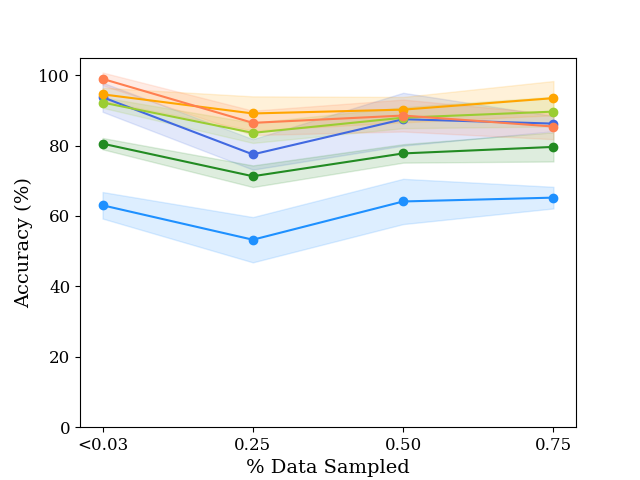}
    \includegraphics[width=0.48\textwidth]            {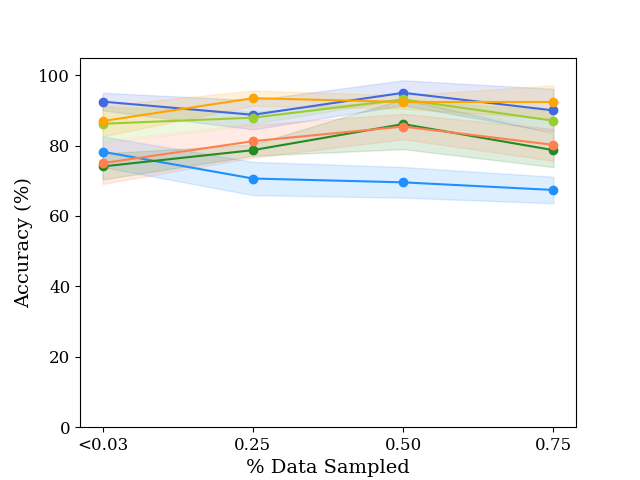}
    \caption{$y_1$}
\end{subfigure}
    \vspace{-2.5mm}
\begin{subfigure}{\linewidth}
    \centering
    \includegraphics[width=0.48\textwidth]            {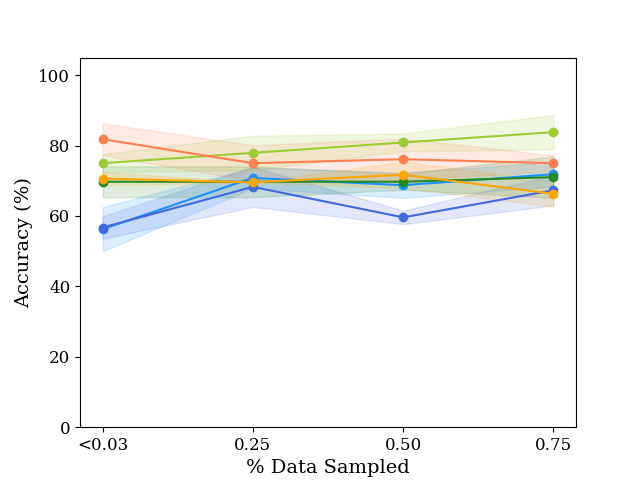}
    \includegraphics[width=0.48\textwidth]            {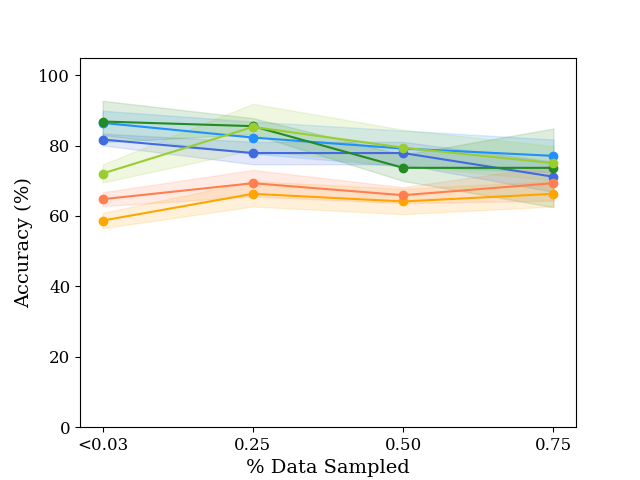}
    \caption{$y_2$}
\end{subfigure}
\end{minipage}
    \hfill
    \begin{minipage}[c]{\linewidth}
        \centering
            \begin{subfigure}{0.4 \linewidth}
            \centering
                        \vspace{2mm}
            \includegraphics[width=\textwidth]{latex/figures/lineplots/legend.pdf}
        \end{subfigure}%
    \end{minipage}
    \hfill
    \vspace{-2.5mm}
\begin{minipage}[c]{\linewidth}
    \caption{RTE dataset (Llama 3 8B)}
\end{minipage}
\end{figure}
\vspace{-7mm}

\begin{figure}[h!]
\vspace{-2.5mm}
\centering
\begin{minipage}[t]{\linewidth}
\begin{subfigure}{\linewidth}
    \centering
    \includegraphics[width=0.48\textwidth]            {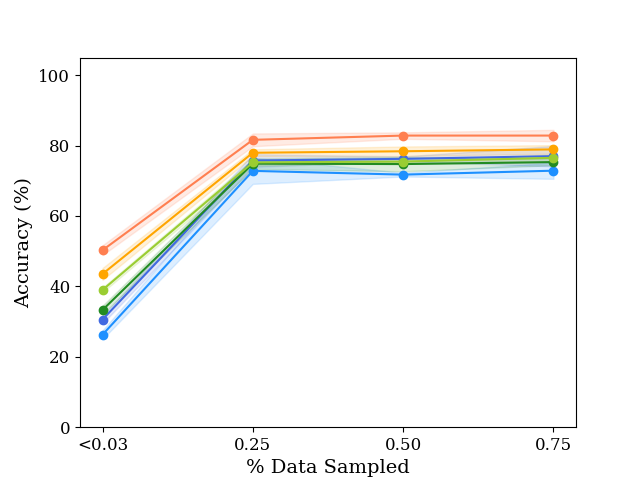}
    \includegraphics[width=0.48\textwidth]            {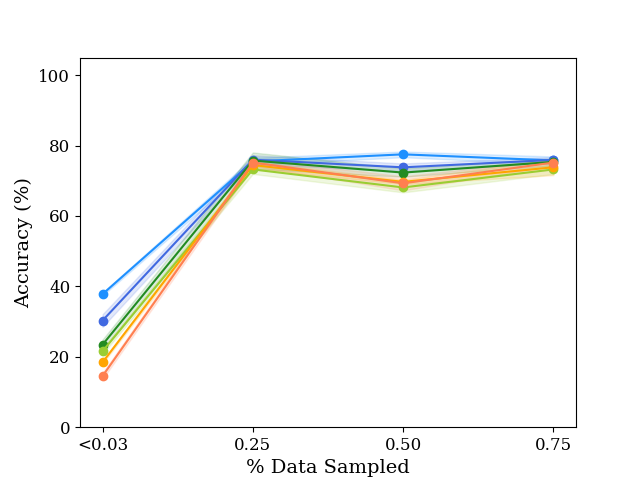}
    \caption{$y_1$}
\end{subfigure}
    \vspace{-2.5mm}
\begin{subfigure}{\linewidth}
    \centering
    \includegraphics[width=0.48\textwidth]            {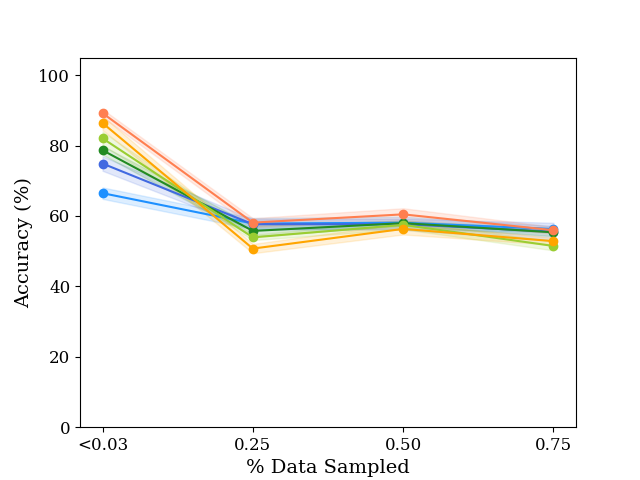}
    \includegraphics[width=0.48\textwidth]            {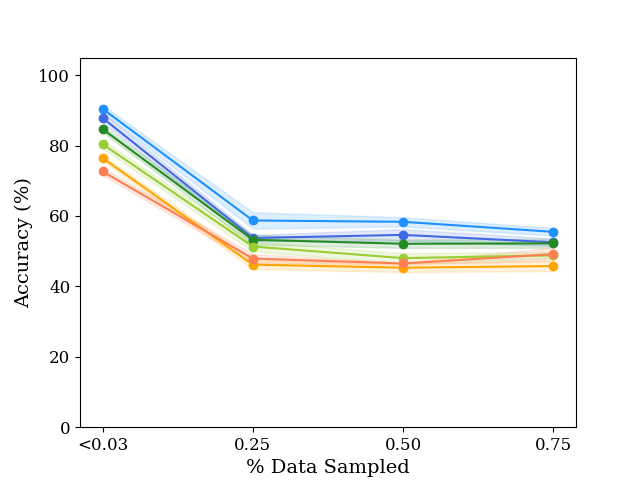}
    \caption{$y_2$}
\end{subfigure}
\end{minipage}
    \hfill
    \begin{minipage}[c]{\linewidth}
        \centering
            \begin{subfigure}{0.4 \linewidth}
            \centering
                        \vspace{2mm}
            \includegraphics[width=\textwidth]{latex/figures/lineplots/legend.pdf}
        \end{subfigure}%
    \end{minipage}
    \hfill
\vspace{-2.5mm}
\begin{minipage}[c]{\linewidth}
    \caption{QNLI dataset (Llama 3 8B)}
\end{minipage}
\end{figure}
\vspace{-7mm}

\begin{figure}[h!]
\vspace{-2.5mm}
\centering
\begin{minipage}[t]{\linewidth}
\begin{subfigure}{\linewidth}
    \centering
    \includegraphics[width=0.48\textwidth]            {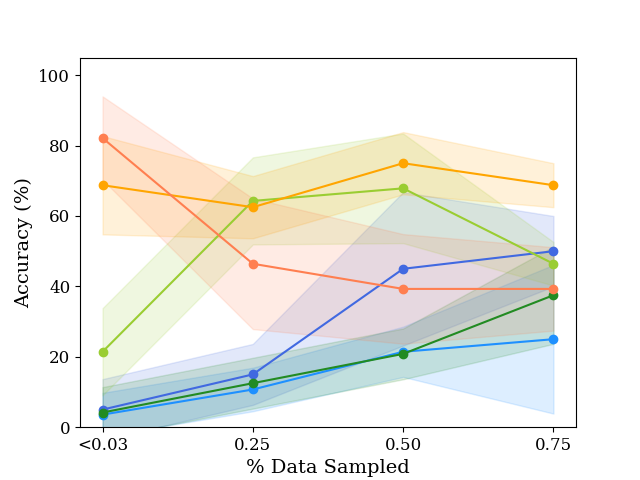}
    \includegraphics[width=0.48\textwidth]            {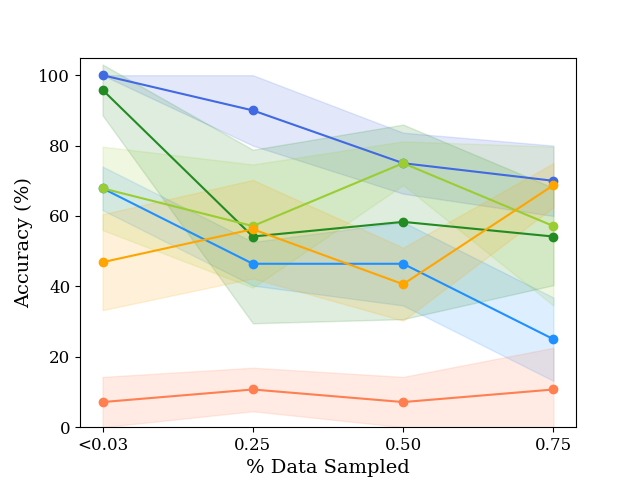}
    \caption{$y_1$}
\end{subfigure}
    \vspace{-2.5mm}
\begin{subfigure}{\linewidth}
    \centering
    \includegraphics[width=0.48\textwidth]            {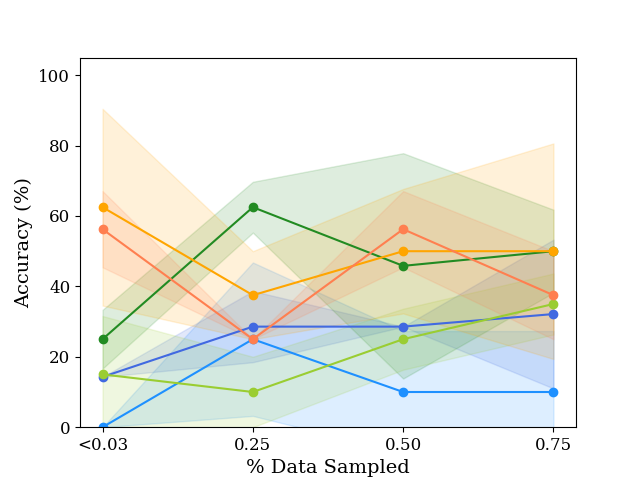}
    \includegraphics[width=0.48\textwidth]            {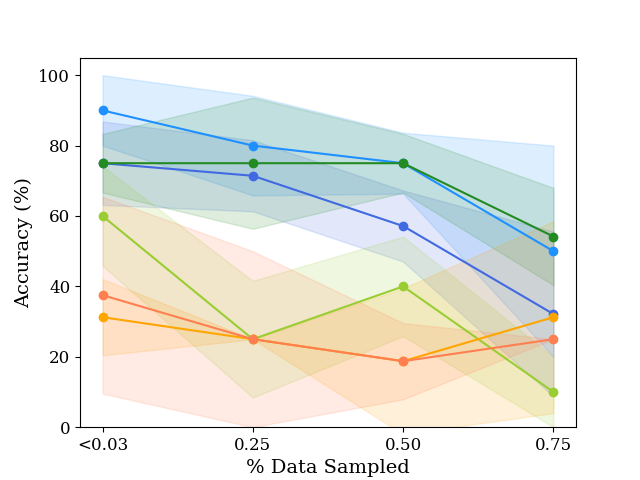}
    \caption{$y_2$}
\end{subfigure}
\end{minipage}
    \hfill
    \begin{minipage}[c]{\linewidth}
        \centering
            \begin{subfigure}{0.4 \linewidth}
            \centering
                        \vspace{2mm}
            \includegraphics[width=\textwidth]{latex/figures/lineplots/legend.pdf}
        \end{subfigure}%
    \end{minipage}
    \hfill
    \vspace{-2.5mm}
\begin{minipage}[c]{\linewidth}
    \caption{WNLI dataset (Llama 3 8B)}
\end{minipage}
\end{figure}
\vspace{-7mm}

\begin{figure}[h!]
\vspace{-2.5mm}
\centering
\begin{minipage}[t]{\linewidth}
\begin{subfigure}{\linewidth}
    \centering
    \includegraphics[width=0.48\textwidth]            {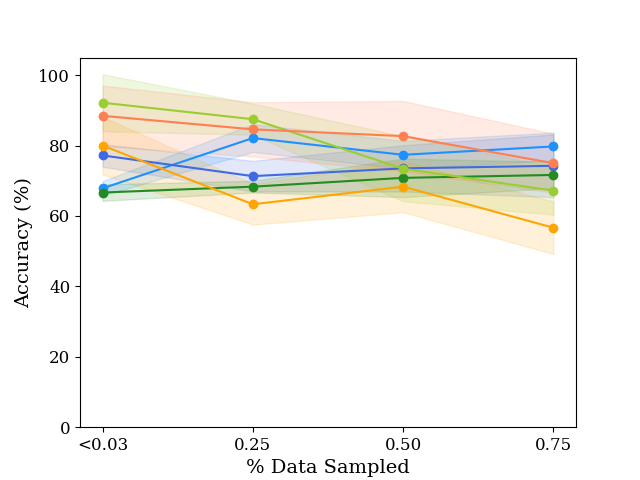}
    \includegraphics[width=0.48\textwidth]            {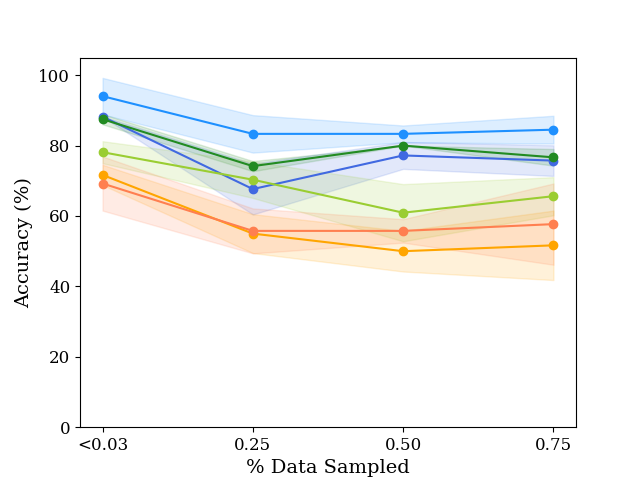}
    \caption{$y_1$}
\end{subfigure}
    \vspace{-2.5mm}
\begin{subfigure}{\linewidth}
    \centering
    \includegraphics[width=0.48\textwidth]            {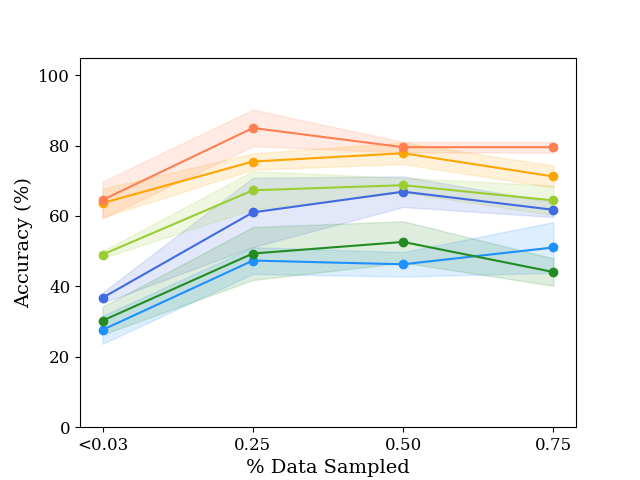}
    \includegraphics[width=0.48\textwidth]            {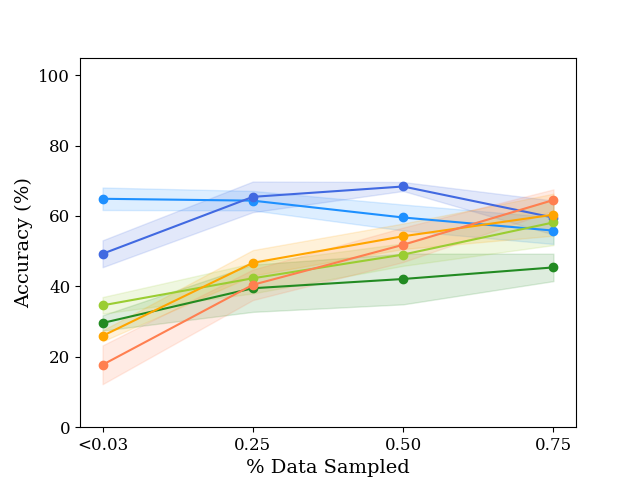}
    \caption{$y_2$}
\end{subfigure}
\end{minipage}
    \hfill
    \begin{minipage}[c]{\linewidth}
        \centering
            \begin{subfigure}{0.4 \linewidth}
            \centering
            \vspace{2mm}
            \includegraphics[width=\textwidth]{latex/figures/lineplots/legend.pdf}
        \end{subfigure}%
    \end{minipage}
    \hfill
    \vspace{-2.5mm}
\begin{minipage}[c]{\linewidth}
    \caption{MPRC dataset (Llama 3 8B)}
\end{minipage}
\end{figure}
\vspace{-7mm}

\begin{figure}[h!]
\vspace{-2.5mm}
\centering
\begin{minipage}[t]{\linewidth}
\begin{subfigure}{\linewidth}
    \centering
    \includegraphics[width=0.48\textwidth]            {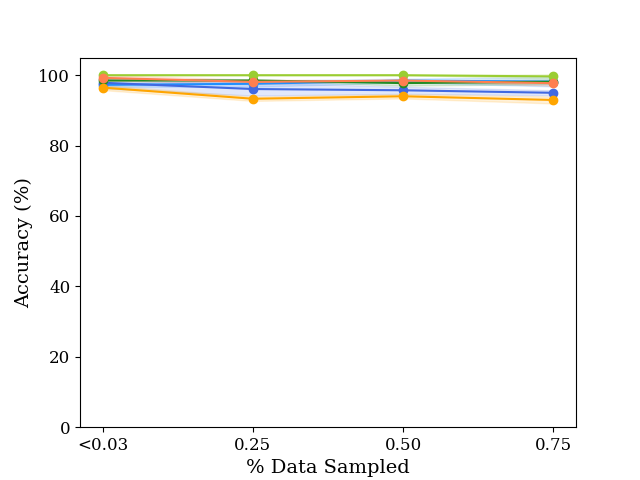}
    \includegraphics[width=0.48\textwidth]            {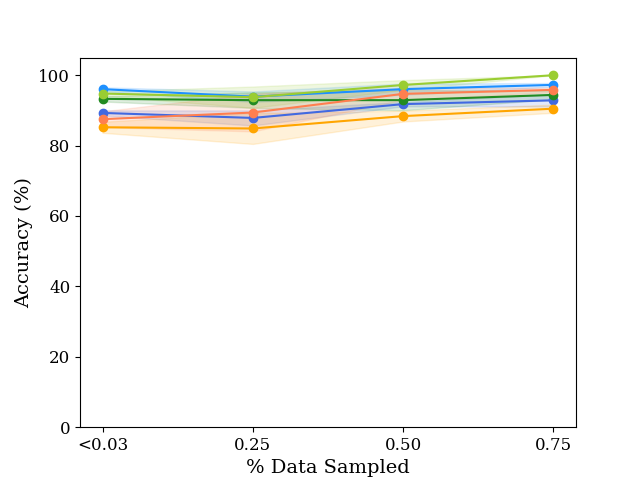}
    \caption{$y_1$}
\end{subfigure}
    \vspace{-2.5mm}
\begin{subfigure}{\linewidth}
    \centering
    \includegraphics[width=0.48\textwidth]            {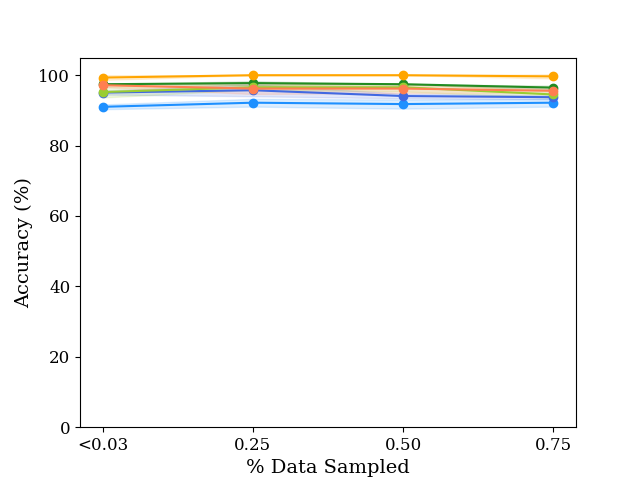}
    \includegraphics[width=0.48\textwidth]            {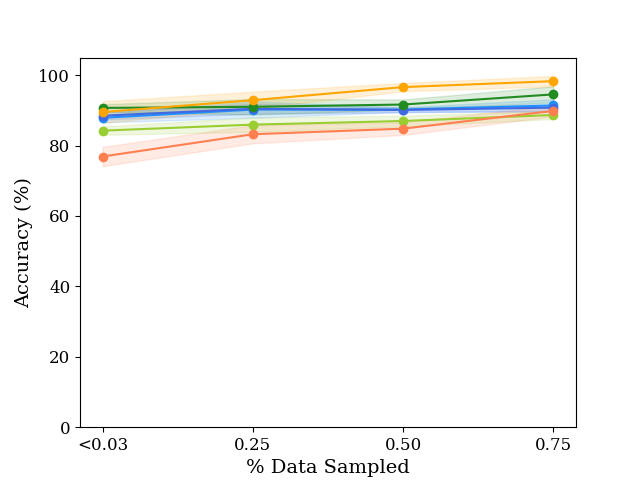}
    \caption{$y_2$}
\end{subfigure}
\end{minipage}
    \hfill
    \begin{minipage}[c]{\linewidth}
        \centering
            \begin{subfigure}{0.4 \linewidth}
            \centering
            \vspace{2mm}
            \includegraphics[width=\textwidth]{latex/figures/lineplots/legend.pdf}
        \end{subfigure}%
    \end{minipage}
    \hfill
    \vspace{-2.5mm}
\begin{minipage}[c]{\linewidth}
    \caption{SST-2 dataset (Llama 3 8B)}
\end{minipage}
\end{figure}
\vspace{-7mm}
\begin{figure}[h!]
\vspace{-2.5mm}
\centering
\begin{minipage}[t]{\linewidth}
\begin{subfigure}{\linewidth}
    \centering
    \includegraphics[width=0.48\textwidth]            {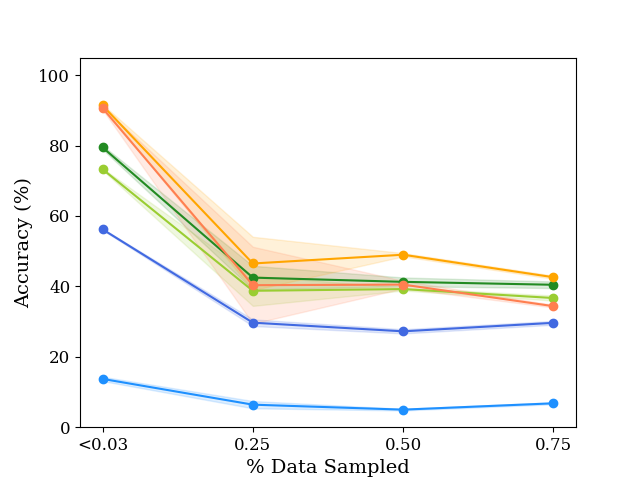}
    \includegraphics[width=0.48\textwidth]            {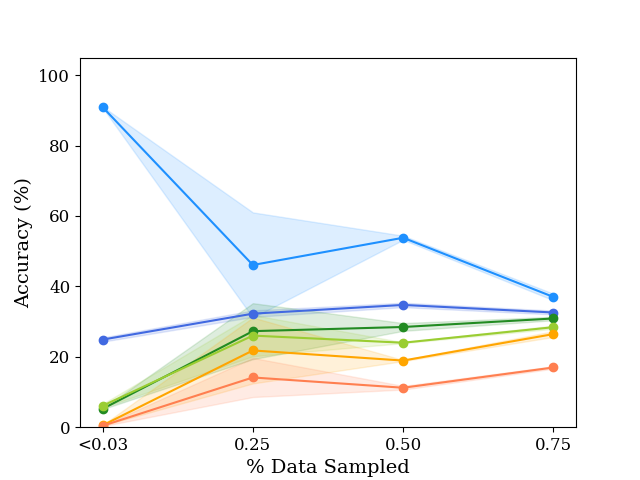}
    \caption{$y_1$}
\end{subfigure}
    \vspace{-2.5mm}
\begin{subfigure}{\linewidth}
    \centering
    \includegraphics[width=0.48\textwidth]            {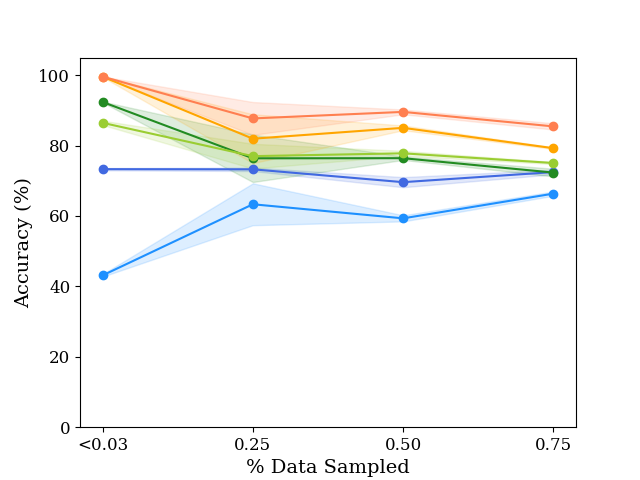}
    \includegraphics[width=0.48\textwidth]            {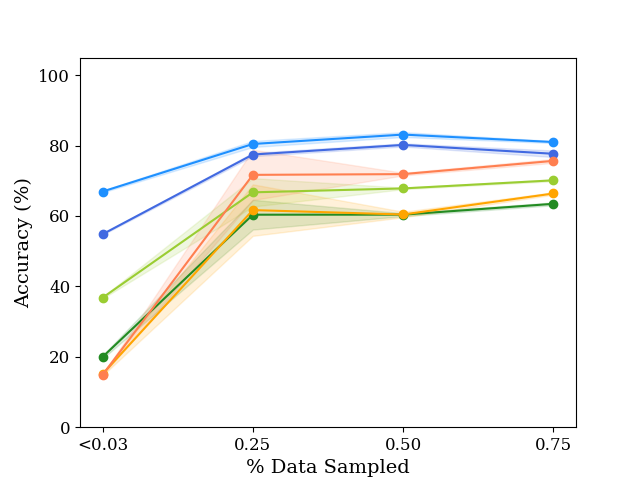}
    \caption{$y_2$}
\end{subfigure}
\end{minipage}
    \hfill
    \begin{minipage}[c]{\linewidth}
        \centering
            \begin{subfigure}{0.4 \linewidth}
            \centering
                        \vspace{2mm}
            \includegraphics[width=\textwidth]{latex/figures/lineplots/legend.pdf}
        \end{subfigure}%
    \end{minipage}
    \hfill
    \vspace{-2.5mm}
\begin{minipage}[c]{\linewidth}
    \caption{Hans dataset (GPT Neo 2.7B)}
\end{minipage}
\end{figure}
\vspace{-5mm}

\begin{figure}[h!]
\vspace{-2.5mm}
\centering
\begin{minipage}[t]{\linewidth}
\begin{subfigure}{\linewidth}
    \centering
    \includegraphics[width=0.48\textwidth]            {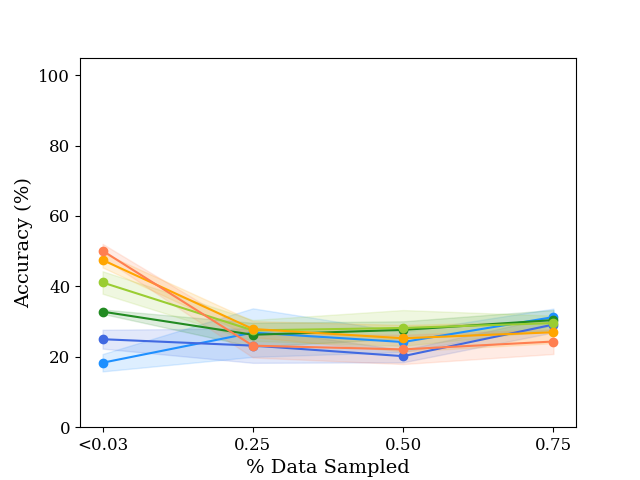}
    \includegraphics[width=0.48\textwidth]            {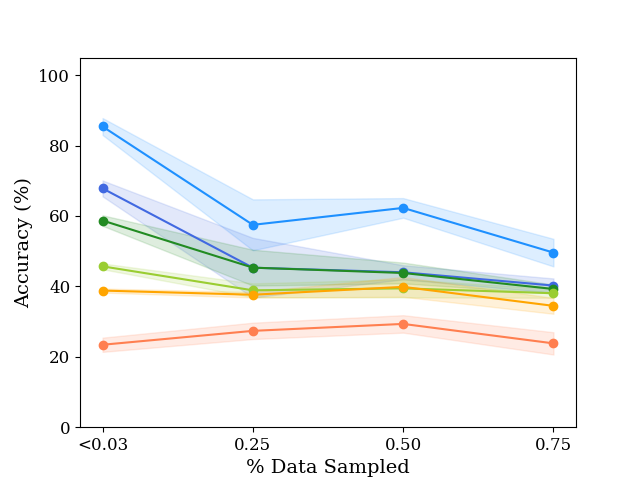}
    \caption{$y_1$}
\end{subfigure}
    \vspace{-2.5mm}
\begin{subfigure}{\linewidth}
    \centering
    \includegraphics[width=0.48\textwidth]            {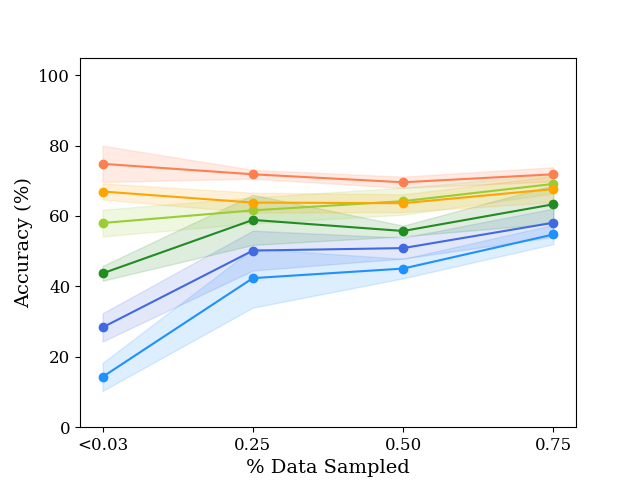}
    \includegraphics[width=0.48\textwidth]            {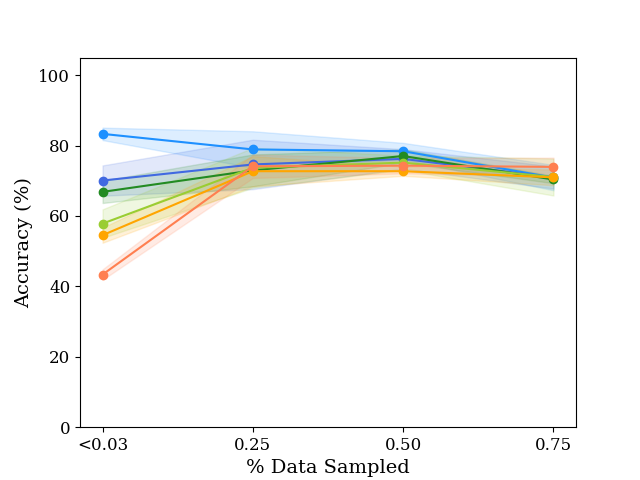}
    \caption{$y_2$}
\end{subfigure}
\end{minipage}
    \hfill
    \begin{minipage}[c]{\linewidth}
        \centering
            \begin{subfigure}{0.4 \linewidth}
            \centering
                        \vspace{2mm}
            \includegraphics[width=\textwidth]{latex/figures/lineplots/legend.pdf}
        \end{subfigure}%
    \end{minipage}
    \hfill
    \vspace{-2.5mm}
\begin{minipage}[c]{\linewidth}
    \caption{PAWS-X$_{\textsc{EN}}$ dataset (GPT Neo 2.7B)}
\end{minipage}
\end{figure}
\vspace{-5mm}

\begin{figure}[h!]
\vspace{-2.5mm}
\centering
\begin{minipage}[t]{\linewidth}
\begin{subfigure}{\linewidth}
    \centering
    \includegraphics[width=0.48\textwidth]            {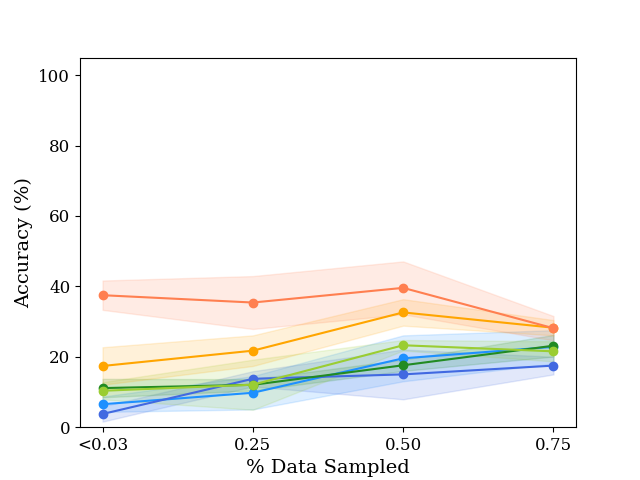}
    \includegraphics[width=0.48\textwidth]            {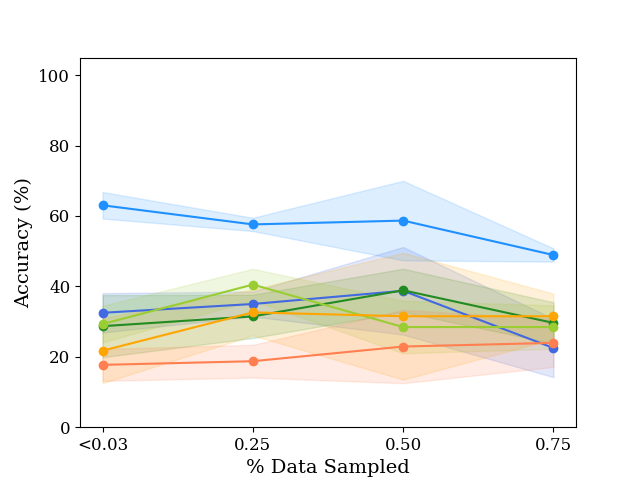}
    \caption{$y_1$}
\end{subfigure}
    \vspace{-2.5mm}
\begin{subfigure}{\linewidth}
    \centering
    \includegraphics[width=0.48\textwidth]            {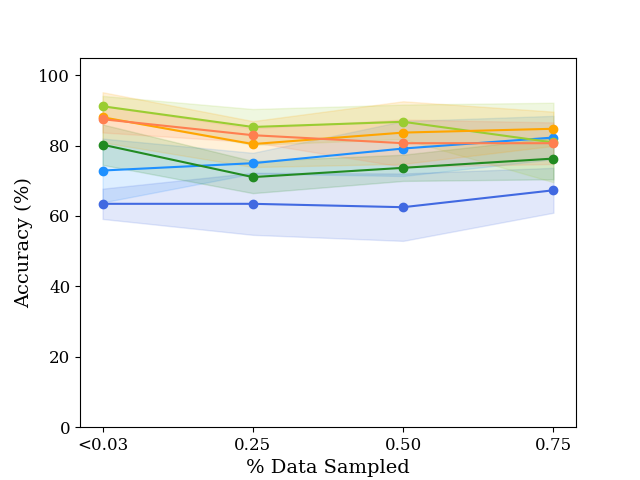}
    \includegraphics[width=0.48\textwidth]            {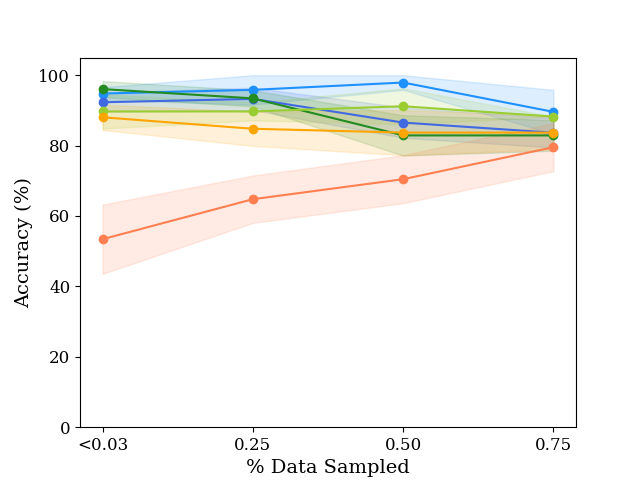}
    \caption{$y_2$}
\end{subfigure}
\end{minipage}
    \hfill
    \begin{minipage}[c]{\linewidth}
        \centering
            \begin{subfigure}{0.4 \linewidth}
            \centering
                        \vspace{2mm}
            \includegraphics[width=\textwidth]{latex/figures/lineplots/legend.pdf}
        \end{subfigure}%
    \end{minipage}
    \hfill
    \vspace{-2.5mm}
\begin{minipage}[c]{\linewidth}
    \caption{RTE dataset (GPT Neo 2.7B)}
\end{minipage}
\end{figure}
\vspace{-7mm}

\begin{figure}[h!]
\vspace{-2.5mm}
\centering
\begin{minipage}[t]{\linewidth}
\begin{subfigure}{\linewidth}
    \centering
    \includegraphics[width=0.48\textwidth]            {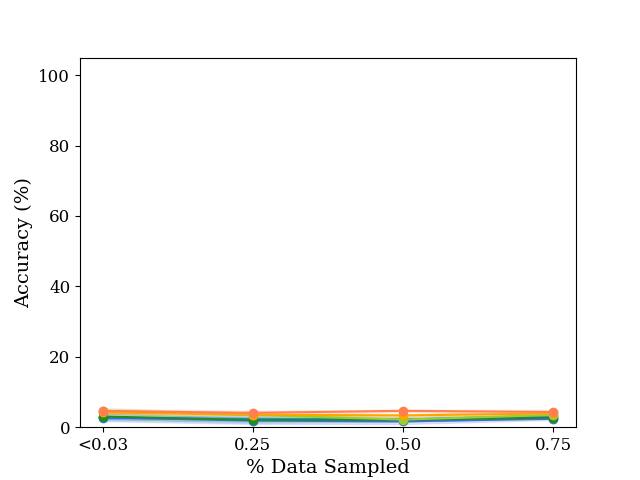}
    \includegraphics[width=0.48\textwidth]            {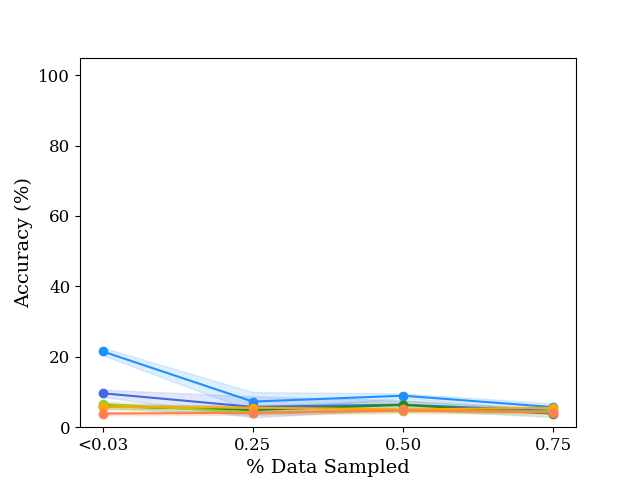}
    \caption{$y_1$}
\end{subfigure}
    \vspace{-2.5mm}
\begin{subfigure}{\linewidth}
    \centering
    \includegraphics[width=0.48\textwidth]            {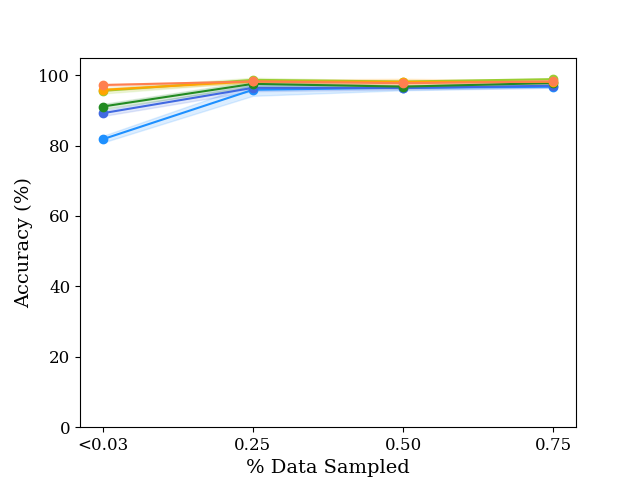}
    \includegraphics[width=0.48\textwidth]            {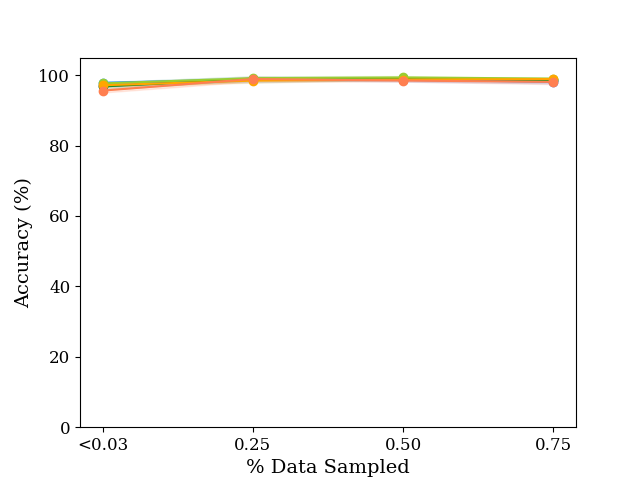}
    \caption{$y_2$}
\end{subfigure}
\end{minipage}
    \hfill
    \begin{minipage}[c]{\linewidth}
        \centering
            \begin{subfigure}{0.4 \linewidth}
            \centering
                        \vspace{2mm}
            \includegraphics[width=\textwidth]{latex/figures/lineplots/legend.pdf}
        \end{subfigure}%
    \end{minipage}
    \hfill
\vspace{-2.5mm}
\begin{minipage}[c]{\linewidth}
    \caption{QNLI dataset (GPT Neo 2.7B)}
\end{minipage}
\end{figure}
\vspace{-7mm}

\begin{figure}[h!]
\vspace{-2.5mm}
\centering
\begin{minipage}[t]{\linewidth}
\begin{subfigure}{\linewidth}
    \centering
    \includegraphics[width=0.48\textwidth]            {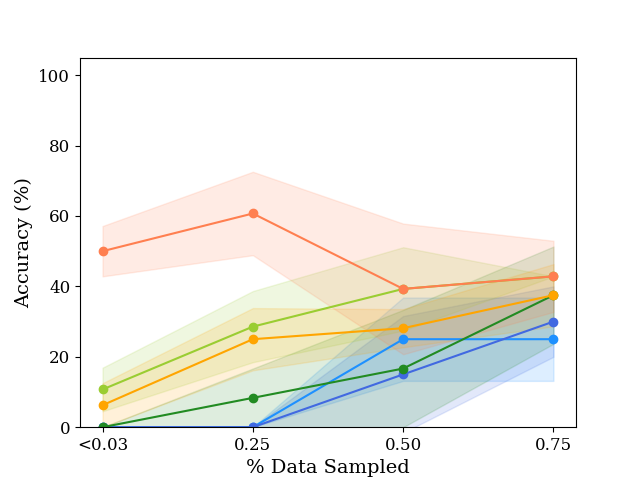}
    \includegraphics[width=0.48\textwidth]            {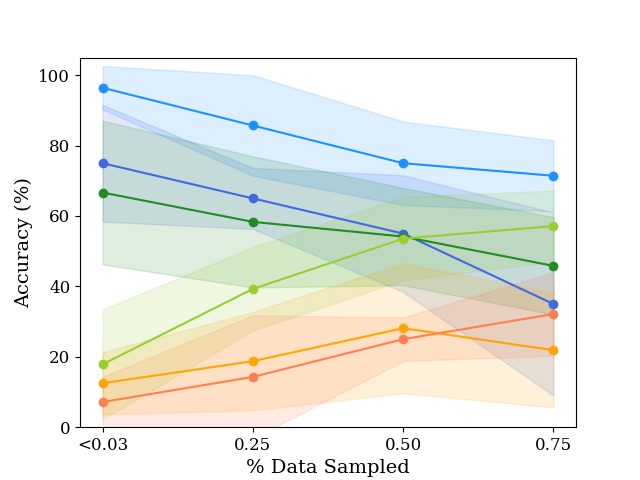}
    \caption{$y_1$}
\end{subfigure}
    \vspace{-2.5mm}
\begin{subfigure}{\linewidth}
    \centering
    \includegraphics[width=0.48\textwidth]            {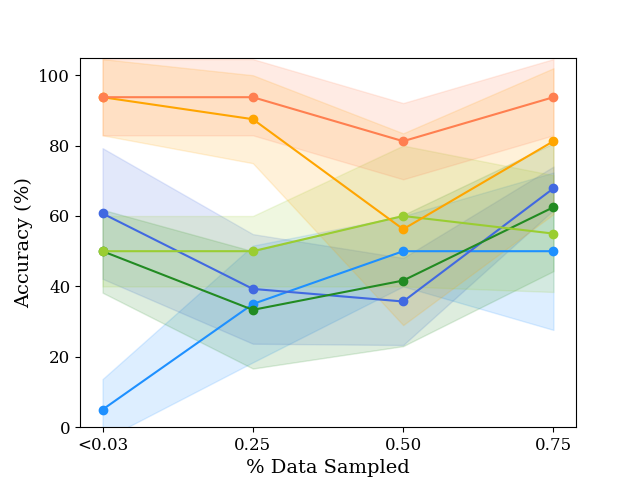}
    \includegraphics[width=0.48\textwidth]            {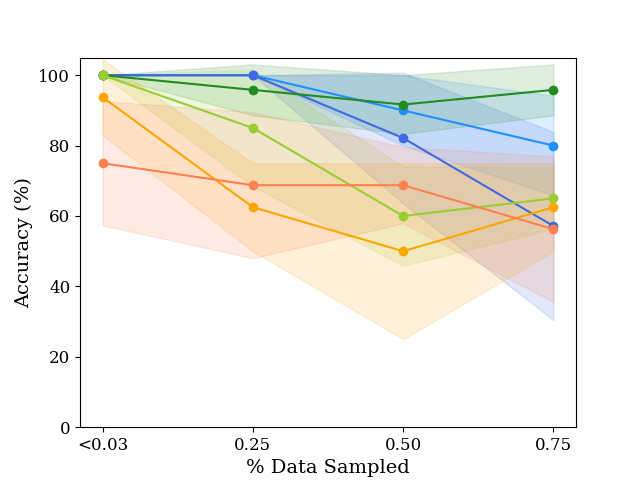}
    \caption{$y_2$}
\end{subfigure}
\end{minipage}
    \hfill
    \begin{minipage}[c]{\linewidth}
        \centering
            \begin{subfigure}{0.4 \linewidth}
            \centering
                        \vspace{2mm}
            \includegraphics[width=\textwidth]{latex/figures/lineplots/legend.pdf}
        \end{subfigure}%
    \end{minipage}
    \hfill
    \vspace{-2.5mm}
\begin{minipage}[c]{\linewidth}
    \caption{WNLI dataset (GPT Neo 2.7B)}
\end{minipage}
\end{figure}
\vspace{-7mm}

\begin{figure}[h!]
\vspace{-2.5mm}
\centering
\begin{minipage}[t]{\linewidth}
\begin{subfigure}{\linewidth}
    \centering
    \includegraphics[width=0.48\textwidth]            {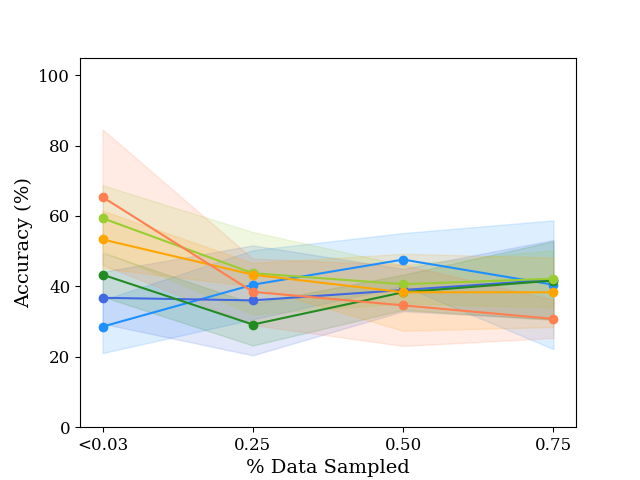}
    \includegraphics[width=0.48\textwidth]            {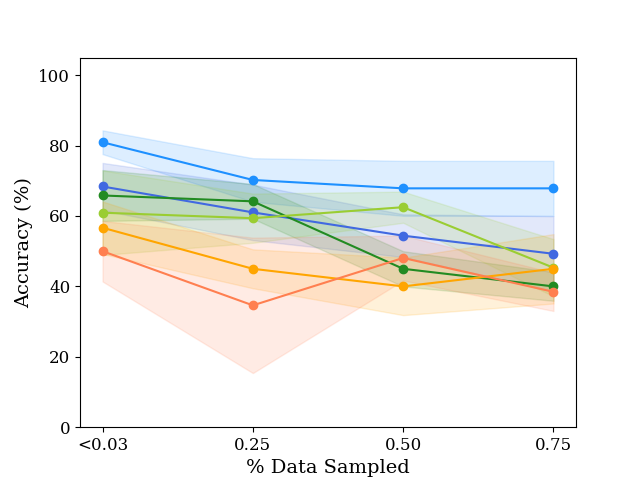}
    \caption{$y_1$}
\end{subfigure}
    \vspace{-2.5mm}
\begin{subfigure}{\linewidth}
    \centering
    \includegraphics[width=0.48\textwidth]            {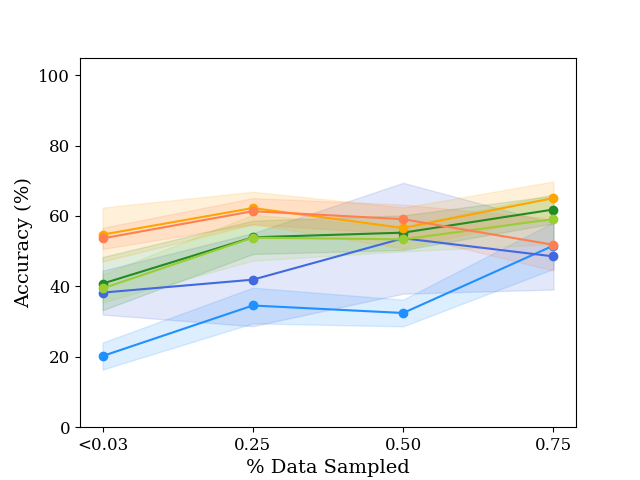}
    \includegraphics[width=0.48\textwidth]            {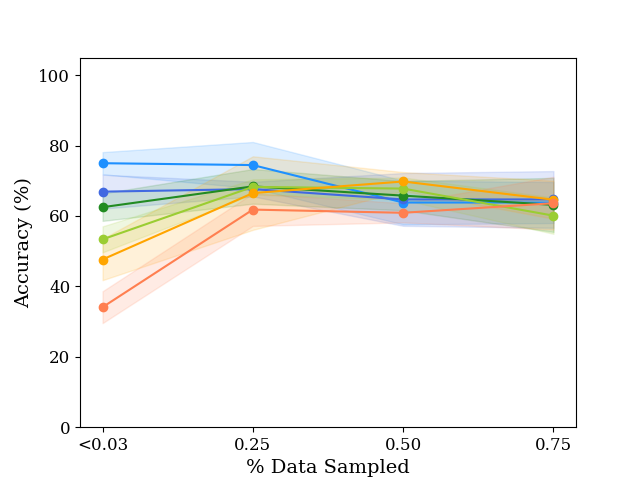}
    \caption{$y_2$}
\end{subfigure}
\end{minipage}
    \hfill
    \begin{minipage}[c]{\linewidth}
        \centering
            \begin{subfigure}{0.4 \linewidth}
            \centering
            \vspace{2mm}
            \includegraphics[width=\textwidth]{latex/figures/lineplots/legend.pdf}
        \end{subfigure}%
    \end{minipage}
    \hfill
    \vspace{-2.5mm}
\begin{minipage}[c]{\linewidth}
    \caption{MPRC dataset (GPT Neo 2.7B)}
\end{minipage}
\end{figure}
\vspace{-7mm}

\begin{figure}[h!]
\vspace{-2.5mm}
\centering
\begin{minipage}[t]{\linewidth}
\begin{subfigure}{\linewidth}
    \centering
    \includegraphics[width=0.48\textwidth]            {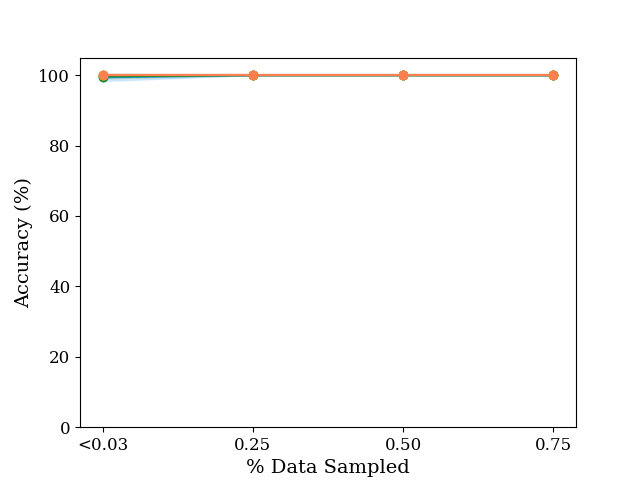}
    \includegraphics[width=0.48\textwidth]            {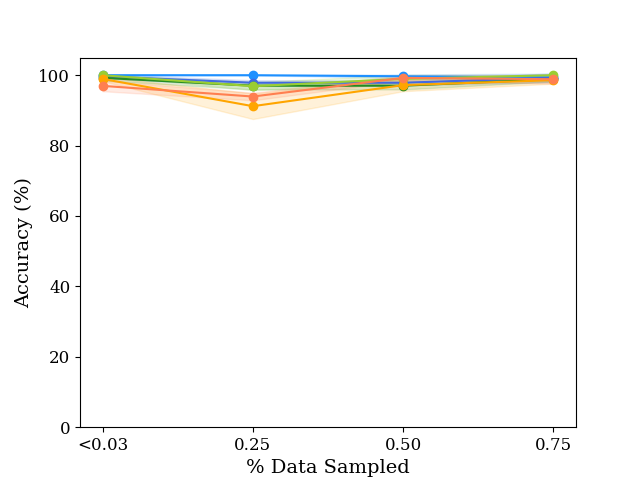}
    \caption{$y_1$}
\end{subfigure}
    \vspace{-2.5mm}
\begin{subfigure}{\linewidth}
    \centering
    \includegraphics[width=0.48\textwidth]            {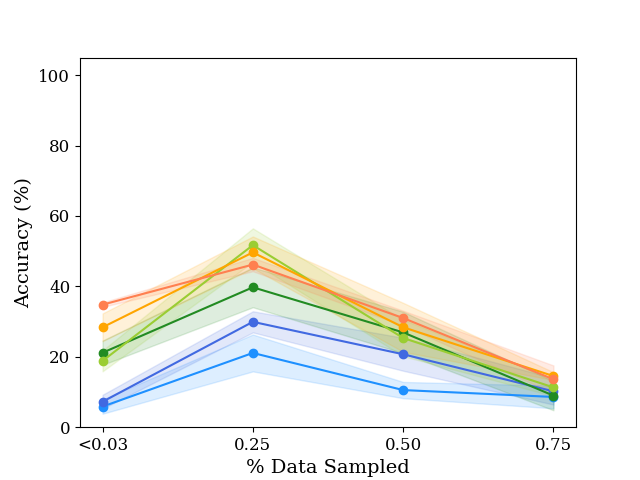}
    \includegraphics[width=0.48\textwidth]            {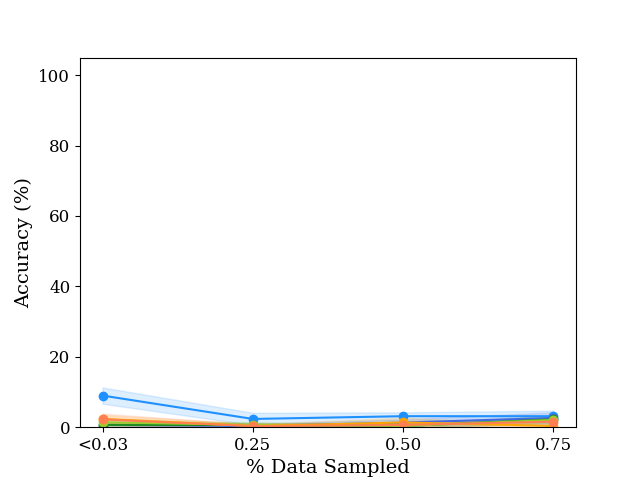}
    \caption{$y_2$}
\end{subfigure}
\end{minipage}
    \hfill
    \begin{minipage}[c]{\linewidth}
        \centering
            \begin{subfigure}{0.4 \linewidth}
            \centering
            \vspace{2mm}
            \includegraphics[width=\textwidth]{latex/figures/lineplots/legend.pdf}
        \end{subfigure}%
    \end{minipage}
    \hfill
    \vspace{-2.5mm}
\begin{minipage}[c]{\linewidth}
    \caption{SST-2 dataset (GPT Neo 2.7B)}
\end{minipage}
\end{figure}
\vspace{-7mm}

\clearpage
\subsection{Additional Intervention Results}\label{app:int-results}
Each of the following figures shows validation set performance on a finetuned Llama 3 8B or GPT Neo 2.7B model exhibiting a length bias. 
For each figure, (a) shows finetuning performance prior to intervention). (b) and (c) show results on two debiasing conditions: ICL demonstrations ($k=16$) sampled from the opposite lengths from what the model saw during finetuning (e.g $y_1$ long demonstrations, $y_2$ short demonstrations), and random sampling, respectively.

\begin{figure*}[t!]
    \centering
    \begin{minipage}[t]{\linewidth}
        \begin{subfigure}{0.31\linewidth}
            \centering
            \includegraphics[width=\textwidth]{latex/figures/ft/rte_200_llama3_8b_class2_6bins.png}
            \caption{Finetuning: $y_1$ (Blue) short demonstrations, $y_2$ (Orange) long demonstrations.}
        \end{subfigure}%
        \hfill
        \begin{subfigure}{0.31\linewidth}
            \centering
            \includegraphics[width=\textwidth]{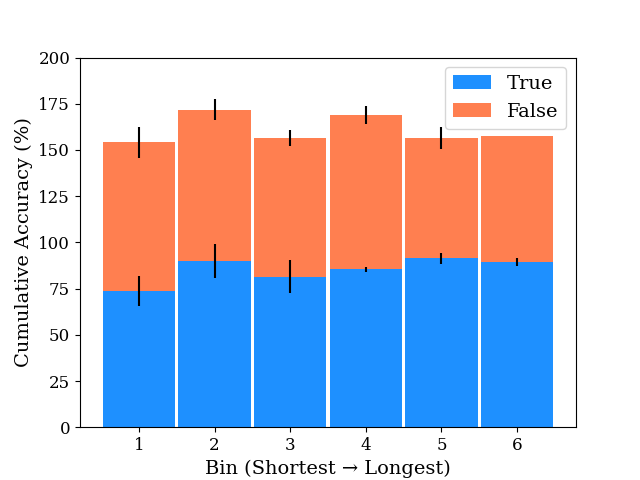}
            \caption{Intervention: $y_1$ (Blue) long demonstrations, $y_2$ (Orange) short demonstrations.}
        \end{subfigure}
        \hfill
        \begin{subfigure}{0.31\linewidth}
            \centering
            \includegraphics[width=\textwidth]{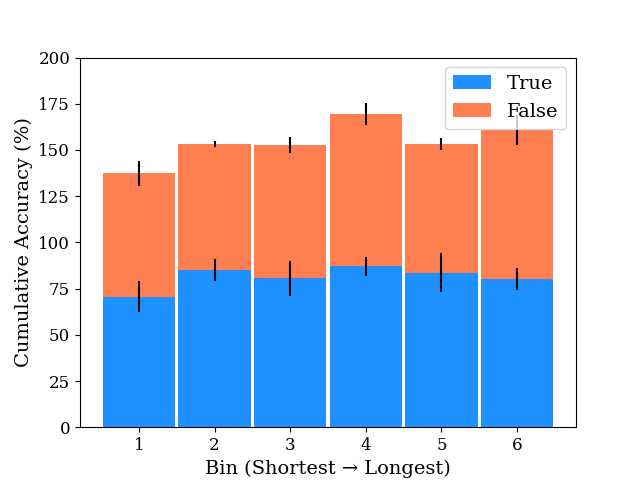}
            \caption{Intervention: $y_1$ (Blue) and $y_2$ (Orange) demonstrations randomly sampled.}
        \end{subfigure}
    \end{minipage}%
    \hfill
    \begin{minipage}[c]{\linewidth}
        \caption{RTE (Llama 3 8B)}
    \end{minipage}
\end{figure*}

\begin{figure*}[t!]
    \centering
    \begin{minipage}[t]{\linewidth}
        \begin{subfigure}{0.31\linewidth}
            \centering
            \includegraphics[width=\textwidth]{latex/figures/ft/qnli_200_llama3_8b_class2_6bins.png}
            \caption{Finetuning: $y_1$ (Blue) short demonstrations, $y_2$ (Orange) long demonstrations.}
        \end{subfigure}%
        \hfill
        \begin{subfigure}{0.31\linewidth}
            \centering
            \includegraphics[width=\textwidth]{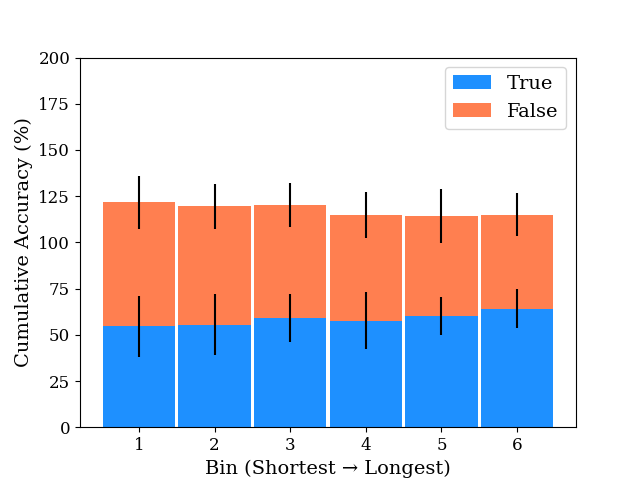}
            \caption{Intervention: $y_1$ (Blue) long demonstrations, $y_2$ (Orange) short demonstrations.}
        \end{subfigure}
        \hfill
        \begin{subfigure}{0.31\linewidth}
            \centering
            \includegraphics[width=\textwidth]{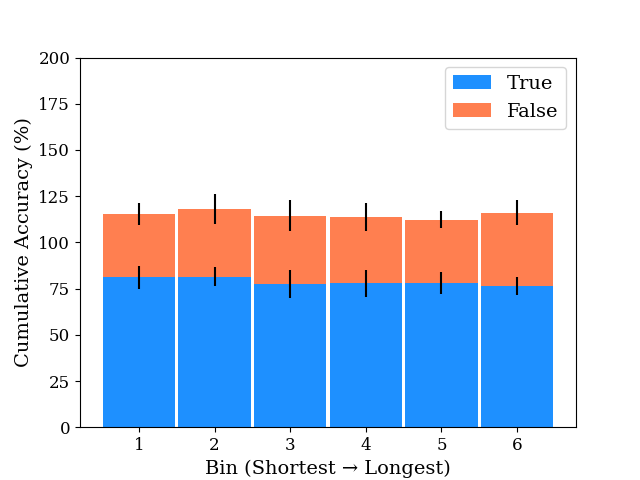}
            \caption{Intervention: $y_1$ (Blue) and $y_2$ (Orange) demonstrations randomly sampled.}
        \end{subfigure}
    \end{minipage}%
    \hfill
    \begin{minipage}[c]{\linewidth}
        \caption{QNLI (Llama 3 8B)}
    \end{minipage}
\end{figure*}

\begin{figure*}[t!]
    \centering
    \begin{minipage}[t]{\linewidth}
        \begin{subfigure}{0.31\linewidth}
            \centering
            \includegraphics[width=\textwidth]{latex/figures/ft/mrpc_200_llama3_8b_class2_6bins.png}
            \caption{Finetuning: $y_1$ (Blue) short demonstrations, $y_2$ (Orange) long demonstrations.}
        \end{subfigure}%
        \hfill
        \begin{subfigure}{0.31\linewidth}
            \centering
            \includegraphics[width=\textwidth]{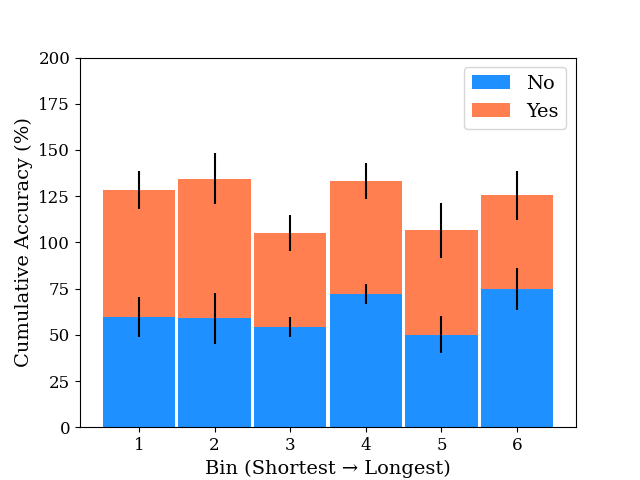}
            \caption{Intervention: $y_1$ (Blue) long demonstrations, $y_2$ (Orange) short demonstrations.}
        \end{subfigure}
        \hfill
        \begin{subfigure}{0.31\linewidth}
            \centering
            \includegraphics[width=\textwidth]{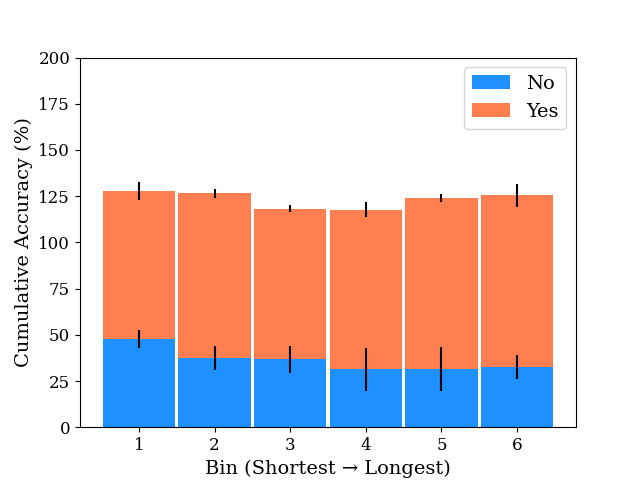}
            \caption{Intervention: $y_1$ (Blue) and $y_2$ (Orange) demonstrations randomly sampled.}
        \end{subfigure}
    \end{minipage}%
    \hfill
    \begin{minipage}[c]{\linewidth}
        \caption{MRPC (Llama 3 8B)}
    \end{minipage}
\end{figure*}

\begin{figure*}[t!]
    \centering
    \begin{minipage}[t]{\linewidth}
        \begin{subfigure}{0.31\linewidth}
            \centering
            \includegraphics[width=\textwidth]{latex/figures/ft/sst2_200_llama3_8b_class2_6bins.png}
            \caption{Finetuning: $y_1$ (Blue) short demonstrations, $y_2$ (Orange) long demonstrations.}
        \end{subfigure}%
        \hfill
        \begin{subfigure}{0.31\linewidth}
            \centering
            \includegraphics[width=\textwidth]{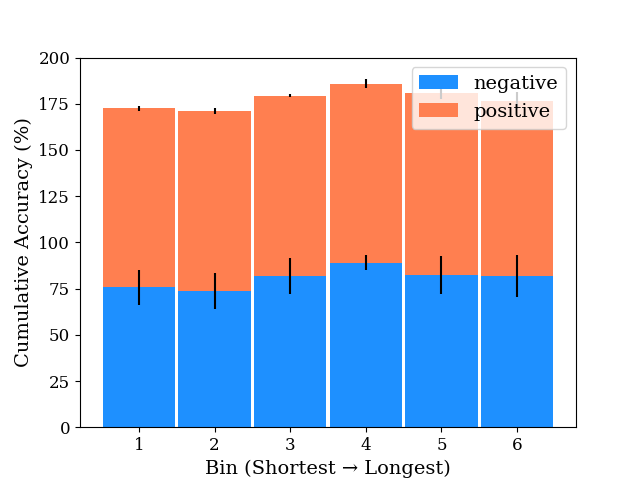}
            \caption{Intervention: $y_1$ (Blue) long demonstrations, $y_2$ (Orange) short demonstrations.}
        \end{subfigure}
        \hfill
        \begin{subfigure}{0.31\linewidth}
            \centering
            \includegraphics[width=\textwidth]{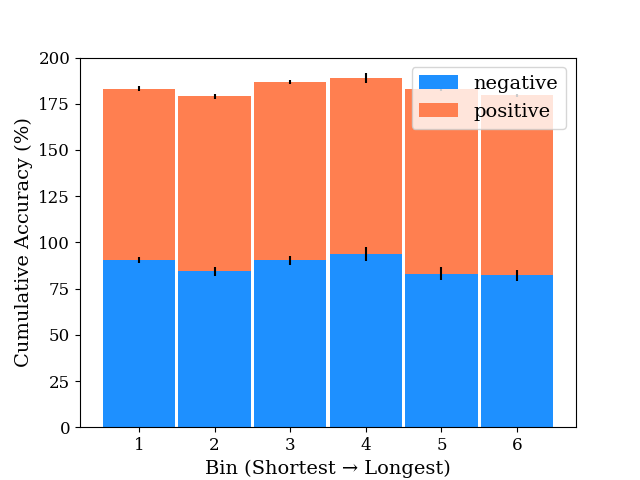}
            \caption{Intervention: $y_1$ (Blue) and $y_2$ (Orange) demonstrations randomly sampled.}
        \end{subfigure}
    \end{minipage}%
    \hfill
    \begin{minipage}[c]{\linewidth}
        \caption{SST-2 (Llama 3 8B)}
    \end{minipage}
\end{figure*}
\begin{figure*}[t!]
    \centering
    \begin{minipage}[t]{\linewidth}
        \begin{subfigure}{0.31\linewidth}
            \centering
            \includegraphics[width=\textwidth]{latex/figures/ft/hans_200_llama3_8b_class1_6bins.png}
            \caption{Finetuning: $y_1$ (Blue) long demonstrations, $y_2$ (Orange) short demonstrations.}
        \end{subfigure}%
        \hfill
        \begin{subfigure}{0.31\linewidth}
            \centering
            \includegraphics[width=\textwidth]{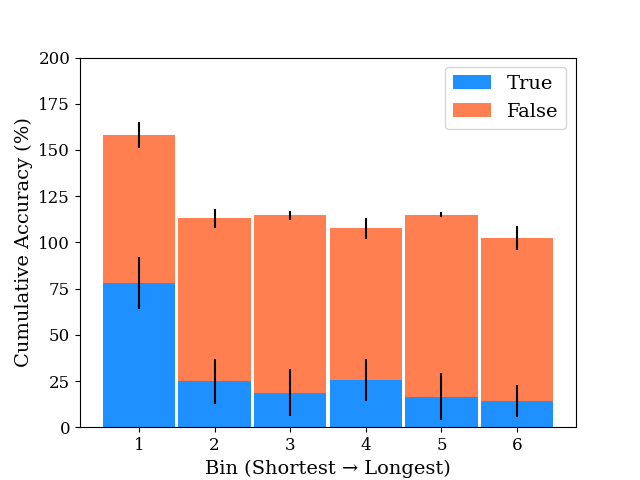}
            \caption{Intervention: $y_1$ (Blue) short demonstrations, $y_2$ (Orange) long demonstrations.}
        \end{subfigure}
        \hfill
        \begin{subfigure}{0.31\linewidth}
            \centering
            \includegraphics[width=\textwidth]{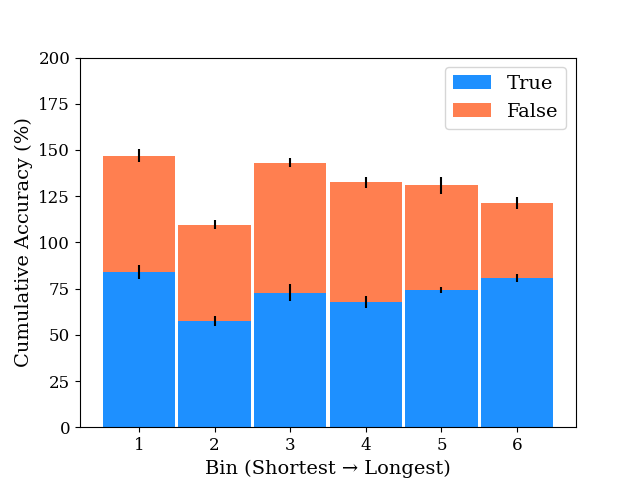}
            \caption{Intervention: $y_1$ (Blue) and $y_2$ (Orange) demonstrations randomly sampled.}
        \end{subfigure}
    \end{minipage}%
    \hfill
    \begin{minipage}[c]{\linewidth}
        \caption{Hans (Llama 3 8B)}
    \end{minipage}
\end{figure*}

\begin{figure*}[t!]
    \centering
    \begin{minipage}[t]{\linewidth}
        \begin{subfigure}{0.31\linewidth}
            \centering
            \includegraphics[width=\textwidth]{latex/figures/ft/paws_200_llama3_8b_class1_6bins.png}
            \caption{Finetuning: $y_1$ (Blue) long demonstrations, $y_2$ (Orange) short demonstrations.}
        \end{subfigure}%
        \hfill
        \begin{subfigure}{0.31\linewidth}
            \centering
            \includegraphics[width=\textwidth]{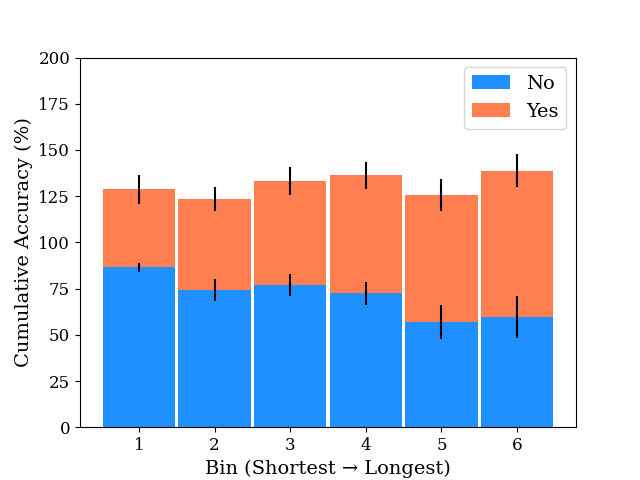}
            \caption{Intervention: $y_1$ (Blue) short demonstrations, $y_2$ (Orange) long demonstrations.}
        \end{subfigure}
        \hfill
        \begin{subfigure}{0.31\linewidth}
            \centering
            \includegraphics[width=\textwidth]{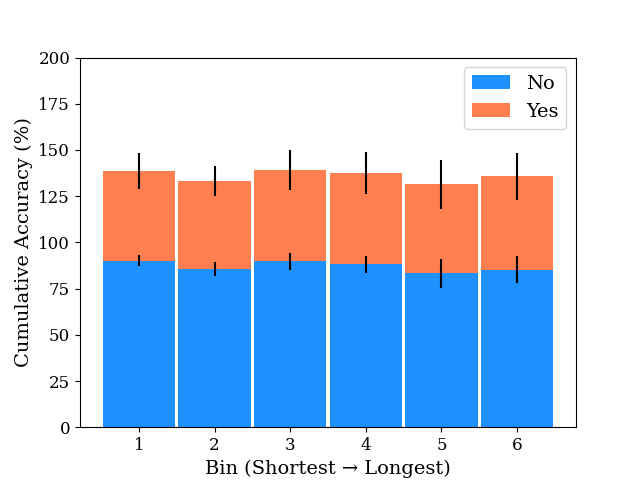}
            \caption{Intervention: $y_1$ (Blue) and $y_2$ (Orange) demonstrations randomly sampled.}
        \end{subfigure}
    \end{minipage}%
    \hfill
    \begin{minipage}[c]{\linewidth}
        \caption{PAWS-X$_{\textsc{EN}}$ (Llama 3 8B)}
    \end{minipage}
\end{figure*}

\begin{figure*}[t!]
    \centering
    \begin{minipage}[t]{\linewidth}
        \begin{subfigure}{0.31\linewidth}
            \centering
            \includegraphics[width=\textwidth]{latex/figures/ft/rte_200_llama3_8b_class1_6bins.png}
            \caption{Finetuning: $y_1$ (Blue) long demonstrations, $y_2$ (Orange) short demonstrations.}
        \end{subfigure}%
        \hfill
        \begin{subfigure}{0.31\linewidth}
            \centering
            \includegraphics[width=\textwidth]{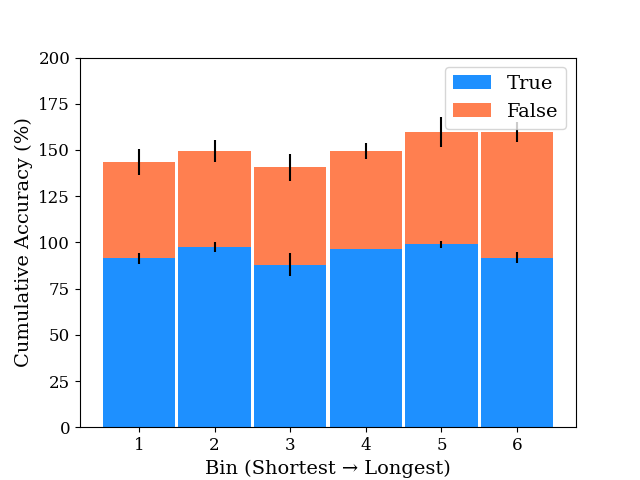}
            \caption{Intervention: $y_1$ (Blue) short demonstrations, $y_2$ (Orange) long demonstrations.}
        \end{subfigure}
        \hfill
        \begin{subfigure}{0.31\linewidth}
            \centering
            \includegraphics[width=\textwidth]{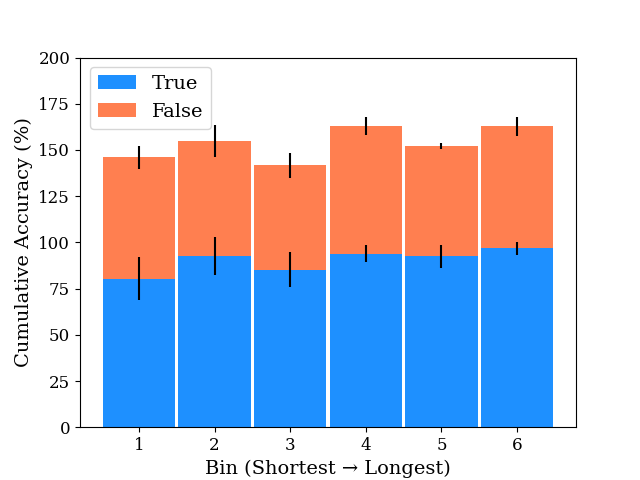}
            \caption{Intervention: $y_1$ (Blue) and $y_2$ (Orange) demonstrations randomly sampled.}
        \end{subfigure}
    \end{minipage}%
    \hfill
    \begin{minipage}[c]{\linewidth}
        \caption{RTE (Llama 3 8B)}
    \end{minipage}
\end{figure*}

\begin{figure*}[t!]
    \centering
    \begin{minipage}[t]{\linewidth}
        \begin{subfigure}{0.31\linewidth}
            \centering
            \includegraphics[width=\textwidth]{latex/figures/ft/qnli_200_llama3_8b_class1_6bins.png}
            \caption{Finetuning: $y_1$ (Blue) long demonstrations, $y_2$ (Orange) short demonstrations.}
        \end{subfigure}%
        \hfill
        \begin{subfigure}{0.31\linewidth}
            \centering
            \includegraphics[width=\textwidth]{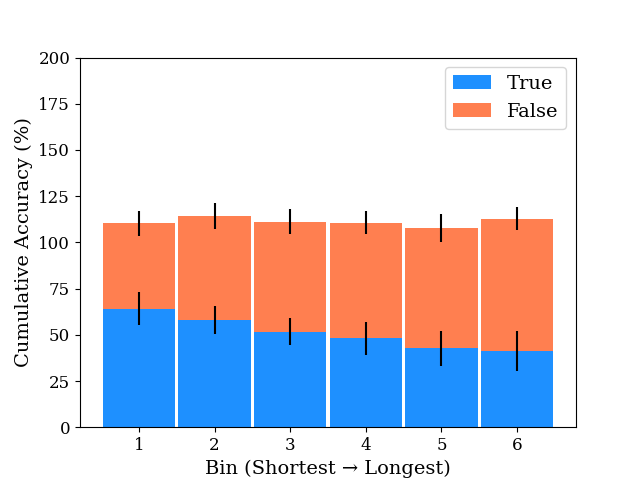}
            \caption{Intervention: $y_1$ (Blue) short demonstrations, $y_2$ (Orange) long demonstrations.}
        \end{subfigure}
        \hfill
        \begin{subfigure}{0.31\linewidth}
            \centering
            \includegraphics[width=\textwidth]{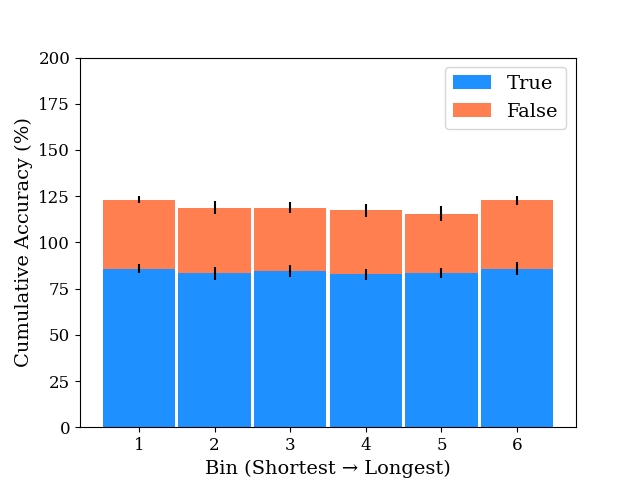}
            \caption{Intervention: $y_1$ (Blue) and $y_2$ (Orange) demonstrations randomly sampled.}
        \end{subfigure}
    \end{minipage}%
    \hfill
    \begin{minipage}[c]{\linewidth}
        \caption{QNLI (Llama 3 8B)}
    \end{minipage}
\end{figure*}

\begin{figure*}[t!]
    \centering
    \begin{minipage}[t]{\linewidth}
        \begin{subfigure}{0.31\linewidth}
            \centering
            \includegraphics[width=\textwidth]{latex/figures/ft/wnli_200_llama3_8b_class1_6bins.png}
            \caption{Finetuning: $y_1$ (Blue) long demonstrations, $y_2$ (Orange) short demonstrations.}
        \end{subfigure}%
        \hfill
        \begin{subfigure}{0.31\linewidth}
            \centering
            \includegraphics[width=\textwidth]{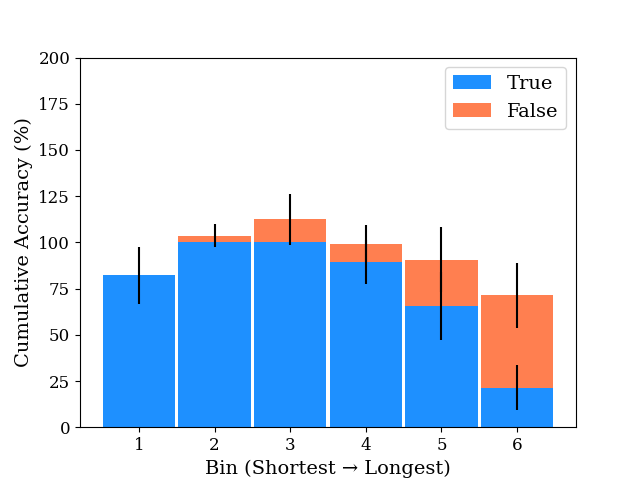}
            \caption{Intervention: $y_1$ (Blue) short demonstrations, $y_2$ (Orange) long demonstrations.}
        \end{subfigure}
        \hfill
        \begin{subfigure}{0.31\linewidth}
            \centering
            \includegraphics[width=\textwidth]{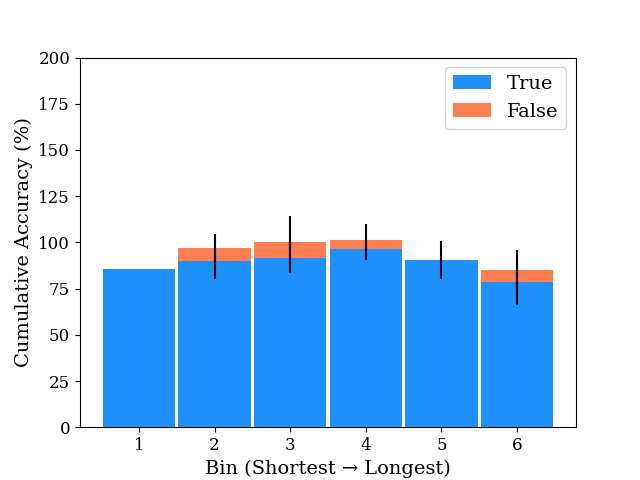}
            \caption{Intervention: $y_1$ (Blue) and $y_2$ (Orange) demonstrations randomly sampled.}
        \end{subfigure}
    \end{minipage}%
    \hfill
    \begin{minipage}[c]{\linewidth}
        \caption{WNLI (Llama 3 8B)}
    \end{minipage}
\end{figure*}

\begin{figure*}[t!]
    \centering
    \begin{minipage}[t]{\linewidth}
        \begin{subfigure}{0.31\linewidth}
            \centering
            \includegraphics[width=\textwidth]{latex/figures/ft/mrpc_200_llama3_8b_class1_6bins.png}
            \caption{Finetuning: $y_1$ (Blue) long demonstrations, $y_2$ (Orange) short demonstrations.}
        \end{subfigure}%
        \hfill
        \begin{subfigure}{0.31\linewidth}
            \centering
            \includegraphics[width=\textwidth]{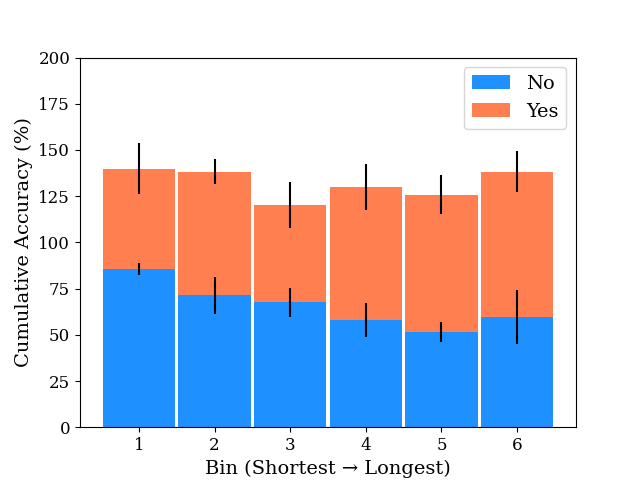}
            \caption{Intervention: $y_1$ (Blue) short demonstrations, $y_2$ (Orange) long demonstrations.}
        \end{subfigure}
        \hfill
        \begin{subfigure}{0.31\linewidth}
            \centering
            \includegraphics[width=\textwidth]{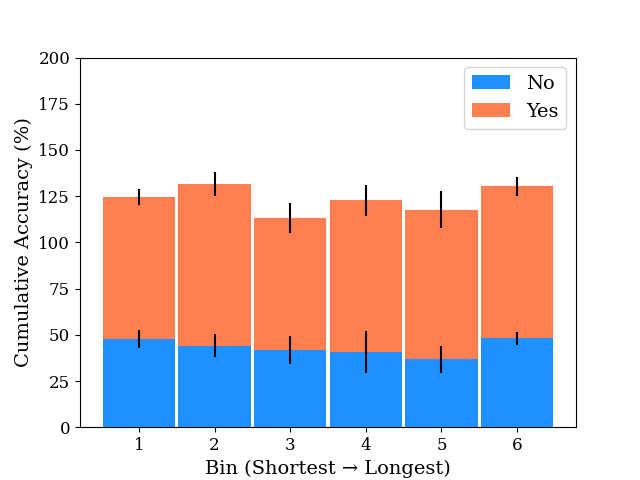}
            \caption{Intervention: $y_1$ (Blue) and $y_2$ (Orange) demonstrations randomly sampled.}
        \end{subfigure}
    \end{minipage}%
    \hfill
    \begin{minipage}[c]{\linewidth}
        \caption{MRPC (Llama 3 8B)}
    \end{minipage}
\end{figure*}

\begin{figure*}[t!]
    \centering
    \begin{minipage}[t]{\linewidth}
        \begin{subfigure}{0.31\linewidth}
            \centering
            \includegraphics[width=\textwidth]{latex/figures/ft/sst2_200_llama3_8b_class1_6bins.png}
            \caption{Finetuning: $y_1$ (Blue) long demonstrations, $y_2$ (Orange) short demonstrations.}
        \end{subfigure}%
        \hfill
        \begin{subfigure}{0.31\linewidth}
            \centering
            \includegraphics[width=\textwidth]{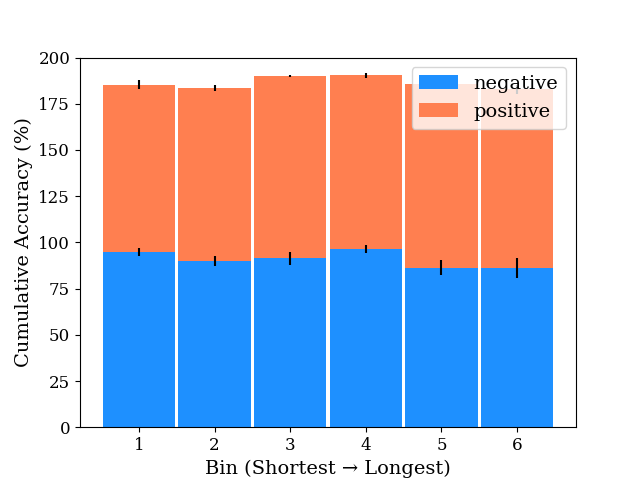}
            \caption{Intervention: $y_1$ (Blue) short demonstrations, $y_2$ (Orange) long demonstrations.}
        \end{subfigure}
        \hfill
        \begin{subfigure}{0.31\linewidth}
            \centering
            \includegraphics[width=\textwidth]{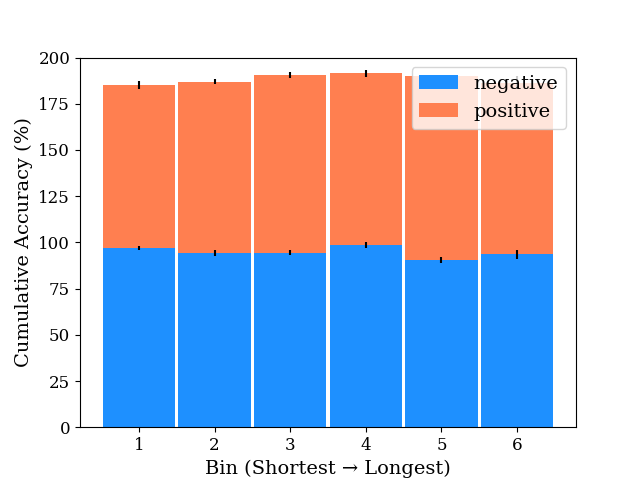}
            \caption{Intervention: $y_1$ (Blue) and $y_2$ (Orange) demonstrations randomly sampled.}
        \end{subfigure}
    \end{minipage}%
    \hfill
    \begin{minipage}[c]{\linewidth}
        \caption{SST-2 (Llama 3 8B)}
    \end{minipage}
\end{figure*}
\begin{figure*}[t!]
    \centering
    \begin{minipage}[t]{\linewidth}
        \begin{subfigure}{0.31\linewidth}
            \centering
            \includegraphics[width=\textwidth]{latex/figures/ft/hans_200_gpt_2dot7b_class2_6bins.png}
            \caption{Finetuning: $y_1$ (Blue) short demonstrations, $y_2$ (Orange) long demonstrations.}
        \end{subfigure}%
        \hfill
        \begin{subfigure}{0.31\linewidth}
            \centering
            \includegraphics[width=\textwidth]{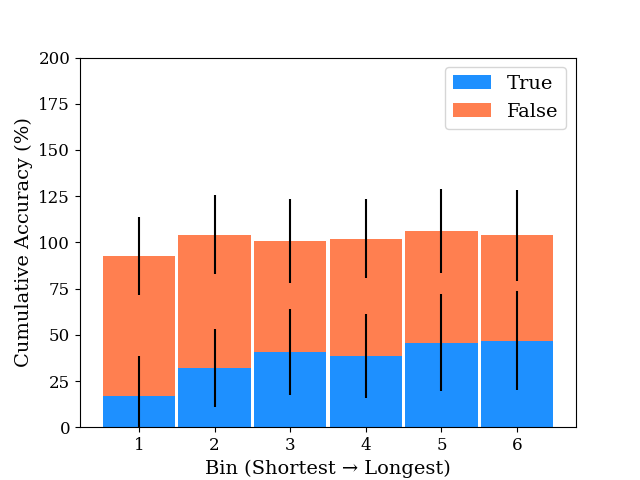}
            \caption{Intervention: $y_1$ (Blue) long demonstrations, $y_2$ (Orange) short demonstrations.}
        \end{subfigure}
        \hfill
        \begin{subfigure}{0.31\linewidth}
            \centering
            \includegraphics[width=\textwidth]{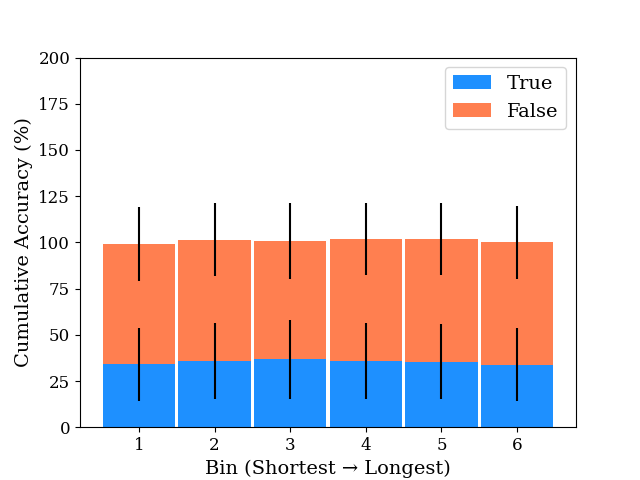}
            \caption{Intervention: $y_1$ (Blue) and $y_2$ (Orange) demonstrations randomly sampled.}
        \end{subfigure}
    \end{minipage}%
    \hfill
    \begin{minipage}[c]{\linewidth}
        \caption{Hans (GPT Neo 2.7B)}
    \end{minipage}
\end{figure*}

\begin{figure*}[t!]
    \centering
    \begin{minipage}[t]{\linewidth}
        \begin{subfigure}{0.31\linewidth}
            \centering
            \includegraphics[width=\textwidth]{latex/figures/ft/paws_200_gpt_2dot7b_class2_6bins.png}
            \caption{Finetuning: $y_1$ (Blue) short demonstrations, $y_2$ (Orange) long demonstrations.}
        \end{subfigure}%
        \hfill
        \begin{subfigure}{0.31\linewidth}
            \centering
            \includegraphics[width=\textwidth]{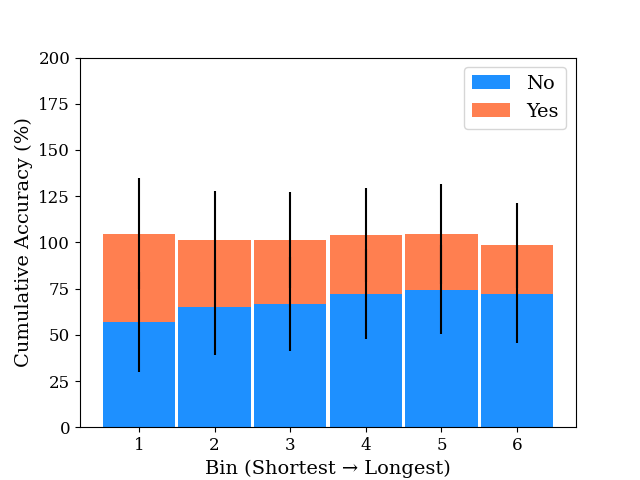}
            \caption{Intervention: $y_1$ (Blue) long demonstrations, $y_2$ (Orange) short demonstrations.}
        \end{subfigure}
        \hfill
        \begin{subfigure}{0.31\linewidth}
            \centering
            \includegraphics[width=\textwidth]{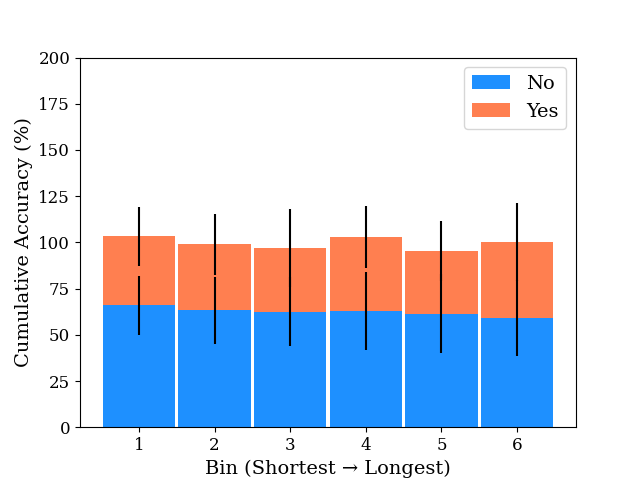}
            \caption{Intervention: $y_1$ (Blue) and $y_2$ (Orange) demonstrations randomly sampled.}
        \end{subfigure}
    \end{minipage}%
    \hfill
    \begin{minipage}[c]{\linewidth}
        \caption{PAWS-X$_{\textsc{EN}}$ (GPT Neo 2.7B)}
    \end{minipage}
\end{figure*}

\begin{figure*}[t!]
    \centering
    \begin{minipage}[t]{\linewidth}
        \begin{subfigure}{0.31\linewidth}
            \centering
            \includegraphics[width=\textwidth]{latex/figures/ft/rte_200_gpt_2dot7b_class2_6bins.png}
            \caption{Finetuning: $y_1$ (Blue) short demonstrations, $y_2$ (Orange) long demonstrations.}
        \end{subfigure}%
        \hfill
        \begin{subfigure}{0.31\linewidth}
            \centering
            \includegraphics[width=\textwidth]{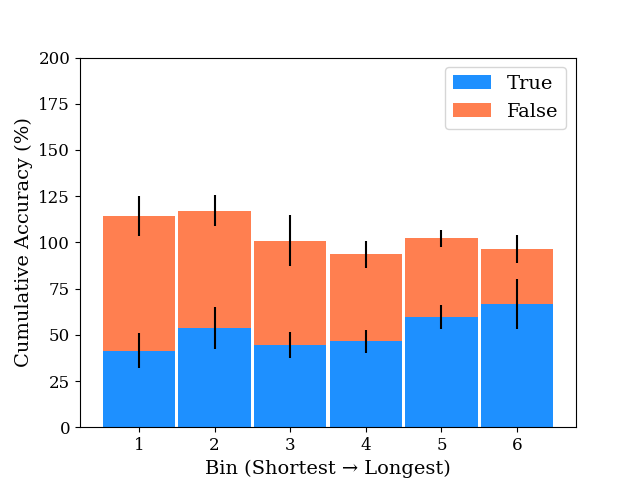}
            \caption{Intervention: $y_1$ (Blue) long demonstrations, $y_2$ (Orange) short demonstrations.}
        \end{subfigure}
        \hfill
        \begin{subfigure}{0.31\linewidth}
            \centering
            \includegraphics[width=\textwidth]{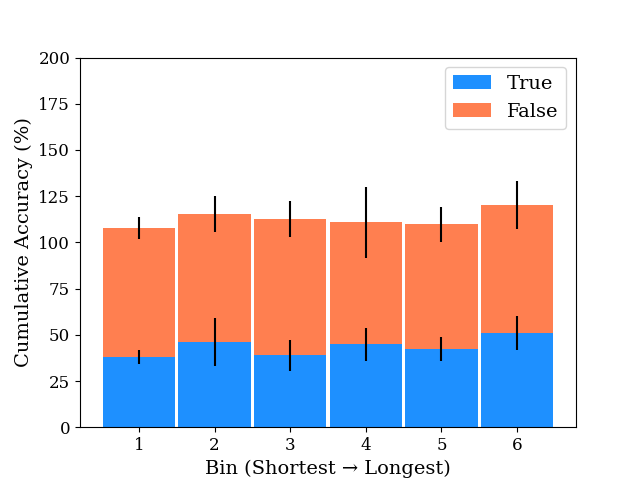}
            \caption{Intervention: $y_1$ (Blue) and $y_2$ (Orange) demonstrations randomly sampled.}
        \end{subfigure}
    \end{minipage}%
    \hfill
    \begin{minipage}[c]{\linewidth}
        \caption{RTE (GPT Neo 2.7B)}
    \end{minipage}
\end{figure*}

\begin{figure*}[t!]
    \centering
    \begin{minipage}[t]{\linewidth}
        \begin{subfigure}{0.31\linewidth}
            \centering
            \includegraphics[width=\textwidth]{latex/figures/ft/qnli_200_gpt_2dot7b_class2_6bins.png}
            \caption{Finetuning: $y_1$ (Blue) short demonstrations, $y_2$ (Orange) long demonstrations.}
        \end{subfigure}%
        \hfill
        \begin{subfigure}{0.31\linewidth}
            \centering
            \includegraphics[width=\textwidth]{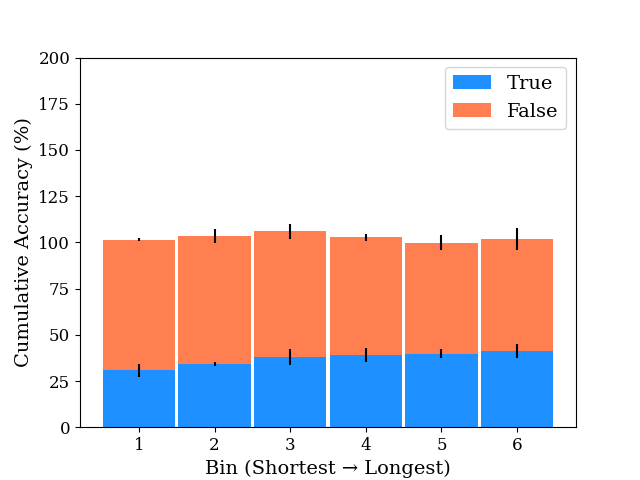}
            \caption{Intervention: $y_1$ (Blue) long demonstrations, $y_2$ (Orange) short demonstrations.}
        \end{subfigure}
        \hfill
        \begin{subfigure}{0.31\linewidth}
            \centering
            \includegraphics[width=\textwidth]{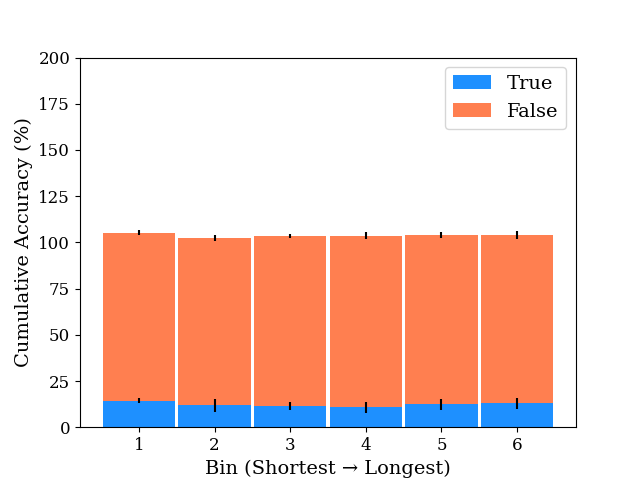}
            \caption{Intervention: $y_1$ (Blue) and $y_2$ (Orange) demonstrations randomly sampled.}
        \end{subfigure}
    \end{minipage}%
    \hfill
    \begin{minipage}[c]{\linewidth}
        \caption{QNLI (GPT Neo 2.7B)}
    \end{minipage}
\end{figure*}

\begin{figure*}[t!]
    \centering
    \begin{minipage}[t]{\linewidth}
        \begin{subfigure}{0.31\linewidth}
            \centering
            \includegraphics[width=\textwidth]{latex/figures/ft/wnli_200_gpt_2dot7b_class2_6bins.png}
            \caption{Finetuning: $y_1$ (Blue) short demonstrations, $y_2$ (Orange) long demonstrations.}
        \end{subfigure}%
        \hfill
        \begin{subfigure}{0.31\linewidth}
            \centering
            \includegraphics[width=\textwidth]{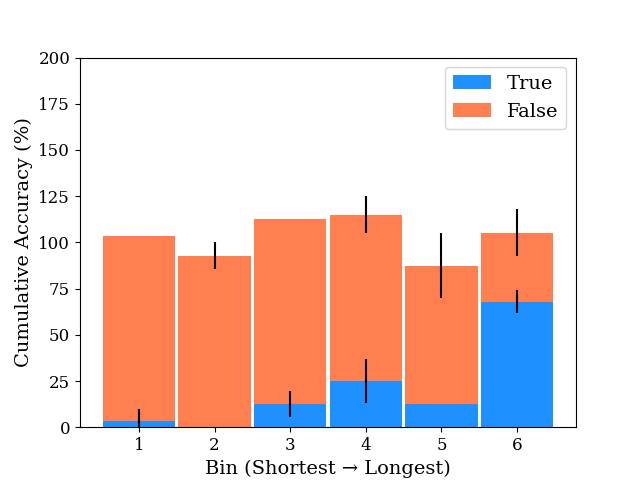}
            \caption{Intervention: $y_1$ (Blue) long demonstrations, $y_2$ (Orange) short demonstrations.}
        \end{subfigure}
        \hfill
        \begin{subfigure}{0.31\linewidth}
            \centering
            \includegraphics[width=\textwidth]{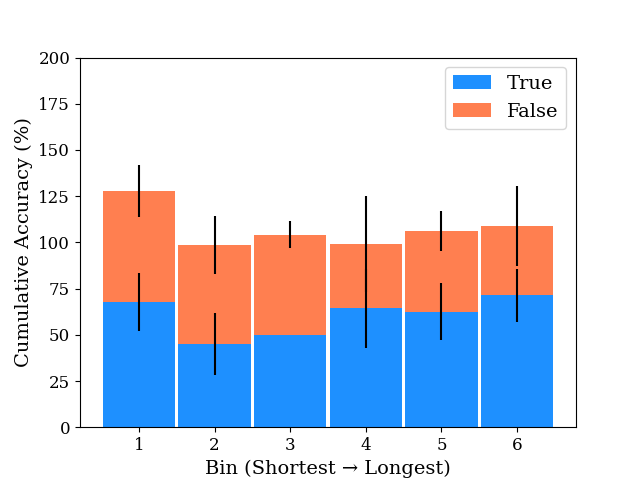}
            \caption{Intervention: $y_1$ (Blue) and $y_2$ (Orange) demonstrations randomly sampled.}
        \end{subfigure}
    \end{minipage}%
    \hfill
    \begin{minipage}[c]{\linewidth}
        \caption{WNLI (GPT Neo 2.7B)}
    \end{minipage}
\end{figure*}

\begin{figure*}[t!]
    \centering
    \begin{minipage}[t]{\linewidth}
        \begin{subfigure}{0.31\linewidth}
            \centering
            \includegraphics[width=\textwidth]{latex/figures/ft/mrpc_200_gpt_2dot7b_class2_6bins.png}
            \caption{Finetuning: $y_1$ (Blue) short demonstrations, $y_2$ (Orange) long demonstrations.}
        \end{subfigure}%
        \hfill
        \begin{subfigure}{0.31\linewidth}
            \centering
            \includegraphics[width=\textwidth]{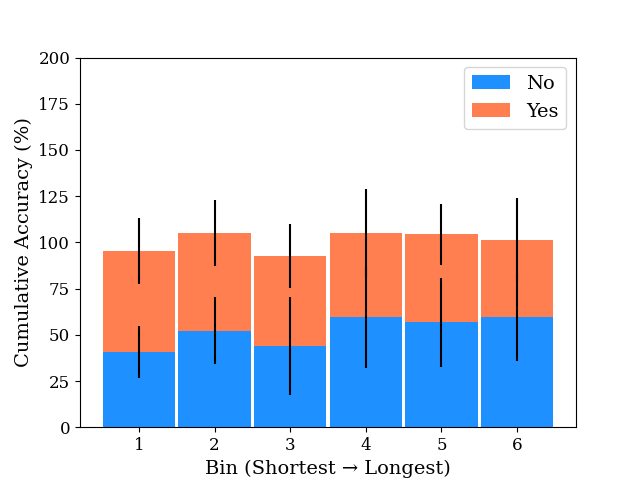}
            \caption{Intervention: $y_1$ (Blue) long demonstrations, $y_2$ (Orange) short demonstrations.}
        \end{subfigure}
        \hfill
        \begin{subfigure}{0.31\linewidth}
            \centering
            \includegraphics[width=\textwidth]{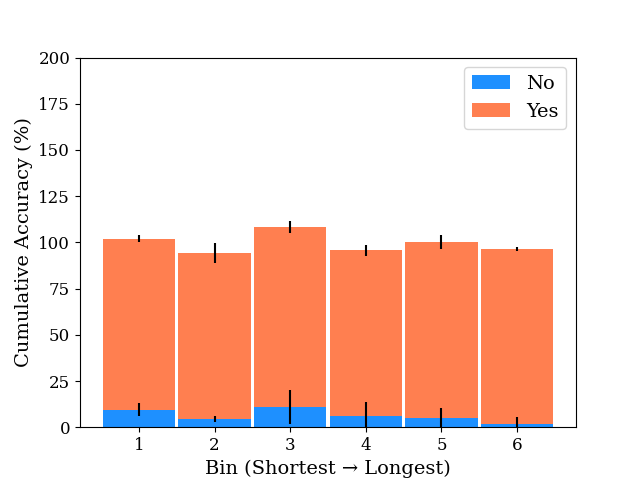}
            \caption{Intervention: $y_1$ (Blue) and $y_2$ (Orange) demonstrations randomly sampled.}
        \end{subfigure}
    \end{minipage}%
    \hfill
    \begin{minipage}[c]{\linewidth}
        \caption{MRPC (GPT Neo 2.7B)}
    \end{minipage}
\end{figure*}

\begin{figure*}[t!]
    \centering
    \begin{minipage}[t]{\linewidth}
        \begin{subfigure}{0.31\linewidth}
            \centering
            \includegraphics[width=\textwidth]{latex/figures/ft/sst2_200_gpt_2dot7b_class2_6bins.png}
            \caption{Finetuning: $y_1$ (Blue) short demonstrations, $y_2$ (Orange) long demonstrations.}
        \end{subfigure}%
        \hfill
        \begin{subfigure}{0.31\linewidth}
            \centering
            \includegraphics[width=\textwidth]{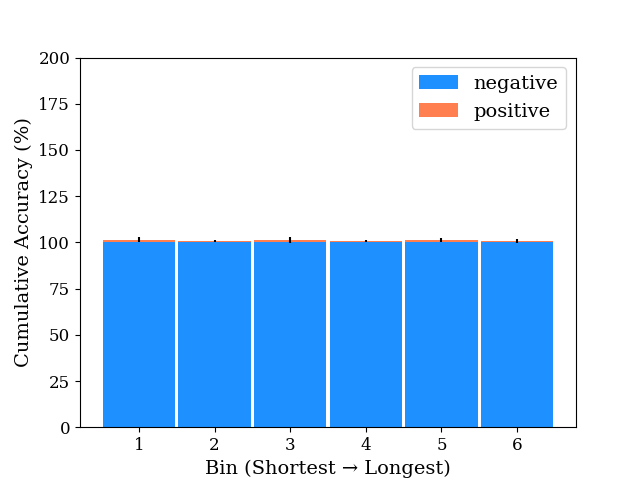}
            \caption{Intervention: $y_1$ (Blue) long demonstrations, $y_2$ (Orange) short demonstrations.}
        \end{subfigure}
        \hfill
        \begin{subfigure}{0.31\linewidth}
            \centering
            \includegraphics[width=\textwidth]{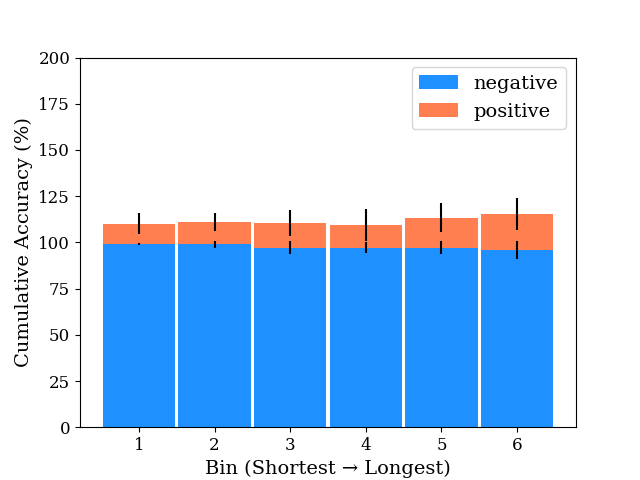}
            \caption{Intervention: $y_1$ (Blue) and $y_2$ (Orange) demonstrations randomly sampled.}
        \end{subfigure}
    \end{minipage}%
    \hfill
    \begin{minipage}[c]{\linewidth}
        \caption{SST-2 (GPT Neo 2.7B)}
    \end{minipage}
\end{figure*}
\begin{figure*}[t!]
    \centering
    \begin{minipage}[t]{\linewidth}
        \begin{subfigure}{0.31\linewidth}
            \centering
            \includegraphics[width=\textwidth]{latex/figures/ft/hans_200_gpt_2dot7b_class1_6bins.png}
            \caption{Finetuning: $y_1$ (Blue) long demonstrations, $y_2$ (Orange) short demonstrations.}
        \end{subfigure}%
        \hfill
        \begin{subfigure}{0.31\linewidth}
            \centering
            \includegraphics[width=\textwidth]{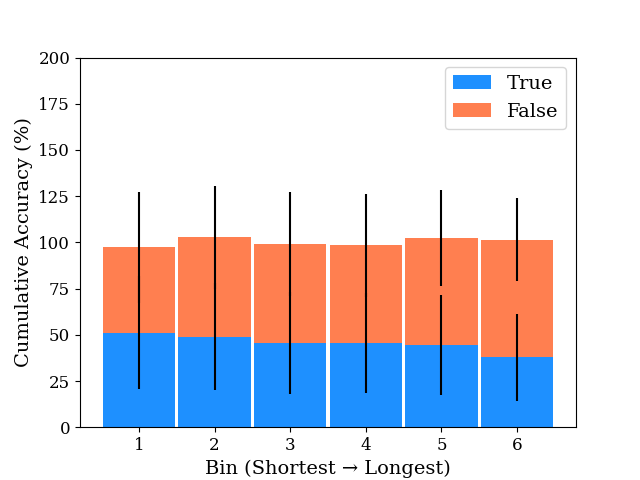}
            \caption{Intervention: $y_1$ (Blue) short demonstrations, $y_2$ (Orange) long demonstrations.}
        \end{subfigure}
        \hfill
        \begin{subfigure}{0.31\linewidth}
            \centering
            \includegraphics[width=\textwidth]{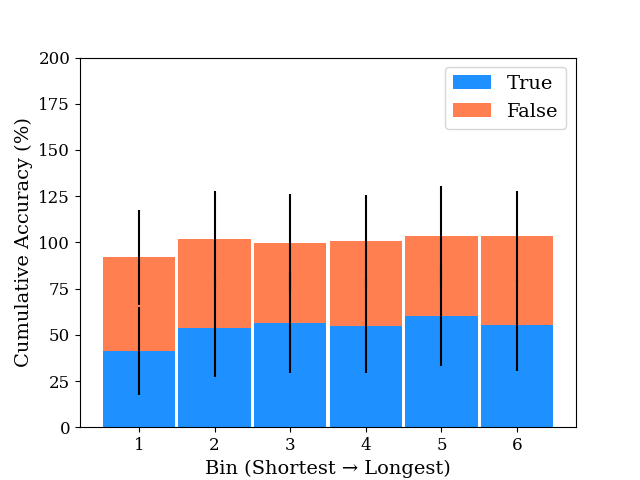}
            \caption{Intervention: $y_1$ (Blue) and $y_2$ (Orange) demonstrations randomly sampled.}
        \end{subfigure}
    \end{minipage}%
    \hfill
    \begin{minipage}[c]{\linewidth}
        \caption{Hans (GPT Neo 2.7B)}
    \end{minipage}
\end{figure*}

\begin{figure*}[t!]
    \centering
    \begin{minipage}[t]{\linewidth}
        \begin{subfigure}{0.31\linewidth}
            \centering
            \includegraphics[width=\textwidth]{latex/figures/ft/paws_200_gpt_2dot7b_class1_6bins.png}
            \caption{Finetuning: $y_1$ (Blue) long demonstrations, $y_2$ (Orange) short demonstrations.}
        \end{subfigure}%
        \hfill
        \begin{subfigure}{0.31\linewidth}
            \centering
            \includegraphics[width=\textwidth]{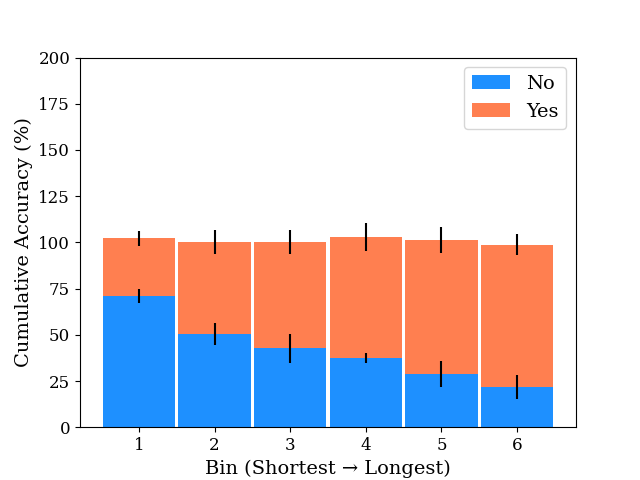}
            \caption{Intervention: $y_1$ (Blue) short demonstrations, $y_2$ (Orange) long demonstrations.}
        \end{subfigure}
        \hfill
        \begin{subfigure}{0.31\linewidth}
            \centering
            \includegraphics[width=\textwidth]{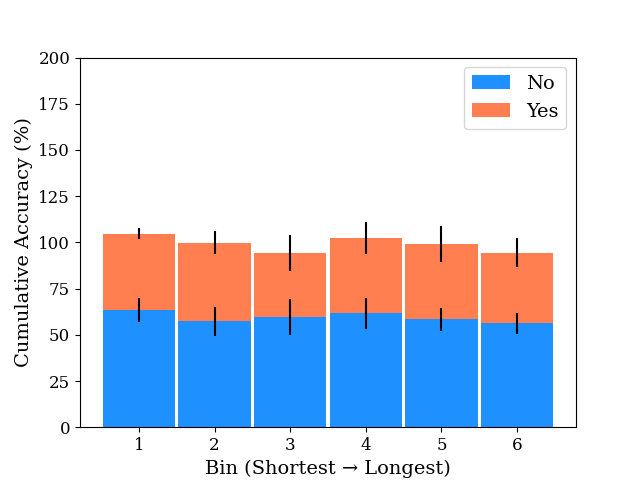}
            \caption{Intervention: $y_1$ (Blue) and $y_2$ (Orange) demonstrations randomly sampled.}
        \end{subfigure}
    \end{minipage}%
    \hfill
    \begin{minipage}[c]{\linewidth}
        \caption{PAWS-X$_{\textsc{EN}}$ (GPT Neo 2.7B)}
    \end{minipage}
\end{figure*}

\begin{figure*}[t!]
    \centering
    \begin{minipage}[t]{\linewidth}
        \begin{subfigure}{0.31\linewidth}
            \centering
            \includegraphics[width=\textwidth]{latex/figures/ft/rte_200_gpt_2dot7b_class1_6bins.png}
            \caption{Finetuning: $y_1$ (Blue) long demonstrations, $y_2$ (Orange) short demonstrations.}
        \end{subfigure}%
        \hfill
        \begin{subfigure}{0.31\linewidth}
            \centering
            \includegraphics[width=\textwidth]{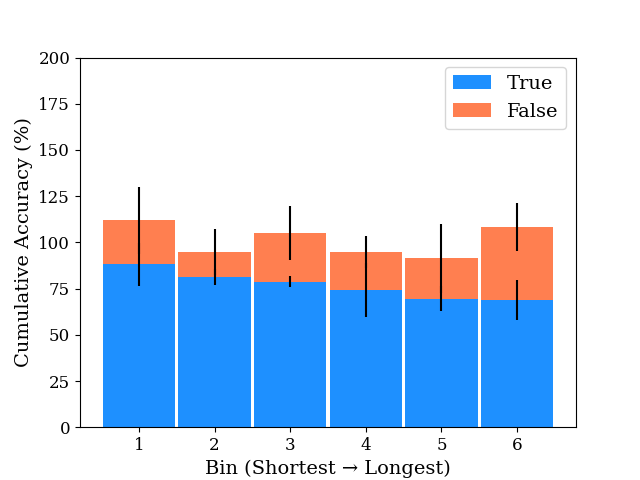}
            \caption{Intervention: $y_1$ (Blue) short demonstrations, $y_2$ (Orange) long demonstrations.}
        \end{subfigure}
        \hfill
        \begin{subfigure}{0.31\linewidth}
            \centering
            \includegraphics[width=\textwidth]{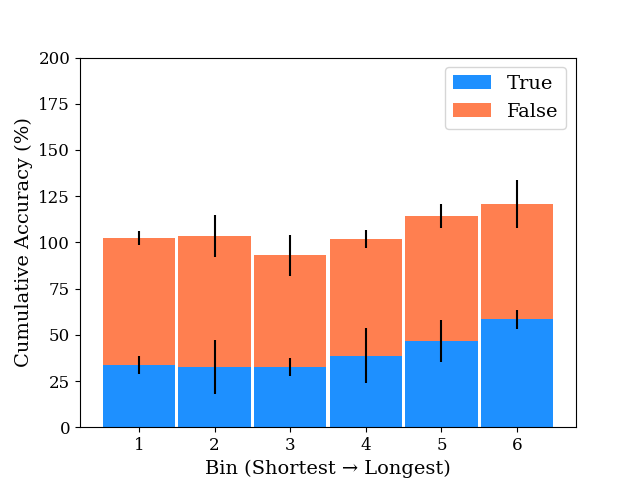}
            \caption{Intervention: $y_1$ (Blue) and $y_2$ (Orange) demonstrations randomly sampled.}
        \end{subfigure}
    \end{minipage}%
    \hfill
    \begin{minipage}[c]{\linewidth}
        \caption{RTE (GPT Neo 2.7B)}
    \end{minipage}
\end{figure*}

\begin{figure*}[t!]
    \centering
    \begin{minipage}[t]{\linewidth}
        \begin{subfigure}{0.31\linewidth}
            \centering
            \includegraphics[width=\textwidth]{latex/figures/ft/qnli_200_gpt_2dot7b_class1_6bins.png}
            \caption{Finetuning: $y_1$ (Blue) long demonstrations, $y_2$ (Orange) short demonstrations.}
        \end{subfigure}%
        \hfill
        \begin{subfigure}{0.31\linewidth}
            \centering
            \includegraphics[width=\textwidth]{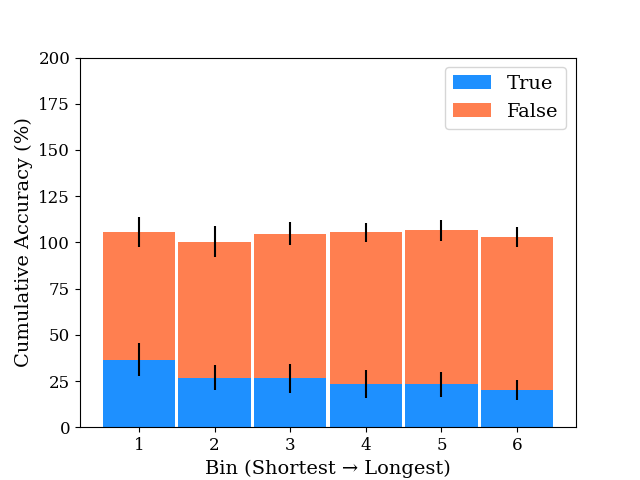}
            \caption{Intervention: $y_1$ (Blue) short demonstrations, $y_2$ (Orange) long demonstrations.}
        \end{subfigure}
        \hfill
        \begin{subfigure}{0.31\linewidth}
            \centering
            \includegraphics[width=\textwidth]{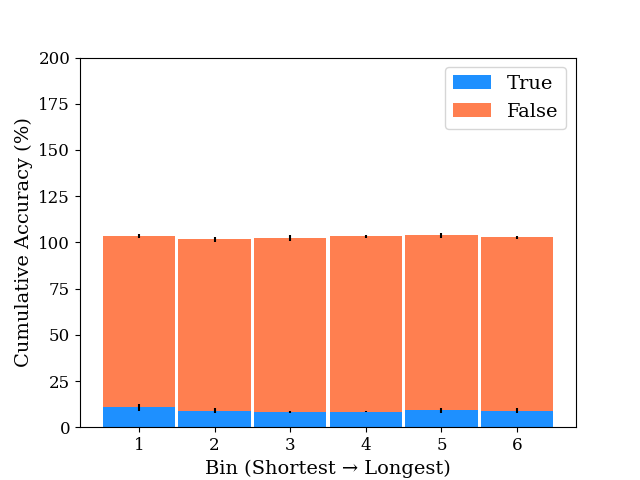}
            \caption{Intervention: $y_1$ (Blue) and $y_2$ (Orange) demonstrations randomly sampled.}
        \end{subfigure}
    \end{minipage}%
    \hfill
    \begin{minipage}[c]{\linewidth}
        \caption{QNLI (GPT Neo 2.7B)}
    \end{minipage}
\end{figure*}

\begin{figure*}[t!]
    \centering
    \begin{minipage}[t]{\linewidth}
        \begin{subfigure}{0.31\linewidth}
            \centering
            \includegraphics[width=\textwidth]{latex/figures/ft/wnli_200_gpt_2dot7b_class1_6bins.png}
            \caption{Finetuning: $y_1$ (Blue) long demonstrations, $y_2$ (Orange) short demonstrations.}
        \end{subfigure}%
        \hfill
        \begin{subfigure}{0.31\linewidth}
            \centering
            \includegraphics[width=\textwidth]{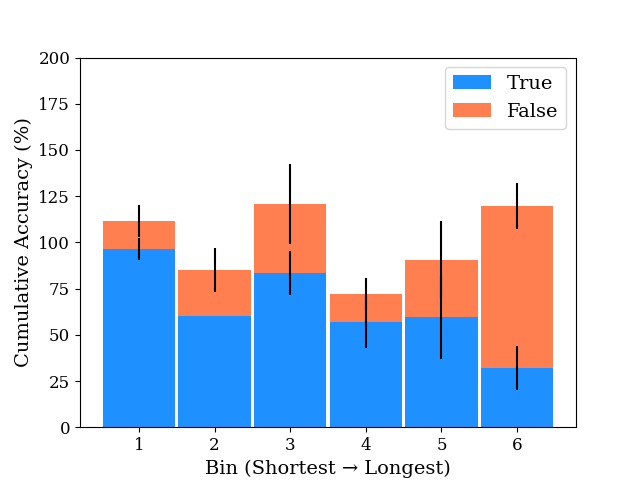}
            \caption{Intervention: $y_1$ (Blue) short demonstrations, $y_2$ (Orange) long demonstrations.}
        \end{subfigure}
        \hfill
        \begin{subfigure}{0.31\linewidth}
            \centering
            \includegraphics[width=\textwidth]{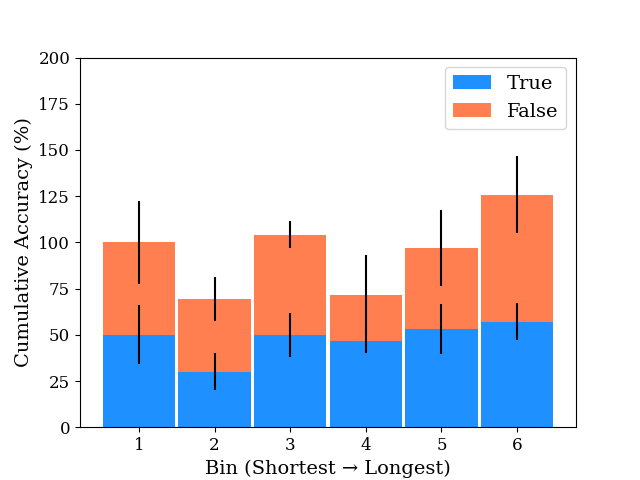}
            \caption{Intervention: $y_1$ (Blue) and $y_2$ (Orange) demonstrations randomly sampled.}
        \end{subfigure}
    \end{minipage}%
    \hfill
    \begin{minipage}[c]{\linewidth}
        \caption{WNLI (GPT Neo 2.7B)}
    \end{minipage}
\end{figure*}

\begin{figure*}[t!]
    \centering
    \begin{minipage}[t]{\linewidth}
        \begin{subfigure}{0.31\linewidth}
            \centering
            \includegraphics[width=\textwidth]{latex/figures/ft/mrpc_200_gpt_2dot7b_class1_6bins.png}
            \caption{Finetuning: $y_1$ (Blue) long demonstrations, $y_2$ (Orange) short demonstrations.}
        \end{subfigure}%
        \hfill
        \begin{subfigure}{0.31\linewidth}
            \centering
            \includegraphics[width=\textwidth]{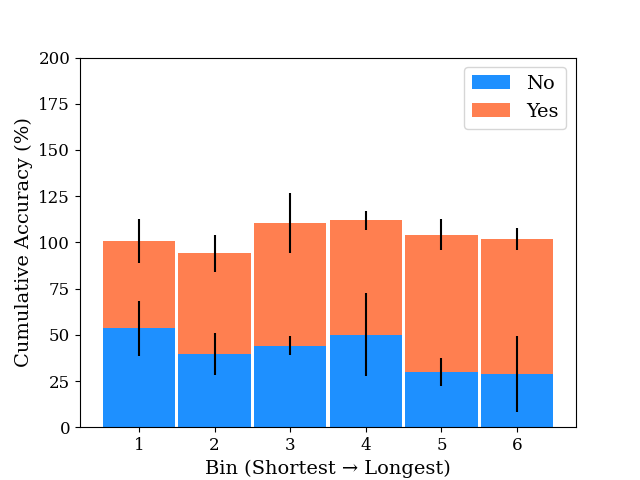}
            \caption{Intervention: $y_1$ (Blue) short demonstrations, $y_2$ (Orange) long demonstrations.}
        \end{subfigure}
        \hfill
        \begin{subfigure}{0.31\linewidth}
            \centering
            \includegraphics[width=\textwidth]{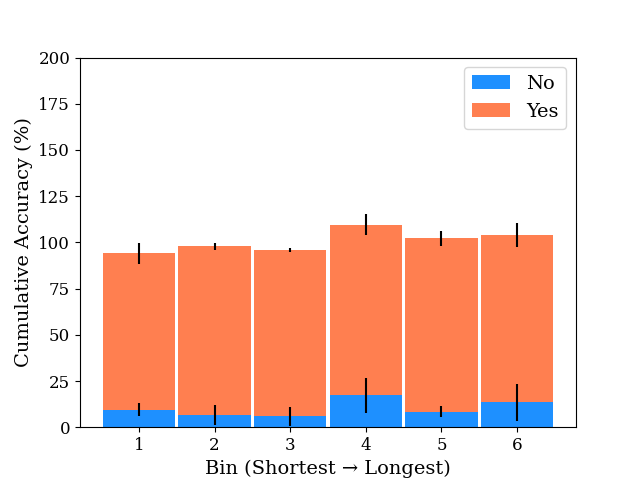}
            \caption{Intervention: $y_1$ (Blue) and $y_2$ (Orange) demonstrations randomly sampled.}
        \end{subfigure}
    \end{minipage}%
    \hfill
    \begin{minipage}[c]{\linewidth}
        \caption{MRPC (GPT Neo 2.7B)}
    \end{minipage}
\end{figure*}

\begin{figure*}[t!]
    \centering
    \begin{minipage}[t]{\linewidth}
        \begin{subfigure}{0.31\linewidth}
            \centering
            \includegraphics[width=\textwidth]{latex/figures/ft/sst2_200_gpt_2dot7b_class1_6bins.png}
            \caption{Finetuning: $y_1$ (Blue) long demonstrations, $y_2$ (Orange) short demonstrations.}
        \end{subfigure}%
        \hfill
        \begin{subfigure}{0.31\linewidth}
            \centering
            \includegraphics[width=\textwidth]{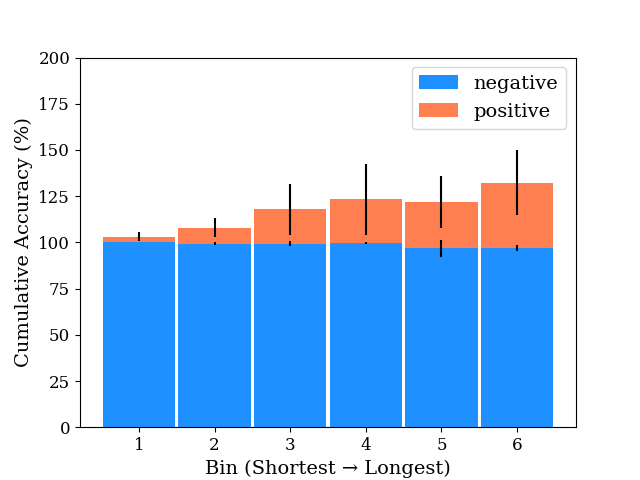}
            \caption{Intervention: $y_1$ (Blue) short demonstrations, $y_2$ (Orange) long demonstrations.}
        \end{subfigure}
        \hfill
        \begin{subfigure}{0.31\linewidth}
            \centering
            \includegraphics[width=\textwidth]{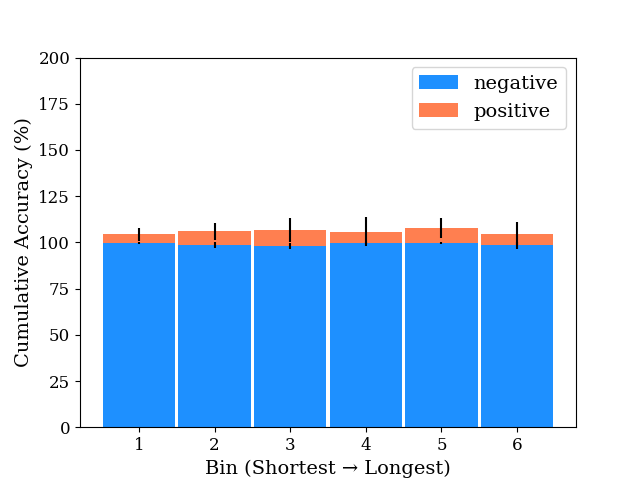}
            \caption{Intervention: $y_1$ (Blue) and $y_2$ (Orange) demonstrations randomly sampled.}
        \end{subfigure}
    \end{minipage}%
    \hfill
    \begin{minipage}[c]{\linewidth}
        \caption{SST-2 (GPT Neo 2.7B)}
    \end{minipage}
\end{figure*}

\end{document}